\documentclass[10pt,conference]{IEEEtran}
\IEEEoverridecommandlockouts

\DeclareRobustCommand{\IEEEauthorrefmark}[1]{\smash{\textsuperscript{\footnotesize #1}}}

\usepackage{amsmath,amssymb,amsfonts}
\usepackage{algorithmic}
\usepackage{graphicx}
\usepackage{textcomp}
\usepackage{xcolor}
\usepackage{pgfplotstable}
\usepgfplotslibrary{groupplots}
\usepackage{booktabs}
\usepackage[dvipsnames]{xcolor}
\usepackage{adjustbox}
\usepackage{array}
\usepackage[ruled,vlined,linesnumbered]{algorithm2e}
\usepackage{dblfloatfix} 
\usepackage{placeins}

\usepackage{graphicx}
\usepackage{array}
\usepackage{makecell}
\usepackage{capt-of}

\newcolumntype{M}[1]{>{\centering\arraybackslash}m{#1}}
\usepackage{booktabs,multirow,siunitx}

\newcommand{\imagewidth}{0.15}
\newcommand{\imagewidthsuppl}{0.15}

\newcommand{\colA}[1]{\includegraphics[width=\imagewidth \textwidth, trim=150pt 150pt 120pt 145pt, clip]{#1}} 
\newcommand{\colB}[1]{\includegraphics[width=\imagewidth \textwidth, trim=200pt 160pt 200pt 200pt, clip]{#1}} 
\newcommand{\colC}[1]{\includegraphics[width=\imagewidth \textwidth,  trim=215pt 200pt 90pt 190pt, clip]{#1}} 
\newcommand{\colD}[1]{\includegraphics[width=\imagewidth \textwidth, trim=120pt 200pt 80pt 190pt, clip]{#1}} 
\newcommand{\colE}[1]{\includegraphics[width=\imagewidth \textwidth, trim=235pt 100pt 80pt 90pt, clip]{#1}} 
\newcommand{\colZ}[1]{\includegraphics[width=\imagewidth \textwidth, trim=50pt 50pt 30pt 50pt, clip]{#1}}

\newcommand{\rowB}[1]{\includegraphics[width=\imagewidth \textwidth, trim=20pt 0pt 30pt 50pt, clip]{#1}}
\newcommand{\rowC}[1]{\includegraphics[width=\imagewidth \textwidth, trim=0pt 20pt 30pt 100pt, clip]{#1}}

\newcommand{\supplColA}[1]{\includegraphics[width=\imagewidthsuppl \textwidth, trim=150pt 150pt 90pt 150pt, clip]{#1}} %
\newcommand{\supplColB}[1]{\includegraphics[width=\imagewidthsuppl \textwidth, trim=240pt 200pt 200pt 200pt, clip]{#1}} %
\newcommand{\supplColC}[1]{\includegraphics[width=\imagewidthsuppl \textwidth,  trim=150pt 50pt 150pt 140pt, clip]{#1}} %
\newcommand{\supplColD}[1]{\includegraphics[width=\imagewidthsuppl \textwidth, trim=100pt 200pt 100pt 200pt, clip]{#1}} %
\newcommand{\supplColE}[1]{\includegraphics[width=\imagewidthsuppl \textwidth, trim=240pt 140pt 130pt 100pt, clip]{#1}} %
\newcommand{\supplColZ}[1]{\includegraphics[width=\imagewidthsuppl \textwidth, trim=200pt 200pt 180pt 200pt, clip]{#1}} %

\newcommand{\up}{\textsuperscript{\raisebox{-0.3ex}{$\uparrow$}}}
\newcommand{\down}{\textsuperscript{\raisebox{-0.3ex}{$\downarrow$}}}

\usepackage{cuted}

\definecolor{TripoSG}{named}{Gray}
\definecolor{Trellis}{named}{Lavender}
\definecolor{Hunyuan}{named}{YellowGreen}
\definecolor{Repaint}{named}{YellowOrange}
\definecolor{Morph}{named}{cyan}
\definecolor{RepaintBB}{named}{Orange}
\definecolor{MorphBB}{named}{NavyBlue}

\pgfplotsset{
  myaxis/.style={
    axis lines=left,
    width=0.9\linewidth, height=5cm,
    xlabel={Threshold},
    xtick=data,
    symbolic x coords={all,0.3,0.5,0.7,0.9},
    scaled ticks=false,
    legend style={draw=none, fill=none, font=\small},
    legend cell align=left,
    scale only axis=true,   
  }
}

\def\BibTeX{{\rm B\kern-.05em{\sc i\kern-.025em b}\kern-.08em
    T\kern-.1667em\lower.7ex\hbox{E}\kern-.125emX}}

\usepackage{biblatex} 
\begin{document}

\title{3DMorph: Single-Image-Guided Local 3D Shape Editing and Morphing}



\author{\IEEEauthorblockN{Tobias Preintner\IEEEauthorrefmark{1}\textsuperscript{,}\IEEEauthorrefmark{2},
Yunfei Deng\IEEEauthorrefmark{2},
Phillip Müller\IEEEauthorrefmark{2},
Sebastian Illing\IEEEauthorrefmark{2}, \\
Adrian König\IEEEauthorrefmark{2},
Thomas Bäck\IEEEauthorrefmark{1},
Elena Raponi\IEEEauthorrefmark{1},
Niki van Stein\IEEEauthorrefmark{1}}
\IEEEauthorblockA{\IEEEauthorrefmark{1}Institute of Advanced Computer Science, Leiden University, Leiden, The Netherlands}
\IEEEauthorblockA{\IEEEauthorrefmark{2}BMW Group, Munich, Germany}}%


\let\oldtwocolumn\twocolumn
\renewcommand\twocolumn[1][]{%
    \oldtwocolumn[{#1}{
        \begin{center}
            \centering
            \includegraphics[width= 0.975 \textwidth, trim=10pt 25pt 10pt 390pt, clip]{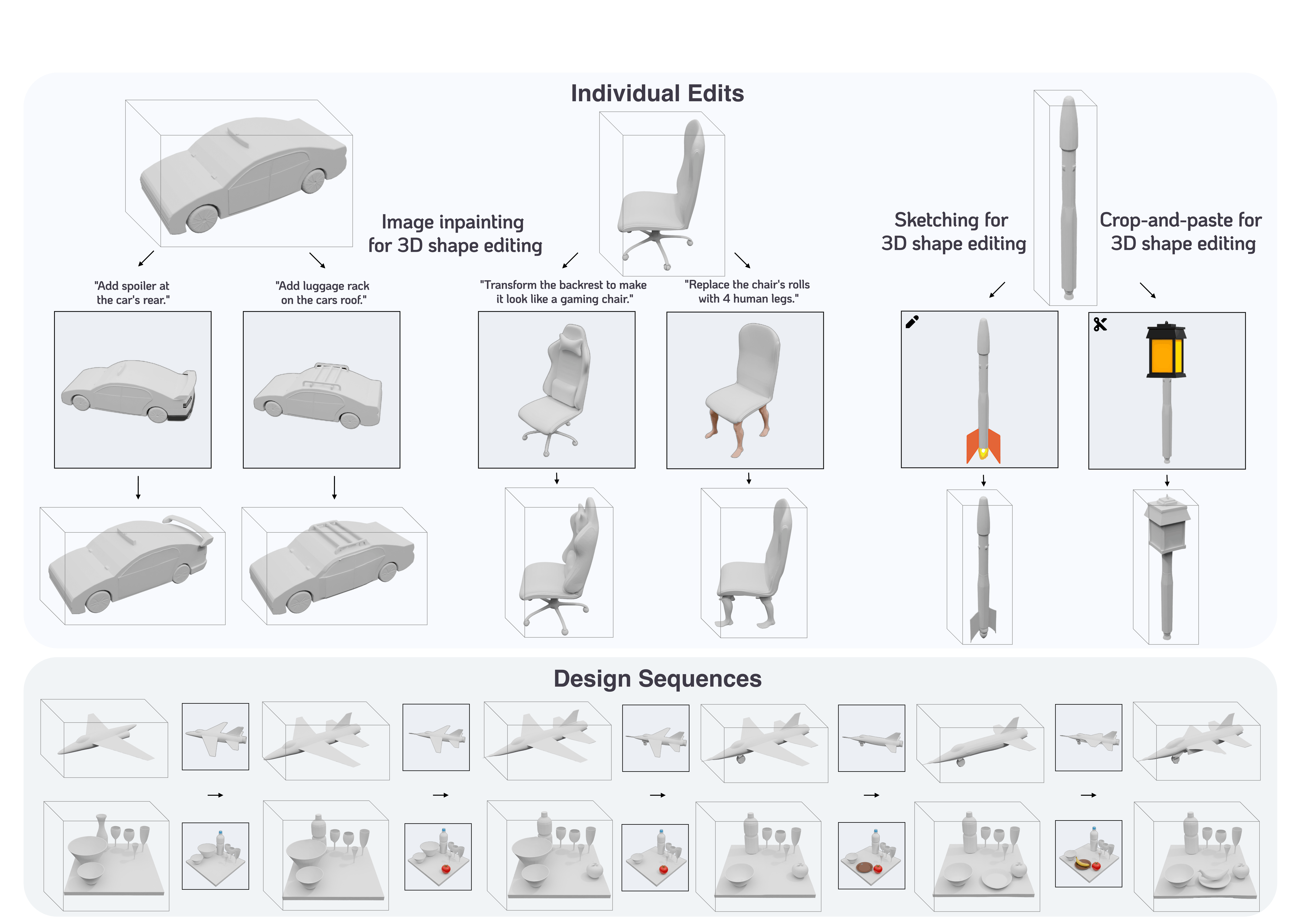}
            \captionof{figure}{High-quality 3D editing results from our method. Using rendered and edited images as input (e.g., via inpainting), 3DMorph lifts 2D modifications into 3D while preserving unedited regions. It can be used for both individual edits and multi-step editing sequences.}
            \label{fig:main-fig}
        \end{center}
    }]
}

\maketitle

\begin{abstract}
Despite recent progress in 3D generation, intuitive editing of existing shapes remains limited.
Unlike images, which benefit from well-established inpainting tools, general 3D objects such as meshes still lack simple and effective methods for local shape editing. Existing approaches are often global, domain-specific, require complex user interaction, or focus on appearance (color and texture) rather than geometry.
We introduce 3DMorph, a training-free framework for single-image-guided local 3D shape editing and morphing.
Given an edited image showing a desired shape modification, our method automatically localizes the relevant 3D region and transfers 2D modifications to 3D while preserving unmodified areas.
3DMorph also enables intermediate shape generation between the original and edited objects, facilitating design exploration.
To benchmark editing quality, we introduce Delta3D, an image-guided local 3D editing benchmark with paired ground-truth edits.
Experimental results show that 3DMorph translates intuitive 2D edits into 3D, outperforming state-of-the-art generative and editing methods.

\end{abstract}


\section{Introduction}
Meshes are the backbone of industrial and design workflows due to their precision and compatibility with established computer-aided design (CAD) and computer-aided engineering (CAE) toolchains. 
%
In practice, meshes used in engineering workflows often originate from CAD models, but even editing only local parts can require re-running the entire mesh-generation pipeline, including modifying the CAD model, re-meshing, and performing manual cleanup. 

Moreover, well-parameterized CAD models are often unavailable: scanned or publicly shared assets typically provide only a static mesh (vertices and faces) without any editable parametric structure.
This limitation has become more salient with recent advances in 3D generative models~\cite{li2025triposg,xiang2025trellis,hunyuan3d2025hunyuan3d}, which can rapidly produce high-quality meshes suitable as starting points for downstream design. However, directly editing meshes remains challenging. Unlike images, with mature inpainting and editing tools, there is a lack of intuitive methods for local mesh modification.

We introduce 3DMorph, a training-free framework for local 3D shape editing that requires only a single rendered and edited view to drive the modification. 
Using standard image-editing operations (e.g., inpainting, sketching, or crop-and-paste) to communicate desired local edits, 3DMorph transfers the specified changes back to the mesh while preserving unedited regions.

Beyond the target edit, 3DMorph can generate intermediate meshes along a continuous morph trajectory. This supports design-space exploration, which is common in engineering and design, where multiple alternatives must be compared (e.g., aerodynamic evaluation of spoiler variants). The candidates differ only within the locally edited region, enabling controlled comparison, optimization, and visualization.

Unlike artistic 3D generation, where color and texture dominate, engineering workflows focus on geometry. We therefore introduce Delta3D, a single-image-guided 3D editing benchmark tailored to geometry-centric evaluation. Each sample includes a source mesh, an edited image that defines the local change, and a 3D ground-truth mesh containing the edit.

Our main contributions are:
\begin{itemize}
\item A training-free method for intuitive local 3D editing of arbitrary shapes from a single edited view.
\item A benchmark dataset with ground-truth local modifications for quantitative 3D editing evaluation.
\item A bounding-box estimation module that automatically localizes the 3D edit region from a single edited image.
\item A local morphing mechanism that generates intermediate shapes for design exploration and visualization.
\end{itemize}

Together, these components form a modular framework for single-image-guided 3D editing and morphing. 
Our code and dataset are available at \mbox{\textit{https://github.com/toprei/3dmorph}}.








\section{Related Work}
\textbf{3D Shape Generation.}
State-of-the-art 3D generation methods produce high-quality geometric assets by adapting diffusion-based approaches~\cite{long2023wonder3d, yi2023gaussiandreamer, hunyuan3d2025hunyuan3d} and rectified flow transformers~\cite{li2025triposg, xiang2025trellis} to the 3D domain.
Although these methods generate high-quality shapes, their applicability in engineering and design is limited, as even minor local edits require a full regeneration, often causing unwanted changes.







\textbf{3D Shape Editing.} 
To avoid full regeneration and increase editing control, a growing line of work has focused on shape editing.
Early approaches learn 3D–language grounding for text-guided editing from object pairs or sequences accompanied by textual difference descriptions~\cite{huang2022ladis, achlioptas2023shapetalk, slim2024shapewalk}, but remain sensitive to subtle grammatical and phrasing variations~\cite{preintner2025why}.

Other text-guided methods leverage text–image losses to edit 3D objects~\cite{sella2023voxe, gao2023textdeformer, chen2024shap, li2024focaldreamer, chen2025partgen}. 
Recently, training-free text-guided methods have emerged that leverage multimodal language models~\cite{xia2025towardsscalable, ye2025nano3d, cai2025native, zhou2025anchorflow}.
However, textual spatial descriptions are often ambiguous and verbose, hindering precise 3D editing in engineering and design.

To mitigate this issue, sketch-based approaches~\cite{binninger2024sens, bandyopadhyay2024doodle, li2025meshpad} capture spatial details but demand significant drawing proficiency and spatial reasoning skills.
%
A more intuitive editing approach mixes parts of existing shapes~\cite{hertz2022spaghetti, koo2023salad}, but is limited by available source objects.

Compared to text, images represent intended 3D modifications more precisely and compactly. 
Some methods~\cite{3DFaceSculptor, li2025voxhammer} use edited images to guide 3D edits, but require additional complex user inputs such as segmentation maps or 3D masks.
%
%
Only a few methods support single-image 3D editing. \cite{gao20243dmeshediting}~formulates shape editing as a conditional reconstruction problem but is limited to edits within the trained canonical view. 
CMD~\cite{li2025cmd} enables 3D editing and progressive generation.
%

Beyond these limitations, most existing methods focus on visual appearance (color and texture) rather than geometric quality, largely due to the lack of ground-truth 3D edits~\cite{li2025cmd}.
Since color and texture can mask geometric artifacts, geometric precision is often under-emphasized in these works.

To address this, we propose a geometry-focused, training-free, single-image-guided 3D editing method and a corresponding benchmark, demonstrating both the benchmark’s utility and the effectiveness of our approach.

\textbf{Shape Morphing.} 
Shape morphing smoothly transforms one object into another by generating a sequence of intermediate shapes. Accordingly, shape editing can be regarded as its single-step variant. 

%
Early geometric methods model deformation using continuum mechanics~\cite{wirth2011continuum}, geometric flows~\cite{brandt2016geometric}, splines in shell space~\cite{heeren2016splines}, or Hamiltonian dynamics~\cite{eisenberger2020hamiltonian}. More recent work, such as 4Derform~\cite{sang20254deform}, learns continuous velocity fields in Euclidean space, enabling morphing directly on point clouds. In contrast, most recent approaches leverage latent spaces: TextDeformer~\cite{gao2023textdeformer} enables text-driven mesh edits with a 3D diffusion prior, NeuroMorph~\cite{eisenberger2021neuromorph} applies graph convolutions to interpolate local features, and SRIF~\cite{sun2024srif} synthesizes intermediate point clouds and images.

Existing methods perform global morphing that alters the entire mesh. In contrast, our approach deforms only the modified region while preserving the rest, supporting sampling of intermediate variations for efficient comparison, optimization, and design-space exploration.



\begin{figure*}[!t]
\centering
    \includegraphics[width= 0.95 \textwidth, trim=80pt 5pt 80pt 75pt, clip]{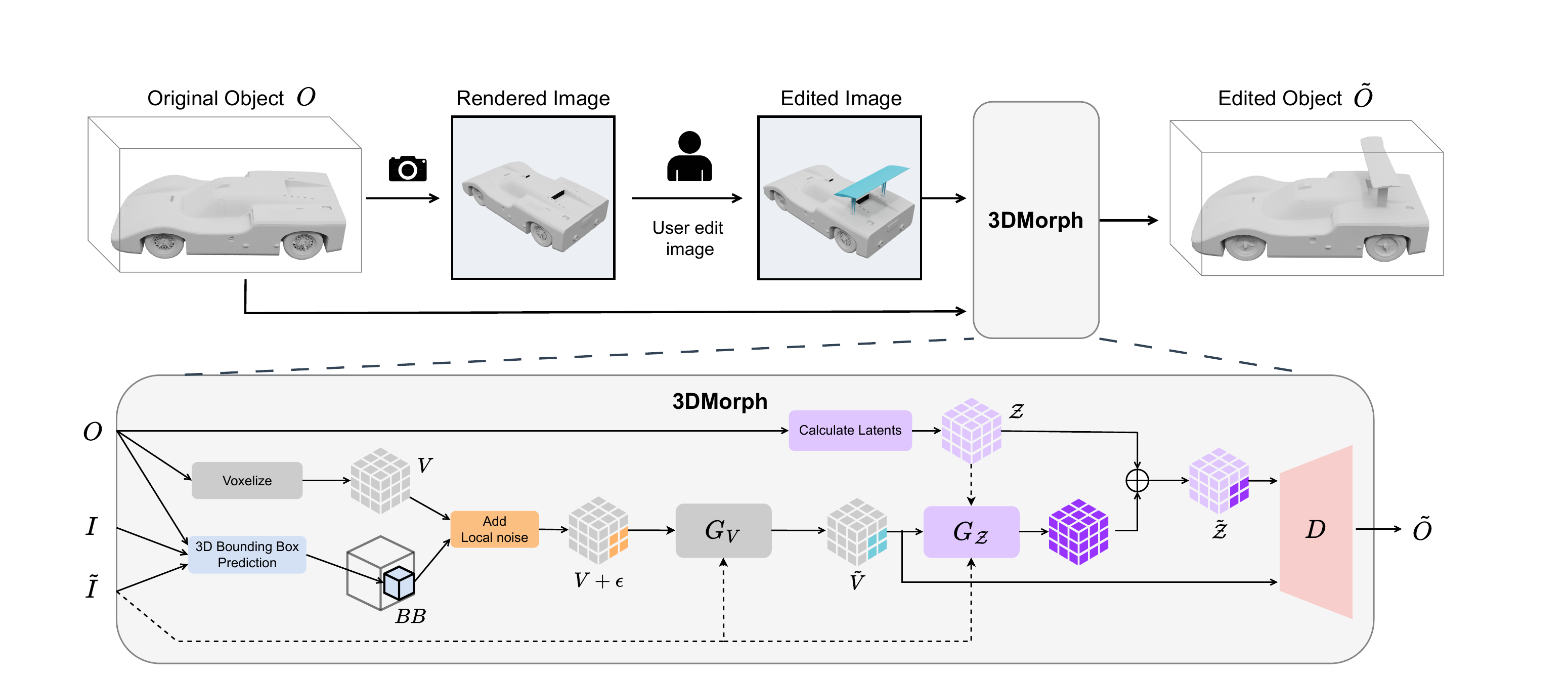}%
    \caption{Overview of 3DMorph. Our method enables intuitive, training-free 3D shape editing guided by 2D image modifications. After a user edits a chosen rendered view, the 2D changes are transferred back to 3D while retaining unedited regions, enabling simple local shape editing and high-quality mesh outputs. \label{fig:method}}
\end{figure*}


\section{Methodology}
We propose 3DMorph, a training-free method for image-guided local 3D shape editing and morphing. From a single edited image, it transfers local 2D edits to the 3D domain while preserving unaltered regions. 
Moreover, it supports smooth morphing from the original to the edited object through intermediate shape generation.

\subsection{Problem Definition}
We formally define a generative model $f(O, \tilde{I})$, that takes as input a 3D object $O$ and an edited image $\tilde{I}$. The image $\tilde{I}$ is obtained by applying local edits to an image $I$, where $I$ is a rendered view of $O$.
A local edit is defined as a modification confined to a specific region in both 2D and 3D, typically involving the addition or adaptation of specific parts.
The goal is to find a model $f$ that transfers the edits from $\tilde{I}$ to a modified 3D object $\tilde{O}$, while preserving the geometry of $O$ in the unaltered regions. 
$\tilde{I}$ serves solely as a medium to specify shape edits, decoupling the formulation from low-level visual attributes such as color.
In this way, intuitive 2D editing techniques, such as inpainting, hand-sketching, or crop-and-paste manipulations, can be leveraged to modify 3D objects.
Moreover, the approach should support local 3D morphing from $O$ to $\tilde{O}$ via intermediate 3D objects.


\subsection{Preliminary} 
\textbf{Image editing.} Image editing tools have evolved for decades, with traditional software manipulating pixels directly or applying user-controlled transformations. However, these tools demand considerable user expertise.

Deep learning–based methods enable local image modifications while preserving surrounding regions. 
Techniques, such as RePaint~\cite{lugmayr2022repaint}, Palette~\cite{saharia2022palette}, Paint-by-Example~\cite{yang2023paint}, Stable Diffusion Inpainting~\cite{Rombach_2022_CVPR}, ControlNet~\cite{zhang2023controlnet}, GeoDrag~\cite{mueller2025geodiffusion}, FLUX.1-Fill~\cite{flux2024} and  Gemini Flash~\cite{comanici2025gemini} make local image editing fast, intuitive, and controllable.






\textbf{Structured latents.} 3D objects can generally be decomposed into geometry and semantics. This decomposition can be represented by a coarse voxel structure $V$ and per-voxel latent features $\mathcal{Z}$, which encode both texture and geometric surface details. Together, they form the structure latent (SLAT) representation, employed by recent generative models~\cite{xiang2025trellis, wu2025amodal3r, xiang2025trellis2} to synthesize new 3D assets from images or text.

Trellis~\cite{xiang2025trellis} utilizes two specialized rectified flow transformers to generate SLAT: one, $G_V$, for the sparse voxel structure and another, $G_{\mathcal{Z}}$, for the latent features. The resulting SLAT is then decoded by a decoder $D$ into various 3D formats, including meshes.
Although Trellis is primarily designed for generation, we incorporate its generative transformer and decoder architecture into our framework and leverage its SLAT representation for editing.
Notably, SLAT can be derived not only through $G_V$ and $G_{\mathcal{Z}}$, but also deterministically from existing 3D objects by voxelizing them and computing per-voxel latents from projections of multi-view renderings.

\subsection{3D Shape Editing}\label{subsec:3DEdit} Our method enables intuitive, training-free local 3D editing from a single edited rendering. We first detect the modified local region between the original and edited images and project it into 3D to obtain an axis-aligned bounding box. This step is non-trivial and critical in our pipeline. Instead of requiring the user to manually estimate a 3D box, we predict it automatically (see Sec.~\ref{subsec:BB-pred}).
Within the bounding box, a coarse voxel grid and latent features are generated by $G_V$ and $G_{\mathcal{Z}}$, respectively, using the edited view.
Outside the bounding box, we retain the original voxel grid and latents, which are deterministically computed from the original object.

Reusing the original voxel structure preserves global geometry, while incorporating the high-quality original latents recovers fine details, including those not visible in the edited view. 
The resulting hybrid structure, combining original and modified features, is decoded into a high-quality 3D mesh that seamlessly integrates the edit while preserving unmodified regions, subject to the limits of the current SLAT resolution and decoder capacity.
This supports long edit sequences without degrading quality in unaffected areas.

A key innovation of this approach is the use of a hard constraint on the voxel structure, in contrast to commonly used flow-consistent conditioning. We describe this design choice in detail in the following subsection.

\textbf{Conditioning Schedule for Local Inpainting.}
The objective of local 3D editing is to modify geometry only within the predicted edit region while preserving the original structure and details elsewhere. During reverse sampling, that is, while synthesizing the edited region, this requires enforcing consistency on the unmodified SLAT to prevent drift and ensure fidelity to the original object.

Classical RePaint conditioning re-injects the known region at every step using its flow-consistent value $\mathcal{F}_{t}(x_0, z)$, preventing covariate shift and matching the model’s training dynamics. However, in our 3D setting this can introduce stochastic variation in the preserved geometry early in the trajectory, which is undesirable when high-fidelity context preservation is required. 

To address this, we adopt a two-stage schedule:
\begin{itemize}
    \item \textbf{Stage-1 (voxel prediction / sparse structure):} enforce a hard constraint on known voxels to stabilize global shape.
    \item \textbf{Stage-2 (latent refinement / per-voxel detail):} use flow-consistent conditioning to smoothly merge the new geometry and details with the preserved shape.
\end{itemize}

\begin{equation}
\Pi_{\text{hard}}(\hat{x}_{t-\Delta})
= (1 - m) \odot x_0 \;+\; m \odot \hat{x}_{t-\Delta}
\label{eq:hard}
\end{equation}

\begin{equation}
\Pi_{\text{flow}}(\hat{x}_{t-\Delta})
= (1 - m) \odot \underbrace{\mathcal F_{t-\Delta}(x_0, z)}_{\substack{\text{path-consistent}\\\text{known region}}}
\;+\; m \odot \hat{x}_{t-\Delta}
\label{eq:flow}
\end{equation}

Here, $m$ denotes the edit mask (in our case, the bounding-box $BB$), $x_0$ is the unedited context (SLAT of $O$), and $z\!\sim\!\pi$ is noise sampled from the prior. $F_t(x_0, z)$ denotes the training-time rectified-flow path that transports the clean context $x_0$ toward a noise sample $z$, evaluated at time $t$, with $F_0(x_0, z)=x_0$ and $F_1(x_0, z)=z$.


In Stage-1, the context voxels are kept fixed, ensuring that preserved regions remain unchanged while the model infers a coarse modified voxel structure. 
In Stage-2 the focus shifts to making the edited region match the original shape and blend in smoothly. We set the known region to its state along the flow-matching path at time $t$. This matches the training setup, yields seamless transitions between edited and preserved regions.

\textbf{Algorithm.} Figure~\ref{fig:method} illustrates our approach, which is detailed in Algorithm~\ref{alg:method}.
Given the original object $O$, its rendered image $I$, and an edited version $\tilde{I}$, our method first applies the 3D bounding box prediction module described in Sec.~\ref{subsec:BB-pred} (line~1).
After voxelizing $O$ (line~2), we add noise within the bounding box region to enable new voxel generation during denoising (line~3). 
This is possible as the stage-1 latent space is also structured, which allows us to inject noise in a way that aligns with the original voxel space.
In combination with $\tilde{I}$, the partially noised voxel grid $\dot{V}$ is input to $G_V$ to generate the edited voxel grid $\tilde{V}$ (line~4).
Following the approach of~\cite{xiang2025trellis}, the object latents $\mathcal{Z}$ are deterministically computed from multiple rendered views for all $L$ voxels, where $L$ denotes the number of surface voxels in the grid (line~5).
Together with $\tilde{V}$ and $\tilde{I}$, these provide context to $G_{\mathcal{Z}}$, which generates new latents $\dot{\mathcal{Z}}$ for the edited voxel grid $\tilde{V}$, including the modified region (line~6).
To preserve quality in the unmodified regions, the new latents are applied only within the edit region defined by $BB$, while the original latents are retained outside $BB$ (line~7).
Finally, $\tilde{V}$ and $\tilde{\mathcal{Z}}$ are decoded to produce the edited object~$\tilde{O}$ (line~8).

\SetKwComment{Comment}{\#}{}
\SetKwInput{KwData}{Given}
\RestyleAlgo{ruled} 
\LinesNumbered

\begin{algorithm}
\caption{Local 3D Editing with 3DMorph}
\label{alg:method}
\SetAlgoLined
\SetKwInOut{Input}{Input}\SetKwInOut{Output}{Output}
\Input{Original object $O$, rendered image $I$, modified image $\tilde{I}$}
\Output{Edited object $\tilde{O}$}
\BlankLine

\tcp{Generate Local Voxels}
$BB\gets \mathrm{getBB}(O, I, \tilde{I})$ \tcp*[r]{$BB \in \{0,1\}^{64^3}$}
$V \gets \mathrm{voxelize}(O)$ \tcp*[r]{$V \in \{0,1\}^{64^3}$}
$\dot{V} \gets V + \epsilon$\ \tcp*[r]{$\epsilon \sim \mathcal{N}(0, I) \odot BB$}
$\tilde{V} \gets G_V(\dot{V},\tilde{I})$\;
\BlankLine

\tcp{Generate Local Latents}
$\mathcal{Z} \gets \mathrm{getLatents}(O, V)$ \tcp*[r]{$\mathcal{Z} \in \mathbb{R}^{L\times d}$}
$\dot{\mathcal{Z}} \gets {G_{\mathcal{Z}}}(\tilde{V},\tilde{I}, \mathcal{Z})$\;
$\tilde{\mathcal{Z}}\gets BB \cdot \dot{\mathcal{Z}} + (1-BB) \cdot \mathcal{Z}$\;

\BlankLine
$\tilde{O}\gets D(\tilde{V}, \tilde{\mathcal{Z}})$\ \tcp*[r]{decode}

\end{algorithm}


\subsection{Bounding Box Prediction}\label{subsec:BB-pred}
%
%
Our method detects structural changes by first identifying 2D differences between image pairs and lifting them into 3D. 
SSIM-based differencing yields 2D change boxes, which we anchor in 3D by projecting their centers into the voxel grid. This procedure assumes accurate camera poses and sufficient visibility of the edited region.

Box extents are estimated by projecting the 2D box into 3D using the camera parameters, the 2D box width and height, and a pixel-to-metric scale factor derived from the top view, which maps pixel distances to 3D object-space units.
Instead of merging all boxes, we compute a tight axis-aligned bounding box that encloses them, yielding a single 3D box in voxel space that localizes the edit region in voxel space. 




\subsection{Local 3D Morphing}\label{Morphing} Our morphing method leverages the SLAT difference between the two objects to generate intermediate shapes between the original object $O$ and the edited object $\tilde{O}$. Voxels inside the modification bounding box are sequentially activated in batches, ordered by their Euclidean distance to $O$, though arbitrary distance functions and activation policies can also be applied. After each batch activation, the corresponding SLAT is decoded into a mesh. This procedure can be interpreted as a form of geometric and latent-space morphing. 
Repeating this process yields a series of intermediate objects between $O$ and $\tilde{O}$, enabling visualization, design-space exploration, and creative discovery. 


\begin{table*}[!t]
\centering
\caption{Quantitative comparison. \textbf{Bold} marks the best results, \underline{underlined} denotes a statistically significant gap (paired t-test, 5\% significance level). $\dagger$ indicates hypothetical availability of multi-view images of the modified object.}
\label{table:quantitative}
    \begin{tabular}{lccccccc}
    \toprule
      Method  & CD\down & HD\down & IoU\up (\%) & DSC\up (\%) & MS-SSIM\up & FSIM\up & GMSD\down \\
    \midrule
TripoSG \cite{li2025triposg}& 0.0454 & 0.126 & 66.2  & 77.5 & 0.721 & 0.675 & 0.218 \\
Trellis \cite{xiang2025trellis} & 0.0458 & 0.123 & 74.0  & 84.1 & 0.771 & 0.690 & 0.206 \\
Trellis-MV$^\dagger$ \cite{xiang2025trellis} & 0.0342 & 0.096 & 77.9  & 86.9 & 0.781 & 0.696 & 0.202 \\
Hunyuan3D \cite{hunyuan3d2025hunyuan3d} & 0.0263 & 0.074 & 82.2  & 89.6 & 0.843 & 0.752 & 0.174 \\
Trellis-RePaint \cite{xiang2025trellis, lugmayr2022repaint} & 0.0129 & 0.036 & 85.2  & 91.0 & 0.861 & 0.795 & 0.162 \\
3DMorph & \textbf{\underline{0.0105}} & \textbf{\underline{0.029}} & \textbf{\underline{88.2}} & \textbf{\underline{93.3}} & \textbf{\underline{0.877}} & \textbf{\underline{0.806}} & \textbf{\underline{0.156}} \\
    \bottomrule
    \end{tabular}
\end{table*}


\section{Experiments}
In this section, we quantitatively and qualitatively evaluate our single-image-guided 3D editing framework for local shape edits using our Delta3D benchmark. We further analyze the impact of the bounding-box prediction module through an ablation study and present results on inpainted image edits as 3D editing guidance, as well as on local shape morphing. 

In summary, 3DMorph consistently outperforms alternative methods both quantitatively and qualitatively, leverages object context more effectively than the baseline editing approach, remains robust with both predicted and ground-truth bounding boxes, and can handle individual edits and multi-step editing sequences.


\subsection{Delta3D benchmark} 
No existing benchmark provides paired 3D ground truth for precise local edits; we therefore introduce Delta3D to fill this gap.
Our Delta3D benchmark is built on the Fusion 360 Gallery Assembly dataset~\cite{willis2022assembly}, a diverse collection of user-submitted CAD models composed of multi-part 3D objects.
Object pairs $(O, \bar{O})$ are generated by removing a single part from an assembly. Pairs are chosen such that both the original object $O$ and the ground-truth edit $\bar{O}$ are meaningful objects on their own, resulting in 74 carefully chosen object pairs. 

For each object pair, we rotate 360° around the ground-truth object in 30° increments, simulating 12 edited images~$\tilde{I}$.
To ensure that the local modification is visible, we compare each image pair $(I, \tilde{I})$ using the SSIM to identify 2D difference regions, from which bounding boxes are computed as described in Section~\ref{subsec:BB-pred}.
Views in which the modified part is fully occluded are excluded, resulting in a final set of 864 evaluation samples. 
This removes unsolvable cases while retaining challenging conditions such as oblique angles, heavy occlusions, and complex depth discontinuities.

\subsection{Experimental Setup} 
We evaluate 3DMorph on our Delta3D benchmark dataset and compare it against state-of-the-art image-to-3D generation methods, including Trellis~\cite{xiang2025trellis}, Hunyuan3D~\cite{hunyuan3d2025hunyuan3d}, and TripoSG~\cite{li2025triposg}.
In addition to generative baselines, we include a 3D editing method built on Trellis and RePaint~\cite{lugmayr2022repaint}, hereafter referred to as Trellis-RePaint, which uses purely flow-consistent conditioning (see Sec.~\ref{subsec:3DEdit}). This approach was briefly mentioned in~\cite{xiang2025trellis} but not evaluated in detail. Since it requires an edit region as input, we pair it with our bounding box prediction module to enable a fair comparison with 3DMorph.

We also conduct an ablation in which both 3DMorph and Trellis-RePaint are given a perfect bounding box that exactly encloses the region of difference between $O$ and $\bar{O}$. This setup (i) enables a direct comparison under identical idealized conditions and (ii) quantifies the effect of bounding box prediction on performance.
All methods are evaluated on the full benchmark dataset.

\subsection{Metrics}\label{subsec:metrics} We evaluate the geometric quality of the edited 3D object by comparing it with the ground truth.
Chamfer Distance (CD) and Hausdorff distance (HD) quantify differences between the generated object $\tilde{O}$ and ground truth $\bar{O}$. 
Here, $d(\bar{o}, \tilde{o})$, denotes the Euclidean distance between surface points $\bar{o} \in \bar{O}$ and $\tilde{o} \in \tilde{O}$. For each object, we sample 20,000 points.
\begin{equation} \label{eq:pcd}
\begin{split}
CD(\bar{O},\tilde{O}) = \frac{1}{2|\bar{O}|}\sum_{\bar{o} \in \bar{O}}\min_{\tilde{o} \in \tilde{O}} d(\bar{o},\tilde{o})   +  \frac{1}{2|\tilde{O}|} \sum_{{\tilde{o}}\in {\tilde{O}}}\min_{{\bar{o}} \in \bar{O}} d(\bar{o}, \tilde{o})
\end{split}
\end{equation}
\begin{equation} \label{eq:hdd}
\begin{split}
HD(\bar{O},\tilde{O}) = \max \{ \sup_{\bar{o}\in \bar{O}}\inf_{\tilde{o} \in \tilde{O}} d(\bar{o},\tilde{o}),\: \sup_{\tilde{o} \in \tilde{O}}\inf_{\bar{o} \in \bar{O}} d(\bar{o}, \tilde{o}) \}
\end{split}
\end{equation}
While CD considers only surfaces, intersection-based metrics capture volume overlap. Thus, we also report Intersection over Union (IoU) and Dice coefficient (DSC).

\noindent
\begin{minipage}{0.45\linewidth}
\begin{equation}
IoU(\bar{O},\tilde{O}) = \frac{|\bar{O}\cap\tilde{O}|}{|\bar{O}\cup\tilde{O}|}
\label{eq:iou}
\end{equation}
\end{minipage}\hfill
\begin{minipage}{0.45\linewidth}
\begin{equation}
DSC(\bar{O},\tilde{O}) = \frac{2|\bar{O}\cap\tilde{O}|}{|\bar{O}|+|\tilde{O}|}
\label{eq:dsc}
\end{equation}
\end{minipage}

%
%
Unless stated otherwise, metrics are computed on the entire object, covering both preserved and modified regions.

We complement the 3D evaluation with image similarity metrics suited for spatial analysis, including Multi-Scale SSIM (MS-SSIM), Feature Similarity Index (FSIM), and Gradient Magnitude Similarity Deviation (GMSD).
For all image similarity evaluations, both the generated and ground-truth objects are rendered from the same viewing angle as $\tilde{I}$ for comparison.
These traditional metrics emphasize structural properties, such as edges and contours, rather than appearance features like brightness or texture. Deep feature–based metrics such as FID or LPIPS primarily capture appearance differences, which are not the focus of our evaluation.


\subsection{Results}
\textbf{Quantitative evaluation.} Tab.~\ref{table:quantitative} shows that our method consistently and statistically significantly outperforms prior methods across all metrics. The closest competitor is Trellis-RePaint, which, despite using the same predicted bounding boxes, performs notably worse than 3DMorph. Compared to Trellis-RePaint, 3DMorph reduces CD by 19\% and increases IoU by 3 points. Interestingly, Trellis-MV, which has access to twelve hypothetically available multi-view images of the modified object, still fails to match 3DMorph’s performance. 

\begin{table}[!t]
\caption{Bounding-box ablation using ground-truth bounding boxes.}
\centering
\footnotesize
\setlength{\tabcolsep}{2.8pt} 
\renewcommand{\arraystretch}{0.95} 
\label{table:ablation}
    \begin{tabular}{lcccccccc}
    \toprule
Method & \shortstack{GT-\\ BB} & \shortstack{CD \\ \down} & \shortstack{HD \\ \down} & \shortstack{IoU \\ \up (\%)} & \shortstack{DSC \\ \up (\%)}  & \shortstack{MS-\\SSIM\up} & \shortstack{FSIM \\ \up}  & \shortstack{GMSD\\ \down} \\
    \midrule
Trellis- & -- & 0.0129 & 0.036 & 85.2 & 91.0 & 0.861 & 0.795  & 0.162 \\
RePaint & \checkmark & 0.0102 & 0.028 & 88.6  & 93.4 & 0.899 & 0.827  & 0.148 \\
\midrule
\multirow{2}{*}{3DMorph} & -- & 0.0105 & 0.029  & 88.2  & 93.3 & 0.877 & 0.806  & 0.156 \\
& \checkmark  & \textbf{0.0089} & \textbf{0.024} & \textbf{91.0} & \textbf{95.0} & \textbf{0.909} & \textbf{0.836}  & \textbf{0.144} \\
    \bottomrule
    \end{tabular}
\end{table}


\begin{table}[!t]
\centering

\caption{Edit-region evaluation within the target region.}
\footnotesize
\setlength{\tabcolsep}{2.8pt} 
\renewcommand{\arraystretch}{0.95} 
\label{table:inside-BB}
    \begin{tabular}{lccccc}
    \toprule
    Method  & GT-BB & CD\down & HD\down & IoU\up (\%) & DSC\up (\%) \\
    \midrule
Trellis-RePaint & -- & 0.0333 & 0.107 & 56.7 & 70.5  \\
3DMorph         & -- & \textbf{0.0305} & \textbf{0.097} & \textbf{59.6 } & \textbf{72.9} \\
\midrule
Trellis-RePaint & \checkmark & 0.0268 & 0.073 & 61.7  & 75.0 \\

3DMorph         & \checkmark  & \textbf{0.0230} & \textbf{0.063} & \textbf{65.0} & \textbf{77.8} \\
    \bottomrule
    \end{tabular}
\end{table}

\newcommand{\imagecolumnwidth}{2.4 cm}

\begin{figure}[t]
\centering
\begin{adjustbox}{width=\columnwidth}
\begin{tabular}{|>{\small\centering\arraybackslash}m{1.6cm}|M{\imagecolumnwidth}|M{\imagecolumnwidth}|M{\imagecolumnwidth}|M{\imagecolumnwidth}|M{\imagecolumnwidth}|}
\hline
\centering \textbf{Method} & \textbf{Forklift} & \textbf{Loader} & \textbf{Race Car} & \textbf{Hair Dryer} & \textbf{Airplane} \\
\hline

\centering \shortstack{Original \\ object $O$}
& \colA{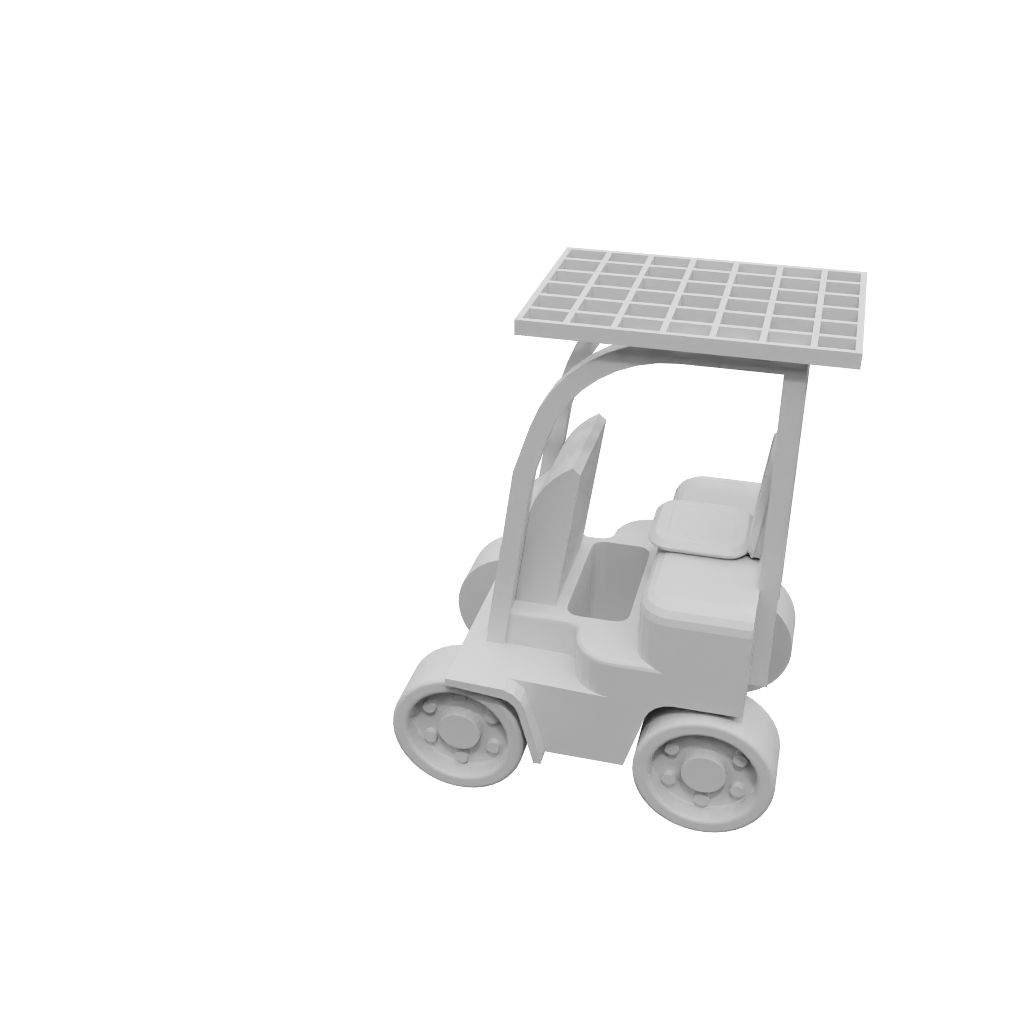}
& \colB{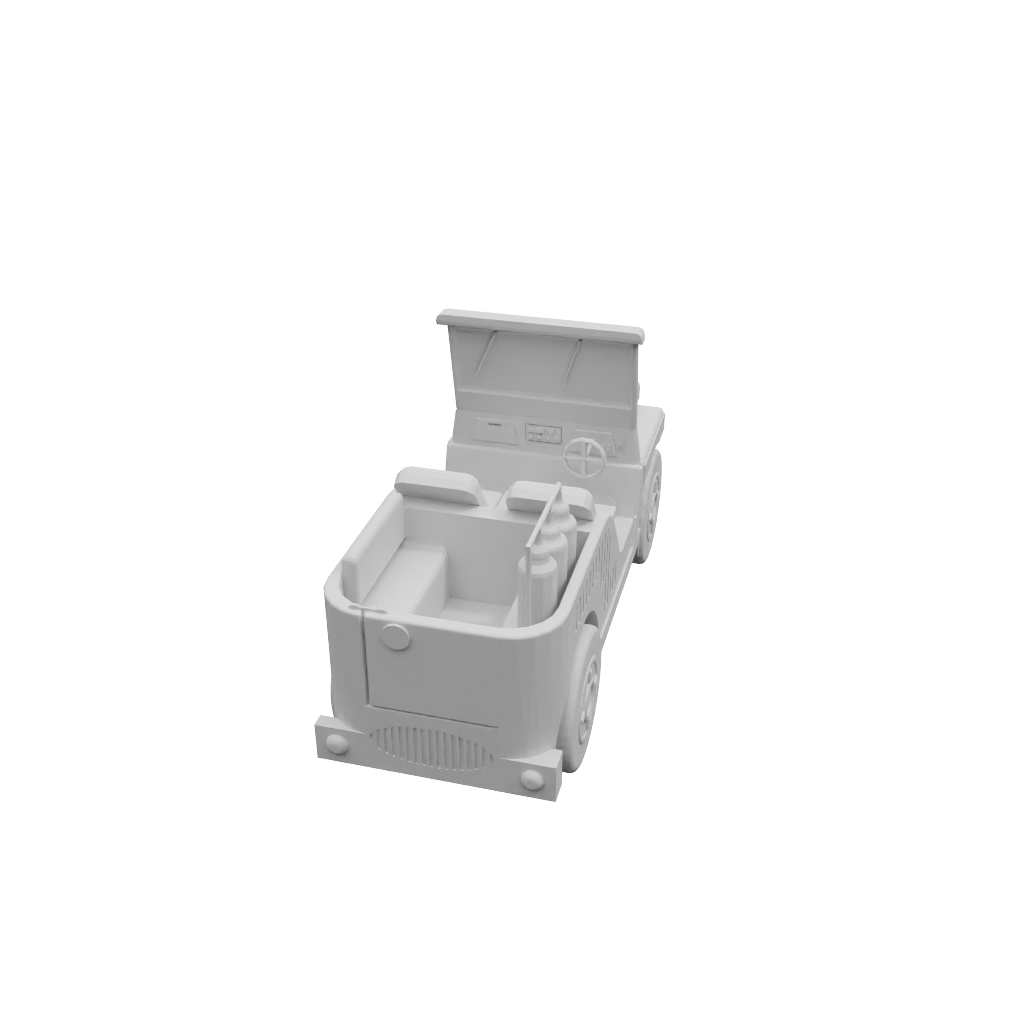}
& \colC{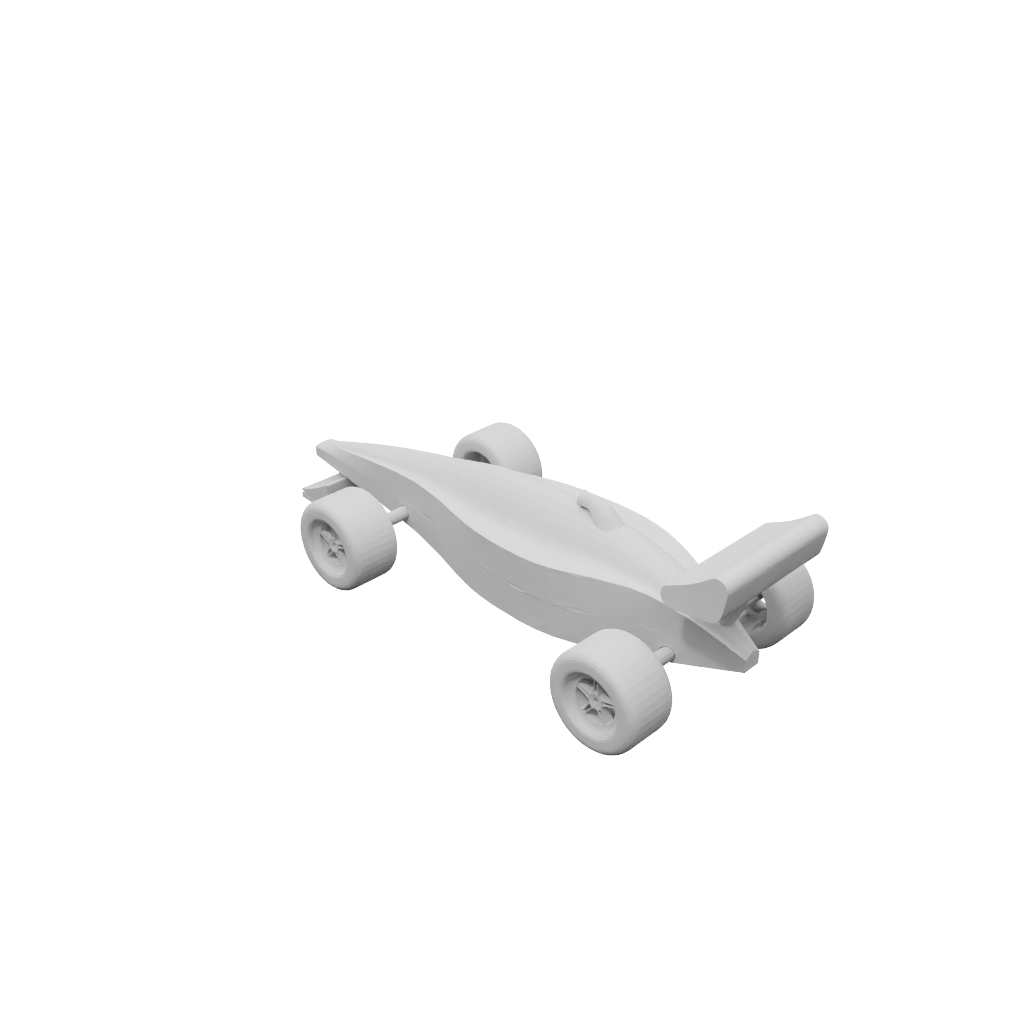}
& \colD{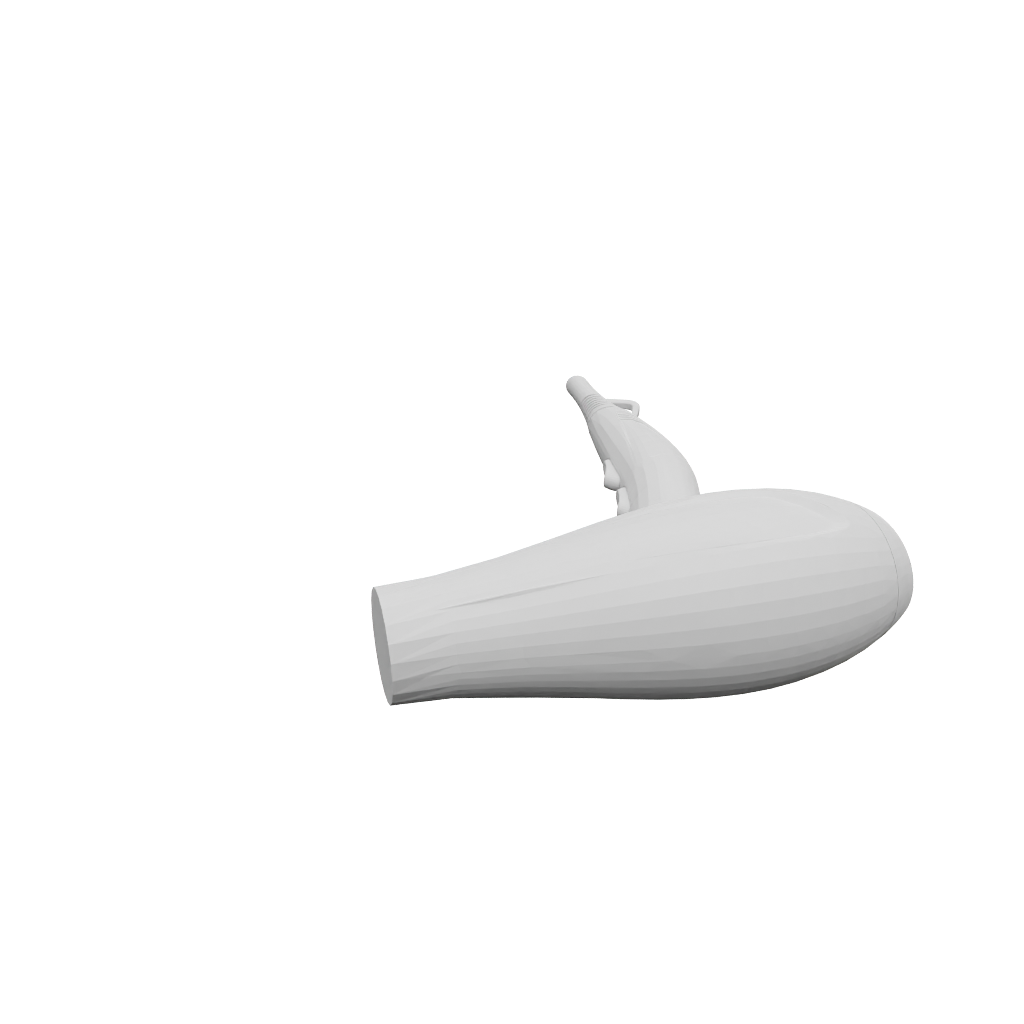}
& \colE{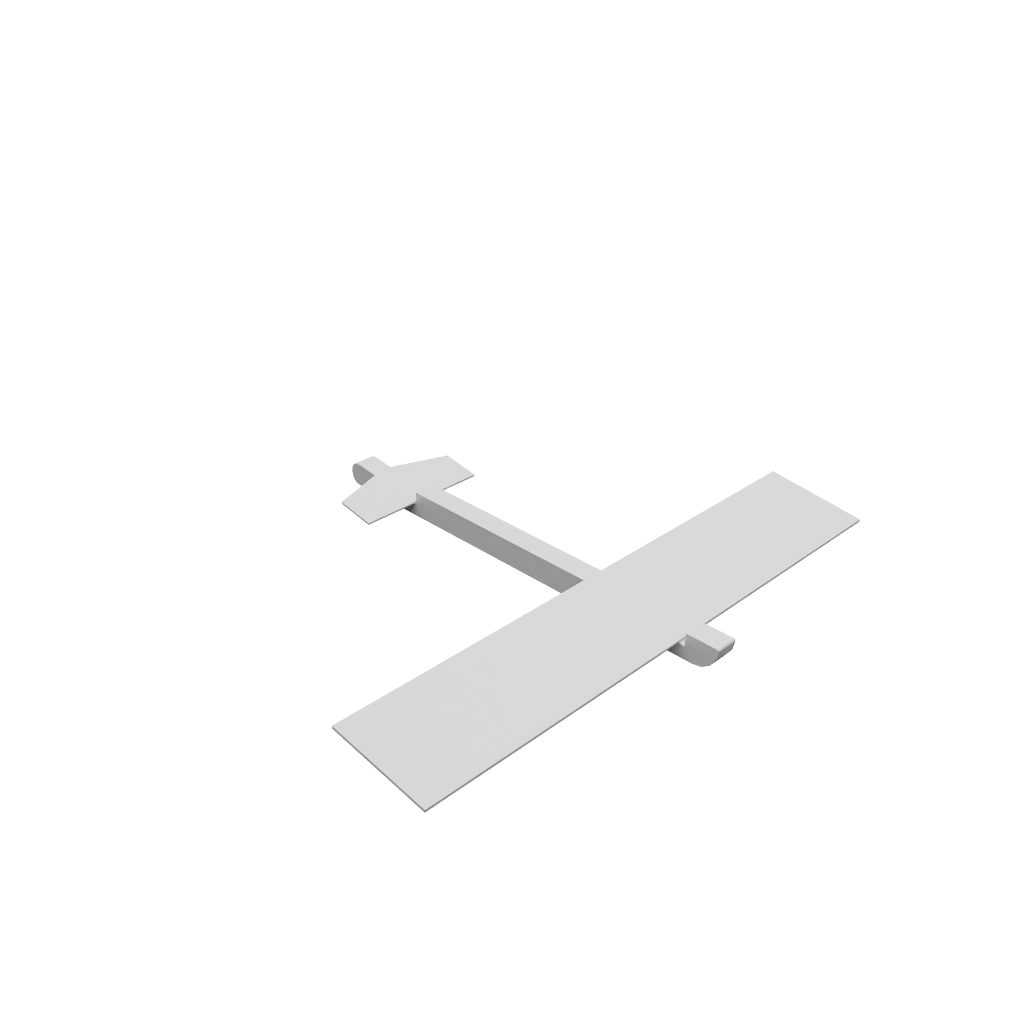} \\
\centering \shortstack{GT edited \\ object $\bar{O}$}
& \colA{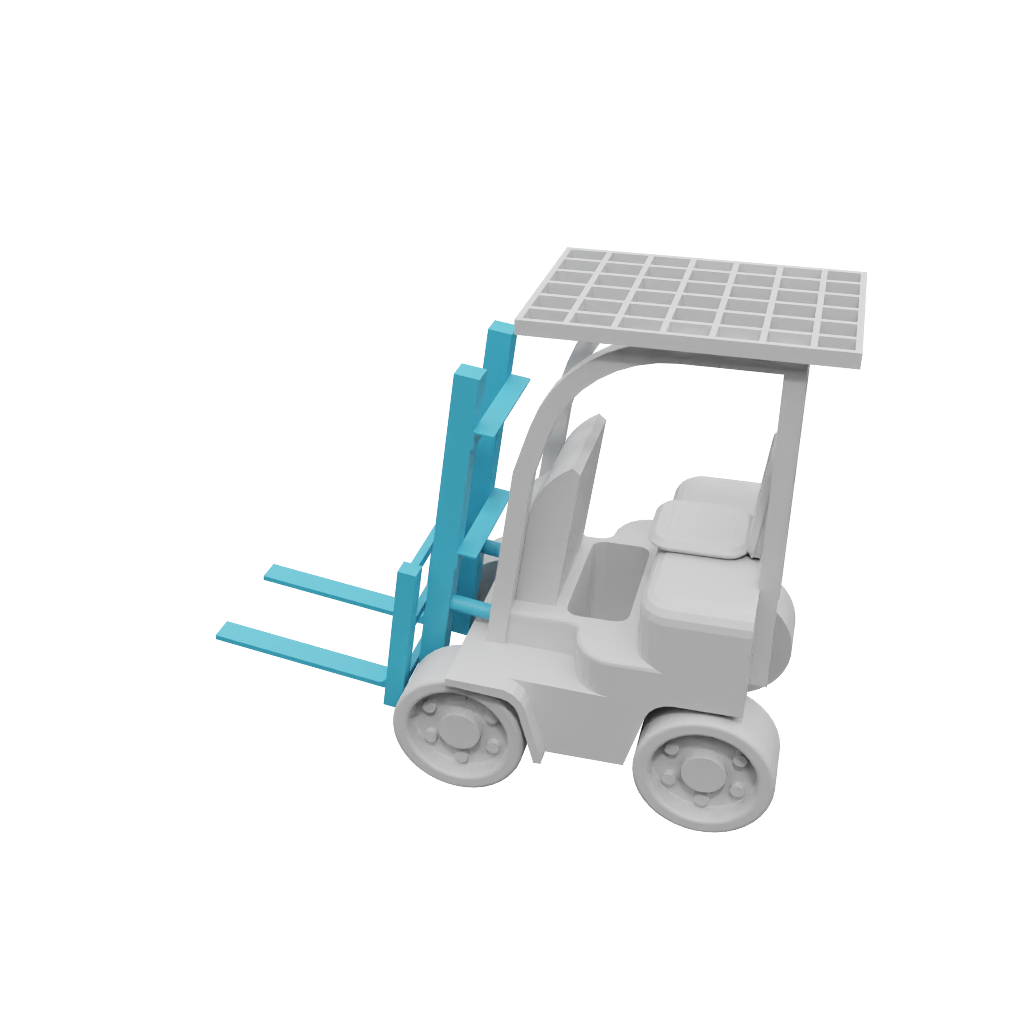}
& \colB{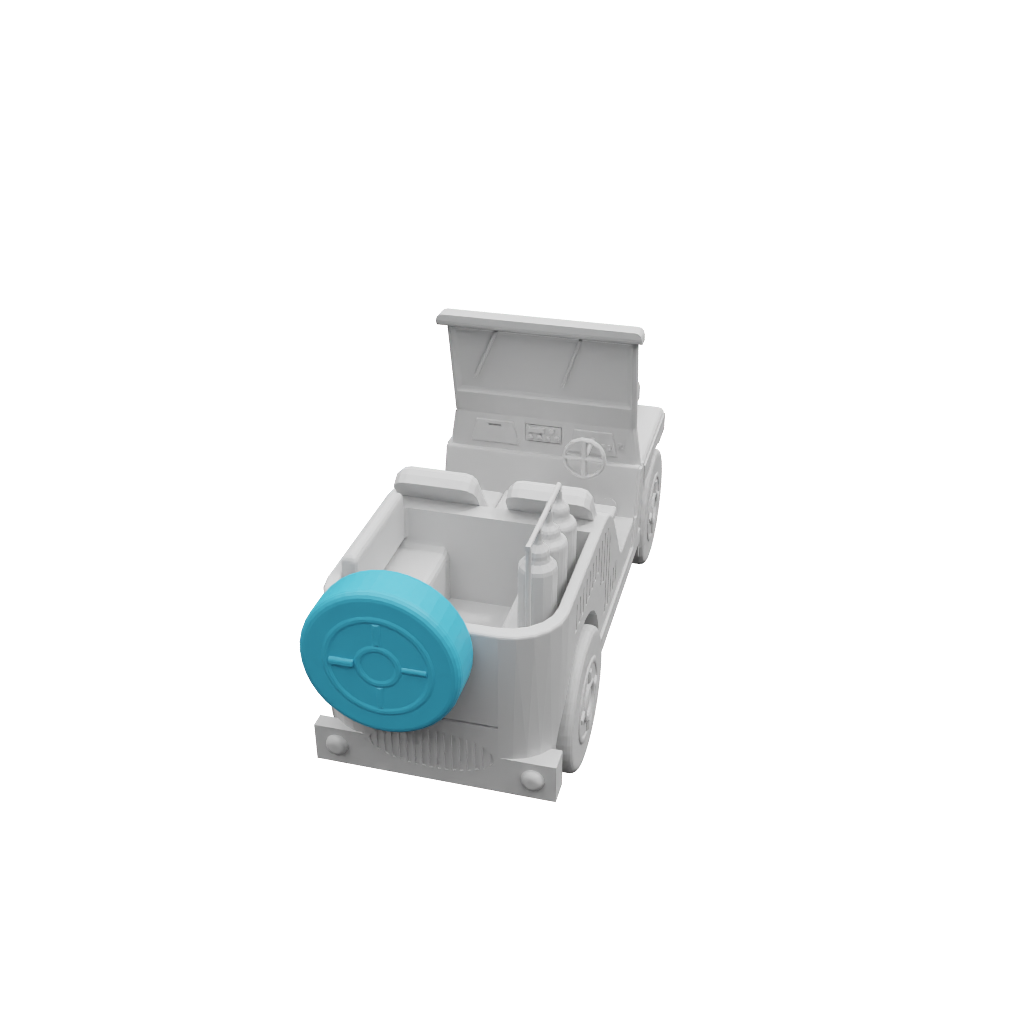}
& \colC{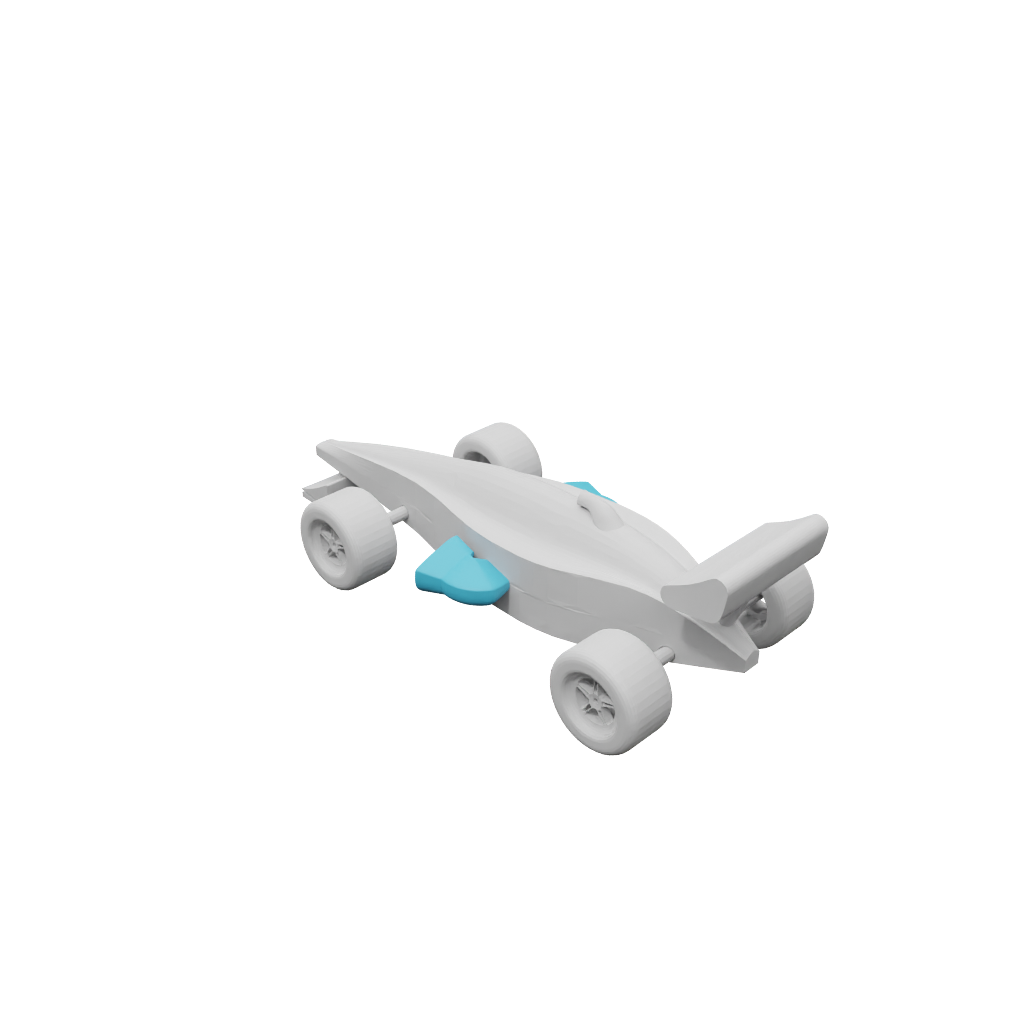}
& \colD{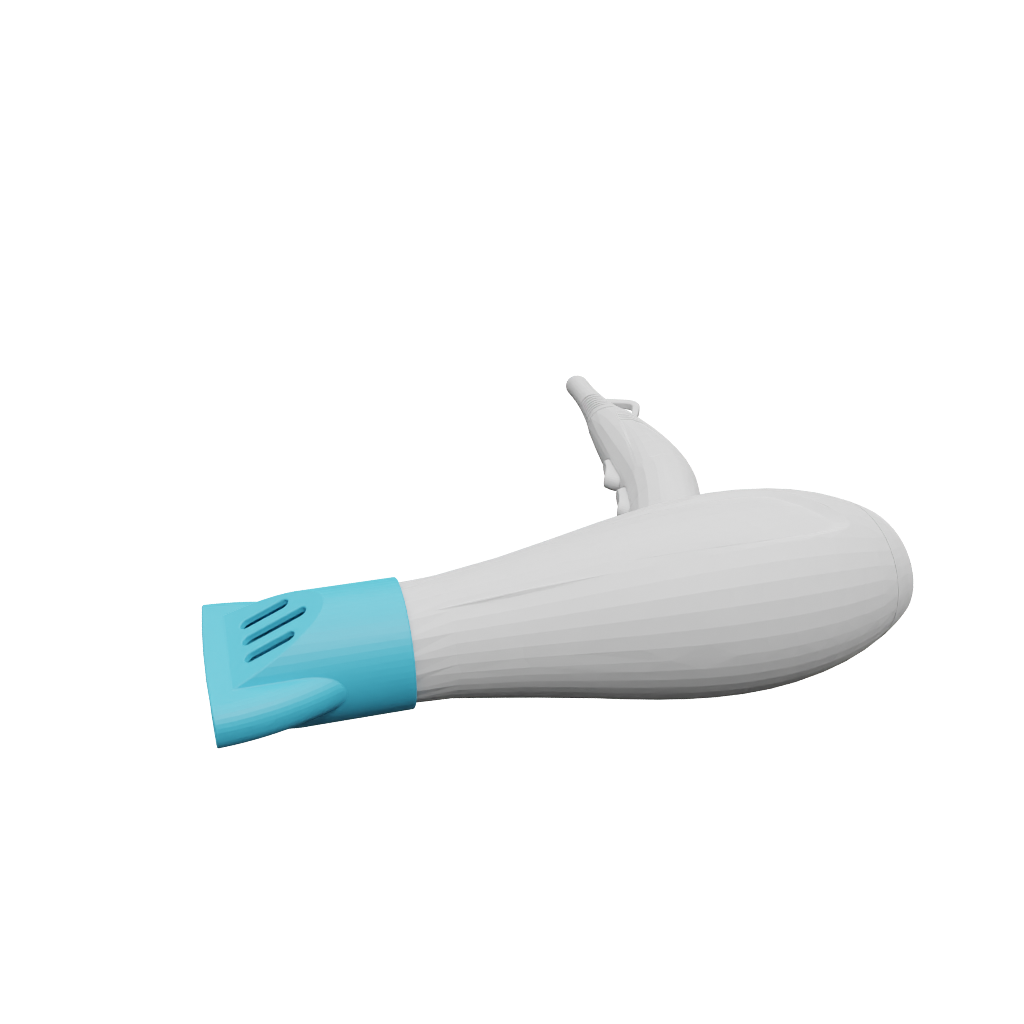}
& \colE{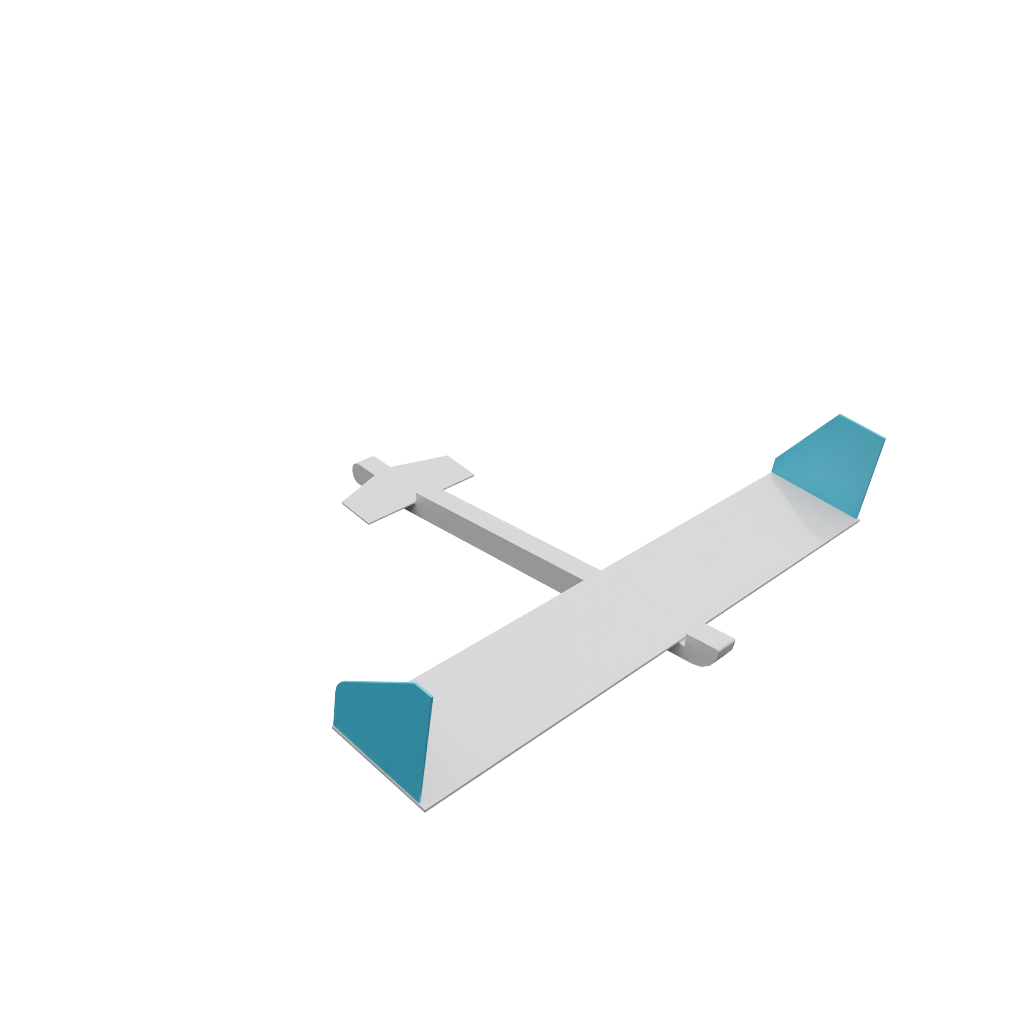} \\
\hline
\centering 3DMorph
& \colA{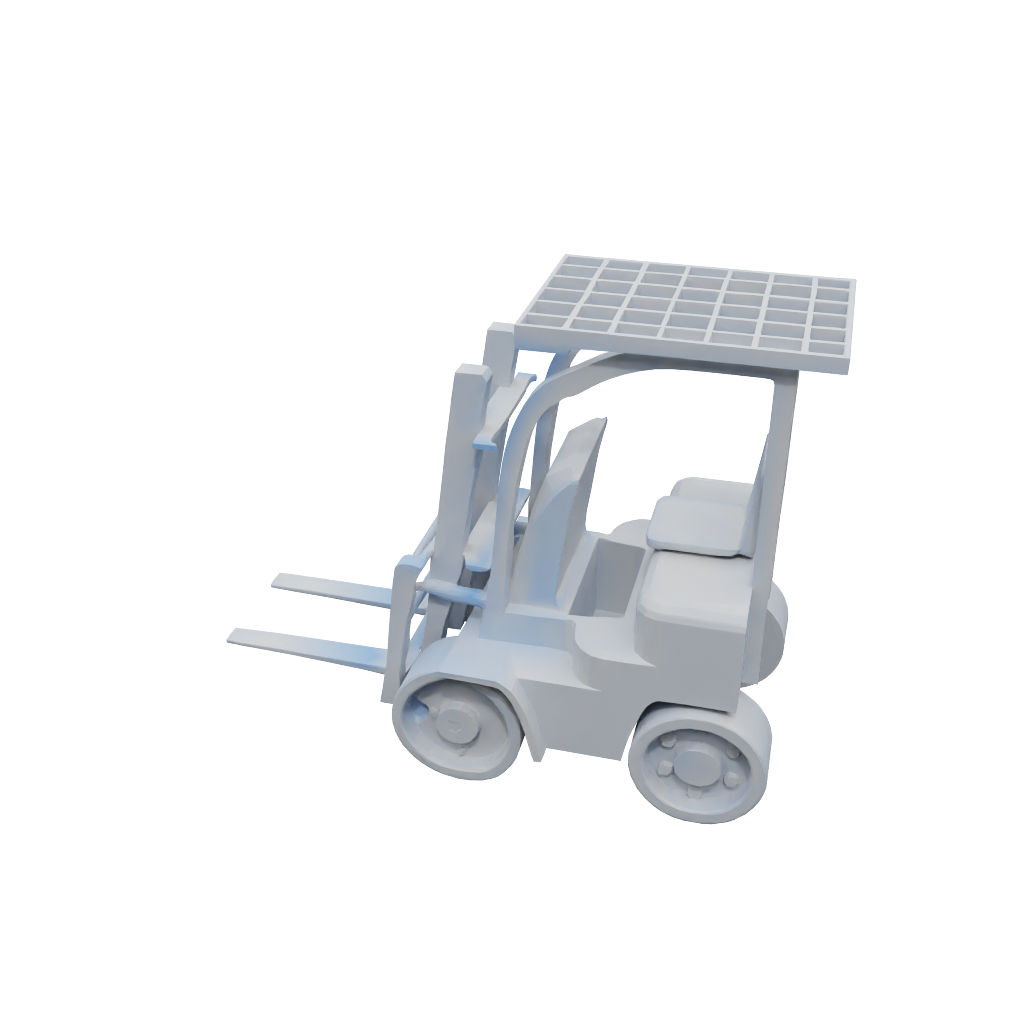}
& \colB{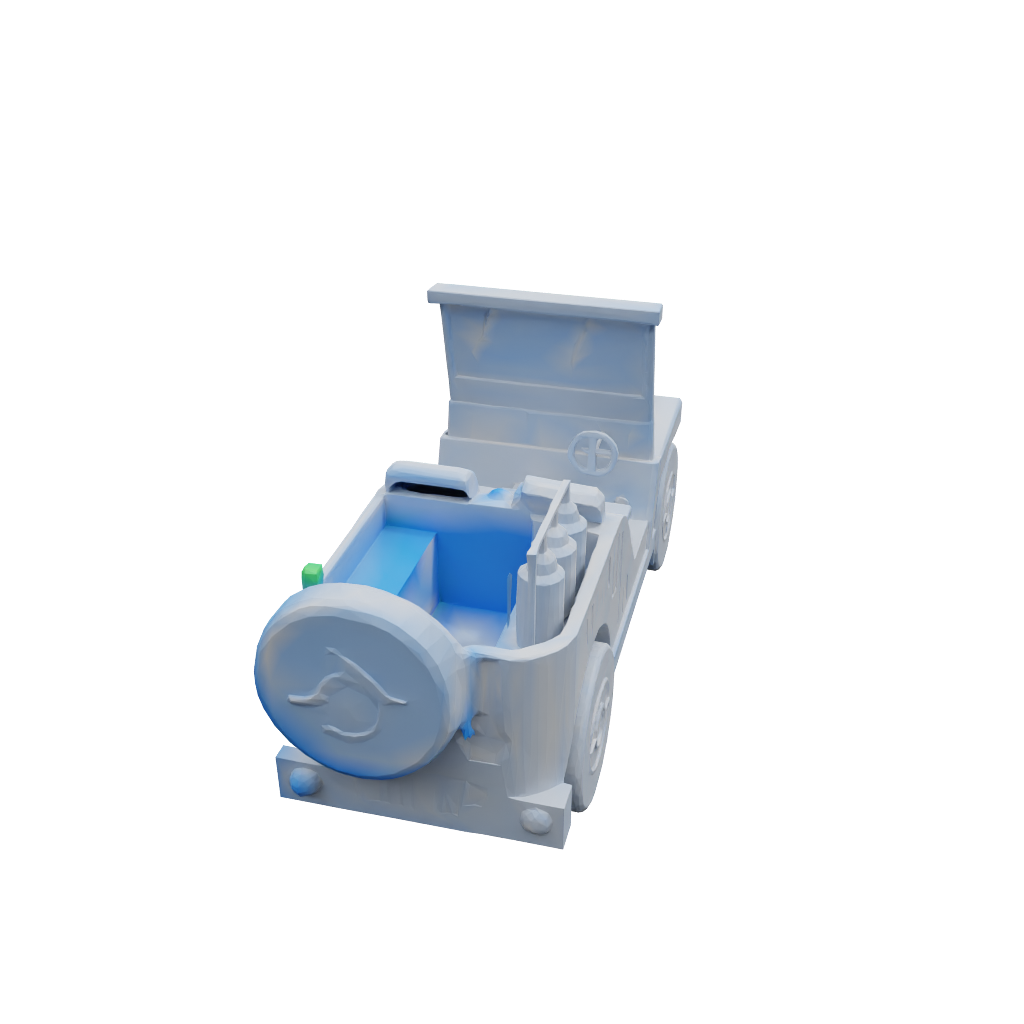}
& \colC{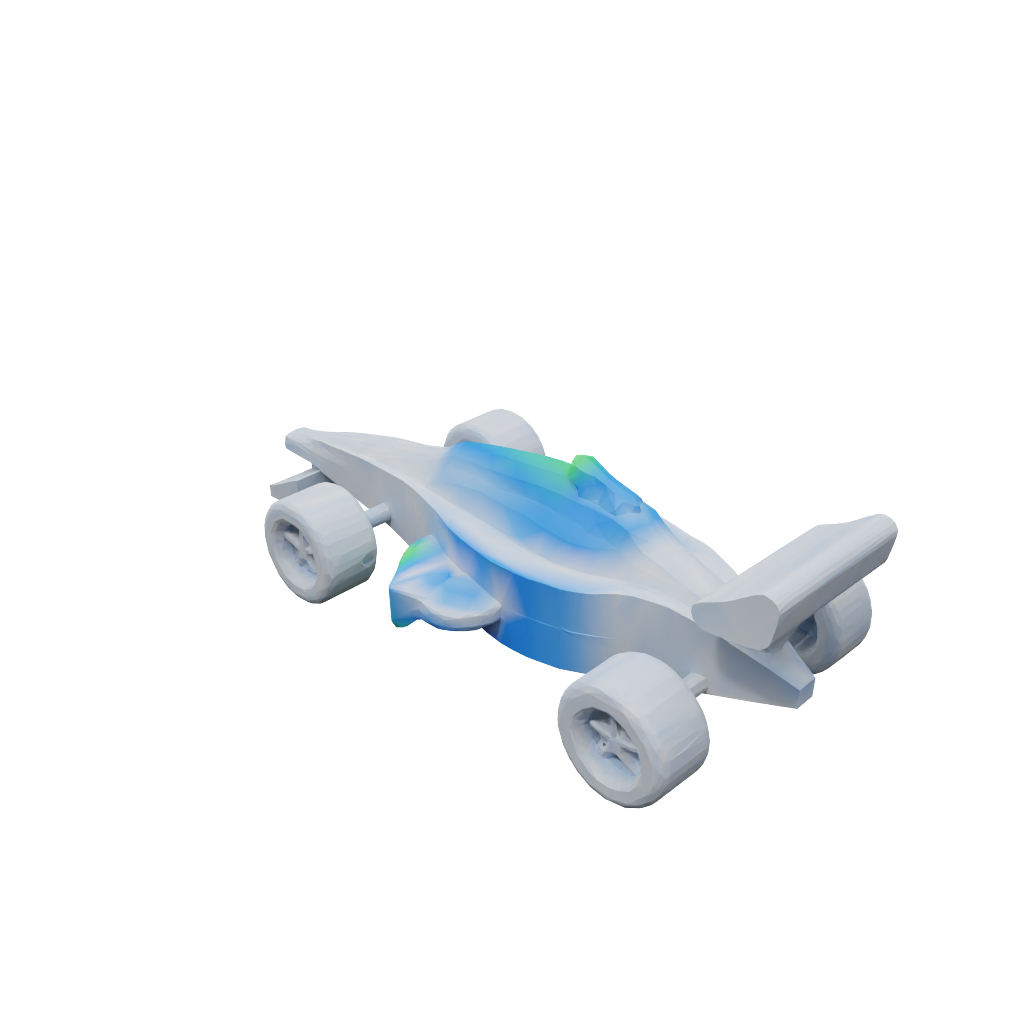}
& \colD{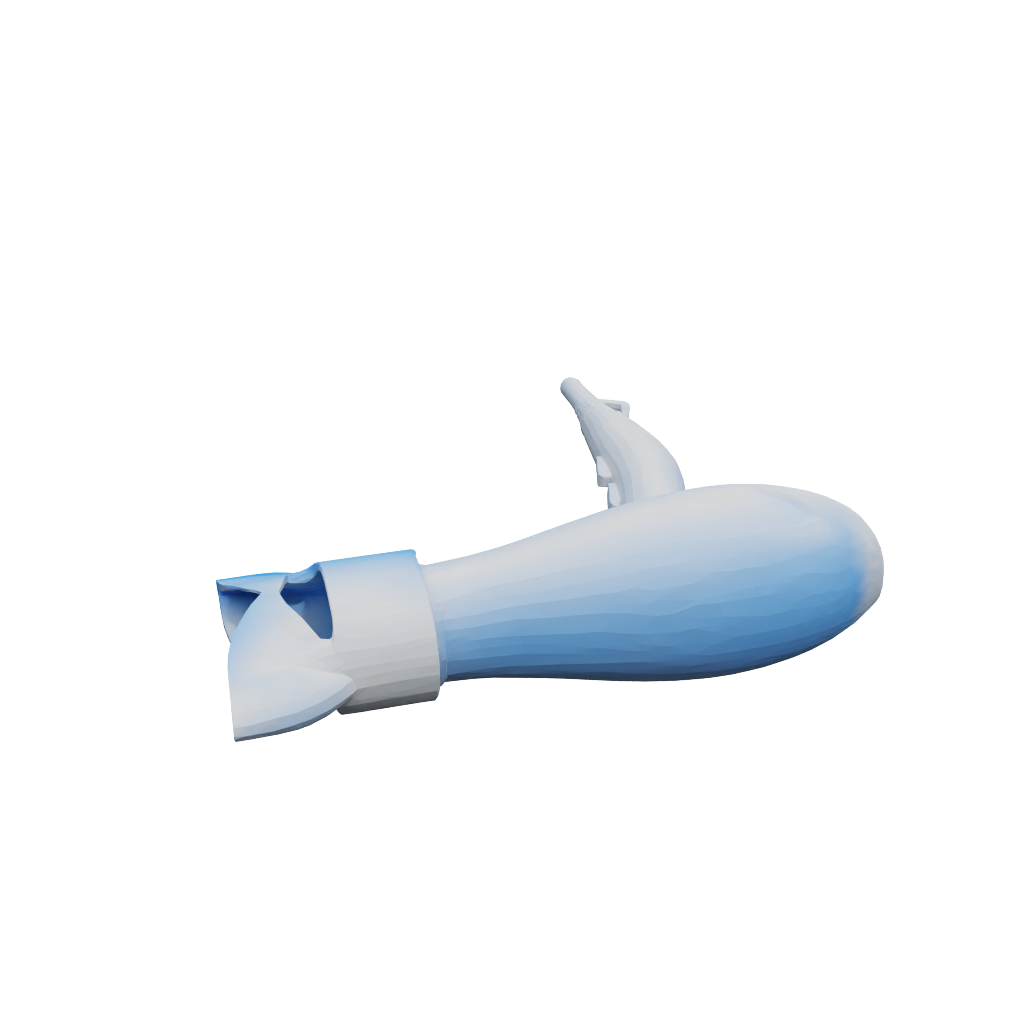}
& \colE{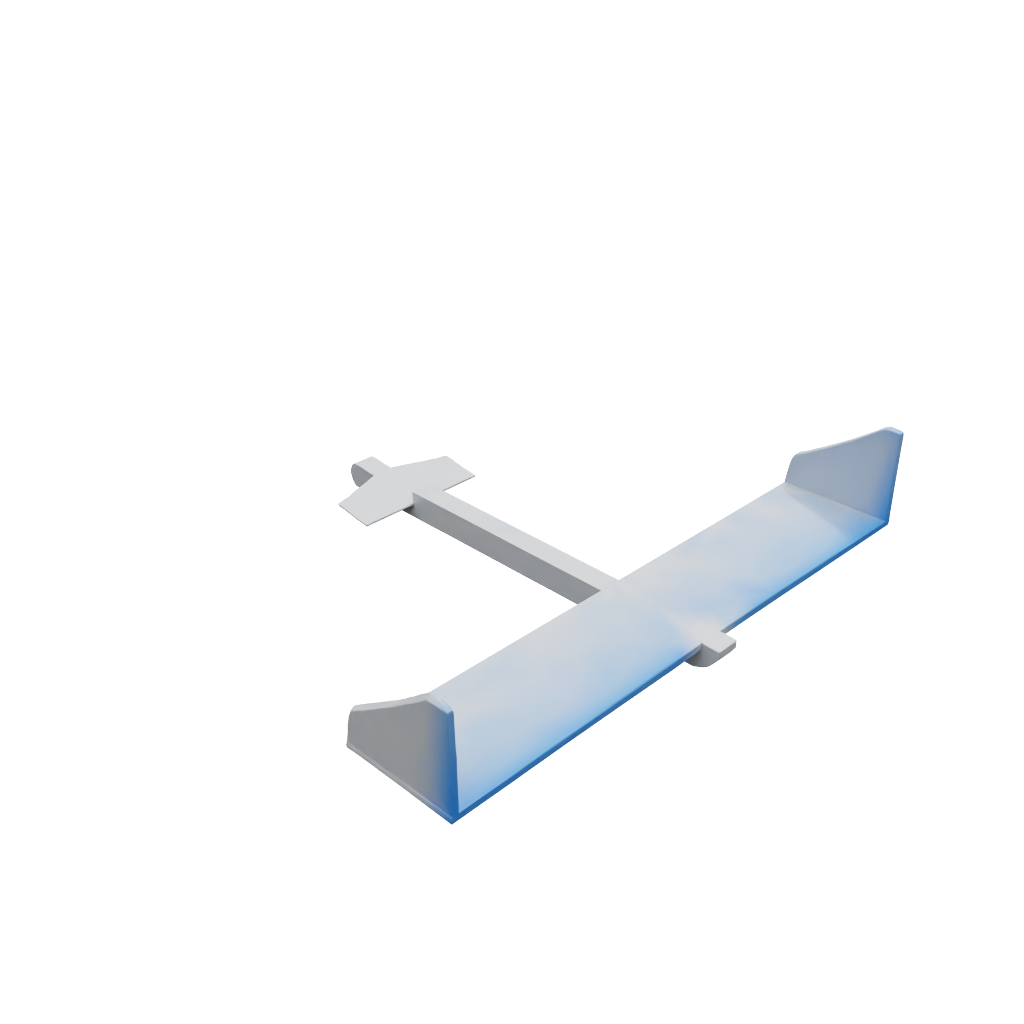} \\
\centering \shortstack{Trellis- \\ RePaint}
& \colA{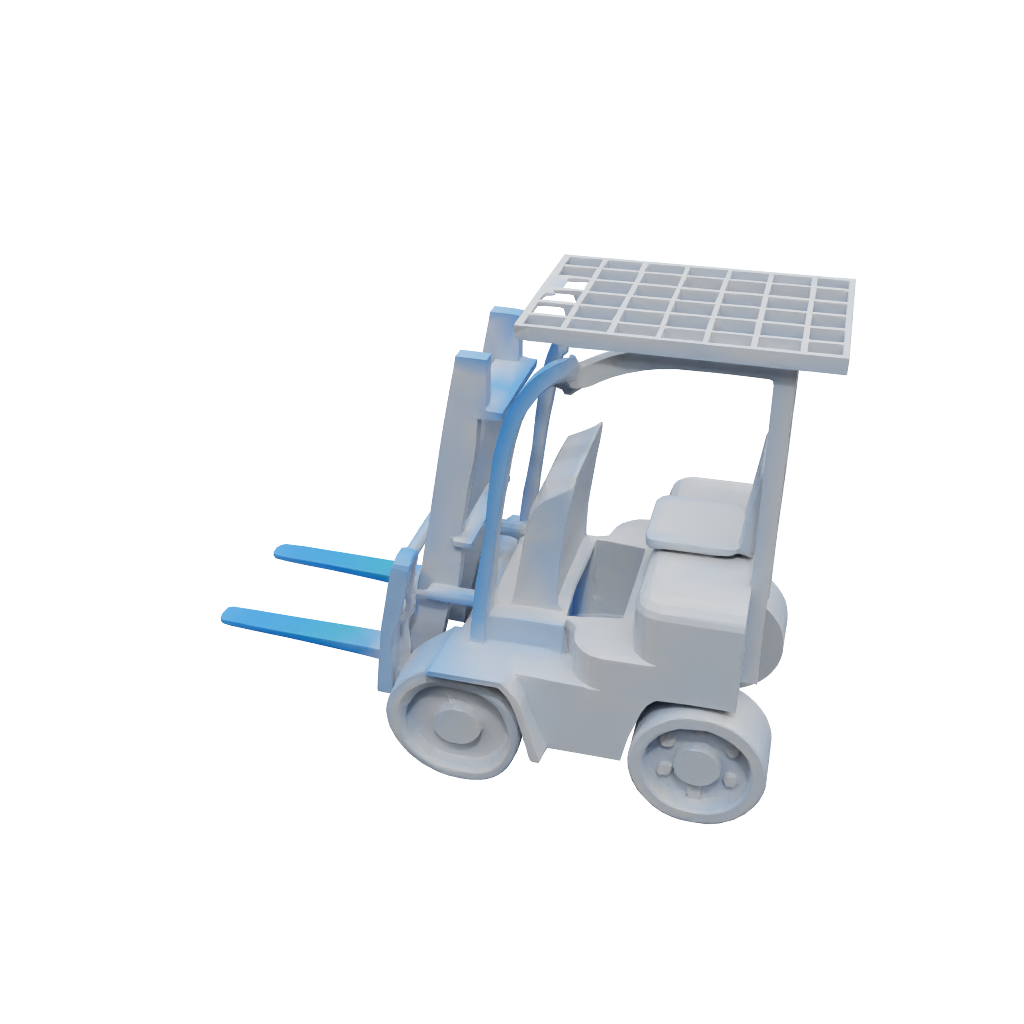}
& \colB{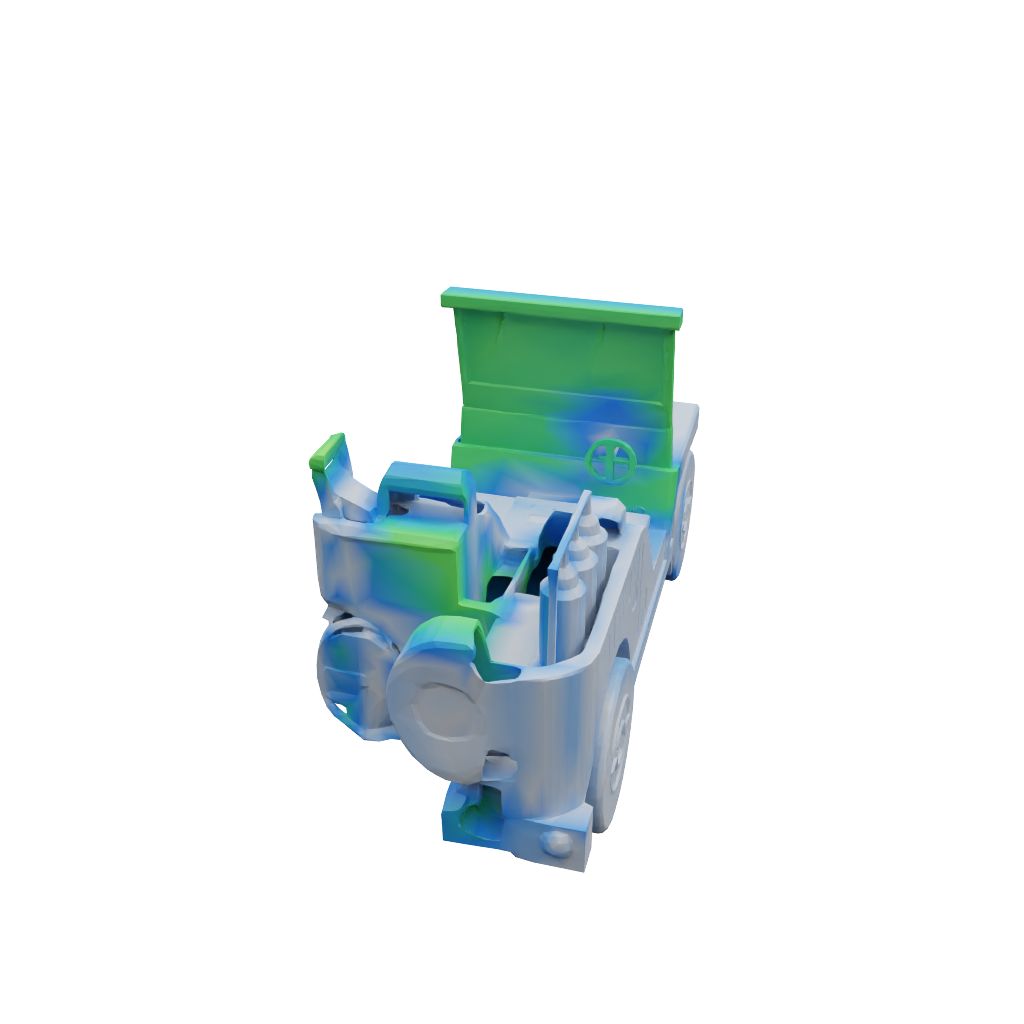}
& \colC{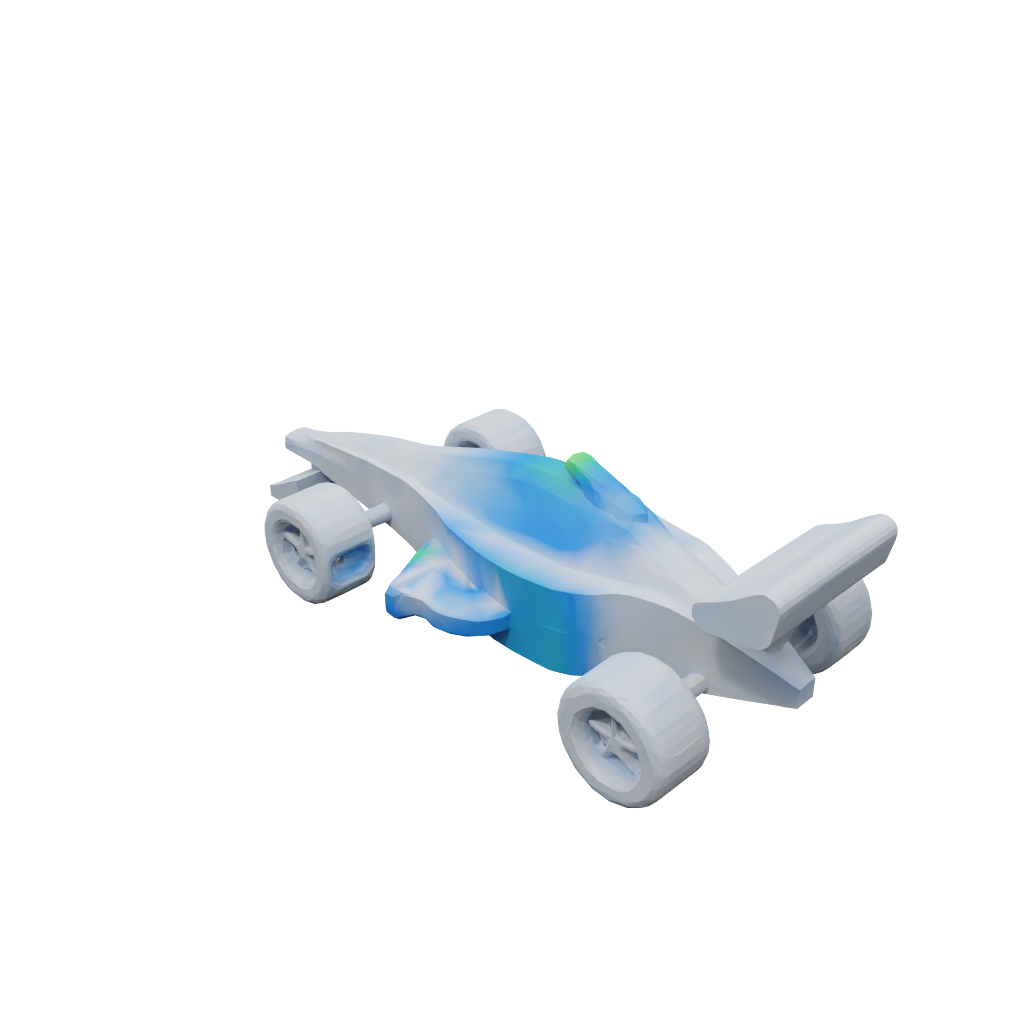}
& \colD{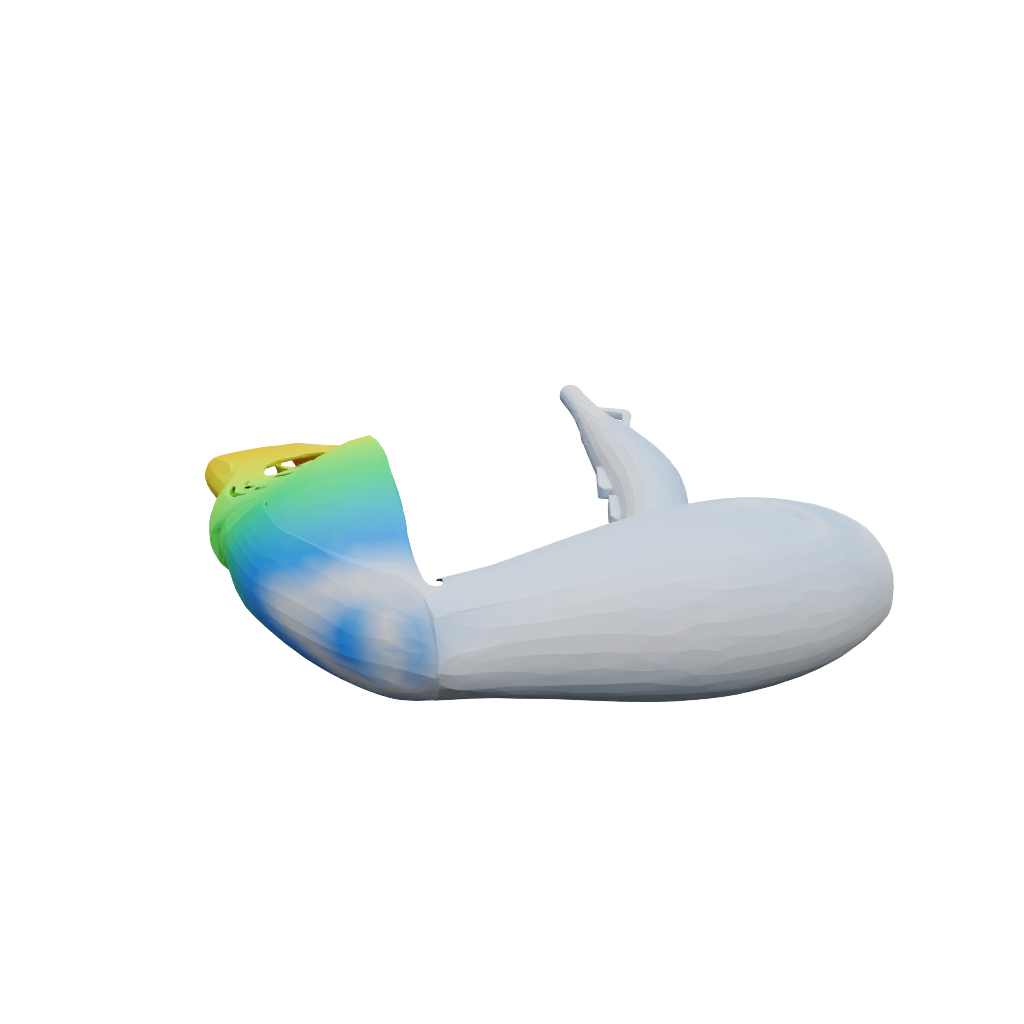}
& \colE{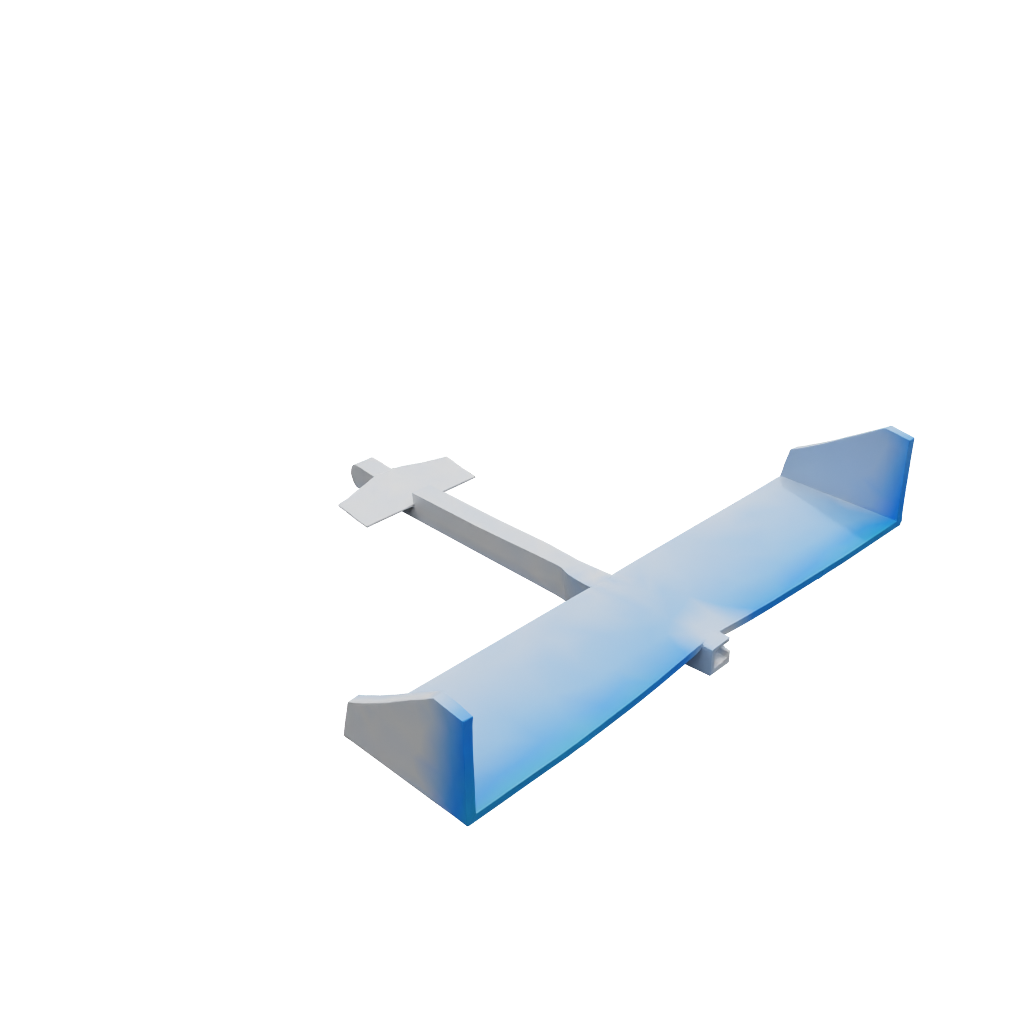} \\
\centering Hunyuan3D
& \colA{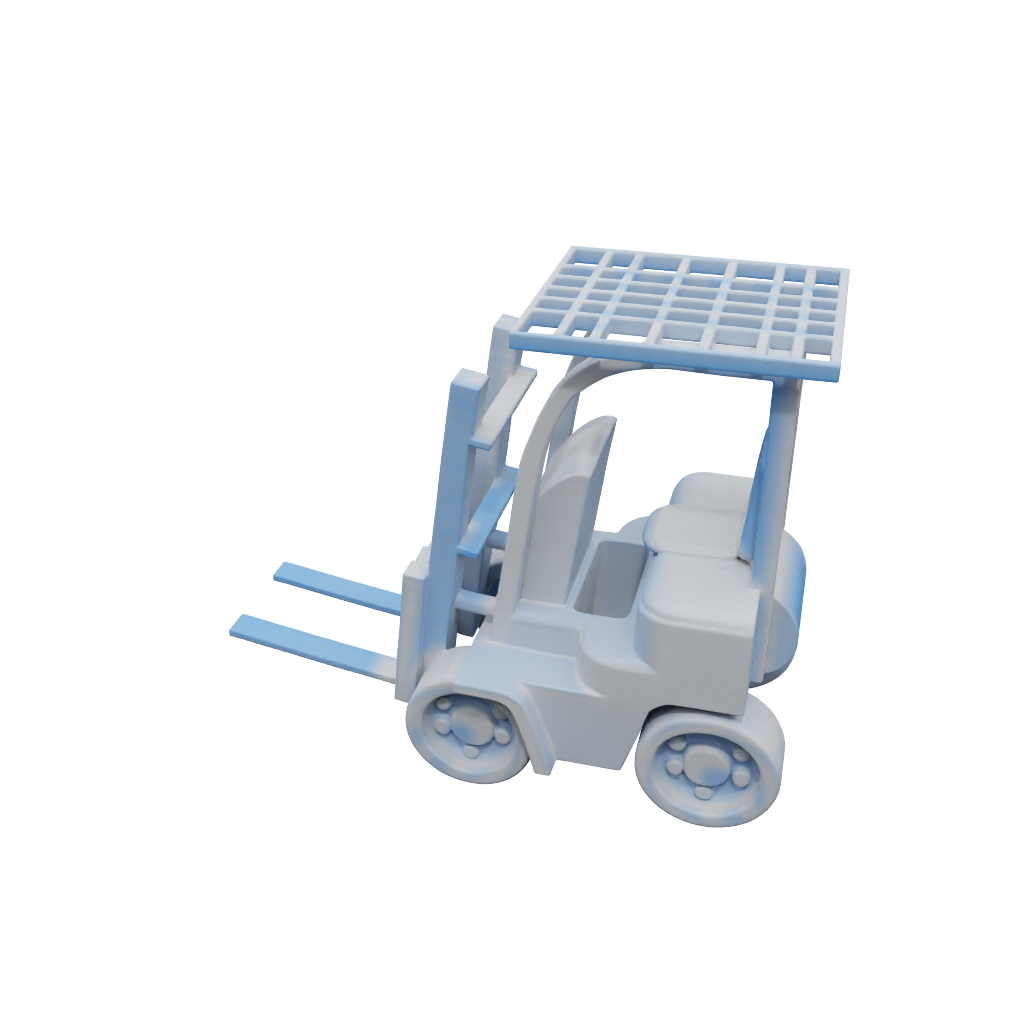}
& \colB{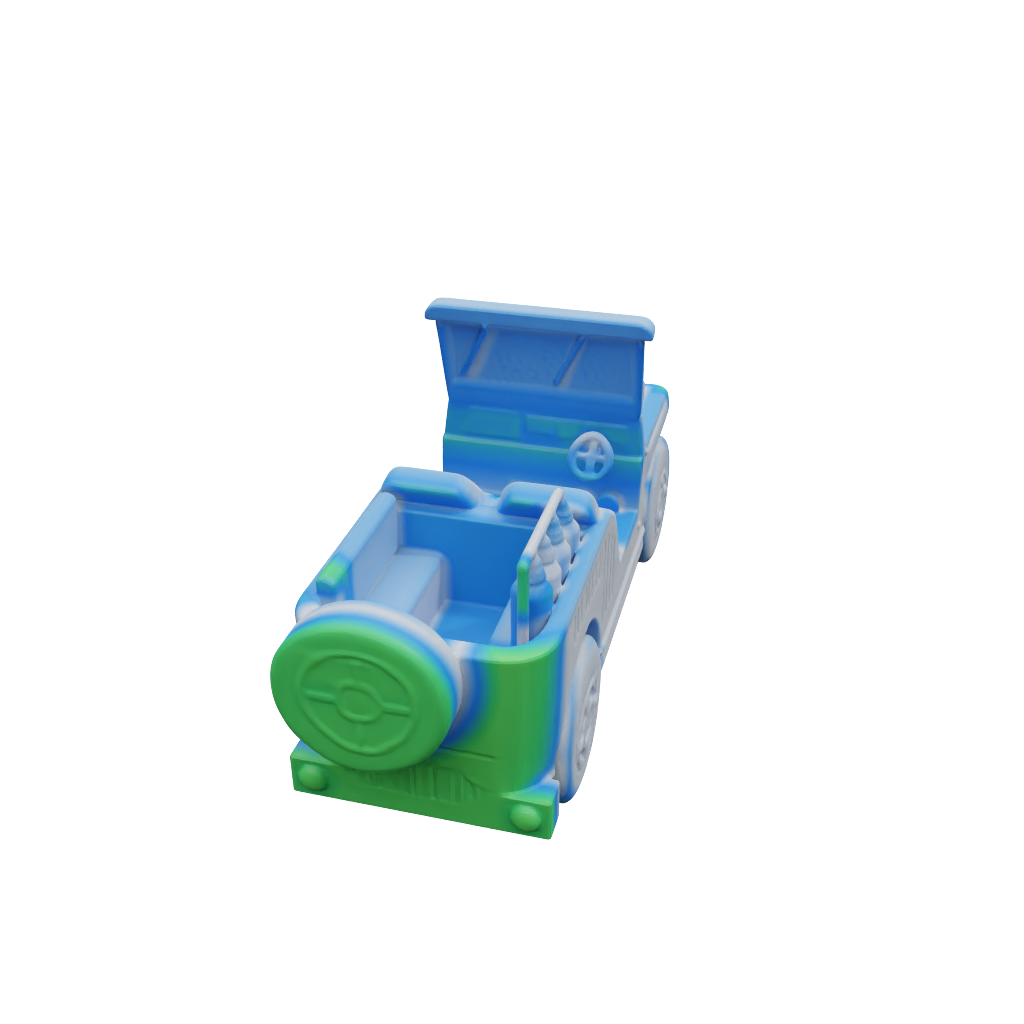}
& \colC{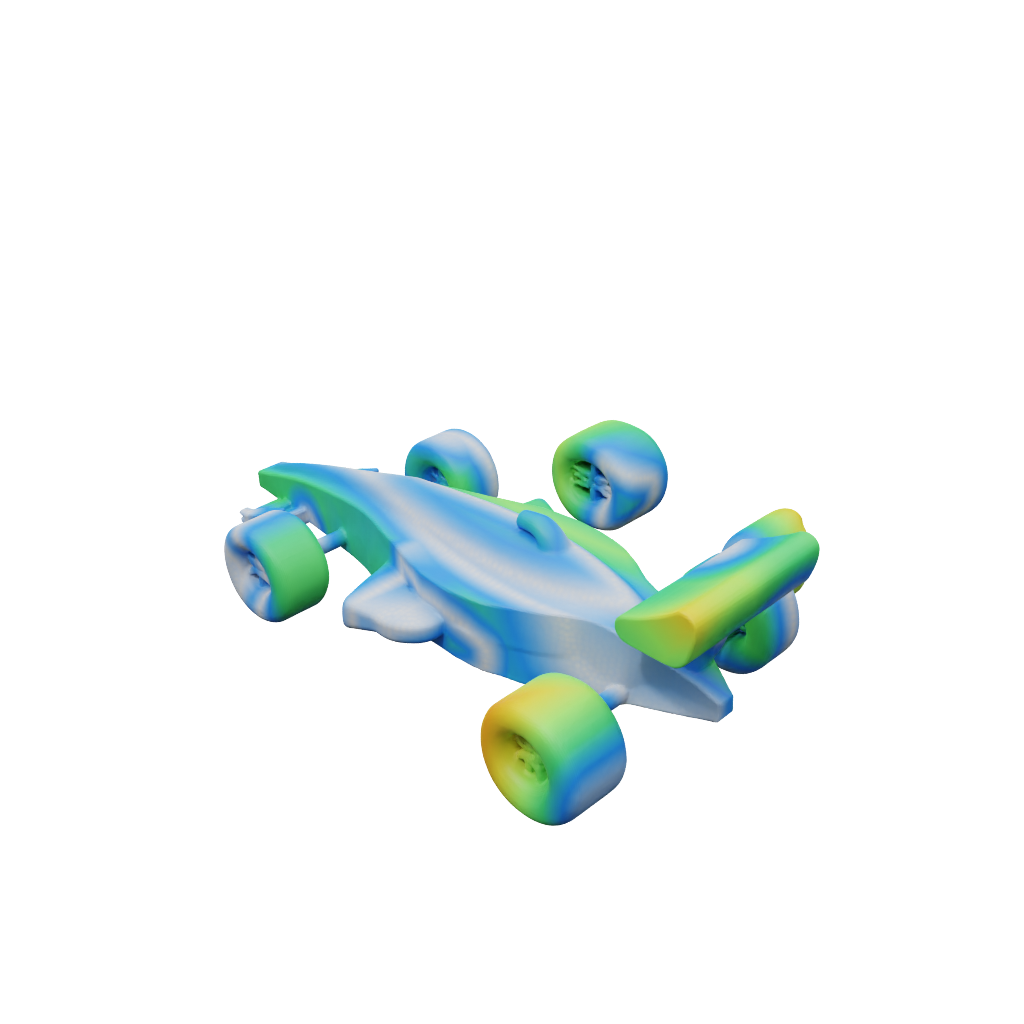}
& \colD{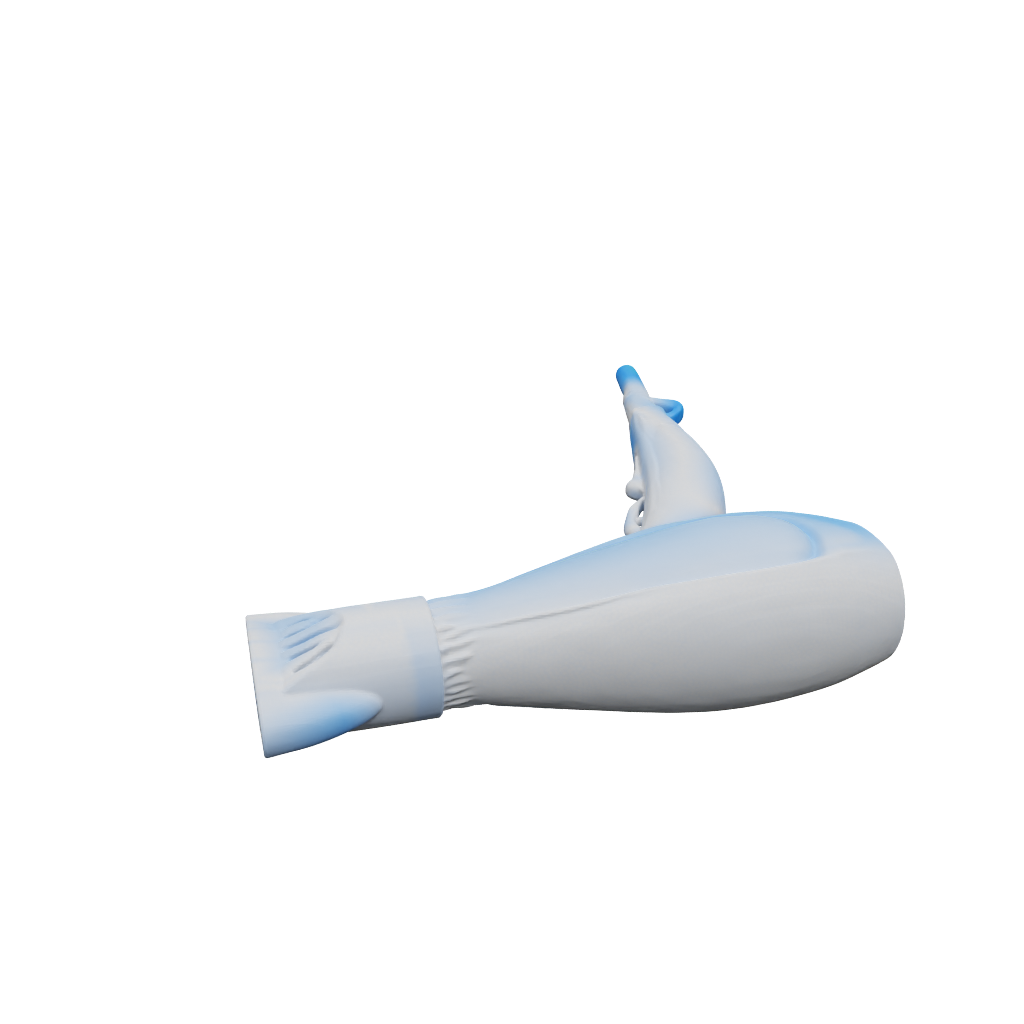}
& \colE{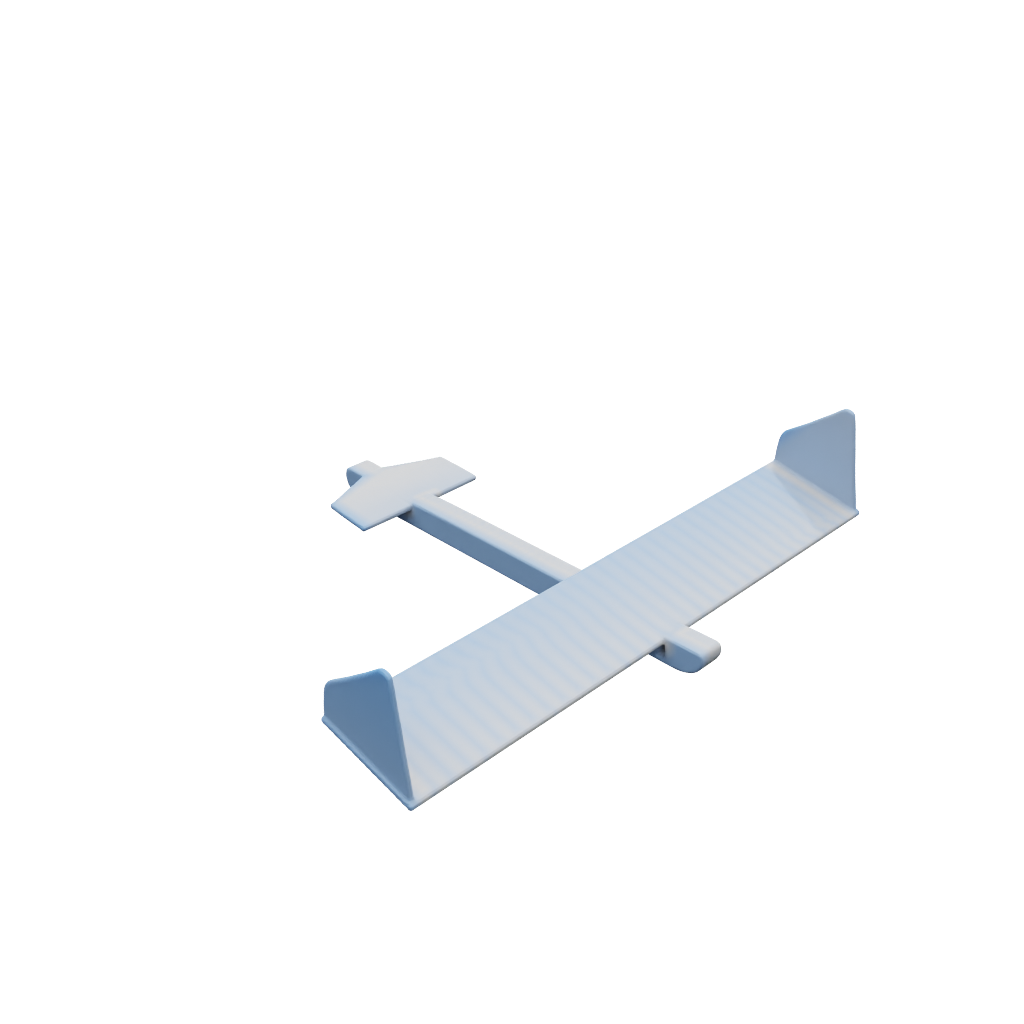} \\
\centering Trellis
& \colA{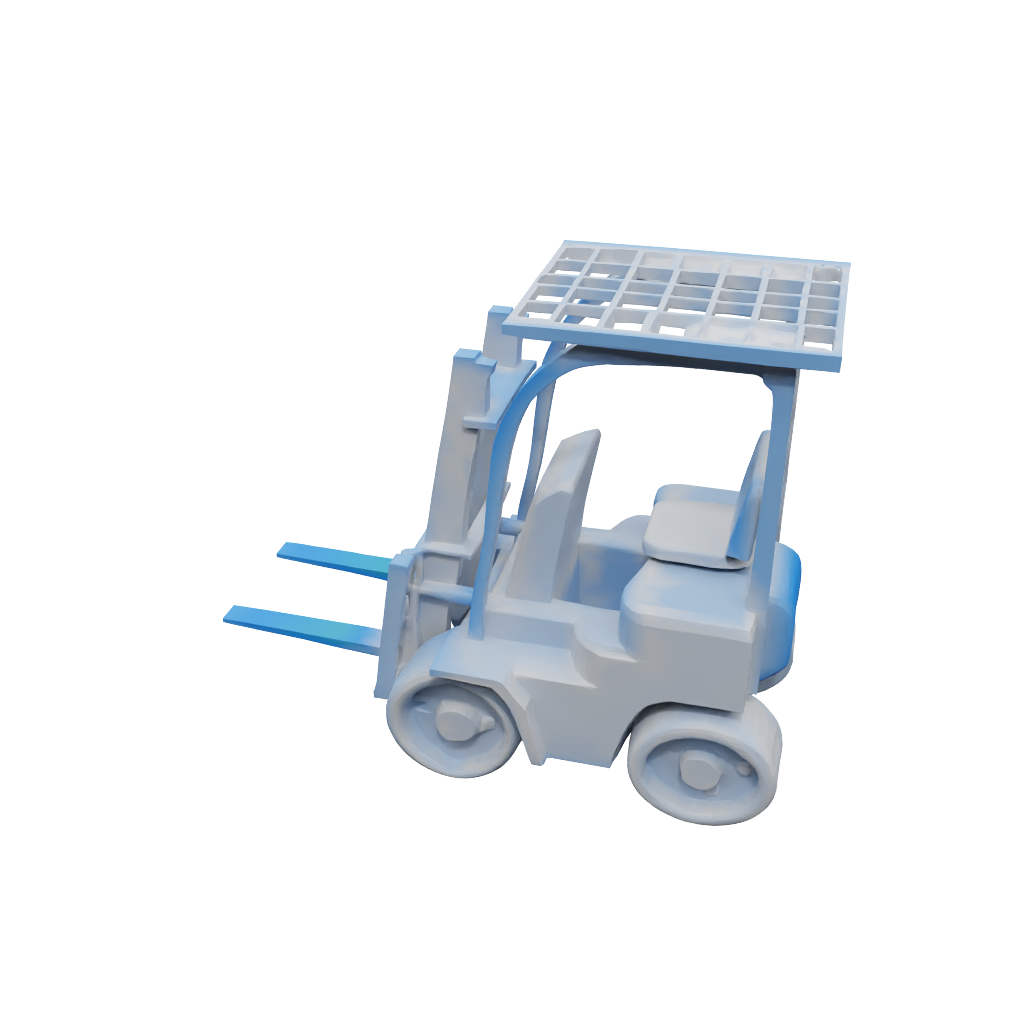}
& \colB{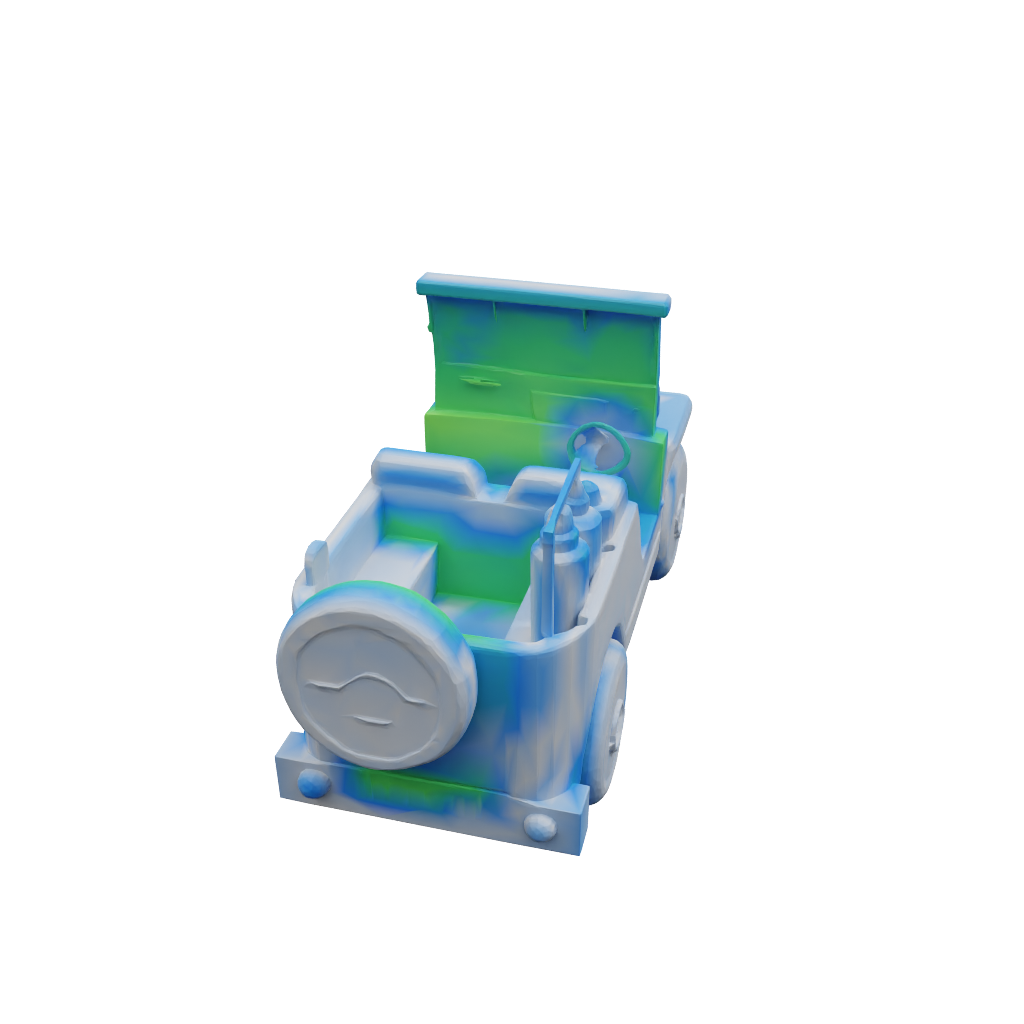}
& \colC{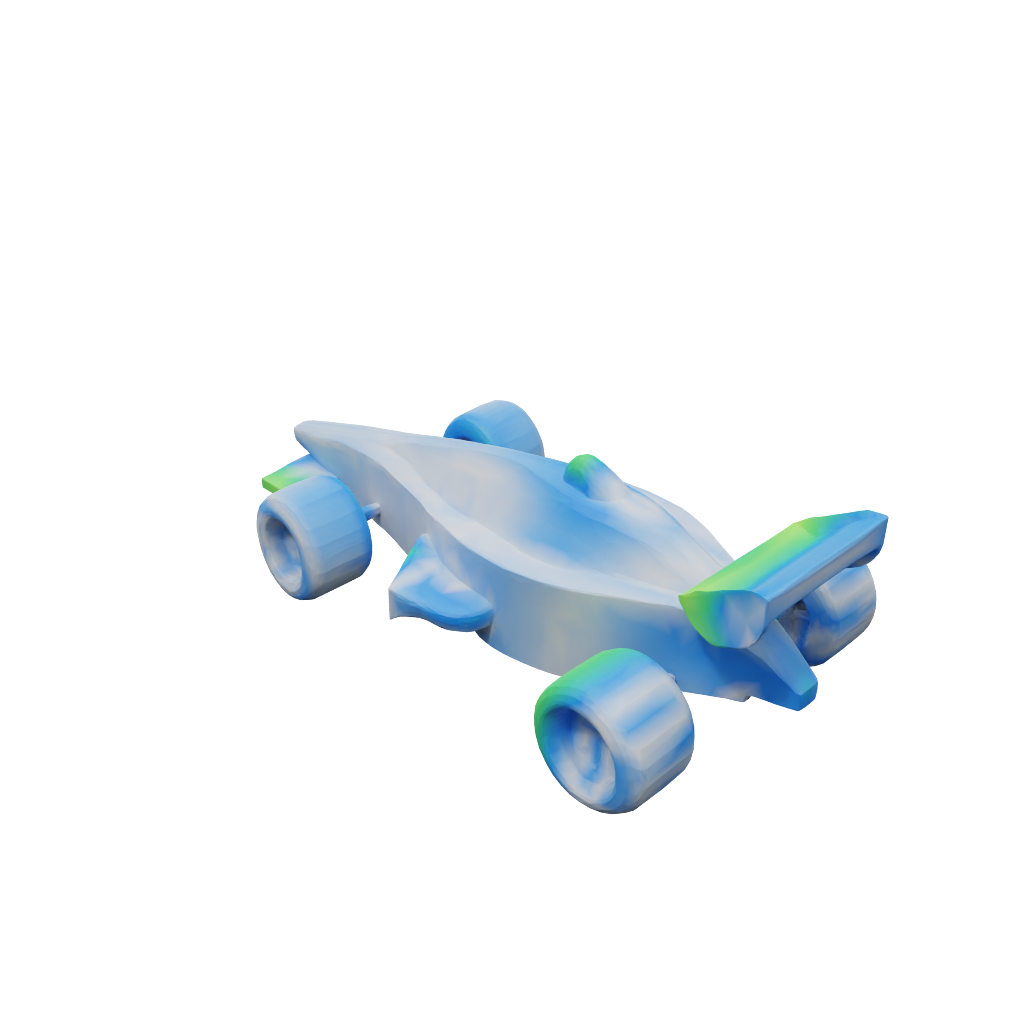}
& \colD{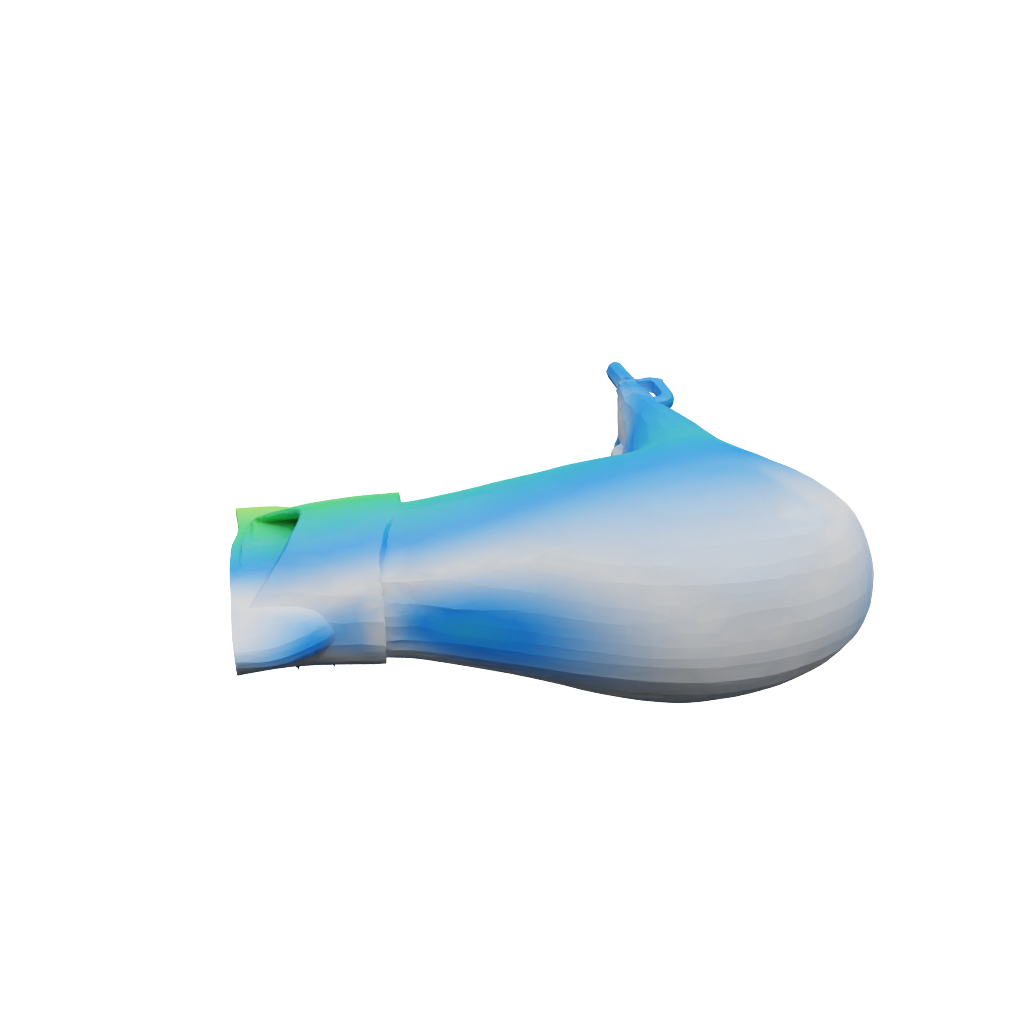}
& \colE{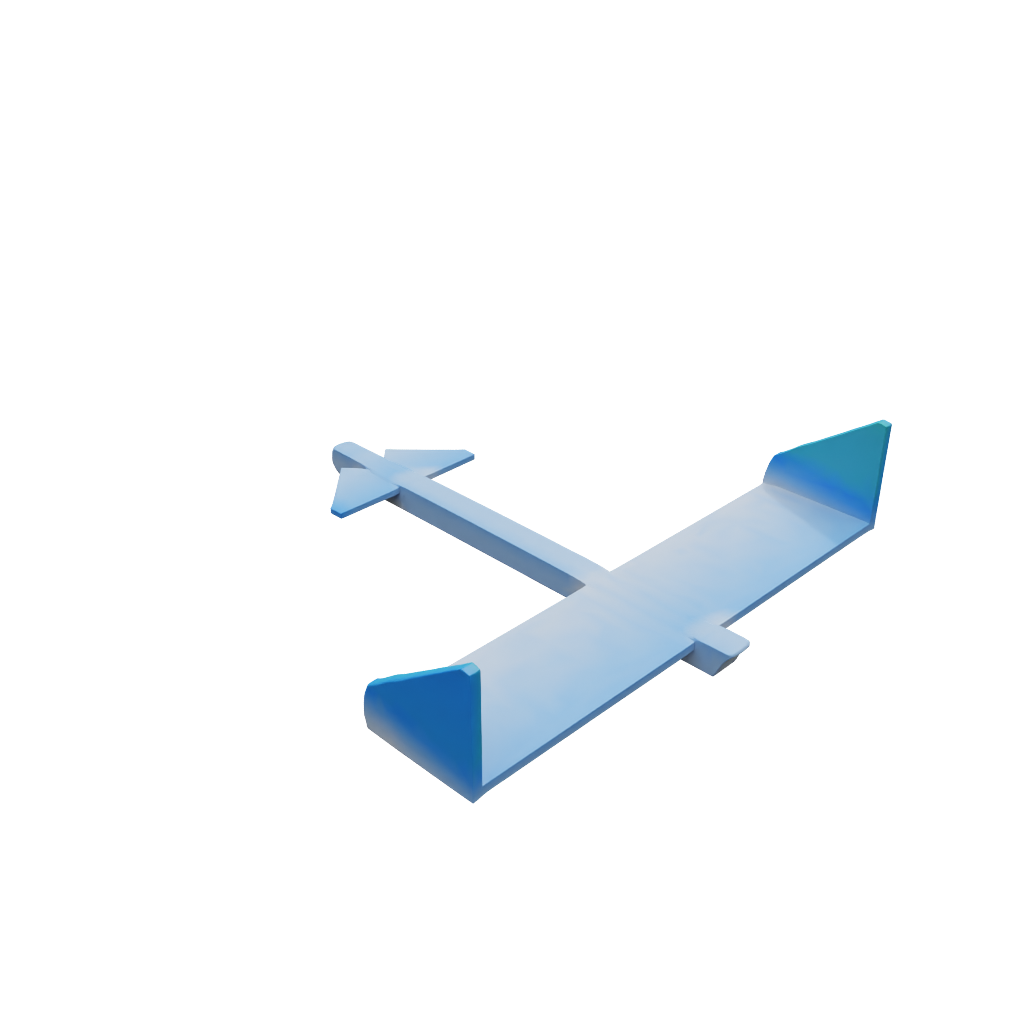} \\
\centering TripoSG
& \colA{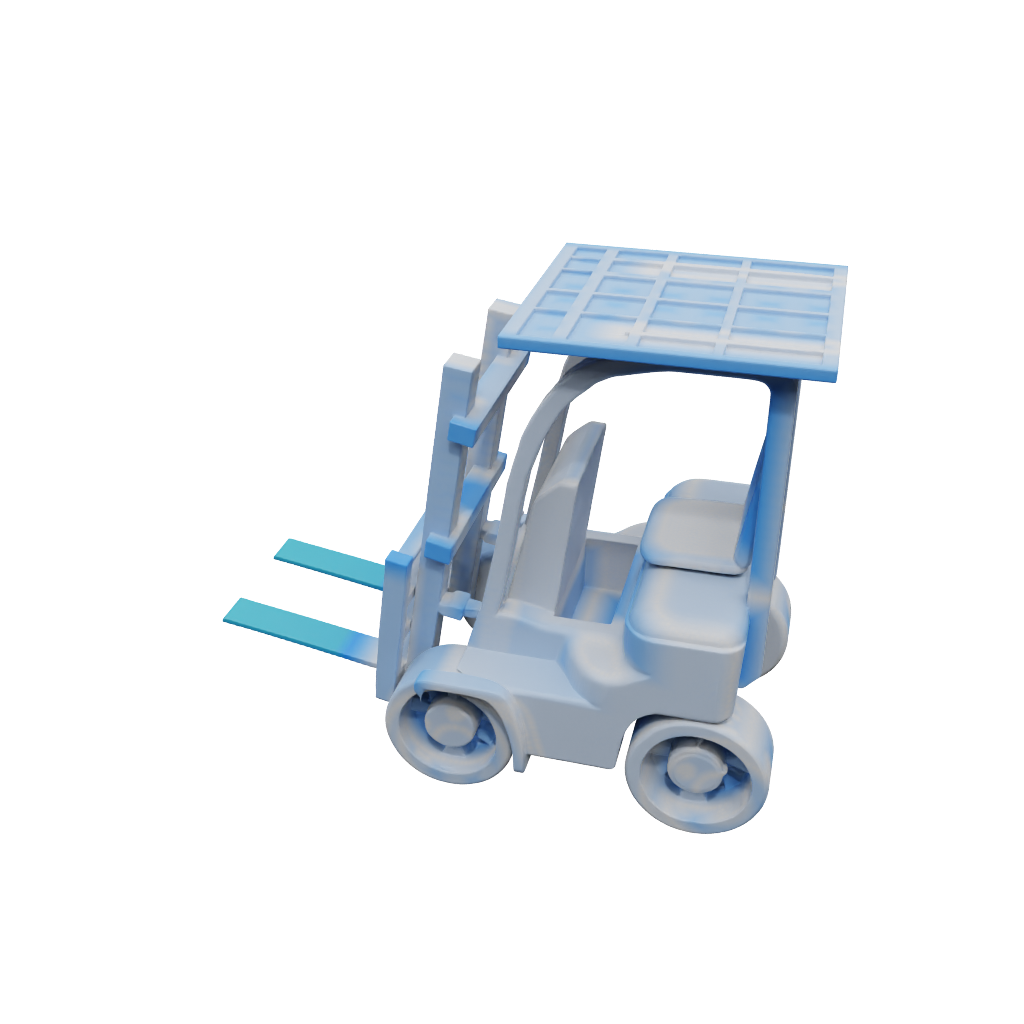}
& \colB{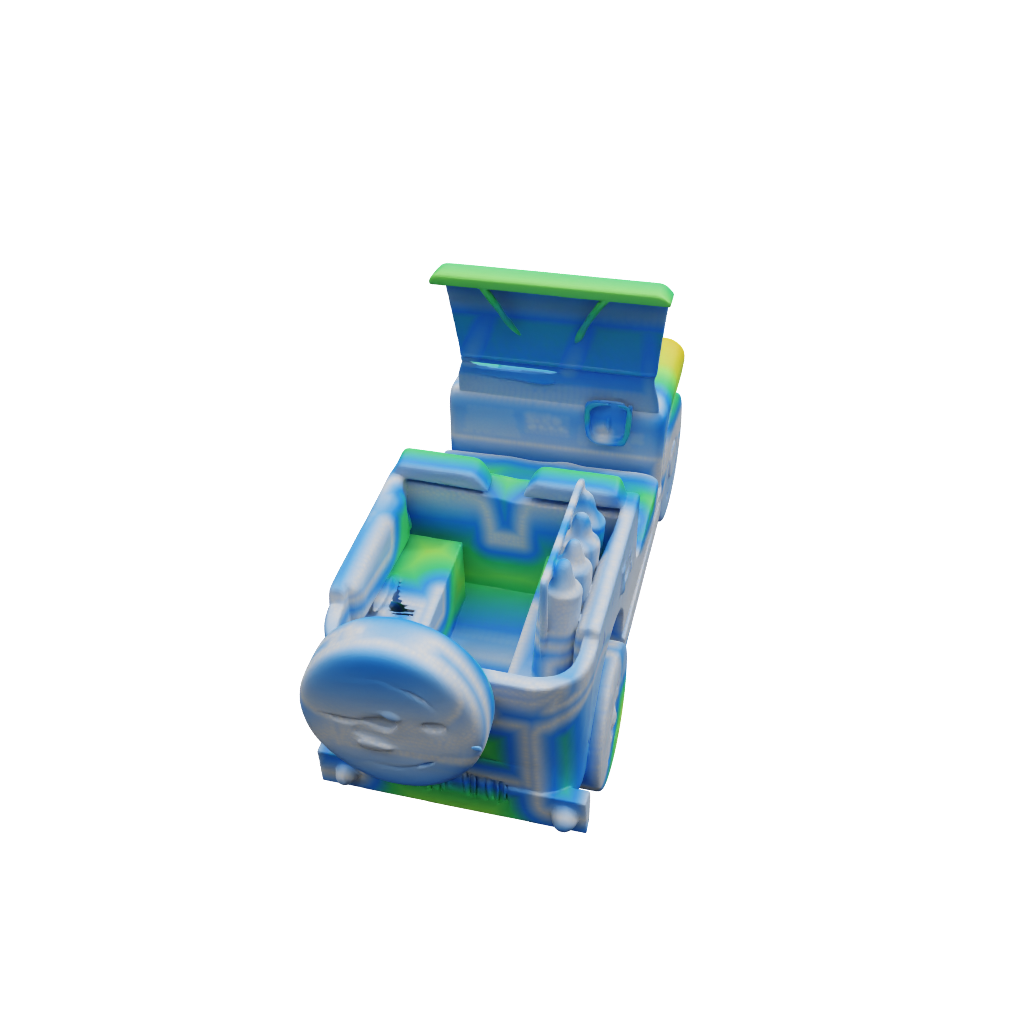}
& \colC{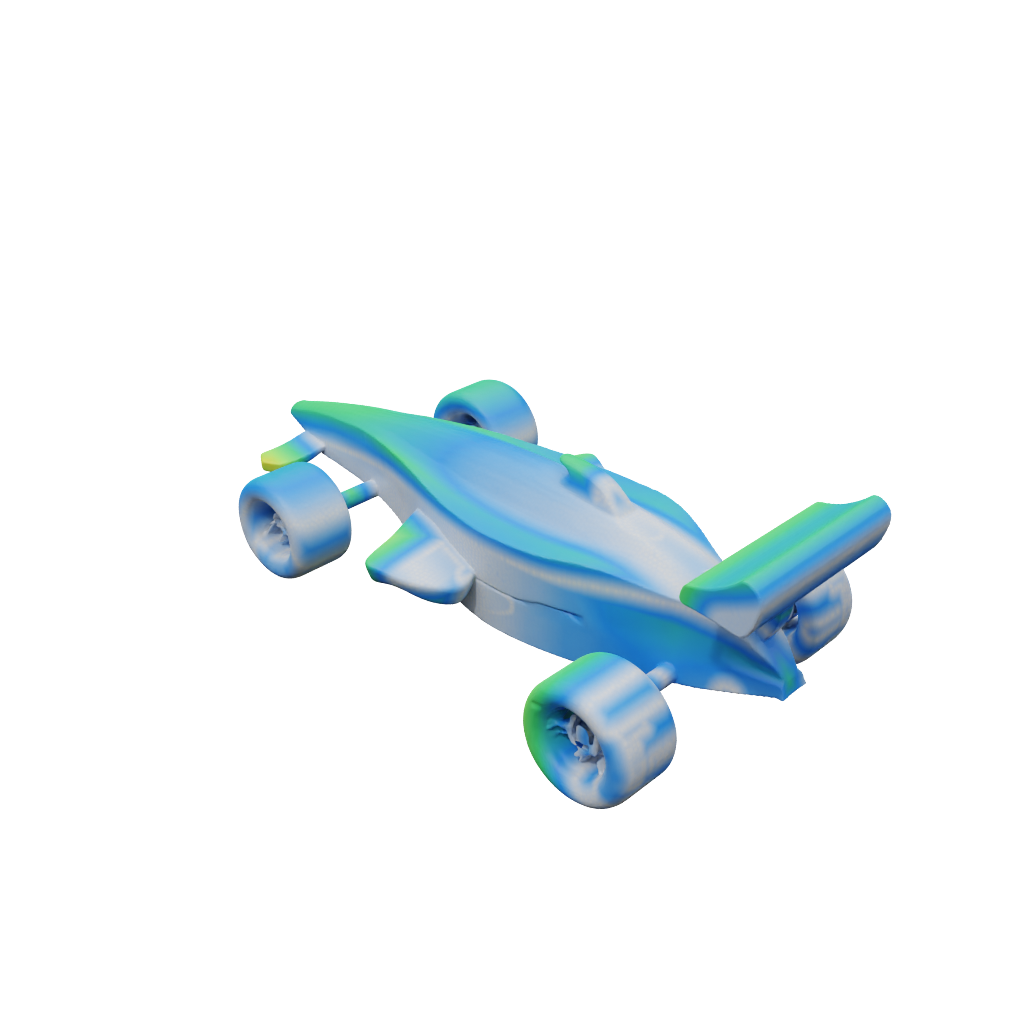}
& \colD{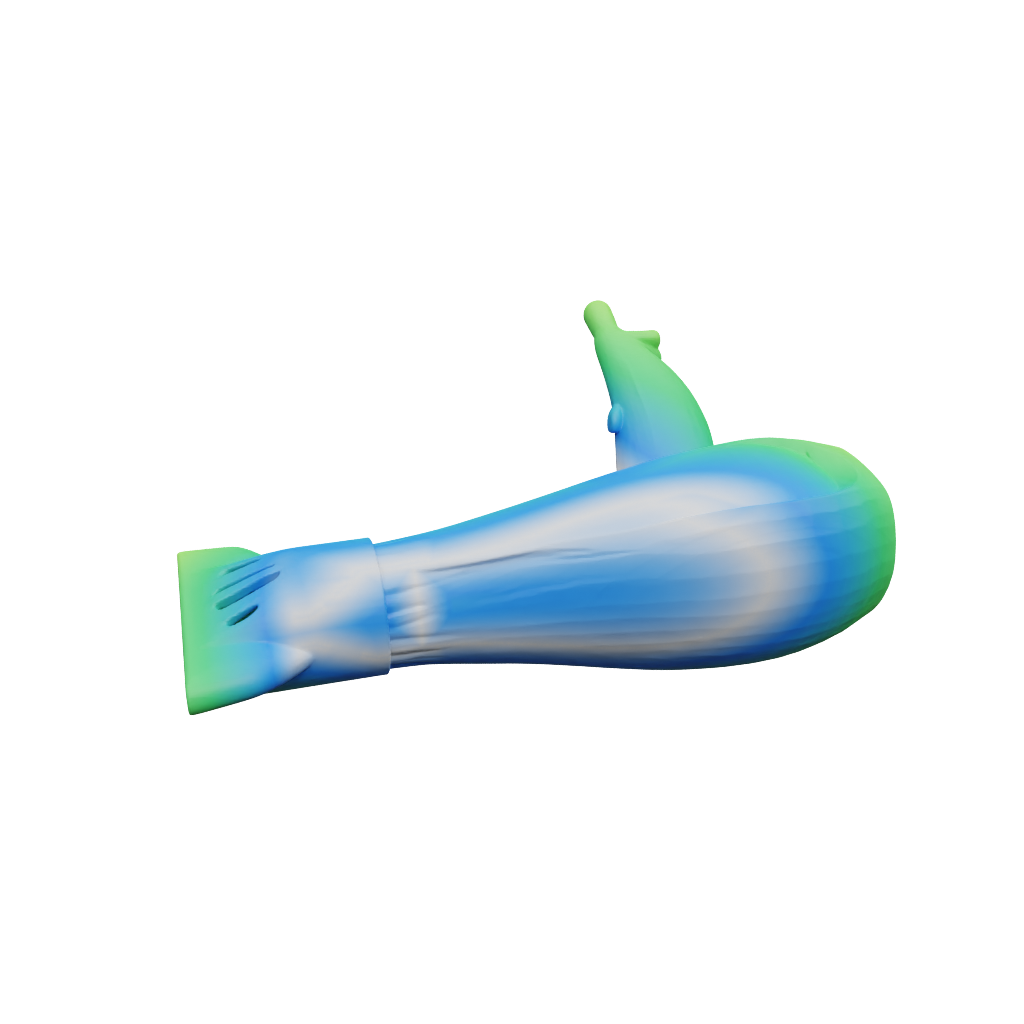}
& \colZ{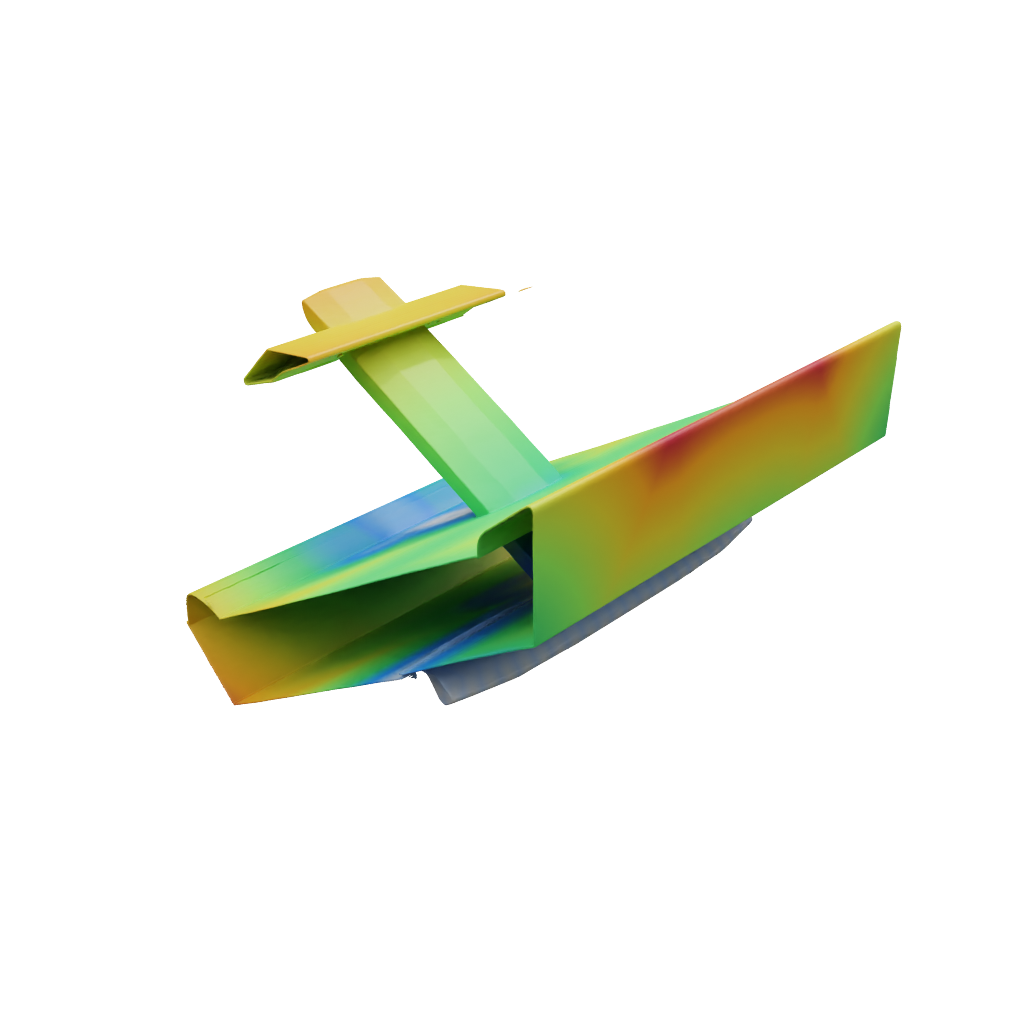} \\
\hline
\end{tabular}
\end{adjustbox}
\caption{Qualitative 3D editing results. The first two rows show the original object and the ground-truth edit, where blue denotes the desired local edit. Differences between the edited object and the GT $\bar{O}$ are visualized, with colors ranging from blue (small deviation) to red (large deviation).}
\label{fig:diff-main}
\end{figure}

\textbf{Bounding box prediction.} We evaluate bounding box quality using Intersection over Ground Truth (IoGT) between predicted and ground-truth boxes. IoGT prioritizes target coverage over penalizing excess area, reflecting the goal of avoiding missing edited regions. 
Ground-truth boxes occupy only 4.1\% of the voxel volume, while the predicted boxes remain compact at 8.2\% and achieve an average IoGT of 58\%. 
The larger predicted boxes stem from assumptions about unknown depth, with modest overestimation helping to mitigate this ambiguity.

\textbf{Bounding box ablation.} We assess the impact of the bounding box prediction module through an ablation study on 3DMorph and Trellis-RePaint, replacing predicted boxes with ground-truth ones obtained by comparing $O$ and $\bar{O}$. 

Table~\ref{table:ablation} shows that this substitution improves performance for both methods, but 3DMorph remains superior to Trellis-RePaint. Notably, 3DMorph without access to ground-truth boxes nearly matches the performance of Trellis-RePaint with ground-truth box access. This demonstrates the robustness of 3DMorph when using predicted boxes and its competitiveness even against methods with perfect bounding box information.


\textbf{Edit-region evaluation.} Our Delta3D benchmark enables quantitative 3D evaluation of edits, filling a gap in the assessment of 3D editing methods noted in prior work~\cite{li2025cmd}. Tab.~\ref{table:inside-BB} compares the editing quality of 3DMorph with Trellis-RePaint, demonstrating consistent superiority of our method when using either predicted or ground-truth bounding boxes.





\textbf{Image edits for 3D shape editing.}
Fig.~\ref{fig:main-fig} demonstrates our method in its intended application scenario: lifting single-image edits into 3D. 
Image edits can be flexibly created by the user, including inpainting, hand-sketching, or crop-and-paste operations using parts from other images. 
Inpainting is performed using FLUX.1 Fill~\cite{flux2024} and Gemini~2.5 Flash Image~\cite{comanici2025gemini}. 
Our method supports diverse 3D modifications, including adding, removing, or altering parts, as well as composing multiple edits sequentially.

\textbf{Qualitative results.} 
Fig.~\ref{fig:diff-main} qualitatively demonstrates the superiority of our method in editing object shapes using only a single image as guidance.
We additionally visualize differences between generated edits and the ground truth $\bar{O}$. This color-coded visualization provides only a rough indication, as it assumes geometric correspondence via surface-distance alignment, which may break down for severe deformations where unrelated surfaces lie close in space.

Single-image generative approaches (last three rows) often introduce unintended edits, such as closing the forklift roof or airplane rear wing, adding or removing components (e.g., a gas bottle in the loader), misinterpreting geometry (e.g., treating the race-car side wing as a fifth wheel), or failing to preserve the global shape (e.g., deforming the hair dryer).

Despite using the same bounding box estimate, Trellis-RePaint either fails entirely (e.g., loader, hair dryer) or struggles near bounding box boundaries, introducing unintended changes such as deforming the race-car front wheel or the airplane nose. In contrast, our method preserves global structure, avoids unwanted changes, and produces high-quality 3D edits.

\begin{figure}[!t]
\centering
\resizebox{\linewidth}{!}{
\begin{tabular}{cccccc}
\toprule
\centering  $O=\mathcal{O}_0$ &  $\mathcal{O}_{\frac{1}{5}M}$ &  $\mathcal{O}_{\frac{2}{5}M}$ &  $\mathcal{O}_{\frac{3}{5}M}$ &  $\mathcal{O}_{\frac{4}{5}M}$ &  $\tilde{O}=\mathcal{O}_M$  \\ 
\midrule
\rowC{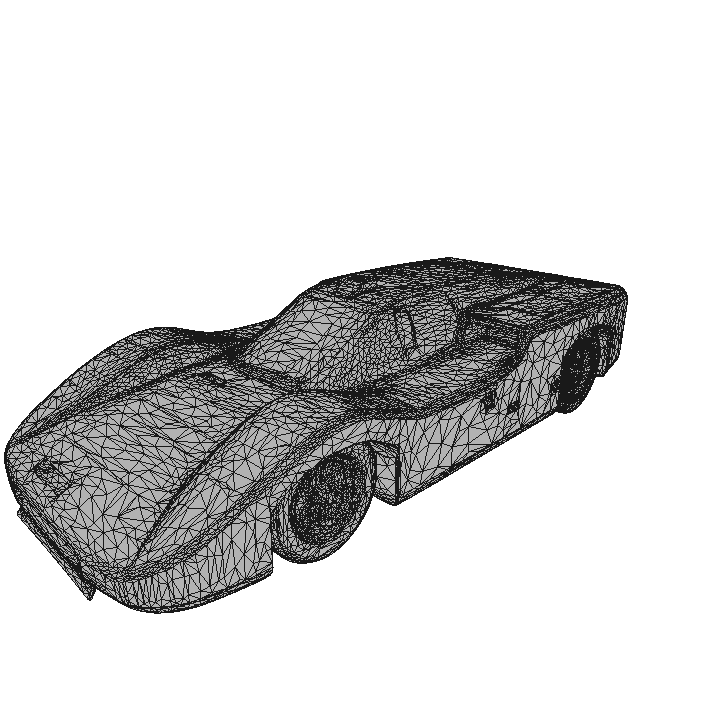}
& \rowC{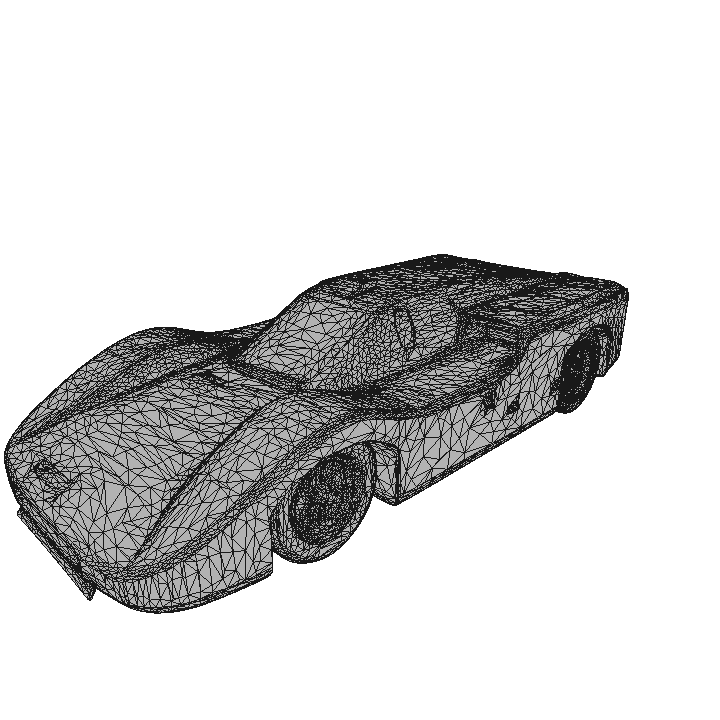}
& \rowC{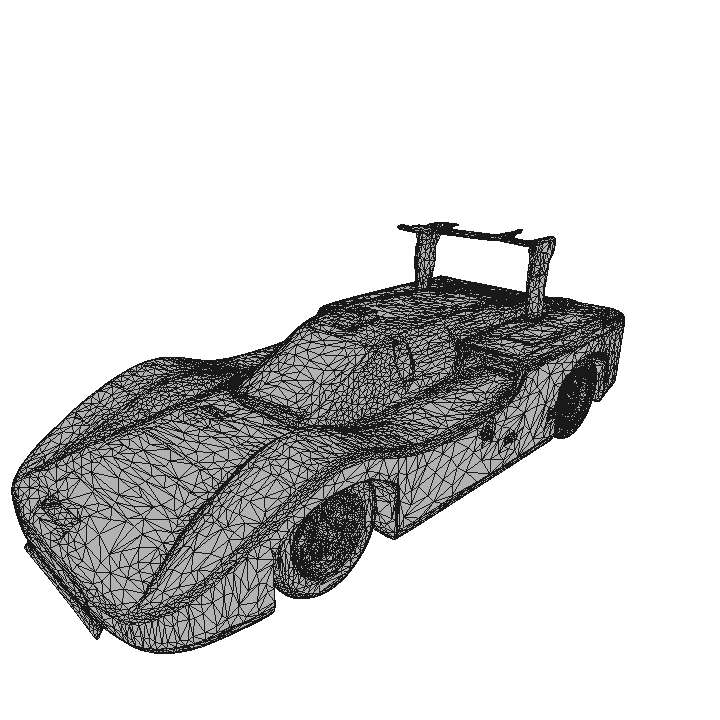}
& \rowC{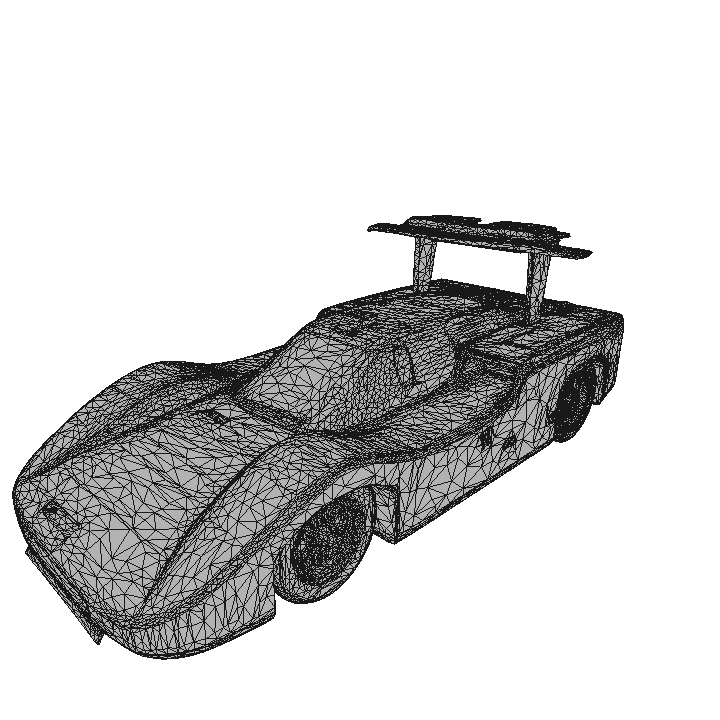}
& \rowC{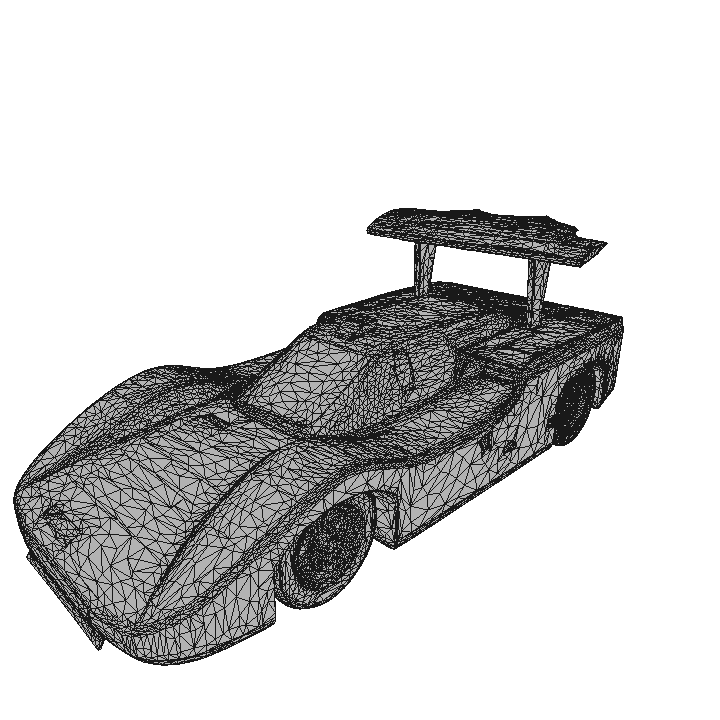}
& \rowC{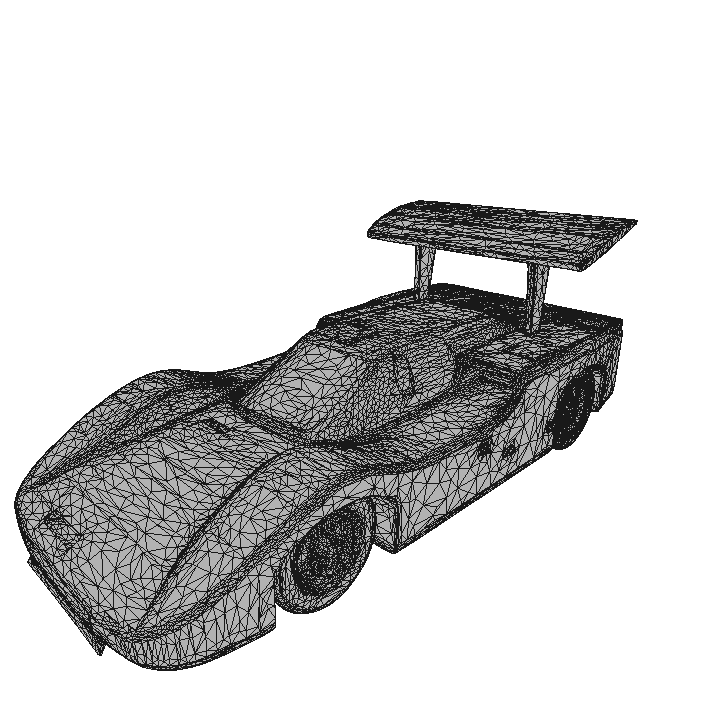} \\
\midrule
\rowB{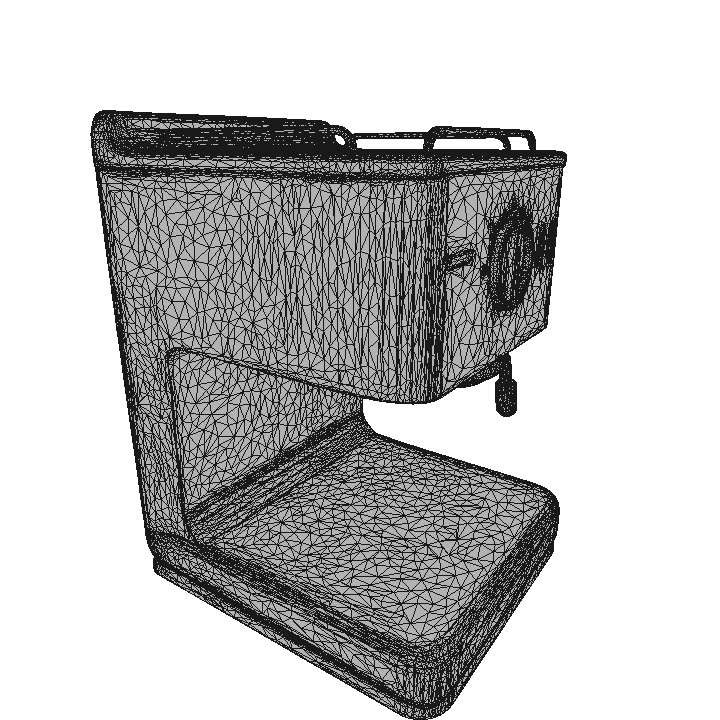}
& \rowB{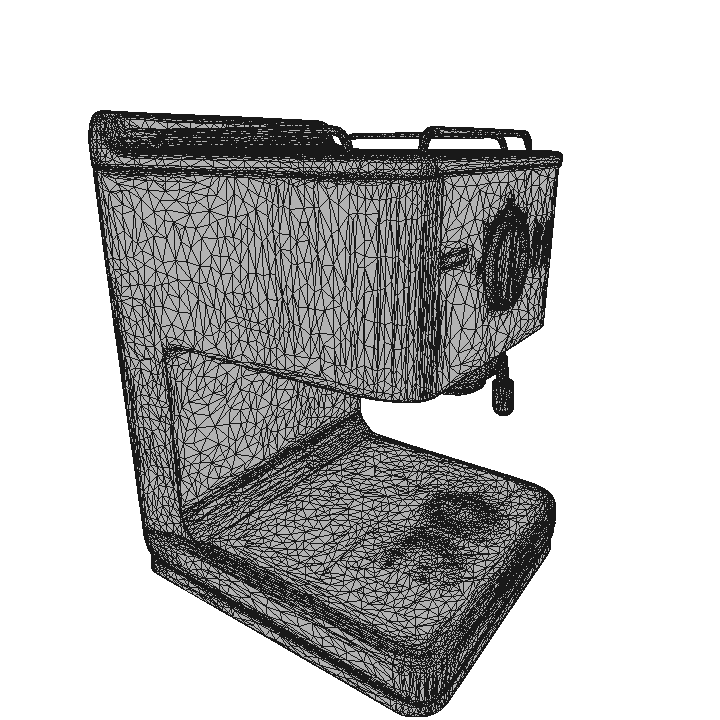}
& \rowB{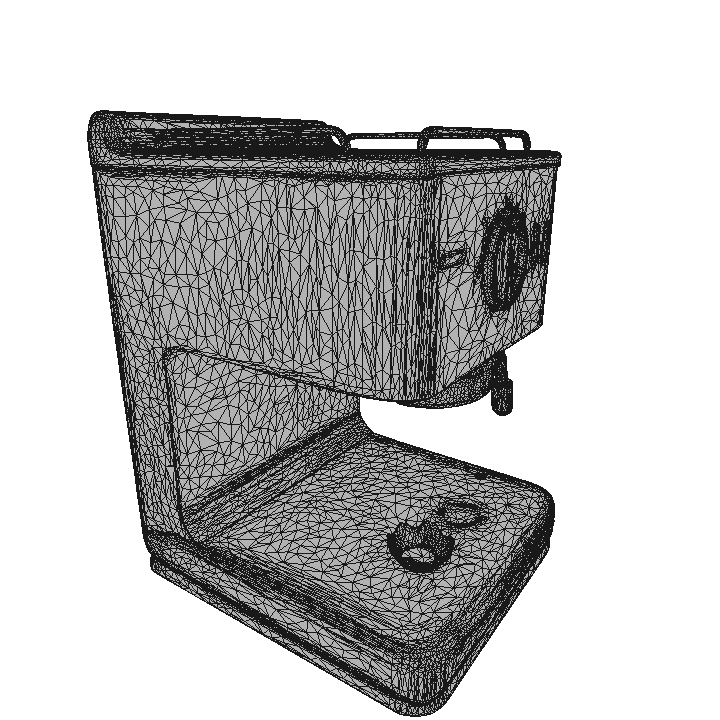}
& \rowB{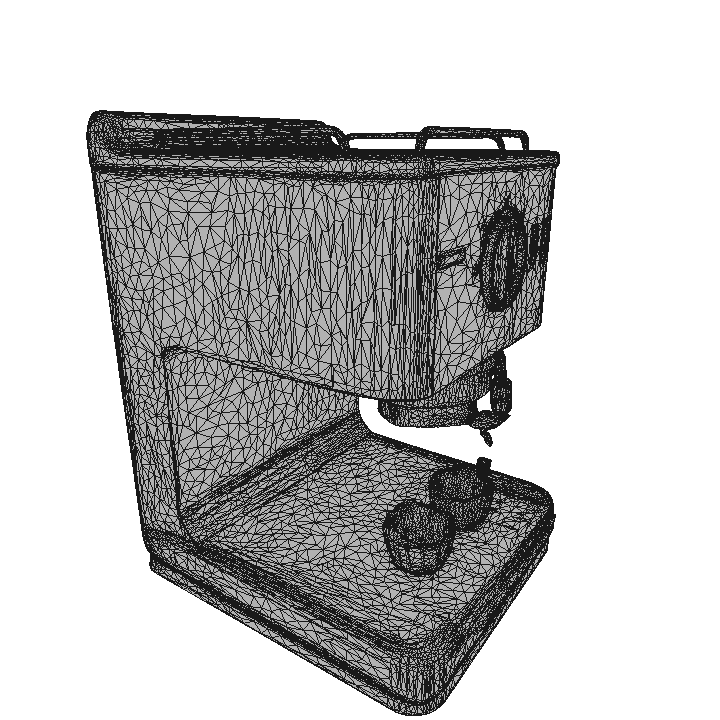}
& \rowB{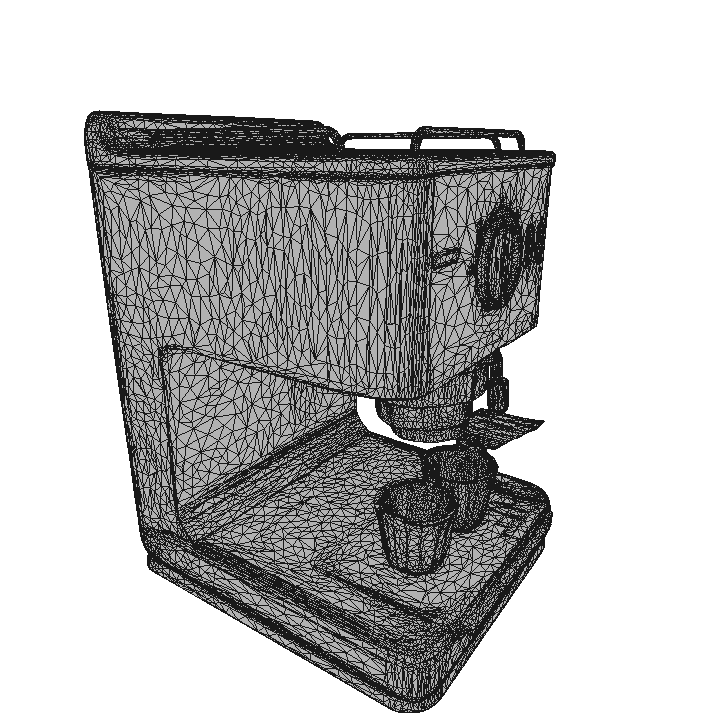}
& \rowB{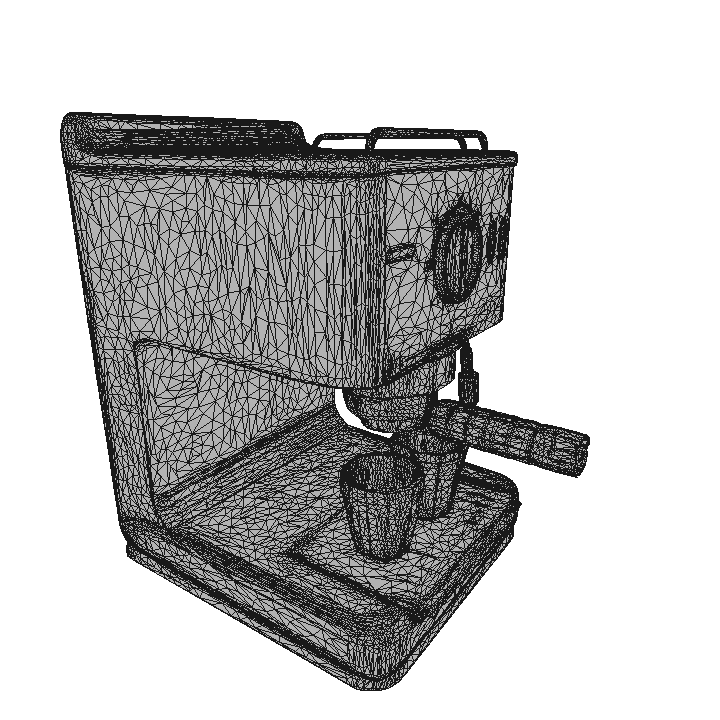} \\
\bottomrule
\end{tabular}
}
\caption{Morphing results. The left and right columns show the original object $O$ and the edited object $\tilde{O}$, respectively. The four middle columns visualize shapes at 20\%, 40\%, 60\%, and 80\% along a series of $M$ morphing steps.}
\label{fig:morphing}
\end{figure}

\textbf{Morphing results.} Fig.~\ref{fig:morphing} qualitatively illustrates local morphing results consisting of a series of $M$ objects $\{\mathcal{O}_i\}_{i=0}^{M}$. We show four intermediate shapes, along with the original object $O$ and the edited object $\tilde{O}$ as the first and last, respectively. For space efficiency, intermediate shapes are displayed at evenly spaced interpolation steps. 
Wireframe visualizations highlight our ability to generate high-quality intermediate 3D objects with changes confined to a single local region

\section{Conclusion}
In this paper, we present 3DMorph, a method for single-image-guided local 3D shape editing and morphing, operating within a confined region, preserving unaltered areas with high quality, and requiring only a single edited image to specify the desired change.
Moreover, we introduce Delta3D, the first geometric benchmark for this task with 3D editing ground truth, enabling geometric evaluation of the entire edited object.
%
Our quantitative and qualitative results demonstrate that our method outperforms state-of-the-art generative and editing approaches on local 3D editing tasks. 

%
%

A current limitation of this work lies in the bounding box prediction module, which can fail when an unfavorable viewing angle is chosen. However, in practical applications, users typically perform edits from an appropriate viewing angle to better express their intended modifications.
%
Moreover, the current level of fine-grained geometric detail is primarily determined by the resolution and capacity of the underlying SLAT decoder. This is a natural property of generative pipelines and is expected to improve as higher-resolution decoders become available, as demonstrated by recent work~\cite{xiang2025trellis2}.
%

In future work, we plan to expand Delta3D and further improve our 2D-to-3D bounding box prediction module, a crucial component of single-image editing.


\printbibliography

@String(CVPR= {IEEE Conf. Comput. Vis. Pattern Recog.})

@String(TOG= {ACM Trans. Graph.})

@String(AAAI = {AAAI})

@String(CVPR  = {CVPR})

@String(TOG   = {ACM TOG})

@article{comanici2025gemini,
  title={Gemini 2.5: Pushing the frontier with advanced reasoning, multimodality, long context, and next generation agentic capabilities},
  author={Comanici, Gheorghe and Bieber, Eric and Schaekermann, Mike and Pasupat, Ice and Sachdeva, Noveen and Dhillon, Inderjit and Blistein, Marcel and Ram, Ori and Zhang, Dan and Rosen, Evan and others},
  journal={arXiv preprint arXiv:2507.06261},
  year={2025}
}

@inproceedings{yi2023gaussiandreamer,
    title={GaussianDreamer: Fast Generation from Text to 3D Gaussians by Bridging 2D and 3D Diffusion Models},
    author={Yi, Taoran and Fang, Jiemin and Wang, Junjie and Wu, Guanjun and Xie, Lingxi and Zhang, Xiaopeng and Liu, Wenyu and Tian, Qi and Wang, Xinggang},
    year = {2024},
    booktitle = {CVPR}
}

@article{long2023wonder3d,
  title={Wonder3D: Single Image to 3D using Cross-Domain Diffusion},
  author={Long, Xiaoxiao and Guo, Yuan-Chen and Lin, Cheng and Liu, Yuan and Dou, Zhiyang and Liu, Lingjie and Ma, Yuexin and Zhang, Song-Hai and Habermann, Marc and Theobalt, Christian and others},
  journal={arXiv preprint arXiv:2310.15008},
  year={2023}
}

@INPROCEEDINGS{preintner2025why,
  author={Preintner, Tobias and Yuan, Weixuan and Huang, Qi and König, Adrian and Bäck, Thomas and Raponi, Elena and van Stein, Niki},
  booktitle={2025 International Joint Conference on Neural Networks (IJCNN)}, 
  title={Why Are You Wrong? Counterfactual Explanations for Language Grounding with 3D Objects}, 
  year={2025},
  doi={10.1109/IJCNN64981.2025.11227256}}

@article{huang2022ladis,
  title={LADIS: Language disentanglement for 3D shape editing},
  author={Huang, Ian and Achlioptas, Panos and Zhang, Tianyi and Tulyakov, Sergey and Sung, Minhyuk and Guibas, Leonidas},
  journal={arXiv preprint arXiv:2212.05011},
  year={2022}
}

@inproceedings{achlioptas2023shapetalk,
  title={ShapeTalk: A language dataset and framework for 3d shape edits and deformations},
  author={Achlioptas, Panos and Huang, Ian and Sung, Minhyuk and Tulyakov, Sergey and Guibas, Leonidas},
  booktitle={Proceedings of the IEEE/CVF conference on computer vision and pattern recognition},
  pages={12685--12694},
  year={2023}
}

@inproceedings{slim2024shapewalk,
  title={Shapewalk: Compositional shape editing through language-guided chains},
  author={Slim, Habib and Elhoseiny, Mohamed},
  booktitle={Proceedings of the IEEE/CVF Conference on Computer Vision and Pattern Recognition},
  pages={22574--22583},
  year={2024}
}

@inproceedings{chen2024shap,
  title={Shap-editor: Instruction-guided latent 3d editing in seconds},
  author={Chen, Minghao and Xie, Junyu and Laina, Iro and Vedaldi, Andrea},
  booktitle={Proceedings of the IEEE/CVF conference on computer vision and pattern recognition},
  pages={26456--26466},
  year={2024}
}

@inproceedings{gao2023textdeformer,
  title={Textdeformer: Geometry manipulation using text guidance},
  author={Gao, William and Aigerman, Noam and Groueix, Thibault and Kim, Vova and Hanocka, Rana},
  booktitle={ACM SIGGRAPH 2023 conference proceedings},
  pages={1--11},
  year={2023}
}

@inproceedings{sella2023voxe,
  title={Vox-e: Text-guided voxel editing of 3d objects},
  author={Sella, Etai and Fiebelman, Gal and Hedman, Peter and Averbuch-Elor, Hadar},
  booktitle={Proceedings of the IEEE/CVF international conference on computer vision},
  pages={430--440},
  year={2023}
}

@inproceedings{chen2025partgen,
  title={Partgen: Part-level 3d generation and reconstruction with multi-view diffusion models},
  author={Chen, Minghao and Shapovalov, Roman and Laina, Iro and Monnier, Tom and Wang, Jianyuan and Novotny, David and Vedaldi, Andrea},
  booktitle={Proceedings of the Computer Vision and Pattern Recognition Conference},
  pages={5881--5892},
  year={2025}
}

@inproceedings{li2024focaldreamer,
  title={Focaldreamer: Text-driven 3d editing via focal-fusion assembly},
  author={Li, Yuhan and Dou, Yishun and Shi, Yue and Lei, Yu and Chen, Xuanhong and Zhang, Yi and Zhou, Peng and Ni, Bingbing},
  booktitle={Proceedings of the AAAI conference on artificial intelligence},
  volume={38},
  number={4},
  pages={3279--3287},
  year={2024}
}

@article{li2025meshpad,
  title={MeshPad: Interactive Sketch-Conditioned Artist-Designed Mesh Generation and Editing},
  author={Li, Haoxuan and Erkoc, Ziya and Li, Lei and Sirigatti, Daniele and Rozov, Vladyslav and Dai, Angela and Nie{\ss}ner, Matthias},
  journal={arXiv preprint arXiv:2503.01425},
  year={2025}
}

@inproceedings{bandyopadhyay2024doodle,
  title={Doodle your 3d: From abstract freehand sketches to precise 3d shapes},
  author={Bandyopadhyay, Hmrishav and Koley, Subhadeep and Das, Ayan and Bhunia, Ayan Kumar and Sain, Aneeshan and Chowdhury, Pinaki Nath and Xiang, Tao and Song, Yi-Zhe},
  booktitle={Proceedings of the IEEE/CVF Conference on Computer Vision and Pattern Recognition},
  pages={9795--9805},
  year={2024}
}

@inproceedings{binninger2024sens,
  title={SENS: Part-Aware Sketch-based Implicit Neural Shape Modeling},
  author={Binninger, Alexandre and Hertz, Amir and Sorkine-Hornung, Olga and Cohen-Or, Daniel and Giryes, Raja},
  booktitle={Computer Graphics Forum},
  volume={43},
  number={2},
  pages={e15015},
  year={2024},
  organization={Wiley Online Library}
}

@article{hertz2022spaghetti,
  title={Spaghetti: Editing implicit shapes through part aware generation},
  author={Hertz, Amir and Perel, Or and Giryes, Raja and Sorkine-Hornung, Olga and Cohen-Or, Daniel},
  journal={ACM Transactions on Graphics (TOG)},
  volume={41},
  number={4},
  pages={1--20},
  year={2022},
  publisher={ACM New York, NY, USA}
}

@inproceedings{koo2023salad,
  title={Salad: Part-level latent diffusion for 3d shape generation and manipulation},
  author={Koo, Juil and Yoo, Seungwoo and Nguyen, Minh Hieu and Sung, Minhyuk},
  booktitle={Proceedings of the IEEE/CVF International Conference on Computer Vision},
  pages={14441--14451},
  year={2023}
}

@ARTICLE{3DFaceSculptor,
author={Su, Hao and Wang, Xuxi and Niu, Jianwei and Liu, Xuefeng and Wu, Xinghao and Wang, Nana},
journal={ IEEE Transactions on Visualization \& Computer Graphics },
title={{ 3DFaceSculptor: A Common Framework for Image-Guided 3D Face Deformation }},
year={5555},
volume={},
number={01},
ISSN={1941-0506},
pages={1-18},
doi={10.1109/TVCG.2025.3596482},
url = {https://doi.ieeecomputersociety.org/10.1109/TVCG.2025.3596482},
publisher={IEEE Computer Society},
address={Los Alamitos, CA, USA},
month=aug}

@misc{gao20243dmeshediting,
      title={3D Mesh Editing using Masked LRMs}, 
      author={Will Gao and Dilin Wang and Yuchen Fan and Aljaz Bozic and Tuur Stuyck and Zhengqin Li and Zhao Dong and Rakesh Ranjan and Nikolaos Sarafianos},
      year={2024},
      eprint={2412.08641},
      archivePrefix={arXiv},
      primaryClass={cs.CV},
      url={https://arxiv.org/abs/2412.08641}, 
}

@inproceedings{li2025cmd,
  title={CMD: Controllable Multiview Diffusion for 3D Editing and Progressive Generation},
  author={Li, Peng and Ma, Suizhi and Chen, Jialiang and Liu, Yuan and Zhang, Congyi and Xue, Wei and Luo, Wenhan and Sheffer, Alla and Wang, Wenping and Guo, Yike},
  booktitle={Proceedings of the Special Interest Group on Computer Graphics and Interactive Techniques Conference Conference Papers},
  pages={1--10},
  year={2025}
}

@article{li2025voxhammer,
  title={VoxHammer: Training-Free Precise and Coherent 3D Editing in Native 3D Space},
  author={Li, Lin and Huang, Zehuan and Feng, Haoran and Zhuang, Gengxiong and Chen, Rui and Guo, Chunchao and Sheng, Lu},
  journal={arXiv preprint arXiv:2508.19247},
  year={2025}
}

@article{wirth2011continuum,
  title={A continuum mechanical approach to geodesics in shape space},
  author={Wirth, Benedikt and Bar, Leah and Rumpf, Martin and Sapiro, Guillermo},
  journal={International Journal of Computer Vision},
  volume={93},
  number={3},
  pages={293--318},
  year={2011},
  publisher={Springer}
}

@inproceedings{brandt2016geometric,
  title={Geometric flows of curves in shape space for processing motion of deformable objects},
  author={Brandt, Christopher and von Tycowicz, Christoph and Hildebrandt, Klaus},
  booktitle={Computer Graphics Forum},
  volume={35},
  number={2},
  pages={295--305},
  year={2016},
  organization={Wiley Online Library}
}

@inproceedings{heeren2016splines,
  title={Splines in the space of shells},
  author={Heeren, Behrend and Rumpf, Martin and Schr{\"o}der, Peter and Wardetzky, Max and Wirth, Benedikt},
  booktitle={Computer Graphics Forum},
  volume={35},
  number={5},
  pages={111--120},
  year={2016},
  organization={Wiley Online Library}
}

@inproceedings{eisenberger2020hamiltonian,
  title={Hamiltonian dynamics for real-world shape interpolation},
  author={Eisenberger, Marvin and Cremers, Daniel},
  booktitle={European conference on computer vision},
  pages={179--196},
  year={2020},
  organization={Springer}
}

@inproceedings{sun2024srif,
  title={Srif: Semantic shape registration empowered by diffusion-based image morphing and flow estimation},
  author={Sun, Mingze and Guo, Chen and Jiang, Puhua and Mao, Shiwei and Chen, Yurun and Huang, Ruqi},
  booktitle={SIGGRAPH Asia 2024 Conference Papers},
  pages={1--11},
  year={2024}
}

@inproceedings{eisenberger2021neuromorph,
  title={Neuromorph: Unsupervised shape interpolation and correspondence in one go},
  author={Eisenberger, Marvin and Novotny, David and Kerchenbaum, Gael and Labatut, Patrick and Neverova, Natalia and Cremers, Daniel and Vedaldi, Andrea},
  booktitle={Proceedings of the IEEE/CVF Conference on Computer Vision and Pattern Recognition},
  pages={7473--7483},
  year={2021}
}

@inproceedings{sang20254deform,
  title={4Deform: Neural Surface Deformation for Robust Shape Interpolation},
  author={Sang, Lu and Canfes, Zehranaz and Cao, Dongliang and Marin, Riccardo and Bernard, Florian and Cremers, Daniel},
  booktitle={Proceedings of the Computer Vision and Pattern Recognition Conference},
  pages={6542--6551},
  year={2025}
}

@inproceedings{xiang2025trellis,
  title={Structured 3d latents for scalable and versatile 3d generation},
  author={Xiang, Jianfeng and Lv, Zelong and Xu, Sicheng and Deng, Yu and Wang, Ruicheng and Zhang, Bowen and Chen, Dong and Tong, Xin and Yang, Jiaolong},
  booktitle={Proceedings of the Computer Vision and Pattern Recognition Conference},
  pages={21469--21480},
  year={2025}
}

@article{xiang2025trellis2,
  title={Native and Compact Structured Latents for 3D Generation},
  author={Xiang, Jianfeng and Chen, Xiaoxue and Xu, Sicheng and Wang, Ruicheng and Lv, Zelong and Deng, Yu and Zhu, Hongyuan and Dong, Yue and Zhao, Hao and Yuan, Nicholas Jing and others},
  journal={arXiv preprint arXiv:2512.14692},
  year={2025}
}

@article{wu2025amodal3r,
  title={Amodal3r: Amodal 3d reconstruction from occluded 2d images},
  author={Wu, Tianhao and Zheng, Chuanxia and Guan, Frank and Vedaldi, Andrea and Cham, Tat-Jen},
  journal={arXiv preprint arXiv:2503.13439},
  year={2025}
}

@inproceedings{willis2022assembly,
  title={Joinable: Learning bottom-up assembly of parametric cad joints},
  author={Willis, Karl DD and Jayaraman, Pradeep Kumar and Chu, Hang and Tian, Yunsheng and Li, Yifei and Grandi, Daniele and Sanghi, Aditya and Tran, Linh and Lambourne, Joseph G and Solar-Lezama, Armando and others},
  booktitle={Proceedings of the IEEE/CVF conference on computer vision and pattern recognition},
  pages={15849--15860},
  year={2022}
}

@misc{hunyuan3d2025hunyuan3d,
    title={Hunyuan3D 2.1: From Images to High-Fidelity 3D Assets with Production-Ready PBR Material},
    author={Tencent Hunyuan3D Team},
    year={2025},
    eprint={2506.15442},
    archivePrefix={arXiv},
    primaryClass={cs.CV}
}

@inproceedings{lugmayr2022repaint,
  title={Repaint: Inpainting using denoising diffusion probabilistic models},
  author={Lugmayr, Andreas and Danelljan, Martin and Romero, Andres and Yu, Fisher and Timofte, Radu and Van Gool, Luc},
  booktitle={Proceedings of the IEEE/CVF conference on computer vision and pattern recognition},
  pages={11461--11471},
  year={2022}
}

@article{li2025triposg,
  title={Triposg: High-fidelity 3d shape synthesis using large-scale rectified flow models},
  author={Li, Yangguang and Zou, Zi-Xin and Liu, Zexiang and Wang, Dehu and Liang, Yuan and Yu, Zhipeng and Liu, Xingchao and Guo, Yuan-Chen and Liang, Ding and Ouyang, Wanli and others},
  journal={arXiv preprint arXiv:2502.06608},
  year={2025}
}

@misc{flux2024,
    author={Black Forest Labs},
    title={FLUX},
    year={2024},
    howpublished={\url{https://github.com/black-forest-labs/flux}},
}

@InProceedings{Rombach_2022_CVPR,
    author    = {Rombach, Robin and Blattmann, Andreas and Lorenz, Dominik and Esser, Patrick and Ommer, Bj\"orn},
    title     = {High-Resolution Image Synthesis With Latent Diffusion Models},
    booktitle = {Proceedings of the IEEE/CVF Conference on Computer Vision and Pattern Recognition (CVPR)},
    month     = {June},
    year      = {2022},
    pages     = {10684-10695}
}

@inproceedings{saharia2022palette,
  title={Palette: Image-to-image diffusion models},
  author={Saharia, Chitwan and Chan, William and Chang, Huiwen and Lee, Chris and Ho, Jonathan and Salimans, Tim and Fleet, David and Norouzi, Mohammad},
  booktitle={ACM SIGGRAPH 2022 conference proceedings},
  pages={1--10},
  year={2022}
}

@inproceedings{yang2023paint,
  title={Paint by example: Exemplar-based image editing with diffusion models},
  author={Yang, Binxin and Gu, Shuyang and Zhang, Bo and Zhang, Ting and Chen, Xuejin and Sun, Xiaoyan and Chen, Dong and Wen, Fang},
  booktitle={Proceedings of the IEEE/CVF conference on computer vision and pattern recognition},
  pages={18381--18391},
  year={2023}
}

@inproceedings{zhang2023controlnet,
  title={Adding conditional control to text-to-image diffusion models},
  author={Zhang, Lvmin and Rao, Anyi and Agrawala, Maneesh},
  booktitle={Proceedings of the IEEE/CVF international conference on computer vision},
  pages={3836--3847},
  year={2023}
}

@inproceedings{mueller2025geodiffusion,
  title={GeoDiffusion: A Training-Free Framework for Accurate 3D Geometric Conditioning in Image Generation},
  author={Mueller, Phillip and Uenlue, Talip and Schmidt, Sebastian and Kollovieh, Marcel and Fan, Jiajie and G{\"u}nnemann, Stephan and Mikelsons, Lars},
  booktitle={Proceedings of the IEEE/CVF International Conference on Computer Vision},
  pages={6374--6384},
  year={2025}
}

@article{xia2025towardsscalable,
  title={Towards Scalable and Consistent 3D Editing},
  author={Xia, Ruihao and Tang, Yang and Zhou, Pan},
  journal={arXiv preprint arXiv:2510.02994},
  year={2025}
}

@article{ye2025nano3d,
  title={NANO3D: A Training-Free Approach for Efficient 3D Editing Without Masks},
  author={Ye, Junliang and Xie, Shenghao and Zhao, Ruowen and Wang, Zhengyi and Yan, Hongyu and Zu, Wenqiang and Ma, Lei and Zhu, Jun},
  journal={arXiv preprint arXiv:2510.15019},
  year={2025}
}

@article{cai2025native,
  title={Native 3D Editing with Full Attention},
  author={Cai, Weiwei and Fang, Shuangkang and Ye, Weicai and Dong, Xin and Yang, Yunhan and Zhang, Xuanyang and Cheng, Wei and Cao, Yanpei and Yu, Gang and Chen, Tao},
  journal={arXiv preprint arXiv:2511.17501},
  year={2025}
}

@article{zhou2025anchorflow,
  title={AnchorFlow: Training-Free 3D Editing via Latent Anchor-Aligned Flows},
  author={Zhou, Zhenglin and Ma, Fan and Gui, Chengzhuo and Xia, Xiaobo and Fan, Hehe and Yang, Yi and Chua, Tat-Seng},
  journal={arXiv preprint arXiv:2511.22357},
  year={2025}
}


\clearpage
\setcounter{page}{1}
\let\twocolumn\oldtwocolumn

\setcounter{section}{0}
\renewcommand{\thesection}{\Alph{section}}
\renewcommand{\thesubsection}{\thesection.\arabic{subsection}}


\section{Extended Methodology Details}
This section provides additional methodological details on the bounding box prediction module introduced in Sec.~\ref{subsec:BB-pred} and the local morphing method described in Sec.~\ref{Morphing}.

\subsection{Bounding Box Prediction}\label{suppl:BB-pred}
Algorithm \ref{alg:BB-pred} describes our 3D bounding box prediction method in detail. First structural differences between the image $I$ and $\tilde{I}$ are detected by using SSIM, and connected contours are extracted (line~1). 
For contours with an area larger than $A_{min}$ 2D bounding boxes are created (line~2). 
To establish correspondence with the 3D object, we compute the center of each voxel and project it onto the image plane to obtain the image coordinates $U$ and corresponding depth $Z$ (line~3-4), from which we estimate a coarse visibility mask that retains only the front-most 3D point per image bin (line~5). 
For each 2D box center $(c_x, c_y)$ the nearest projected visible voxel center is identified to define 3D anchor $P_c$, representing the center of the corresponding 3D bounding box (line~9).
The box extents $L$ and $H$ are estimated by projecting the 2D box into 3D using the camera parameters, the 2D box width $w$ and height $h$, and a pixel-to-metric scale factor $\alpha$, derived from the top view, which determines how much a pixel corresponds to in the 3D object space (line~10).
Using this information, we compute a 3D axis-aligned bounding box and add it to a set $\mathcal{B}$ (line~11). The fusion of all boxes in $\mathcal{B}$ produces the final 3D box $BB$ enclosing the to be edit region (line~13).

\SetKwComment{Comment}{\#}{}
\SetKwInput{KwData}{Given}
\RestyleAlgo{ruled} 
\LinesNumbered

\begin{algorithm}[h]
\SetAlgoLined
\SetKwInOut{Input}{Input}\SetKwInOut{Output}{Output}
\caption{Extract 3D Bounding Box from 2D Changes}
\label{alg:BB-pred}
\Input{Original object $O$, rendered image $I$, modified image $\tilde{I}$, camera $(K,E,\text{yaw},\text{pitch})$, thresholds $(\tau,A_{\min})$}
\Output{3D bounding boxes $\mathcal{B}$}
\BlankLine
\tcp{2D difference boxes}
$\mathcal{C} \gets \mathrm{CC}(\mathrm{SSIM\_Diff}(I,\tilde{I},\tau))$\;
$\mathcal{S} \gets \{\,\mathrm{Box}(c)\mid c\in \mathcal{C},\ \mathrm{area}(c)\ge A_{\min}\,\}$\;
\BlankLine

\tcp{Get 2D-3D pairs}
$P \gets \mathrm{voxelCenters}(O)$ \tcp*[r]{$P\in[-0.5,0.5]^{N \times 3}$}
$(U, Z)\gets \mathrm{Project}(P,K,E)$ \tcp*[r]{$U\in[0,1]^2$}
$\mathrm{Vis} \gets \mathrm{visibilityMask}(U,Z)$\; 
\BlankLine

\tcp{Estimate 3D boxes}
$\mathcal{B} \gets \varnothing$ \;
$\alpha \gets \mathrm{PixelMetricScale}(O)$\;
\ForEach{$(c_x,c_y,w,h)\in\mathcal{S}$}{
  $P_c \gets getNN(c_x, c_y, U, Vis)$\;
  $L \gets \alpha\,\big|\tfrac{w}{\sin(\text{yaw})}\big|$; \quad $H \gets \alpha\,\big|\tfrac{h}{\cos(\text{pitch})}\big|$\;
    $\mathcal{B} \gets \mathcal{B} \cup \mathrm{AABB}(P_c, L, H)$ \;

}
\tcp{Fuse multiple boxes into one}
$ BB \gets \mathrm{AABB}(\mathcal{B})$\;
\Return{$BB$} \;
\end{algorithm}




\subsection{Local 3D Morphing}\label{suppl:morphing}
Algorithm \ref{alg:morph} outlines our morphing method in detail. First the mutual difference $\Delta$ and $\tilde{\Delta}$ between the two object $O$ and $\tilde{O}$ are calculated in voxel space (line~1). After calculating the nearest neighbor distance for each voxel in $\Delta$ and $\tilde{\Delta}$ to the respectively excluded object (line~2-3), these distances are used to subdivide $\Delta$ and $\tilde{\Delta}$ into $M$ voxel subsets depending on their distance (line~4-5). These two subset determine the new to be added voxels $add$ from $\tilde{O}$ as well as the to be removed voxels $remove$ from $O$ (line~7). Together with the to be keep and intersecting voxels, the final set of voxel $V_m$ for step $m$ determined (line~8-10). To get the final step latents $\mathcal{Z}_m$ a step-depending linear interpolation is performed, if a voxel belongs mutually to $O$ or $\tilde{O}$ the original latent is kept (line~11-12). Together with the voxels the final morphing step object $\mathcal{O}_m$ is decoded (line~13). This procedure generates $M-1$ intermediate morphing objects between the original object $O=\mathcal{O}_0$ and the edited object $\tilde{O}=\mathcal{O}_M$.

\SetKwComment{Comment}{\#}{}
\SetKwInput{KwData}{Given}
\RestyleAlgo{ruled} 
\LinesNumbered

\begin{algorithm}[h]
\caption{Local 3D Morphing}
\label{alg:morph}
\SetAlgoLined
\SetKwInOut{Input}{Input}\SetKwInOut{Output}{Output}
\Input{Original object voxels $V$ and latents $\mathcal{Z}$, edited object voxels $\tilde{V}$ and latents $\tilde{\mathcal{Z}}$, number of morphing steps $M$}
\Output{Morphing objects $\{\mathcal{O}_i\}_{i=0}^M$}
\BlankLine
$\tilde{\Delta}  \gets \tilde{V} \setminus V$; \quad $\Delta  \gets V \setminus \tilde{V}$\;
$d_{\tilde{\Delta} \to V} \leftarrow \mathrm{NNdist}(\tilde{\Delta},V)$\;
$d_{\Delta \to \tilde{V}} \leftarrow \mathrm{NNdist}(\Delta ,\tilde{V})$\;
$\{a_i\}_{i=1}^M \leftarrow \mathrm{slice}(\tilde{\Delta},d_{\tilde{\Delta} \to V},N)$\; 
$\{r_i\}_{i=1}^M \leftarrow  \mathrm{slice}(\Delta,d_{\Delta \to \tilde{V}},N)$\;
\BlankLine
\For{$m=0$ \KwTo $M$}{ 
$add \leftarrow \{a_i\}_{i=1}^m$; \quad $remove \leftarrow \{r_i\}_{i=1}^m$\; 
$keep \leftarrow \Delta \setminus remove$\;
$intersect \leftarrow V \cap \tilde{V}$\; 
$V_m \leftarrow intersect \cup keep \cup add$\;
\BlankLine
$\alpha \leftarrow m/M$\;
$\mathcal{Z}_m \leftarrow (1-\alpha)\,\mathcal{Z} + \alpha\,\tilde{\mathcal{Z}}$\;
\BlankLine
$\mathcal{O}_m \leftarrow \mathrm{decode}(V_m,\mathcal{Z}_m)$\;
}
\Return $\{\mathcal{O}_i\}_{i=0}^M $\;
\end{algorithm}

\subsection{Trellis vs. 3DMorph Architectures}\label{suppl:trellis} 
\begin{figure*}[htb]
\centering
    \includegraphics[width=\linewidth]{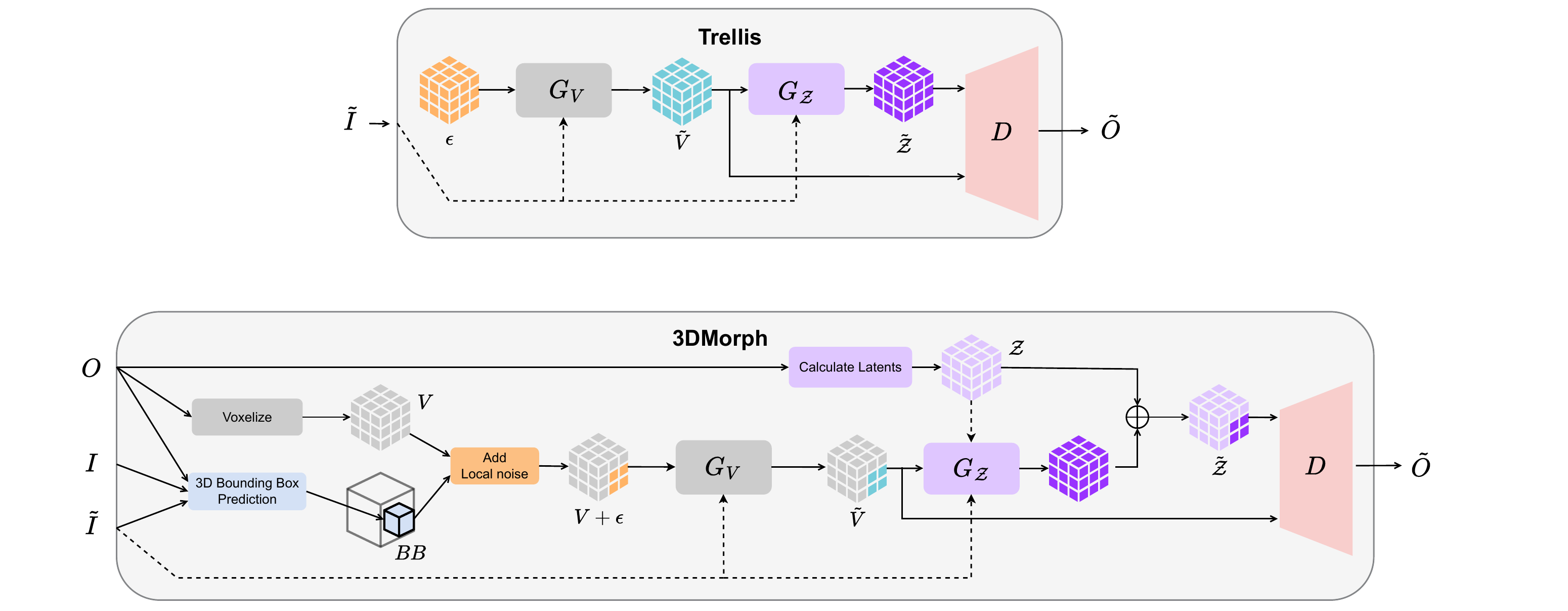}%
    \caption{Architectural differences between Trellis~\cite{xiang2025trellis} and 3DMorph. While Trellis uses SLAT to generate shapes from scratch, 3DMorph employs SLAT for local edits, reusing information from the original object and thereby better preserving global structure and fine details.\label{fig:trellis}}
\end{figure*}

Fig.~\ref{fig:trellis} illustrates the architectural differences between vanilla Trellis~\cite{xiang2025trellis} and 3DMorph. While Trellis generates an object entirely from scratch by denoising a noise-initialized volume, 3DMorph injects noise only within the predicted edit region.

\begin{figure*}[t!]
\centering
    \includegraphics[width= \textwidth, trim=0pt 0pt 0pt 0pt, clip]{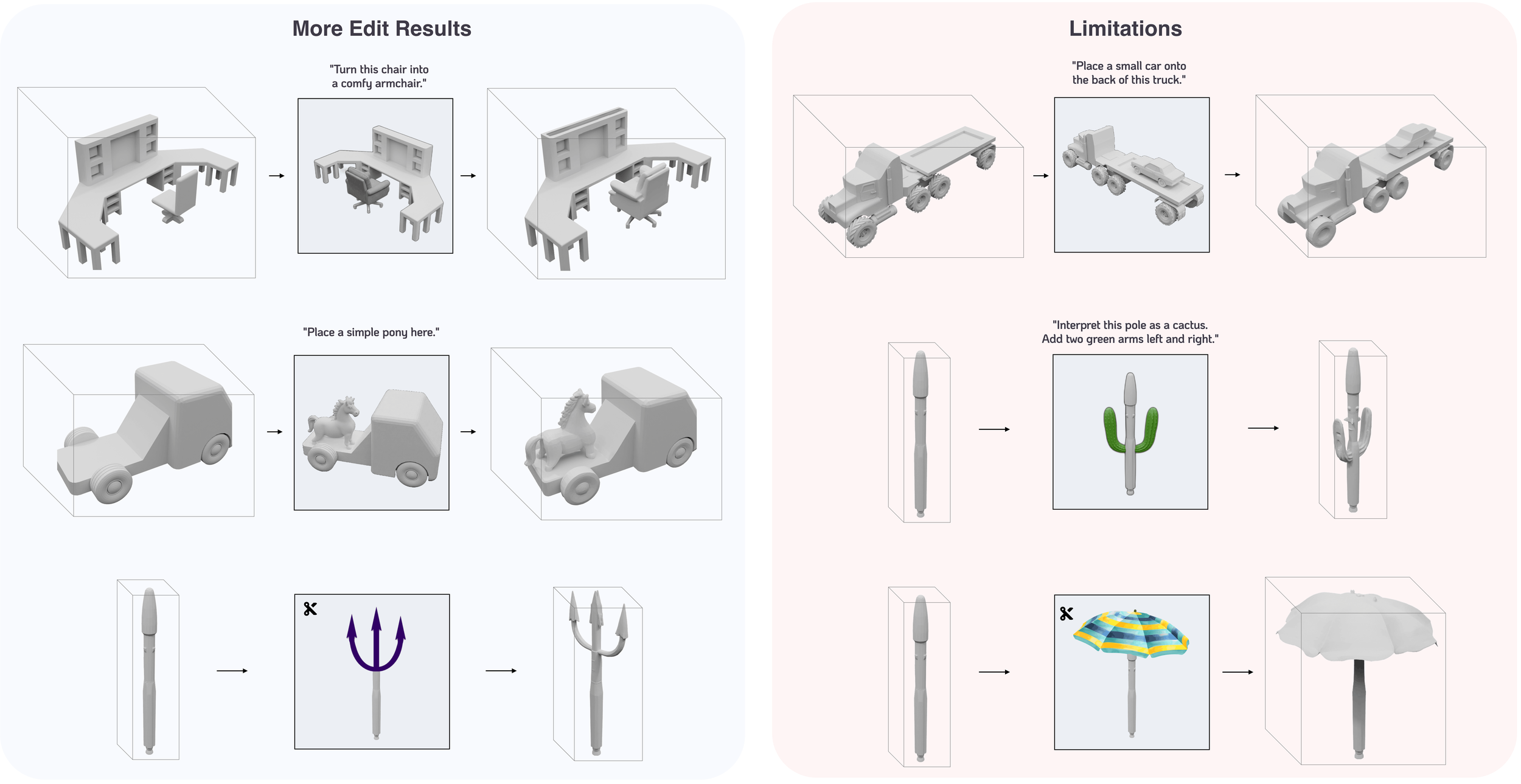}%
    \caption{More editing results obtained using deep or manual inpainting. Limitation cases (right) primarily stem from the finite resolution of SLAT and the constraints of the learned domain. \label{suppl-fig:applied-results}}
\end{figure*}


\newpage
\section{More Qualitative Results} 
\textbf{Edit results and limitations.} Fig.~\ref{suppl-fig:applied-results} shows additional edit results produced using either deep or manual inpainting via a crop-and-paste operation. 

The left side presents successful cases, while the right side highlights previously discussed limitations. Some failures arise from constraints of the SLAT representation: high-fidelity details (e.g., truck wheels) exceed its resolution, and edits involving disjoint regions (e.g., the cactus) cannot preserve intermediate geometry due to the single-box constraint. Edit quality is also limited by the capability and training domain of the SLAT predictor (e.g., the umbrella).


\textbf{More benchmark results.} Fig.~\ref{suppl-fig:diff-suppl} complements Fig.~\ref{fig:diff-main} with additional qualitative comparisons on the Delta3D benchmark, showing that 3DMorph preserves global structure, avoids unintended modifications, and produces high-quality 3D edits.

\newcommand{\imagecolumnwidthsuppl}{2.5 cm}

\begin{figure*}[t]
\centering
\begin{tabular}{|m{2.2cm}|M{\imagecolumnwidthsuppl}|M{\imagecolumnwidthsuppl}|M{\imagecolumnwidthsuppl}|M{\imagecolumnwidthsuppl}|M{\imagecolumnwidthsuppl}|}
\hline
\centering \textbf{Method} & \textbf{Car} & \textbf{Bike} & \textbf{Chess Figures} & \textbf{F1 Car} & \textbf{Helicopter} \\
\hline

\centering \shortstack{Original \\ object $O$}
& \supplColA{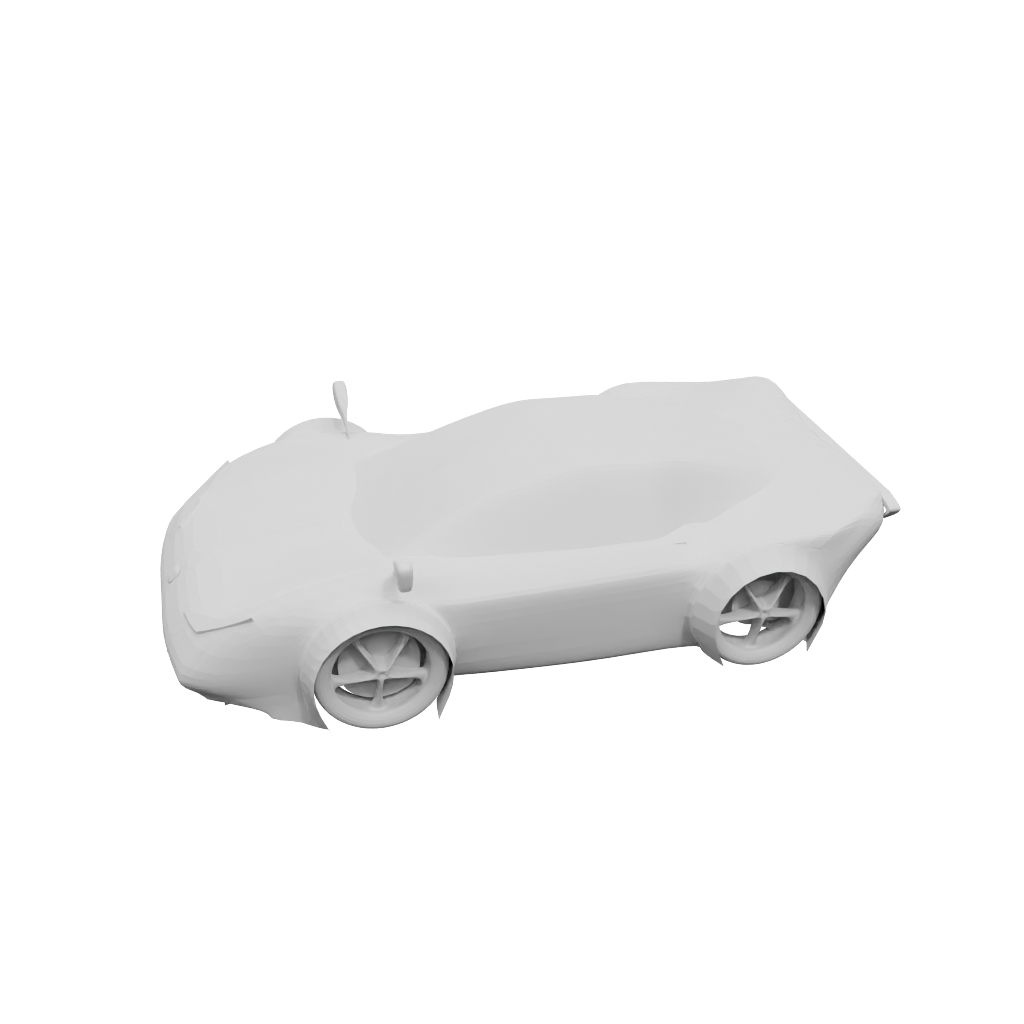}
& \supplColB{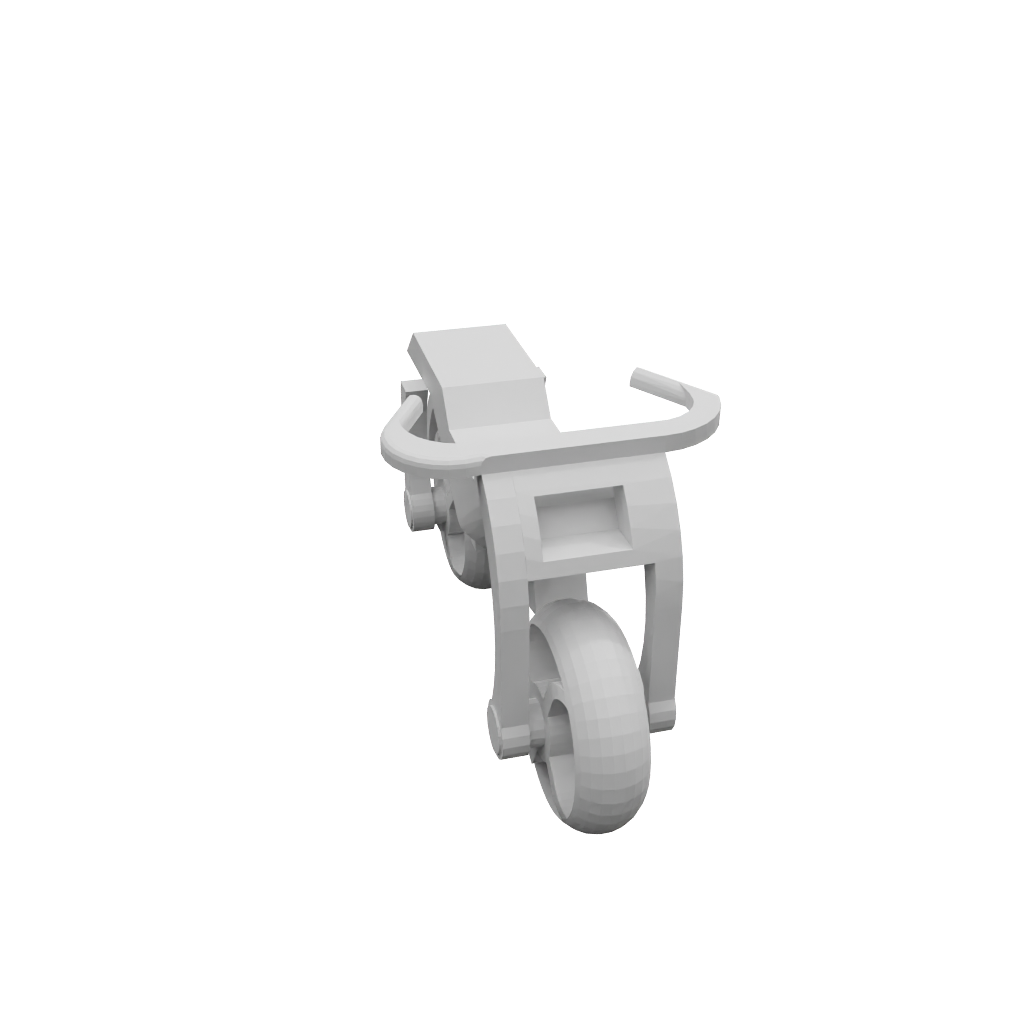}
& \supplColC{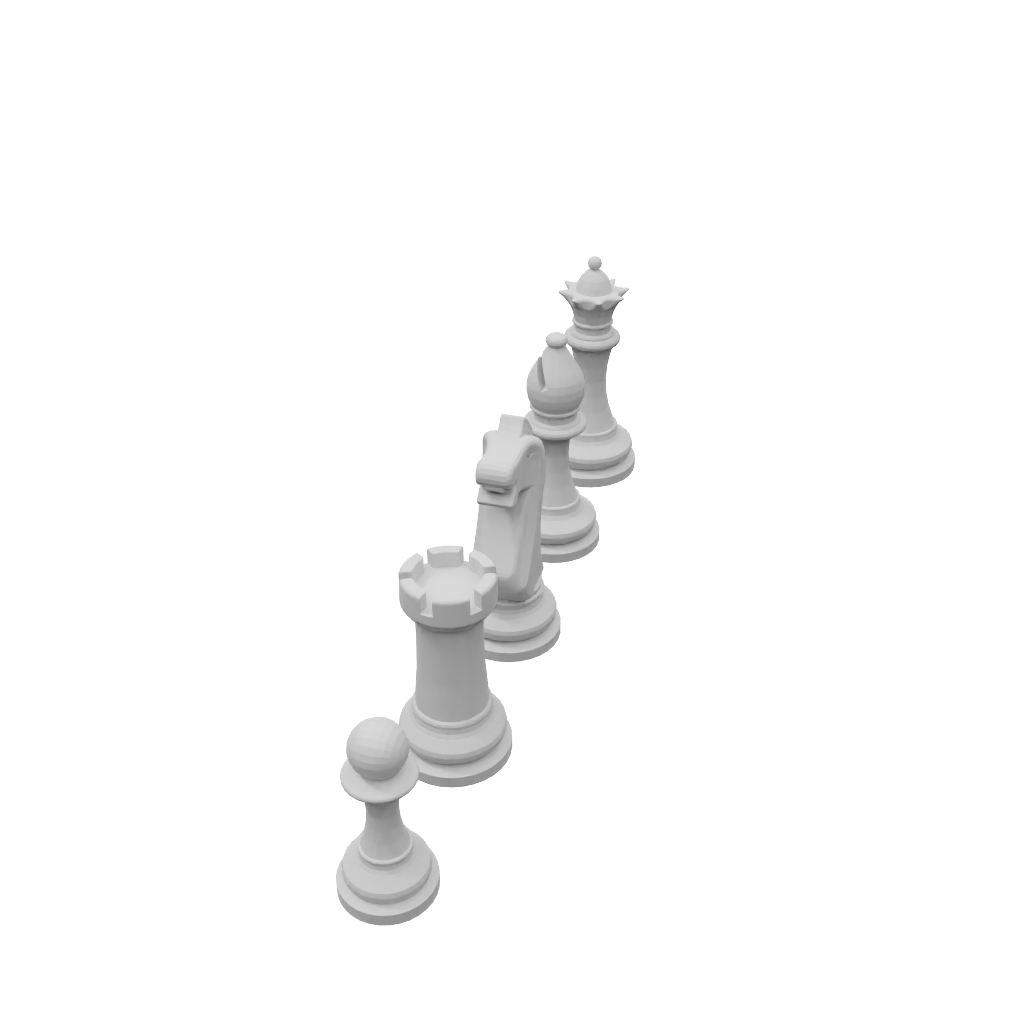}
& \supplColD{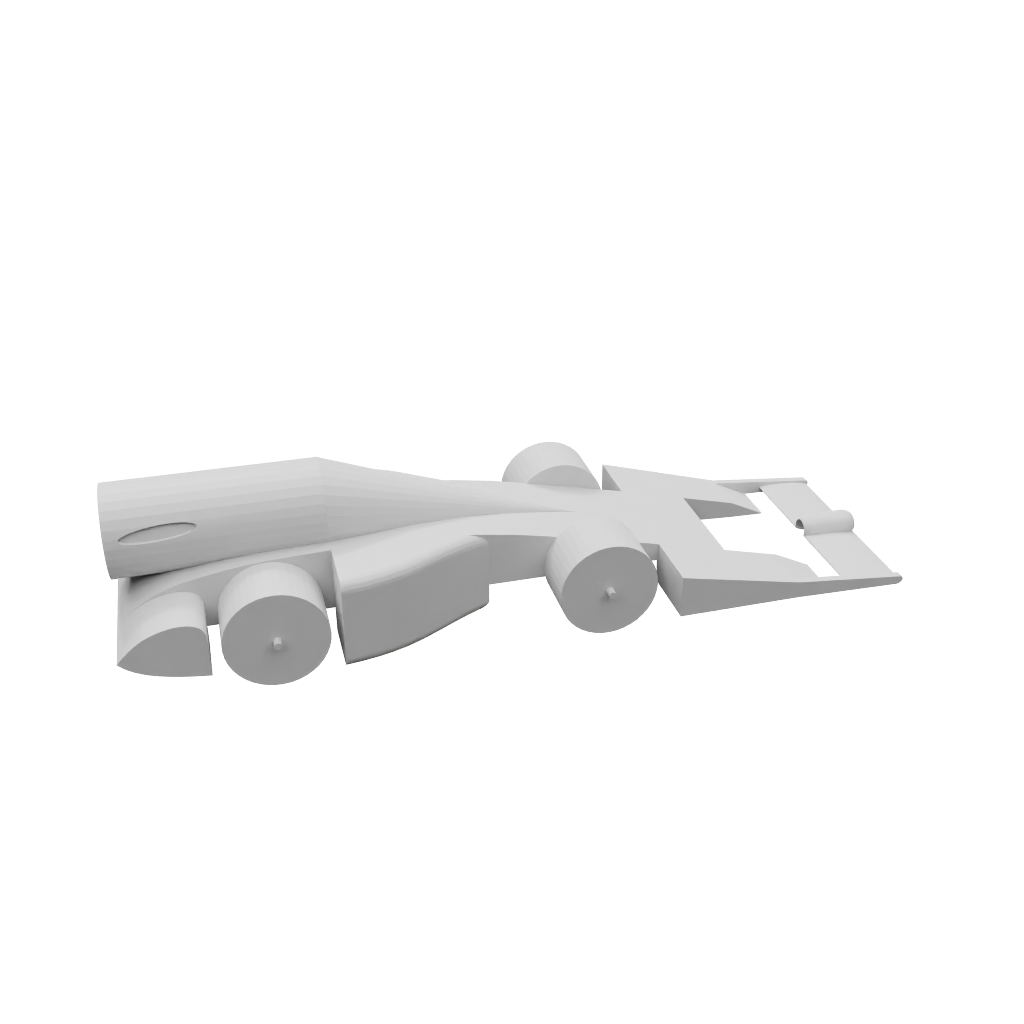}
& \supplColE{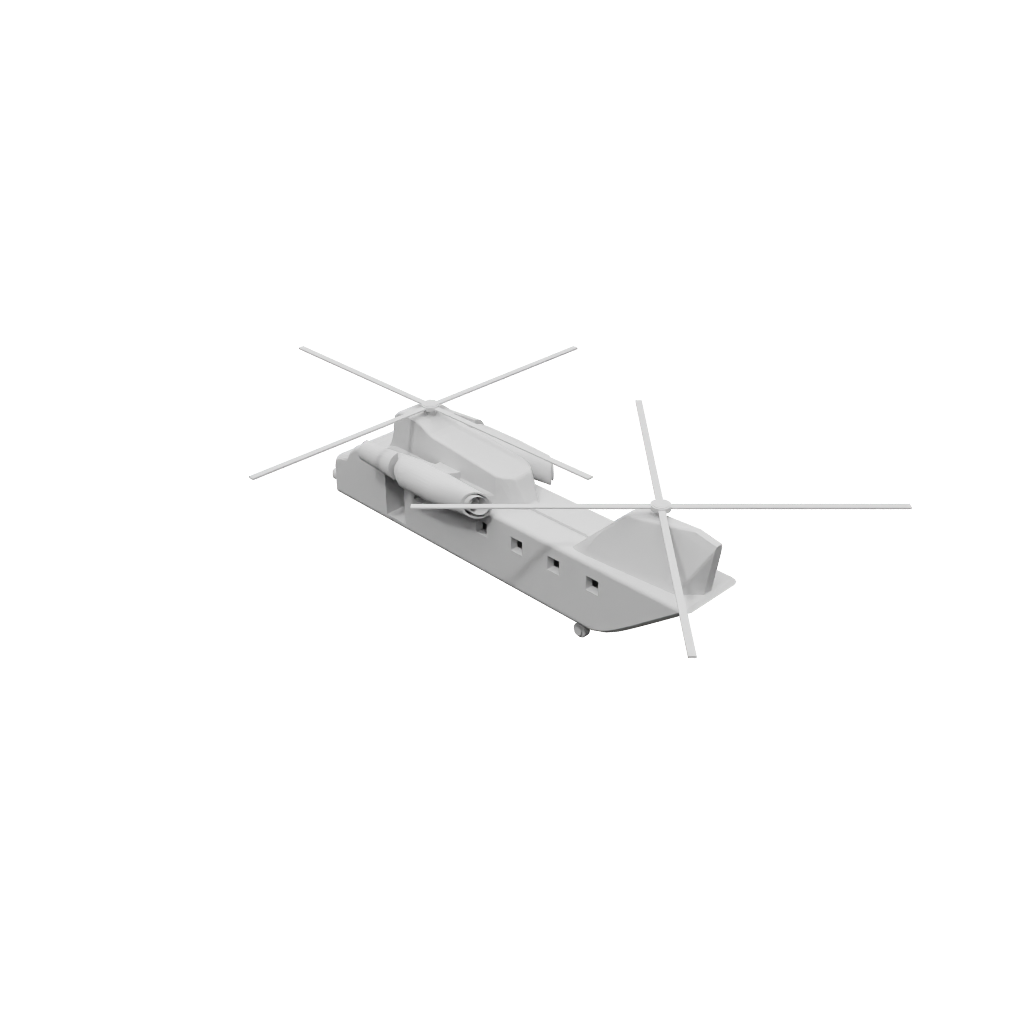} \\
\centering \shortstack{Ground truth \\ edited object $\bar{O}$}
& \supplColA{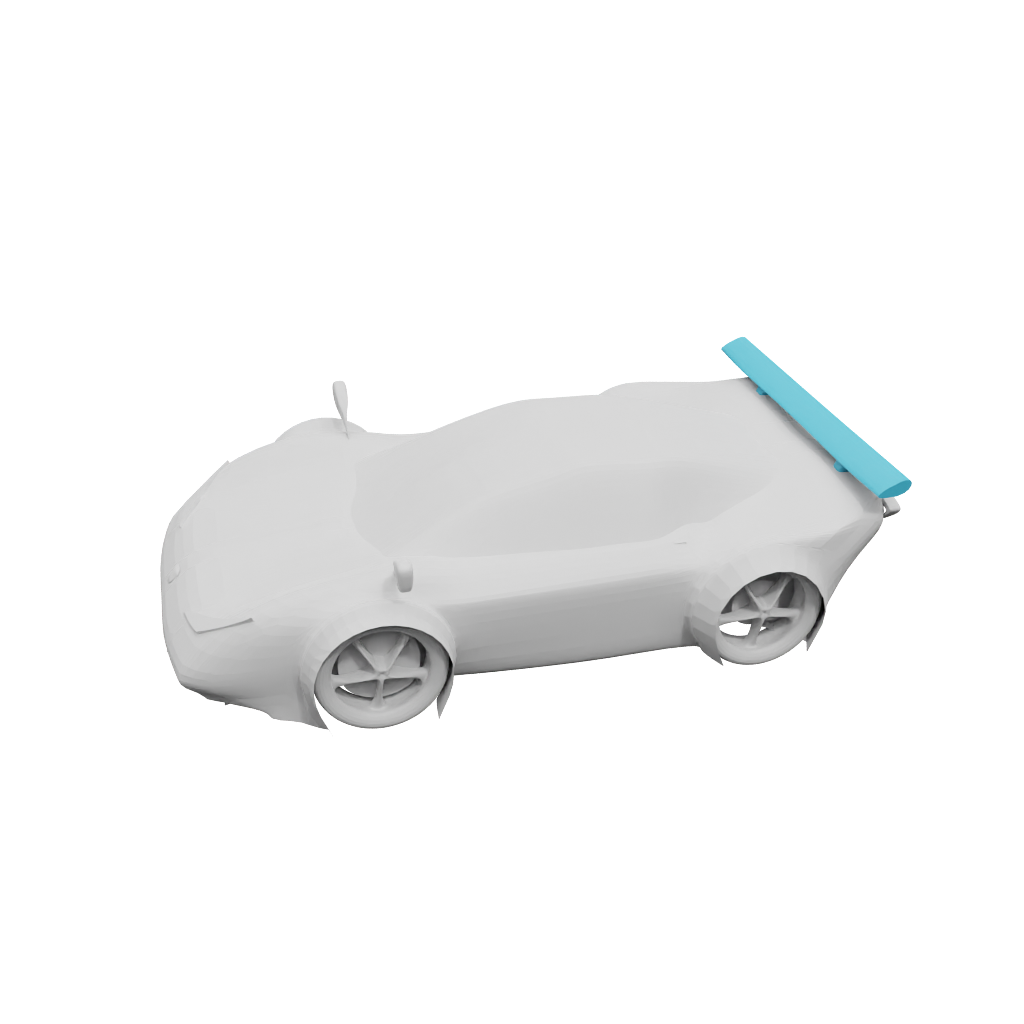}
& \supplColB{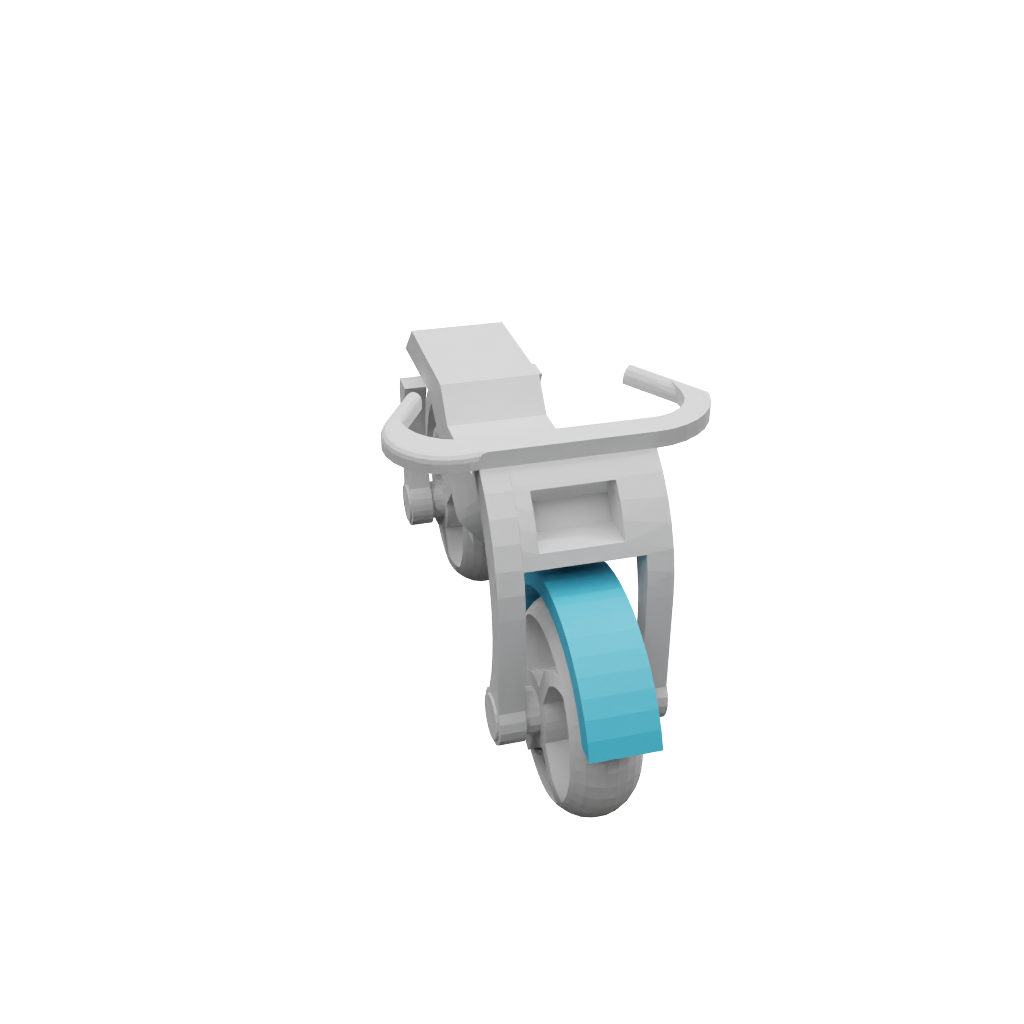}
& \supplColC{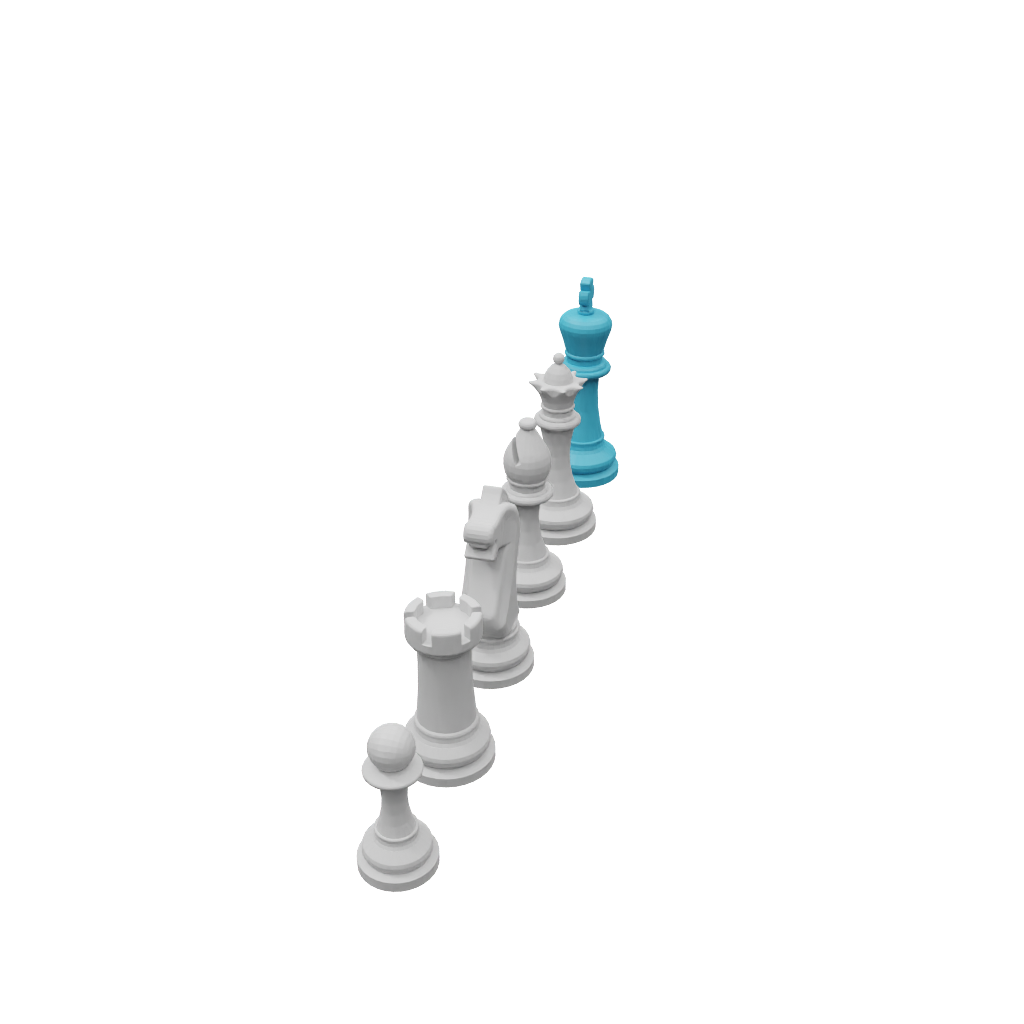}
& \supplColD{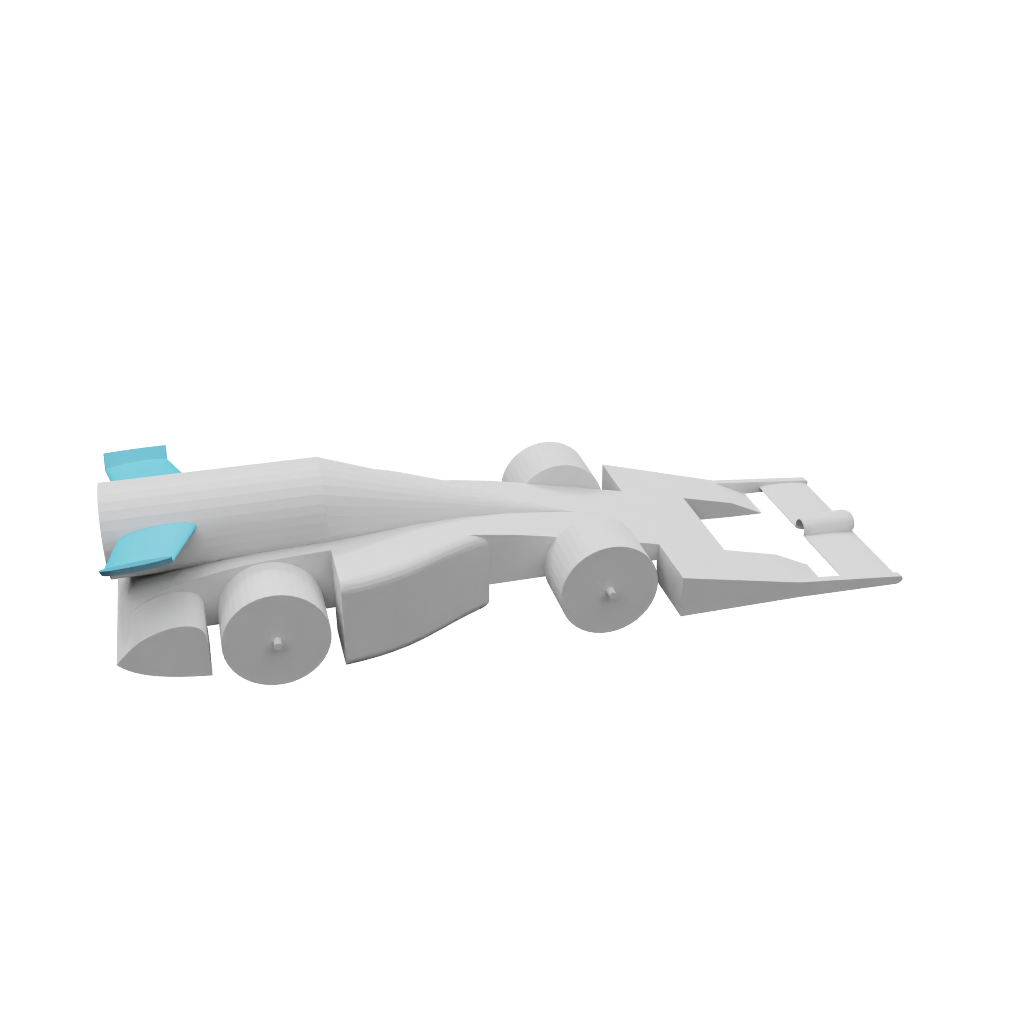}
& \supplColE{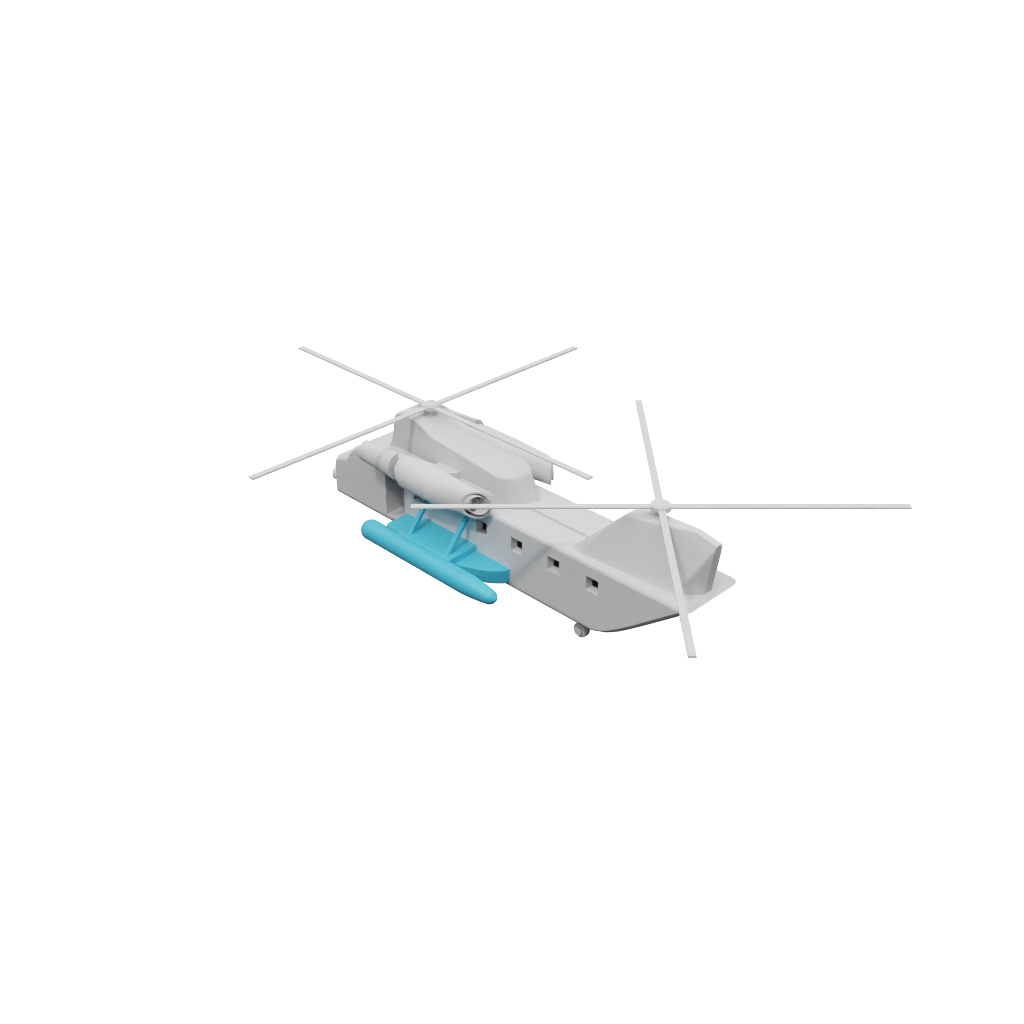} \\
\hline
\centering 3DMorph
& \supplColA{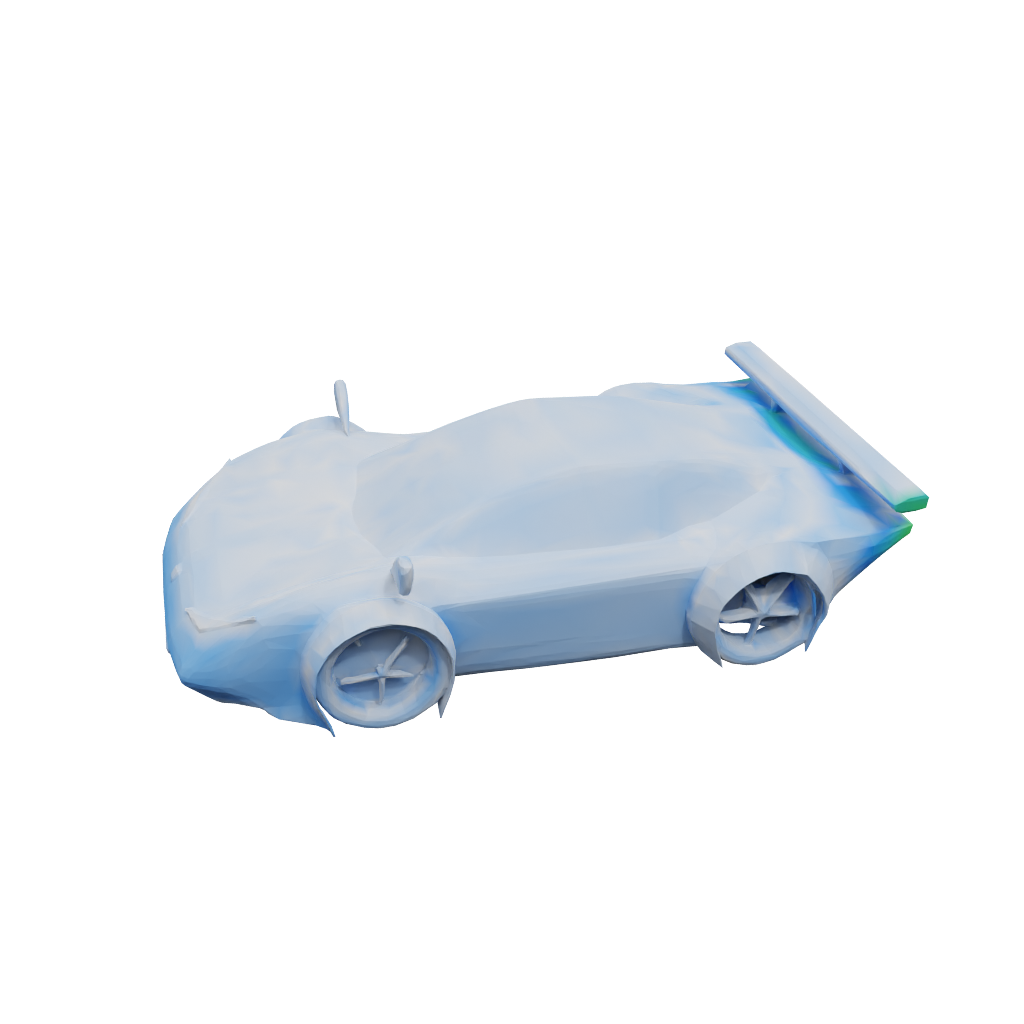}
& \supplColB{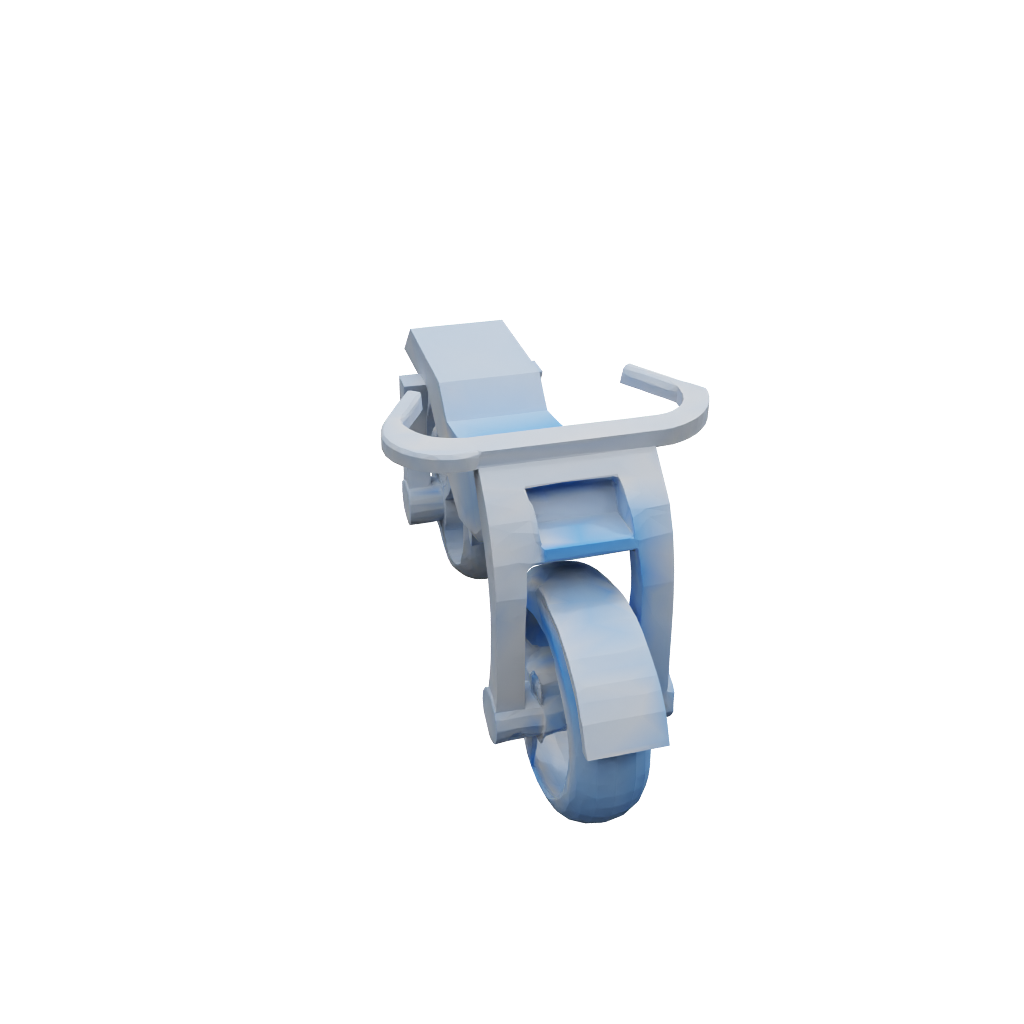}
& \supplColC{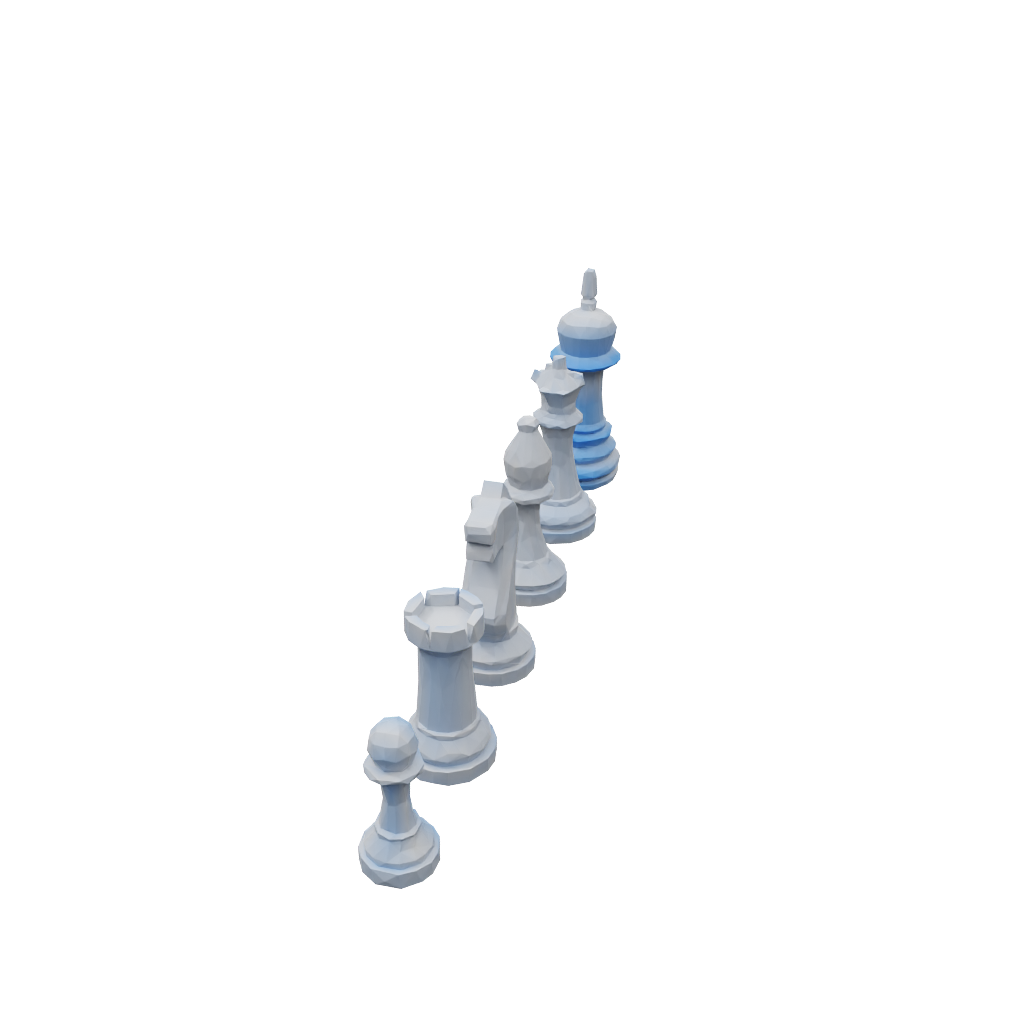}
& \supplColD{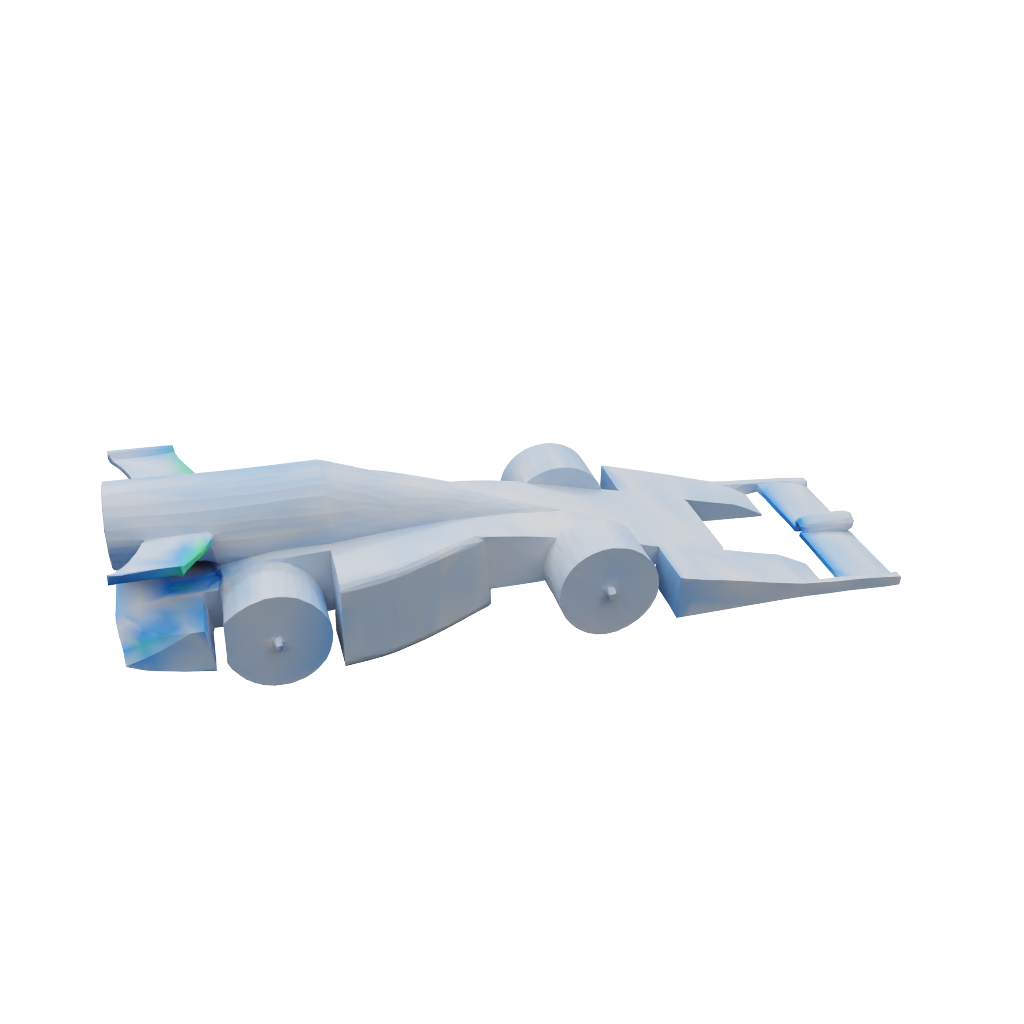}
& \supplColE{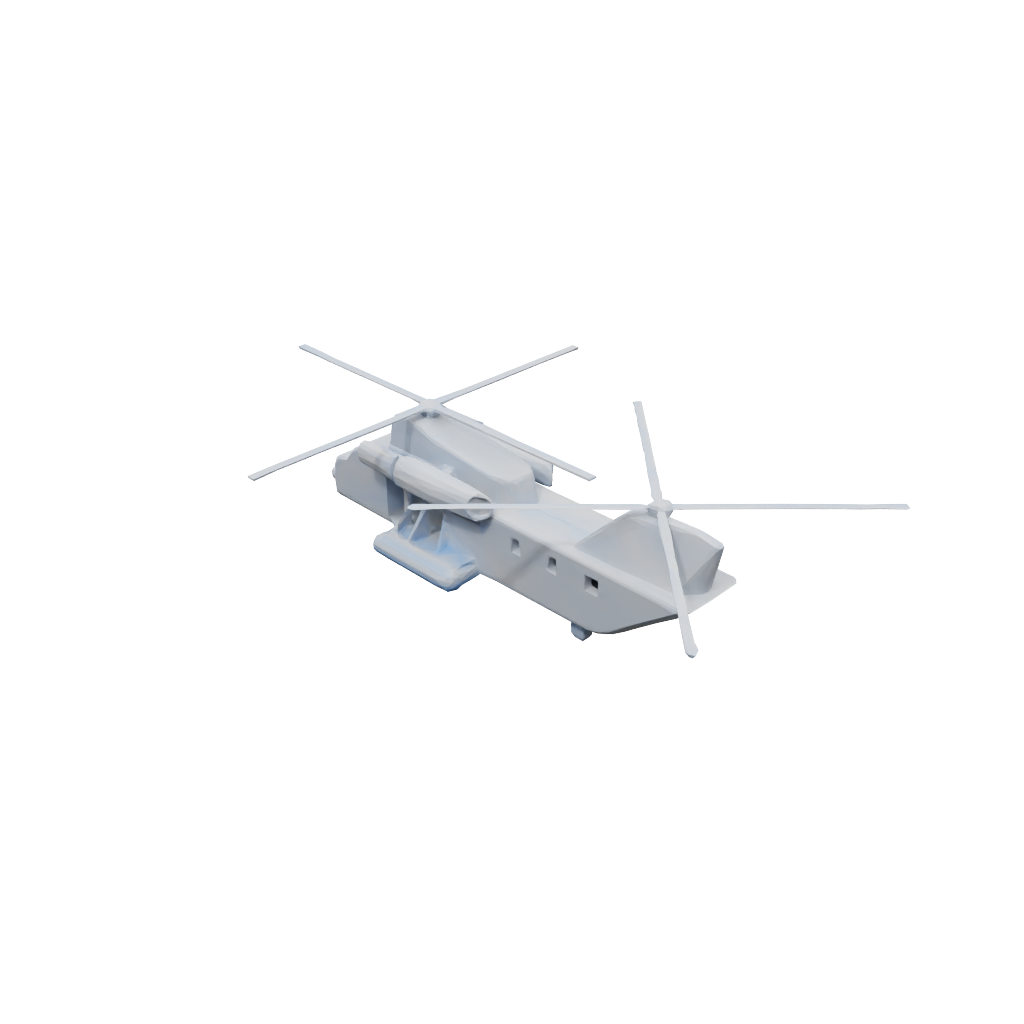} \\
\centering Trellis-RePaint
& \supplColA{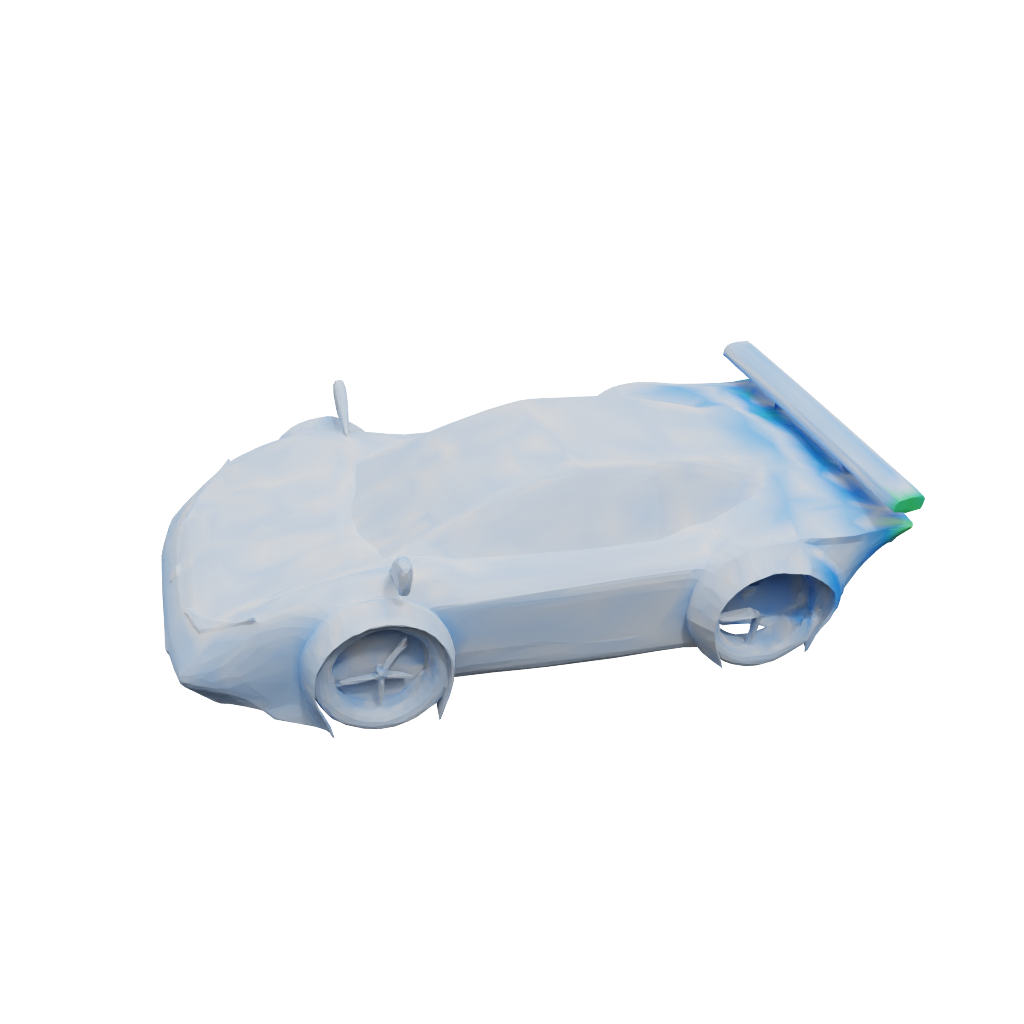}
& \supplColB{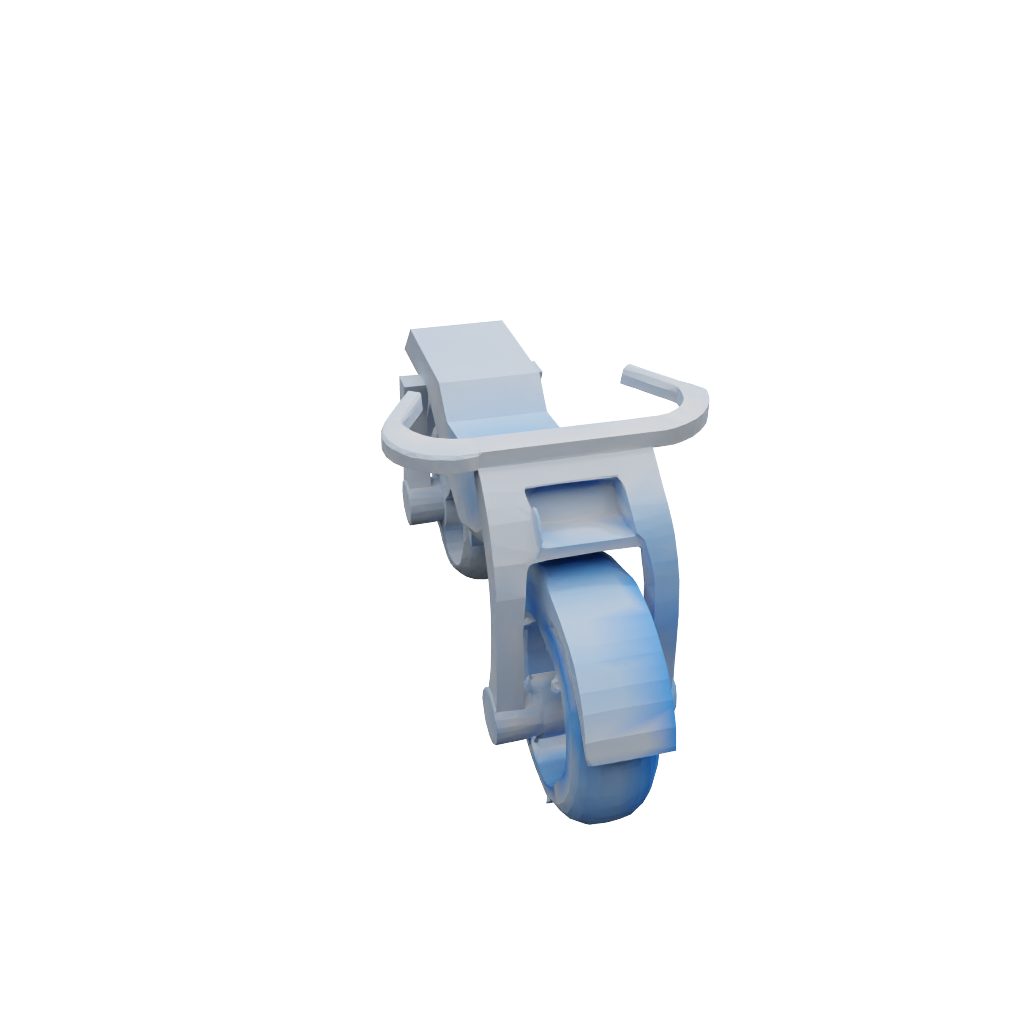}
& \supplColC{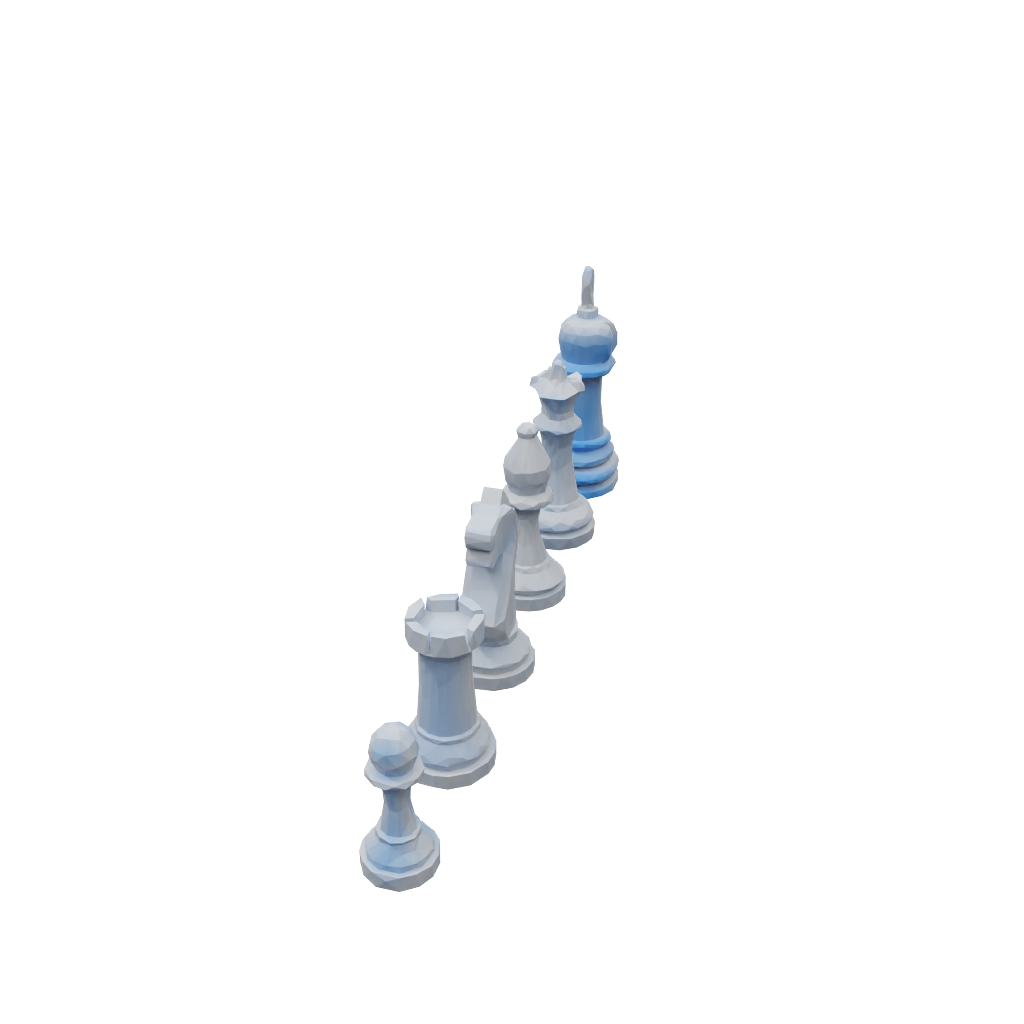}
& \supplColD{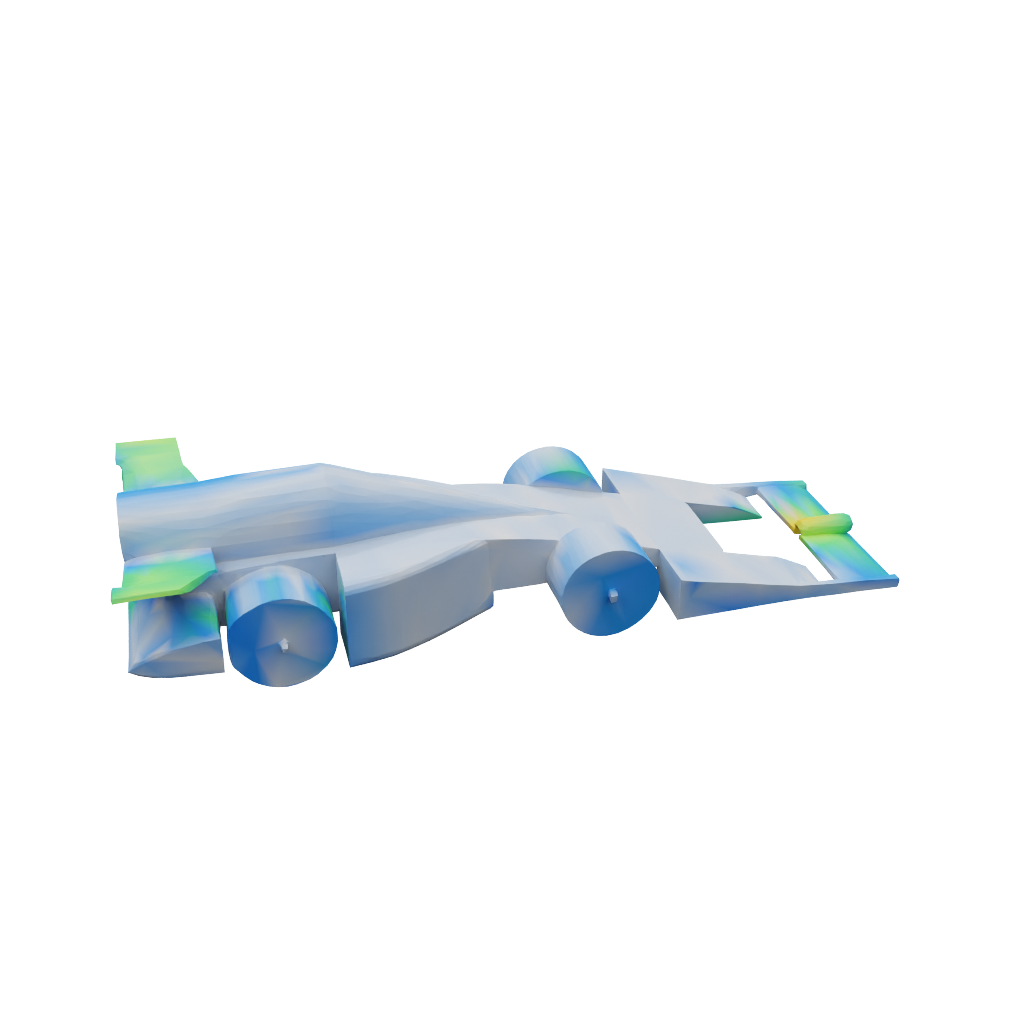}
& \supplColE{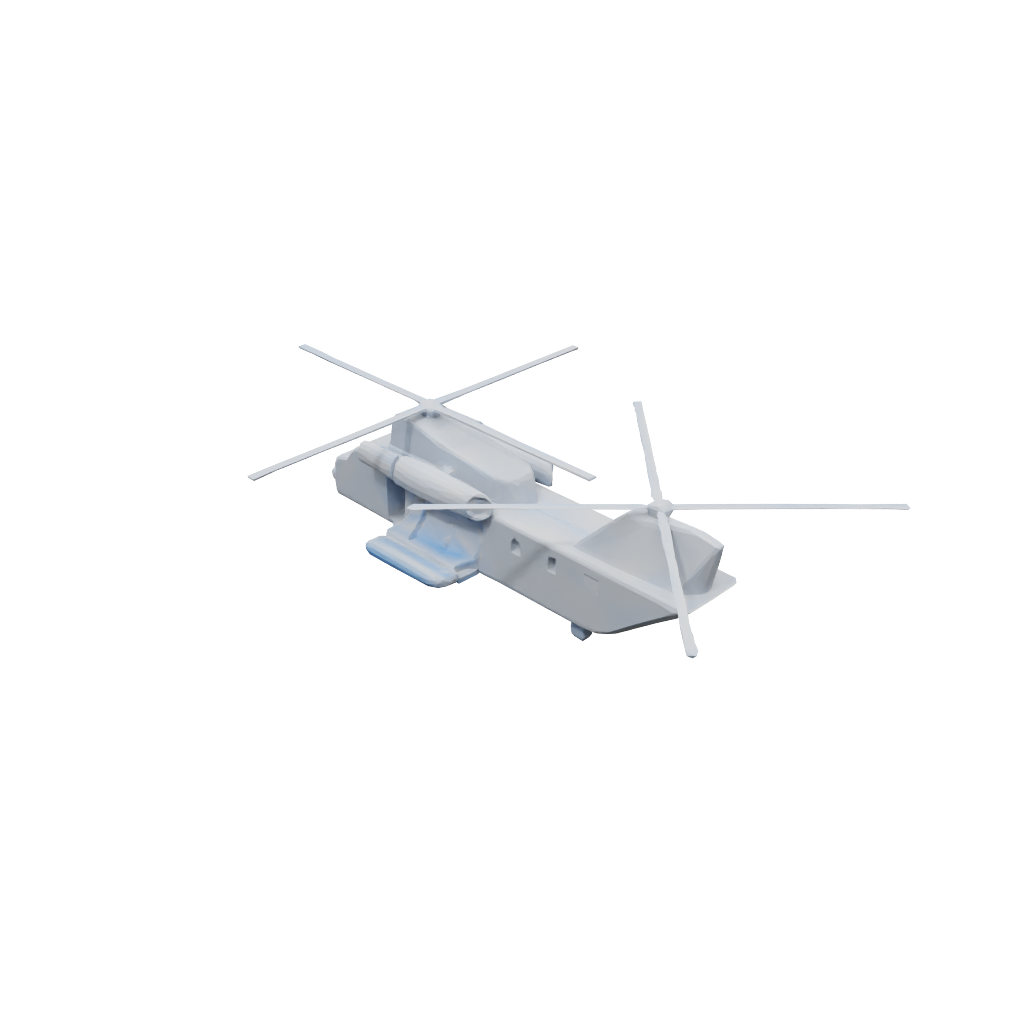} \\
\centering Hunyuan3D
& \supplColA{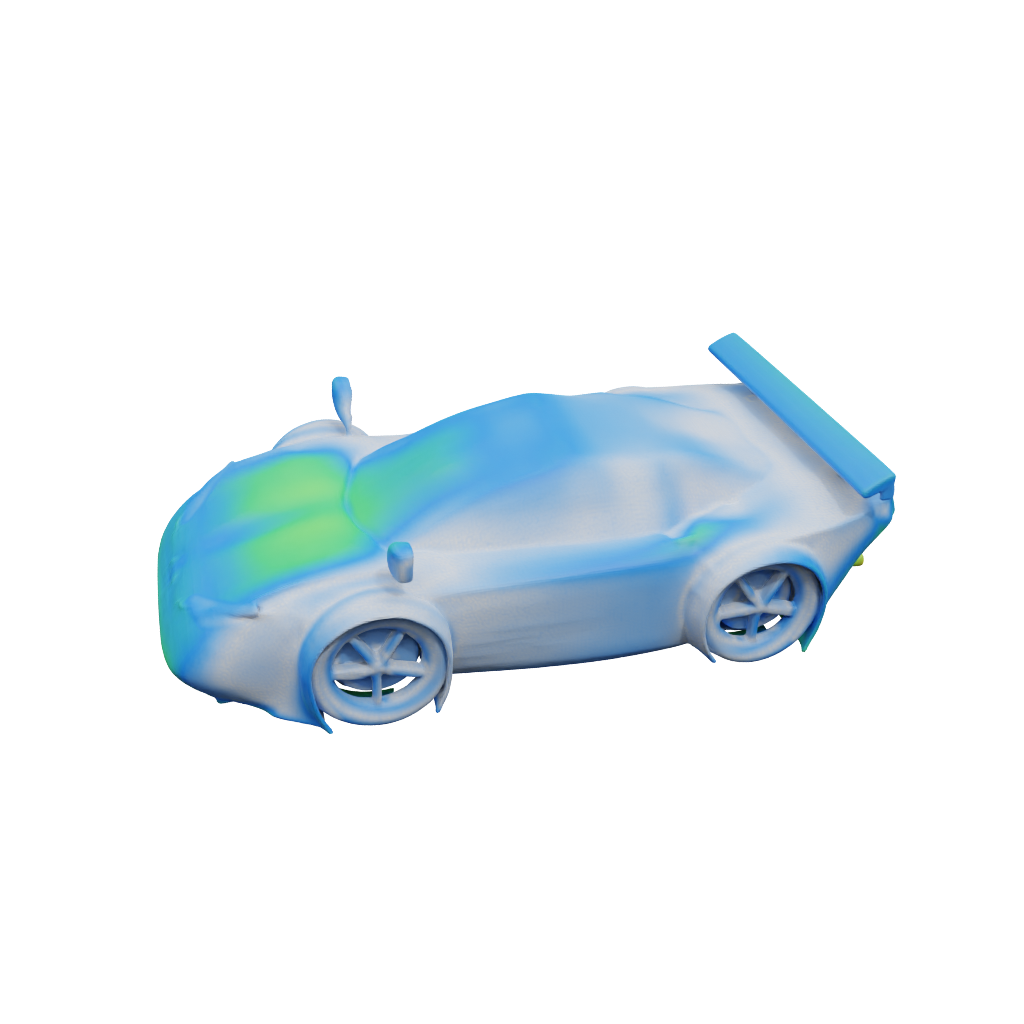}
& \supplColB{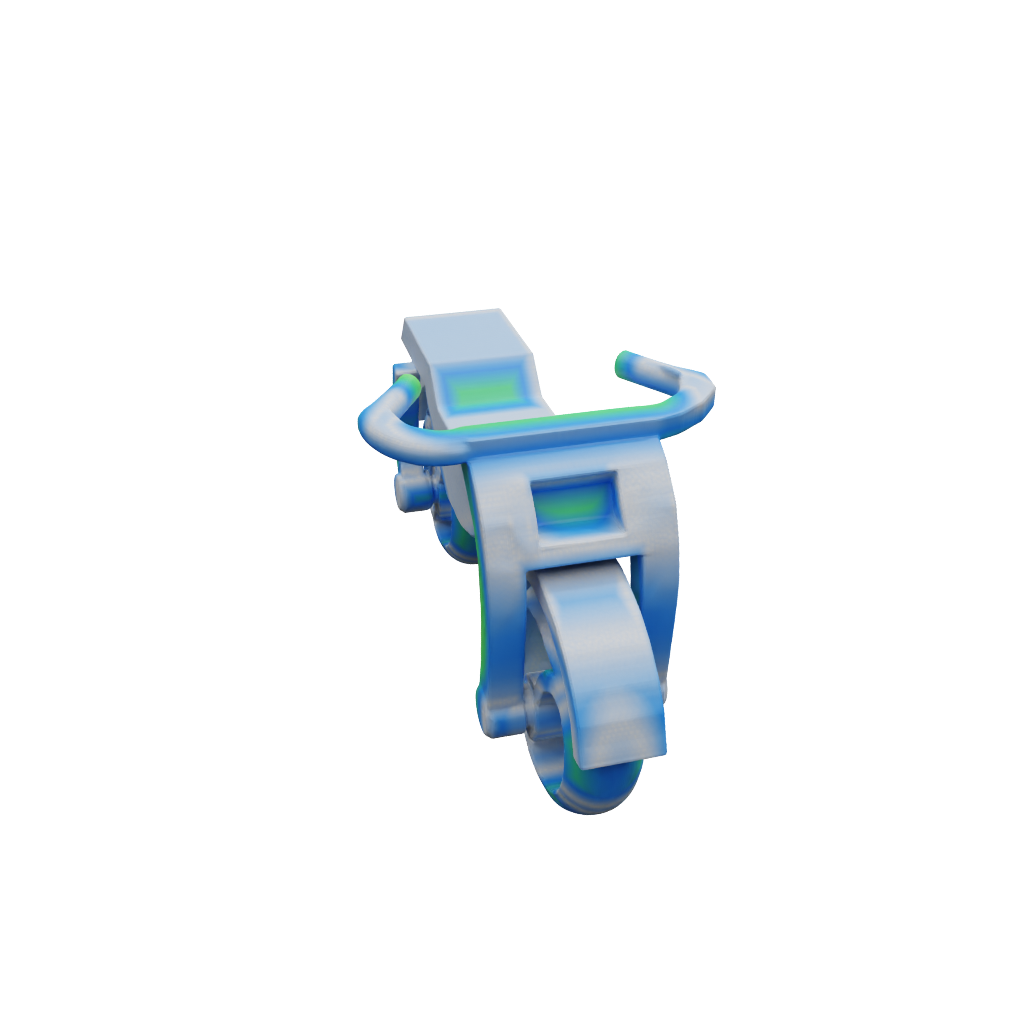}
& \supplColC{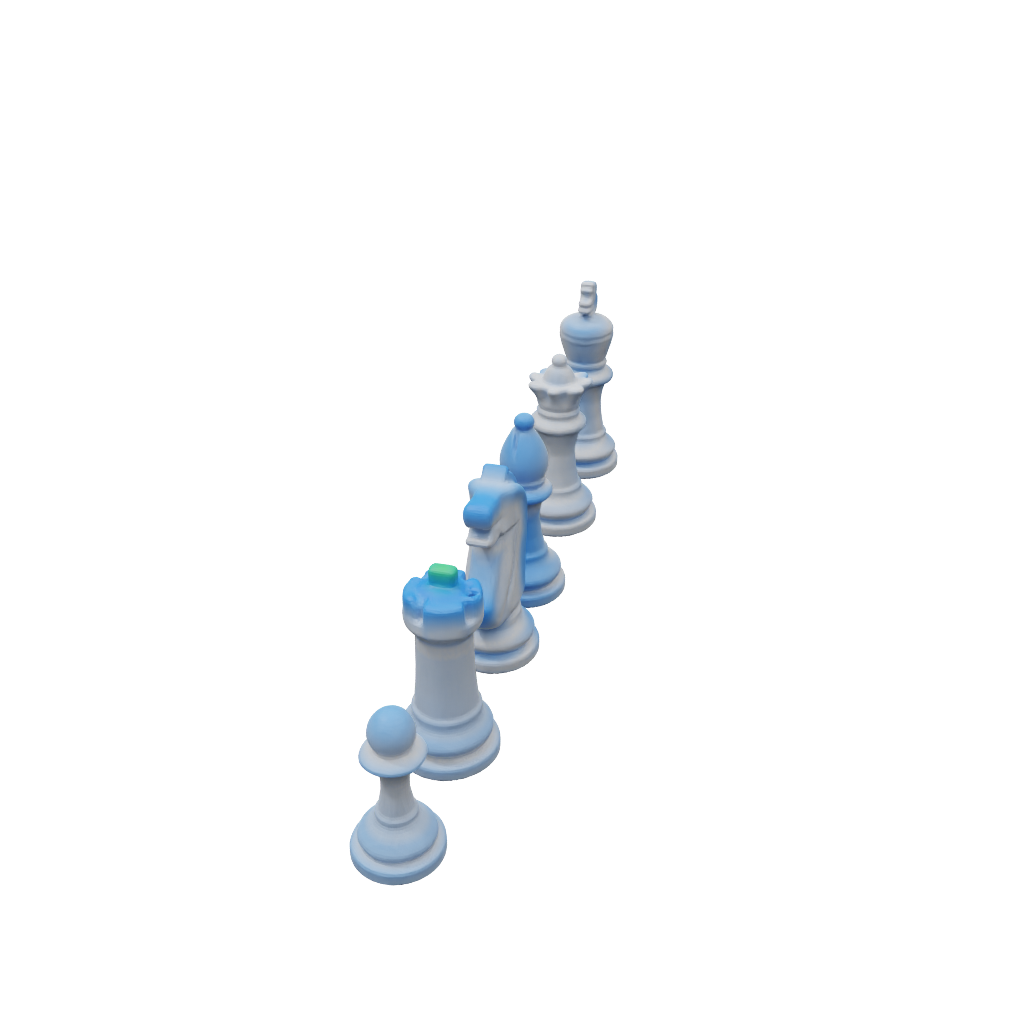}
& \supplColD{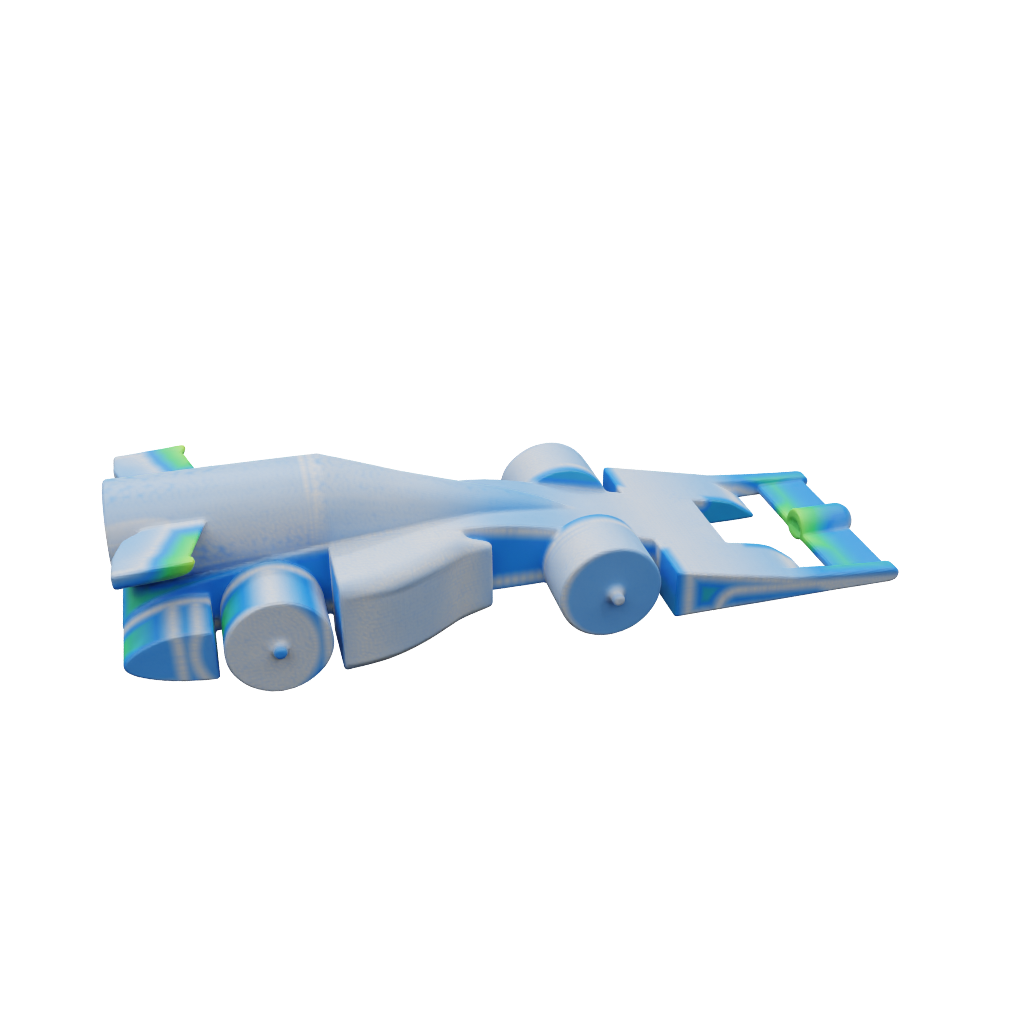}
& \supplColE{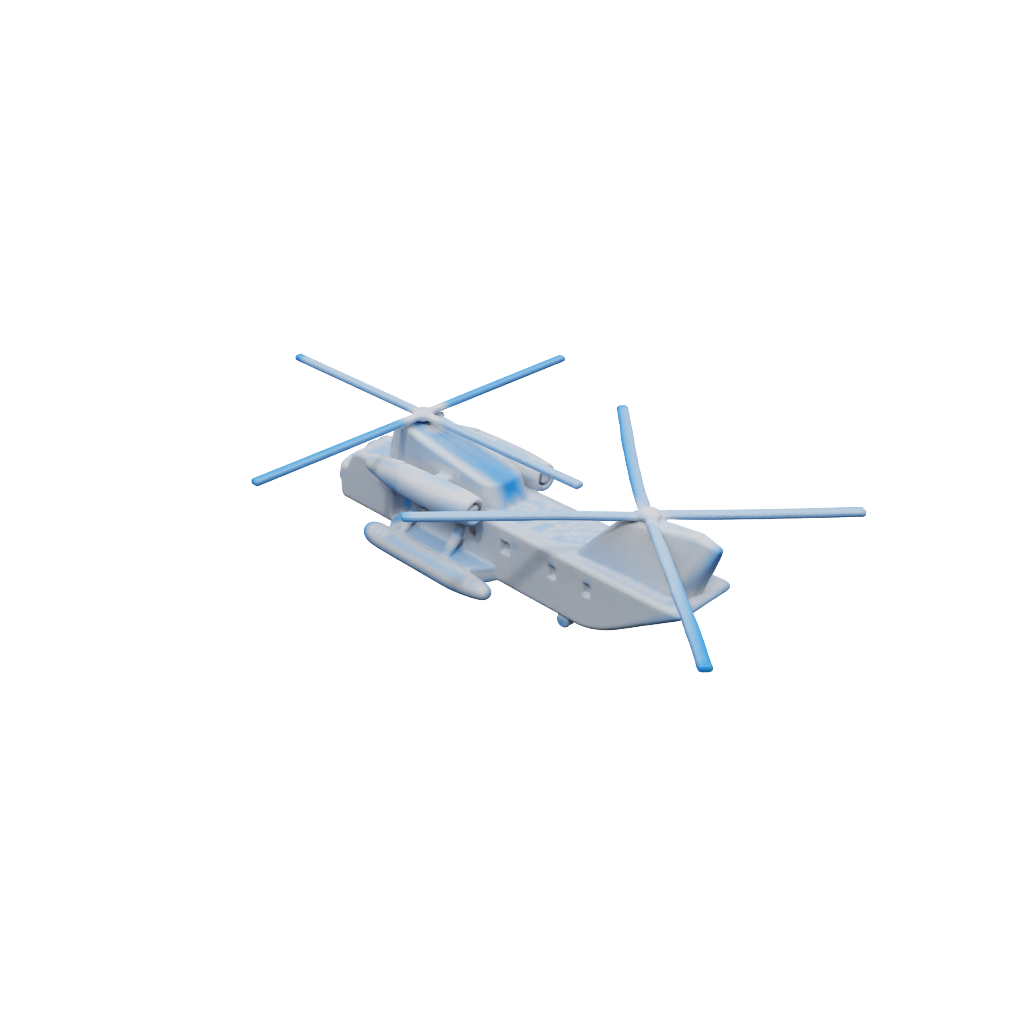} \\
\centering Trellis
& \supplColA{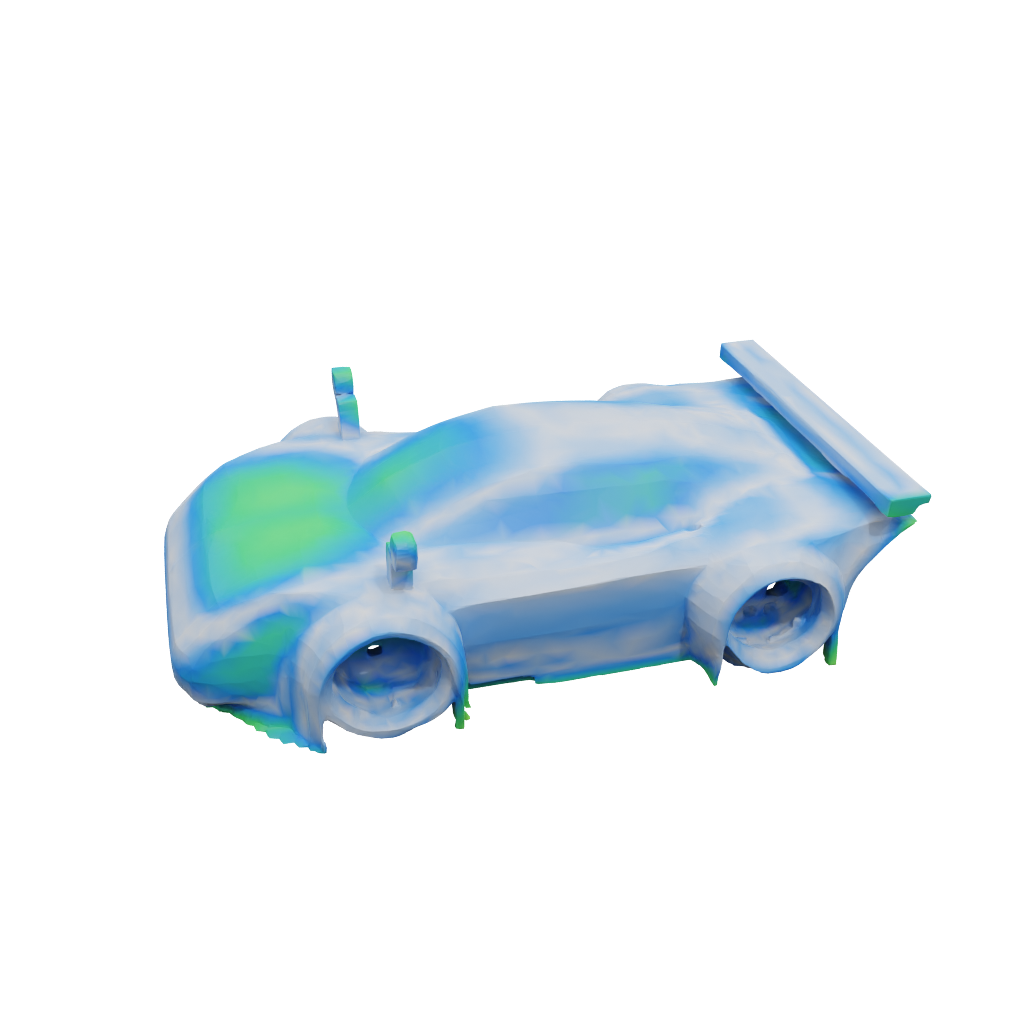}
& \supplColB{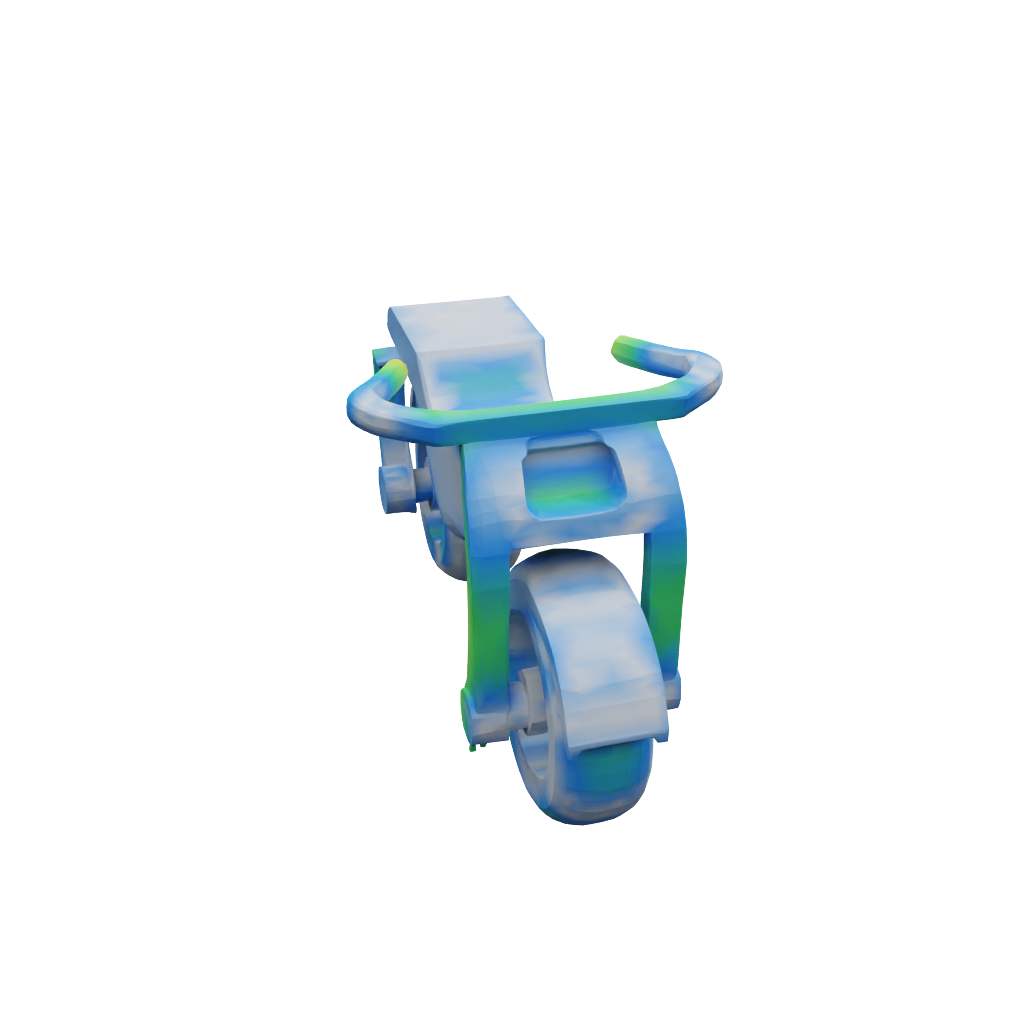}
& \supplColC{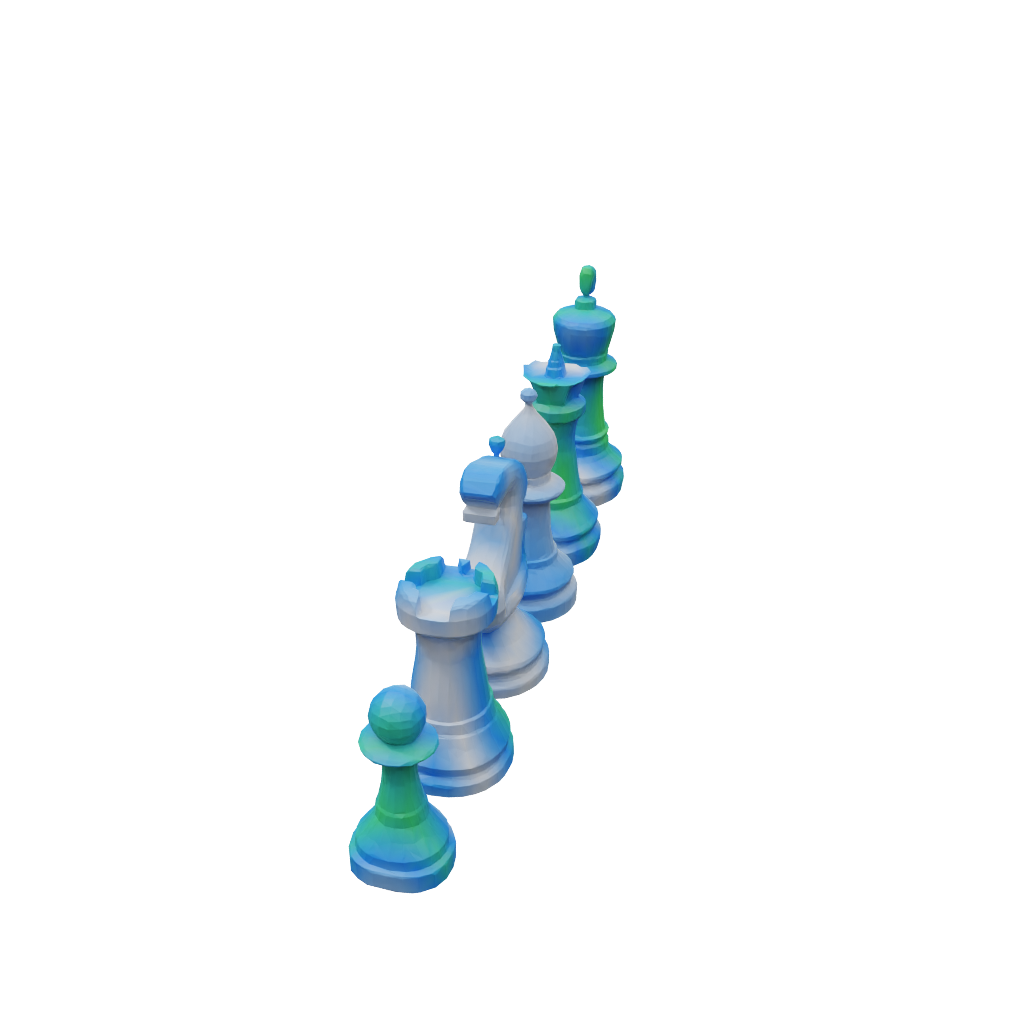}
& \supplColD{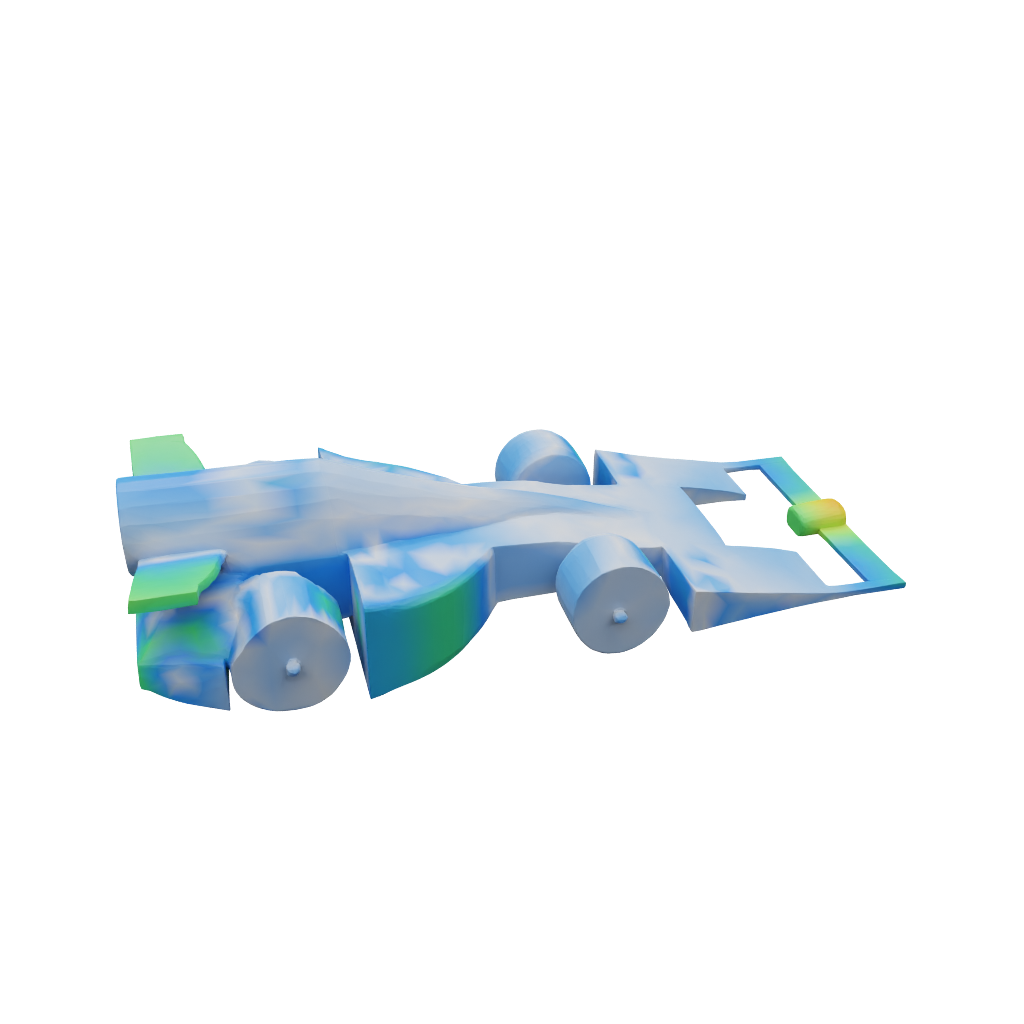}
& \supplColE{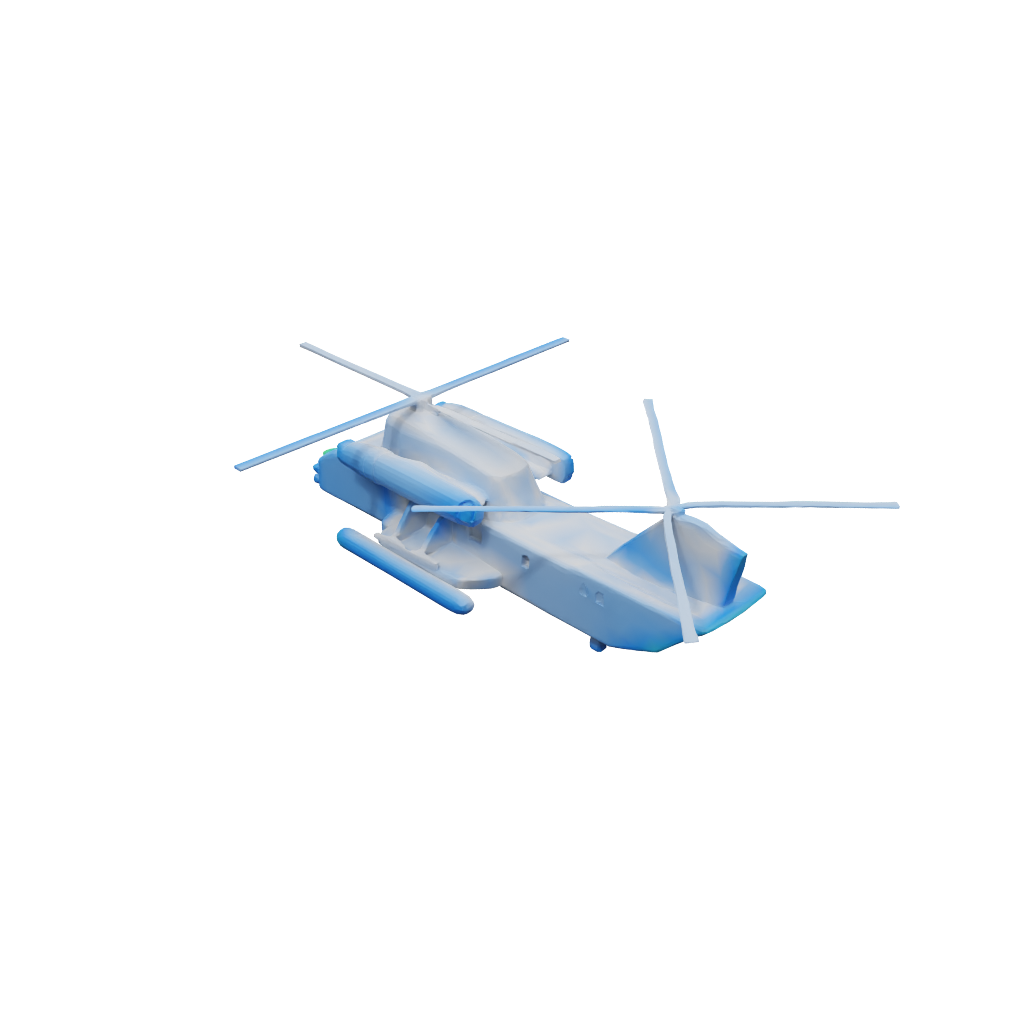} \\
\centering TripoSG
& \supplColA{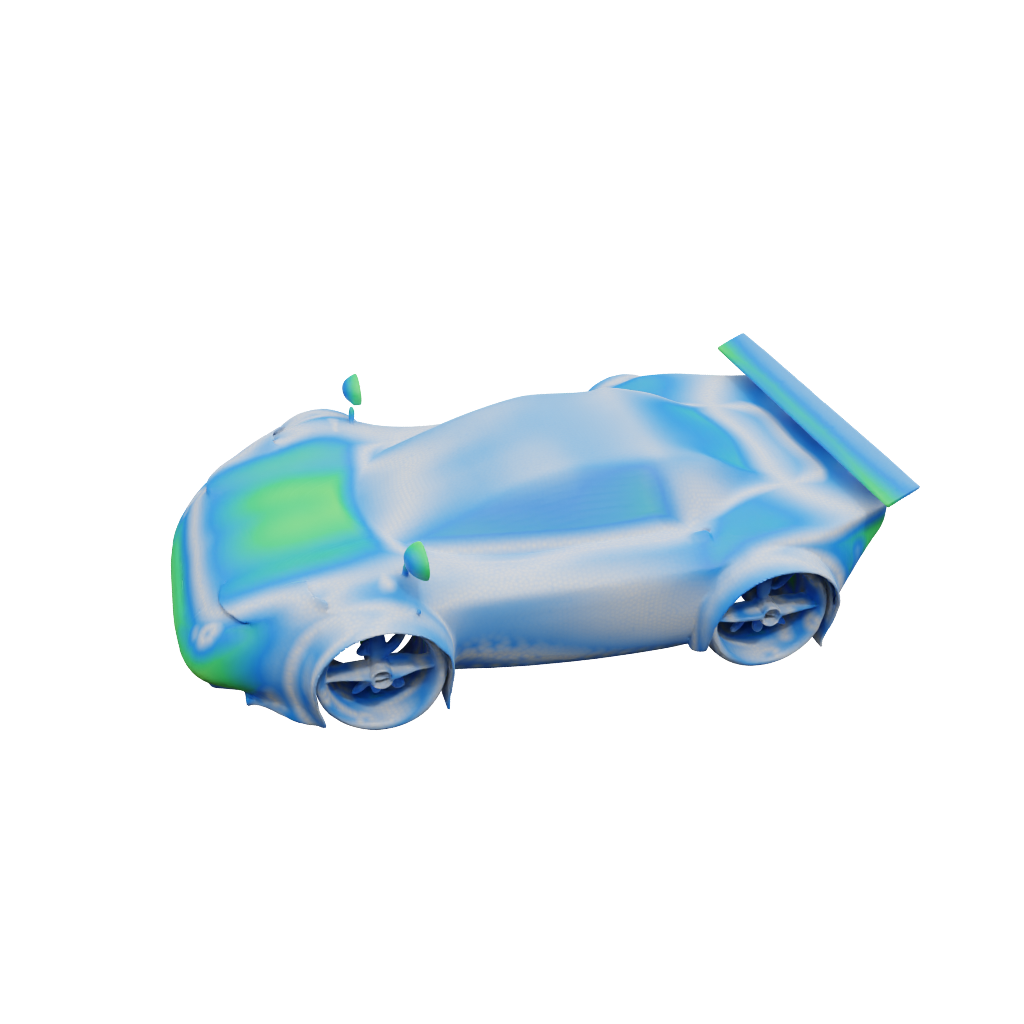}
& \supplColB{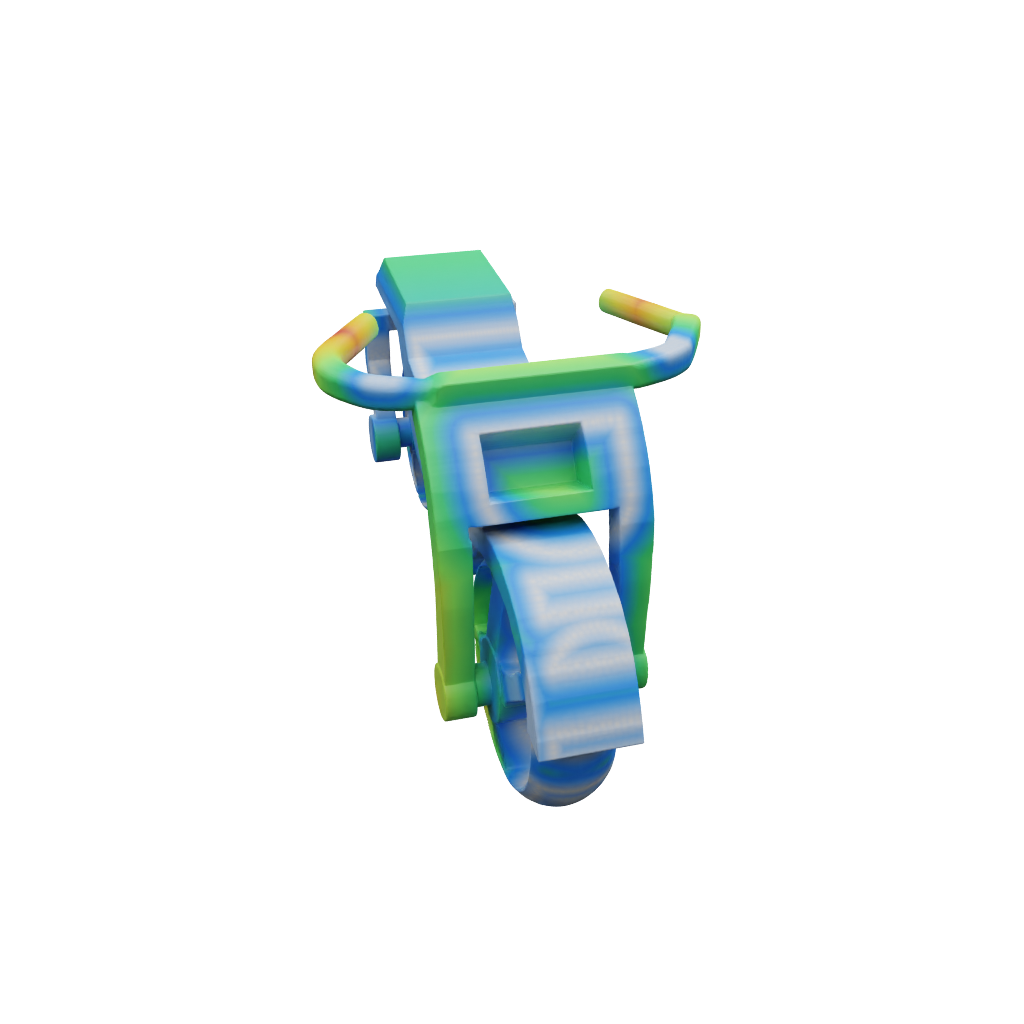}
& \supplColC{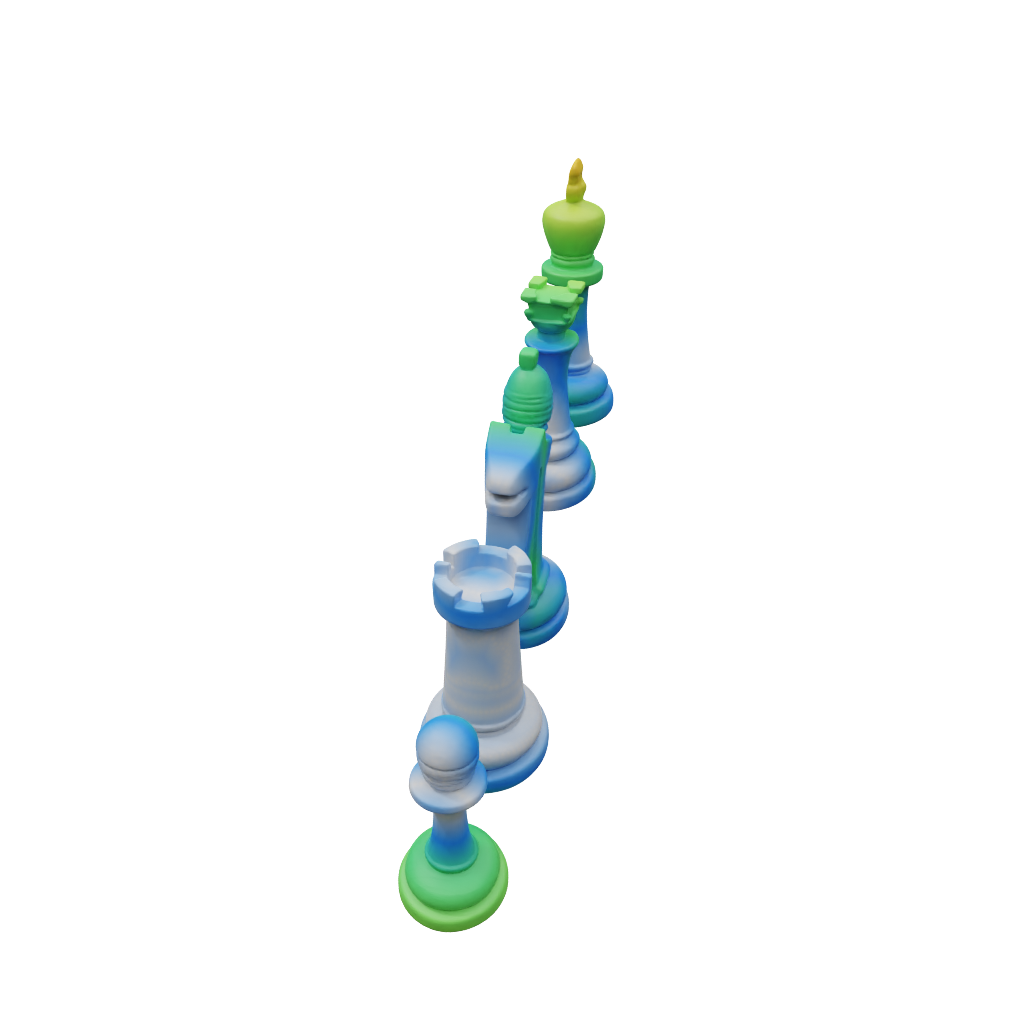}
& \supplColD{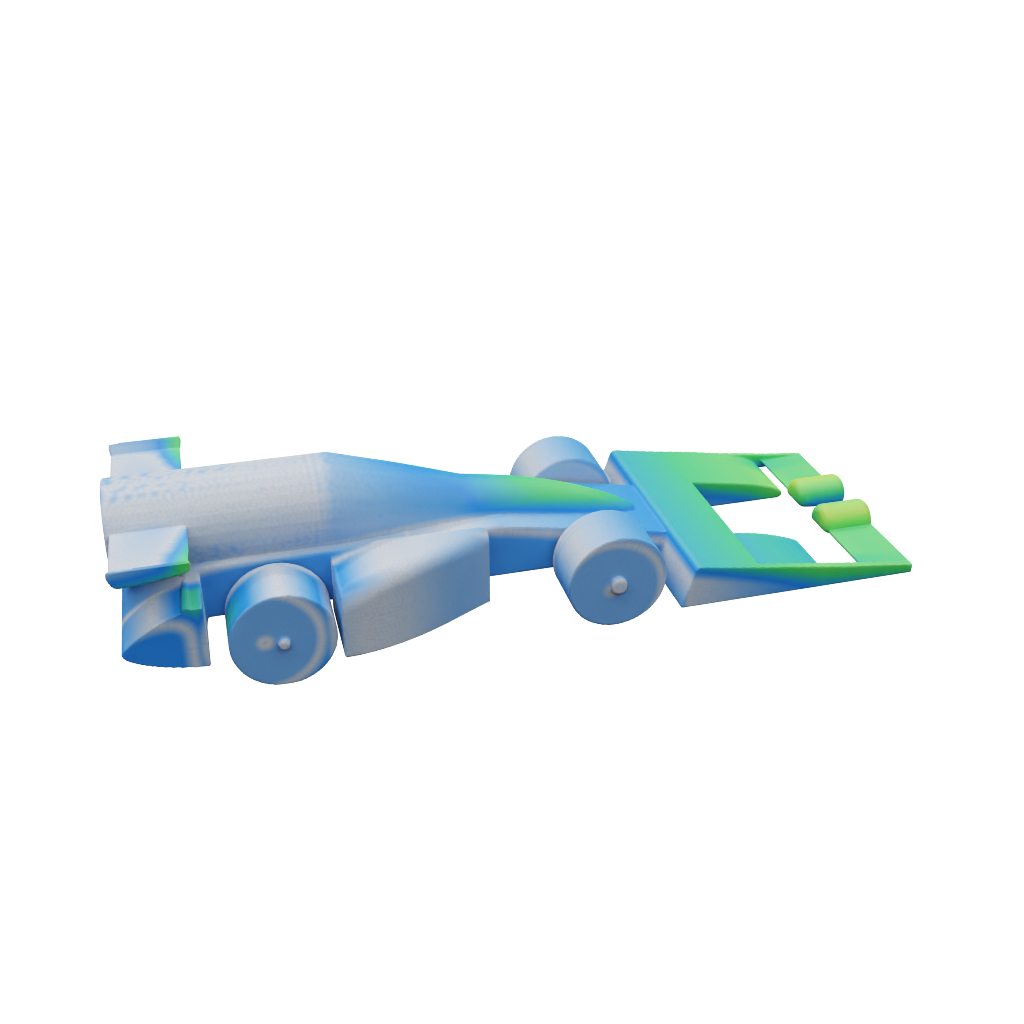}
& \supplColZ{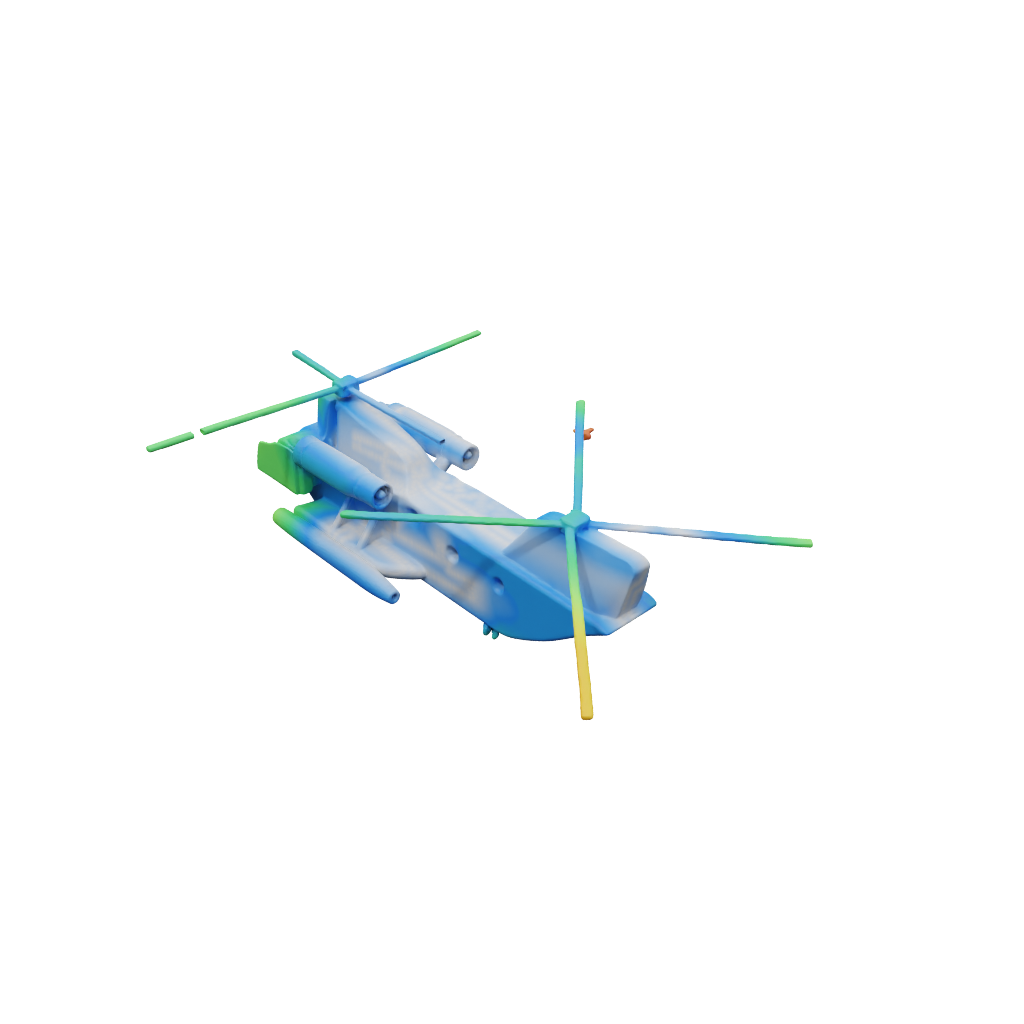} \\
\hline
\end{tabular}
\caption{Additional qualitative editing results. Differences are visualized between the edited object and the ground-truth edit $\tilde{O}$. Colors range from blue (small deviation) to red (large deviation).}
\label{suppl-fig:diff-suppl}
\end{figure*}


\clearpage
\twocolumn[{

\newcommand{\distrocolumnwidth}{3.25 cm}

\begin{center}

\begin{tabular}{m{2.2cm}M{\distrocolumnwidth}M{\distrocolumnwidth}M{\distrocolumnwidth}M{\distrocolumnwidth}}
\toprule
\centering Method &  CD\down & HD\down  & IoU\up (\%)  & Dice\up (\%) \\
\toprule

\centering TripoSG
& \includegraphics[width=\distrocolumnwidth]{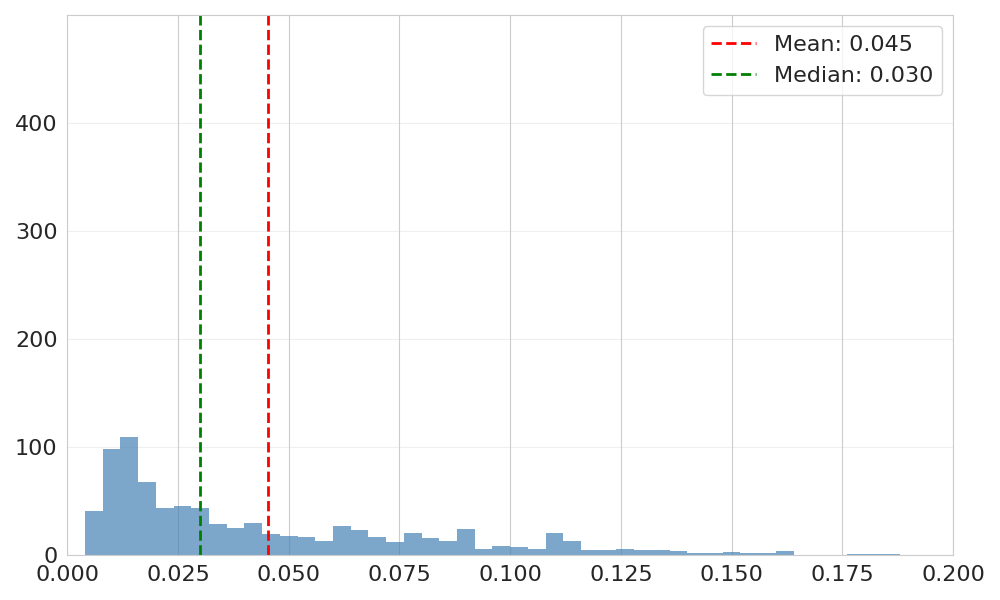}
& \includegraphics[width=\distrocolumnwidth]{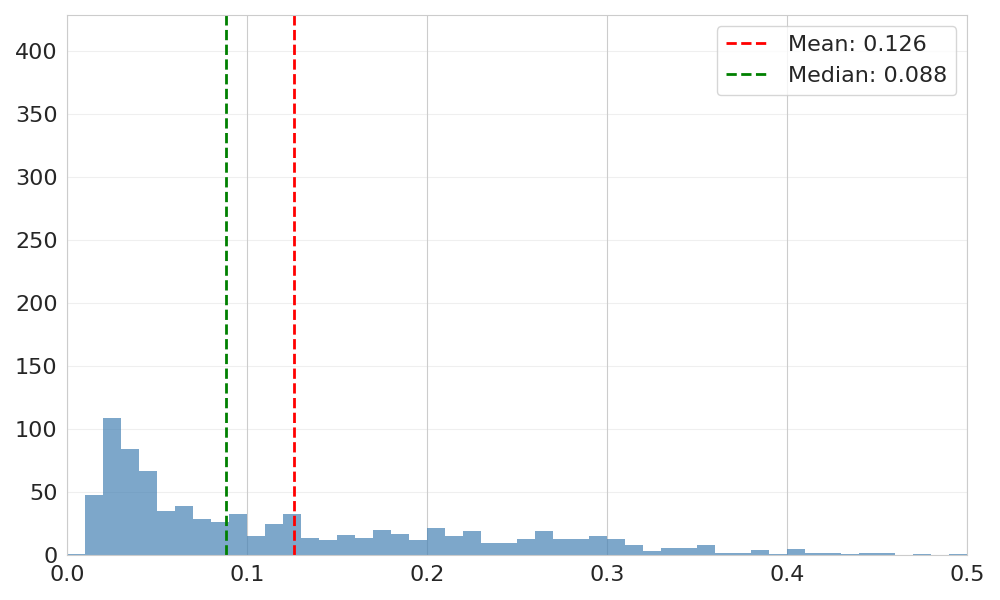}
& \includegraphics[width=\distrocolumnwidth]{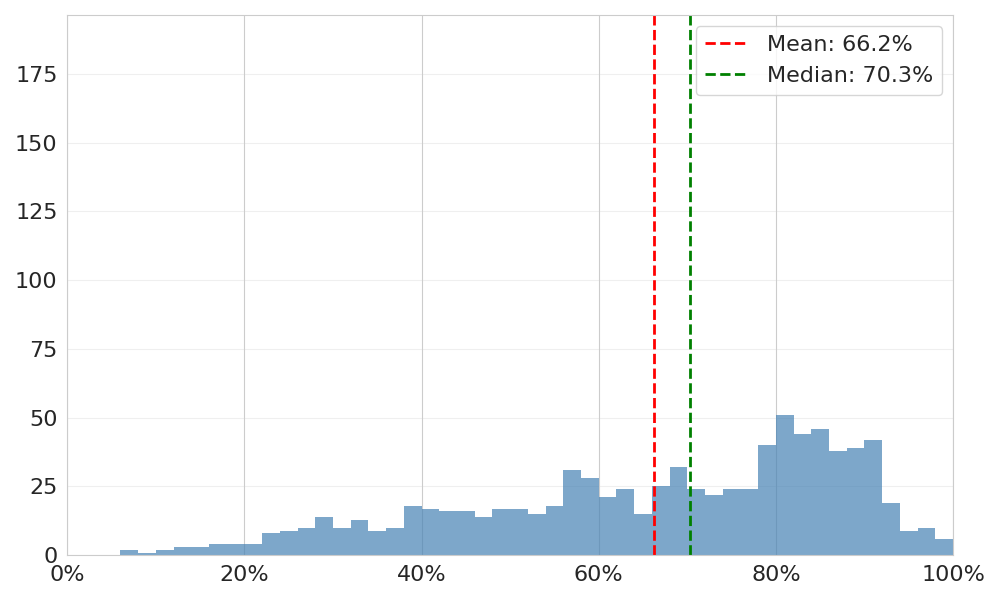}
& \includegraphics[width=\distrocolumnwidth]{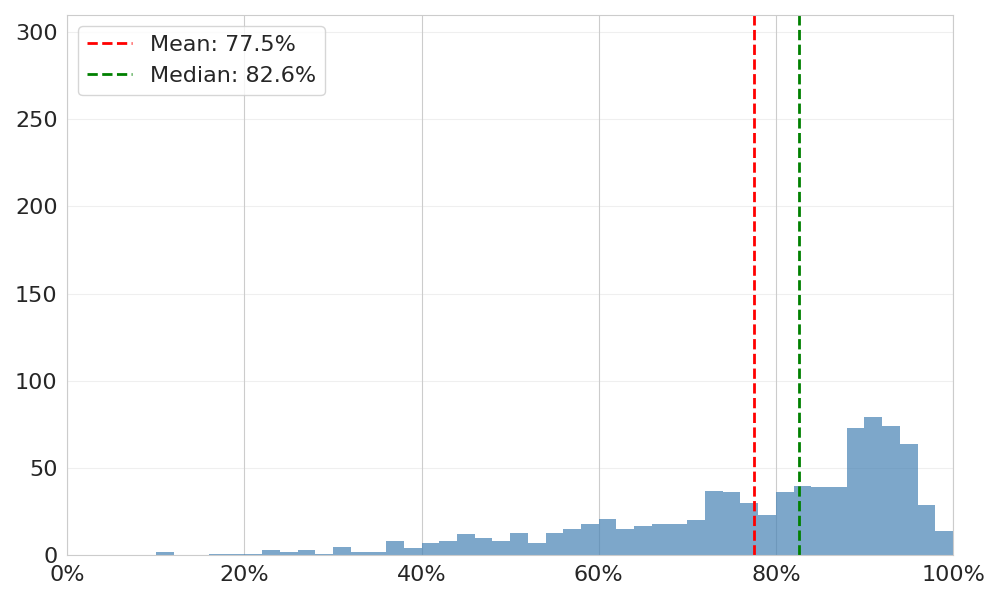} \\

\centering Trellis
& \includegraphics[width=\distrocolumnwidth]{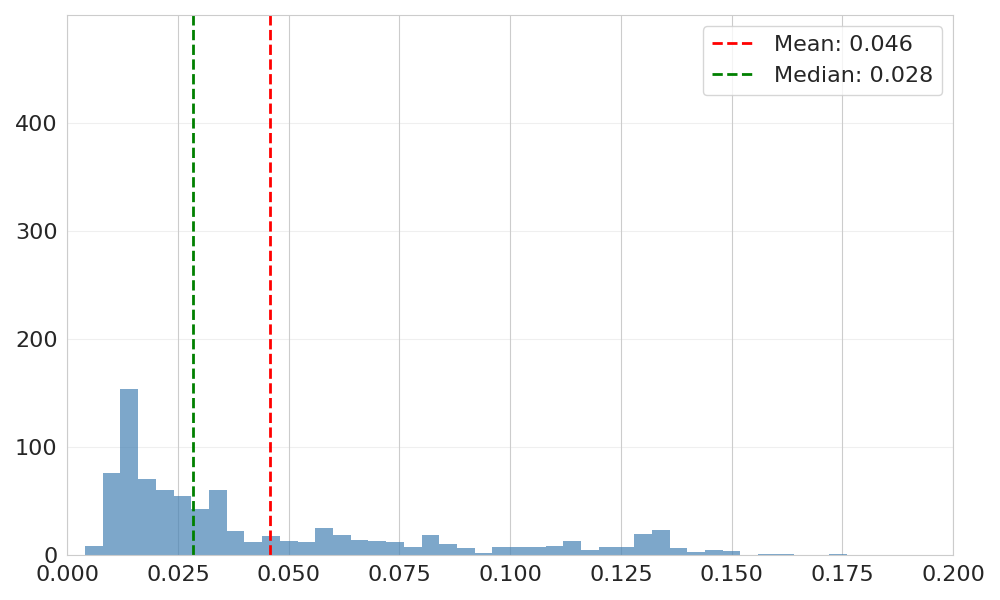}
& \includegraphics[width=\distrocolumnwidth]{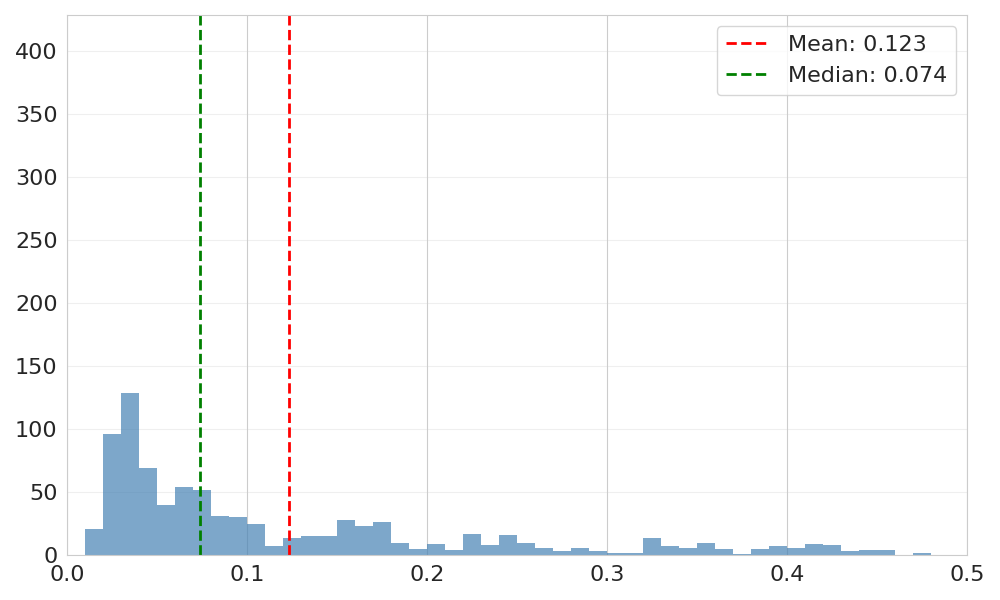}
& \includegraphics[width=\distrocolumnwidth]{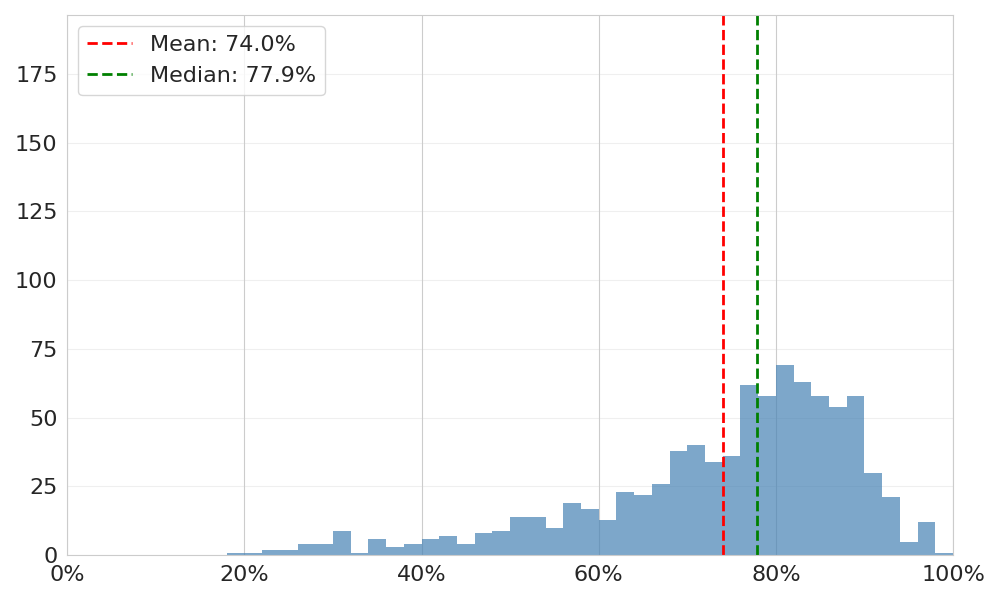}
& \includegraphics[width=\distrocolumnwidth]{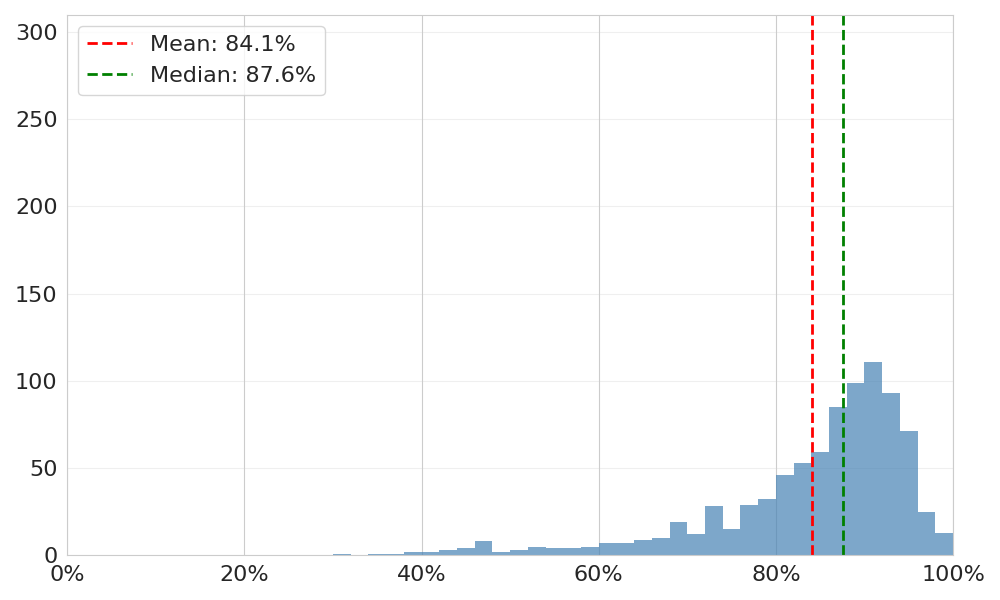} \\

\centering Trellis-MV
& \includegraphics[width=\distrocolumnwidth]{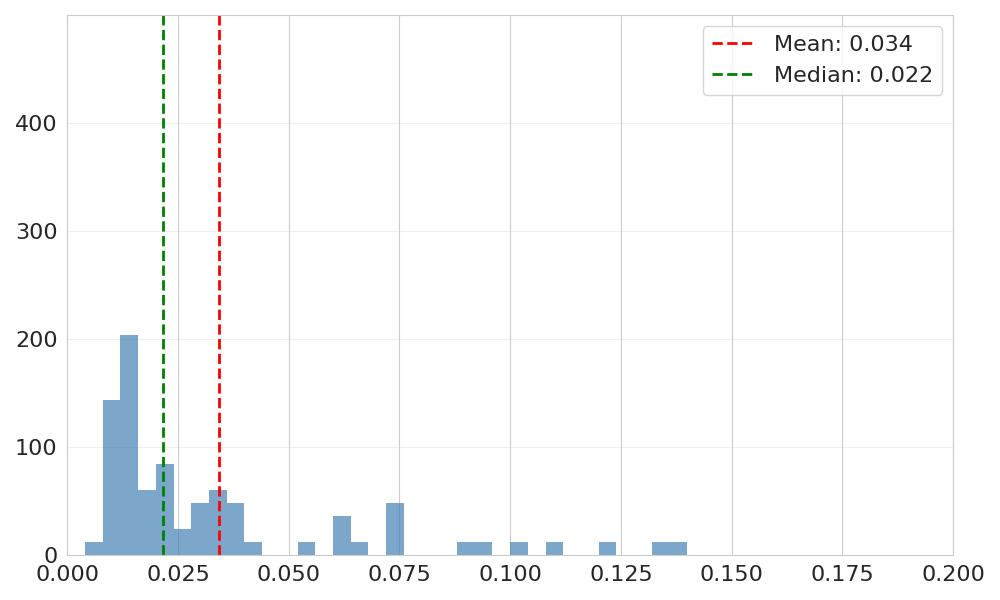}
& \includegraphics[width=\distrocolumnwidth]{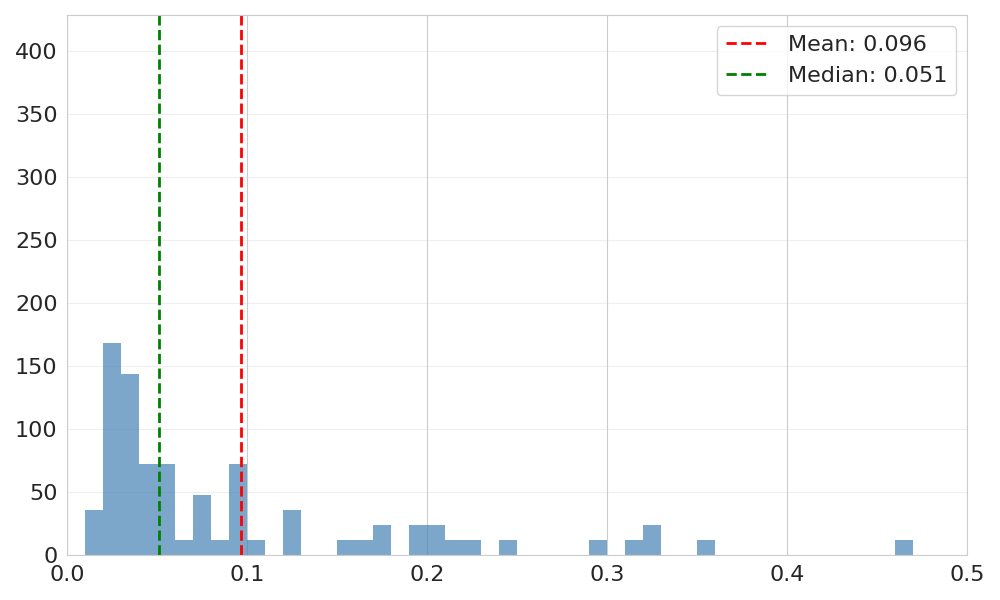}
& \includegraphics[width=\distrocolumnwidth]{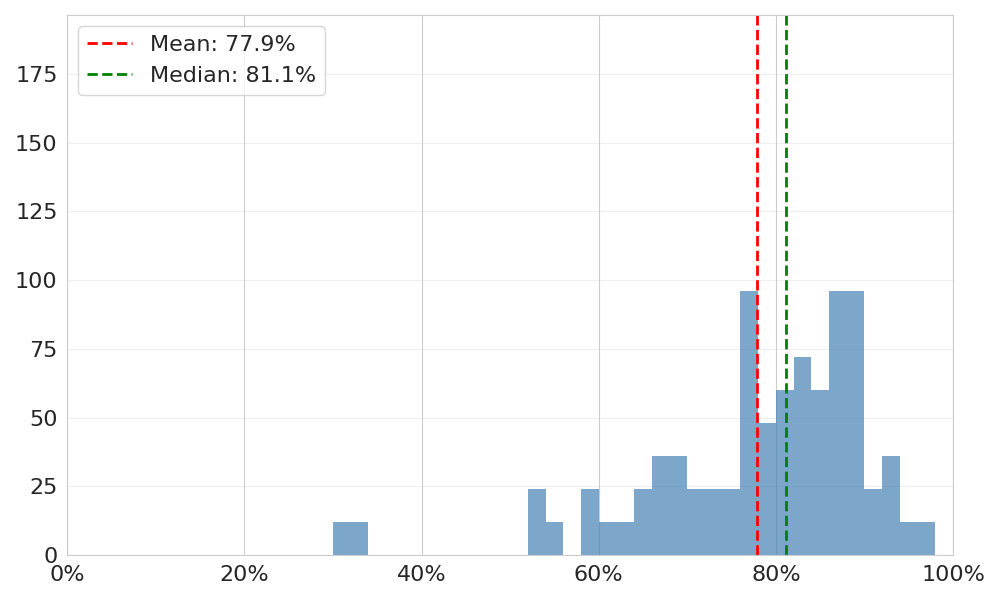}
& \includegraphics[width=\distrocolumnwidth]{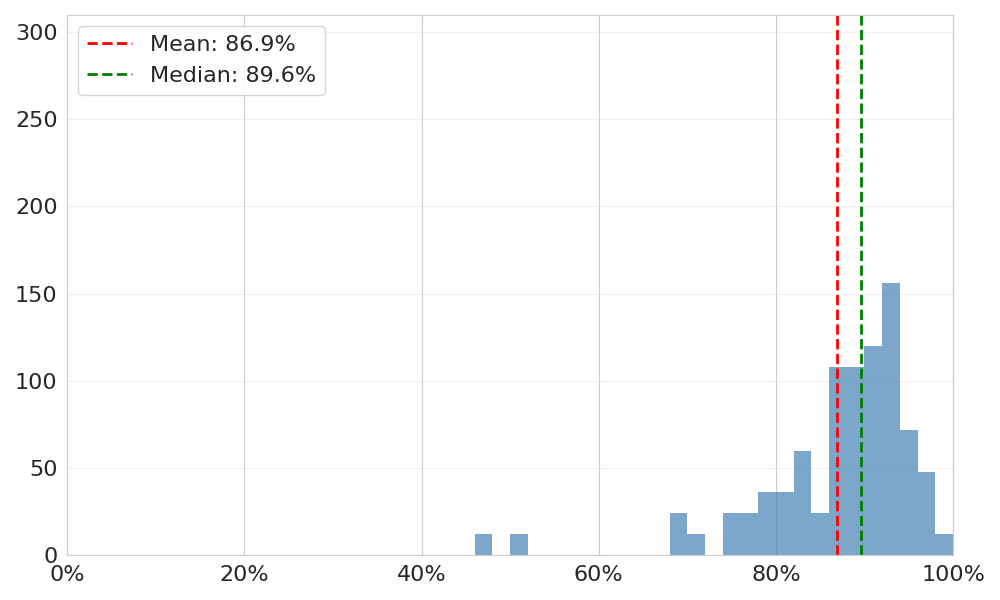} \\

\centering Hunyuan3D
& \includegraphics[width=\distrocolumnwidth]{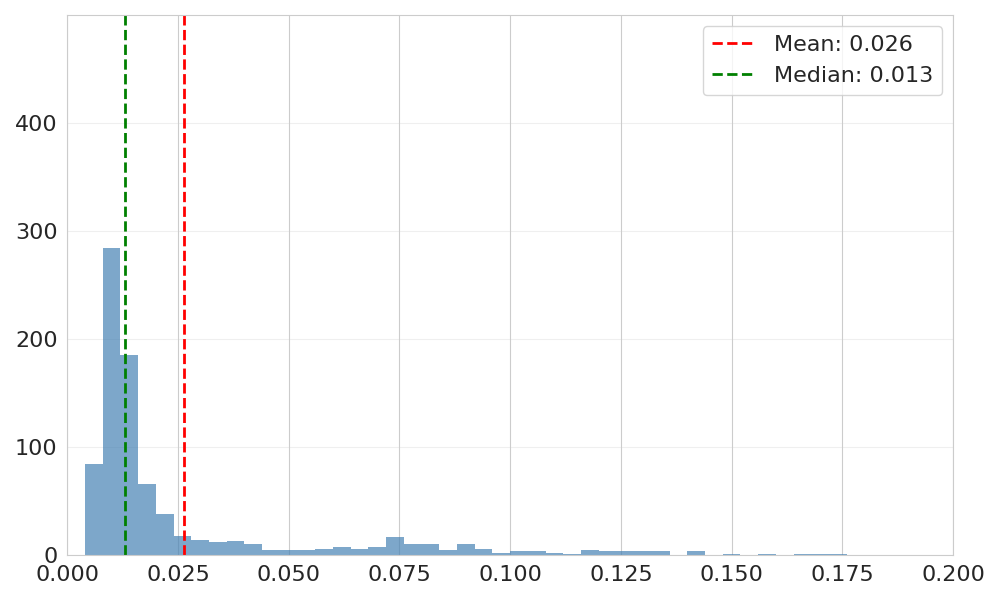}
& \includegraphics[width=\distrocolumnwidth]{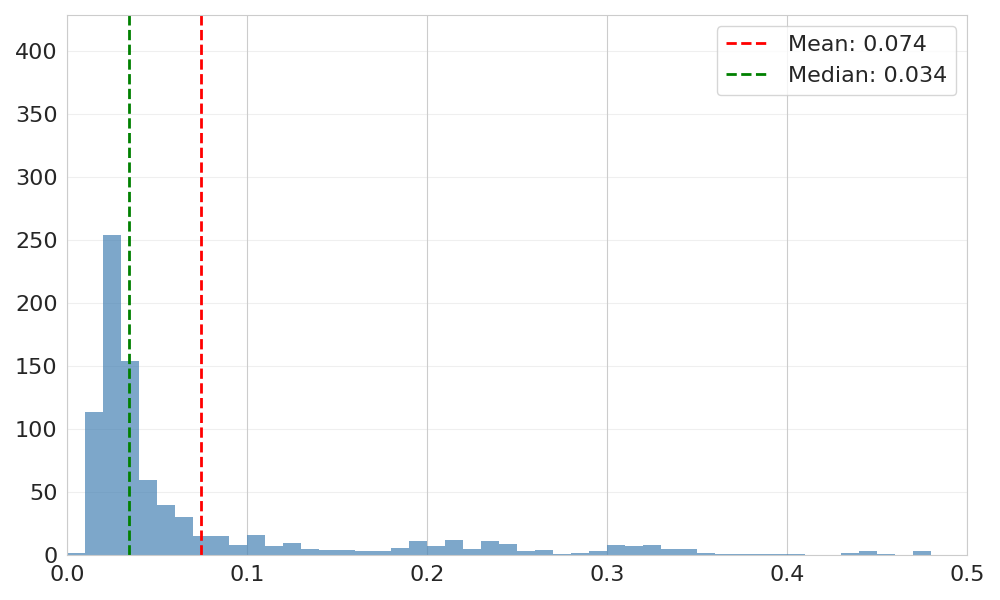}
& \includegraphics[width=\distrocolumnwidth]{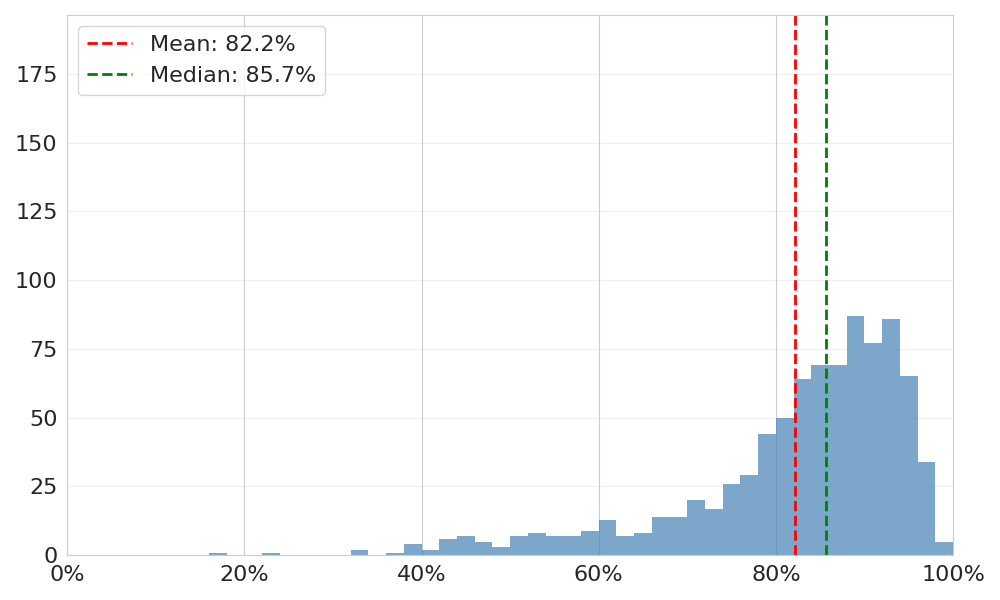}
& \includegraphics[width=\distrocolumnwidth]{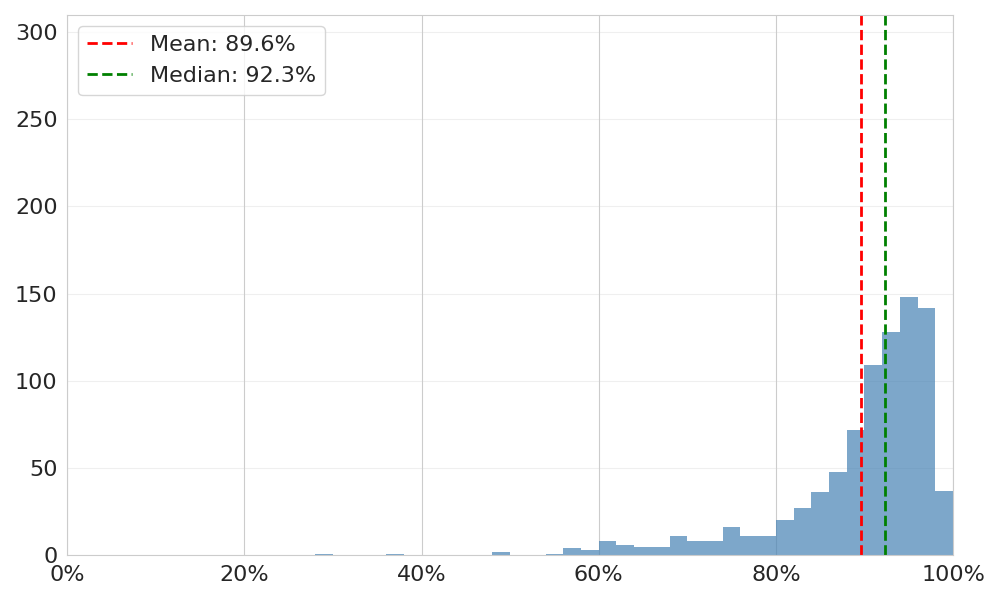} \\

\centering Trellis-RePaint
& \includegraphics[width=\distrocolumnwidth]{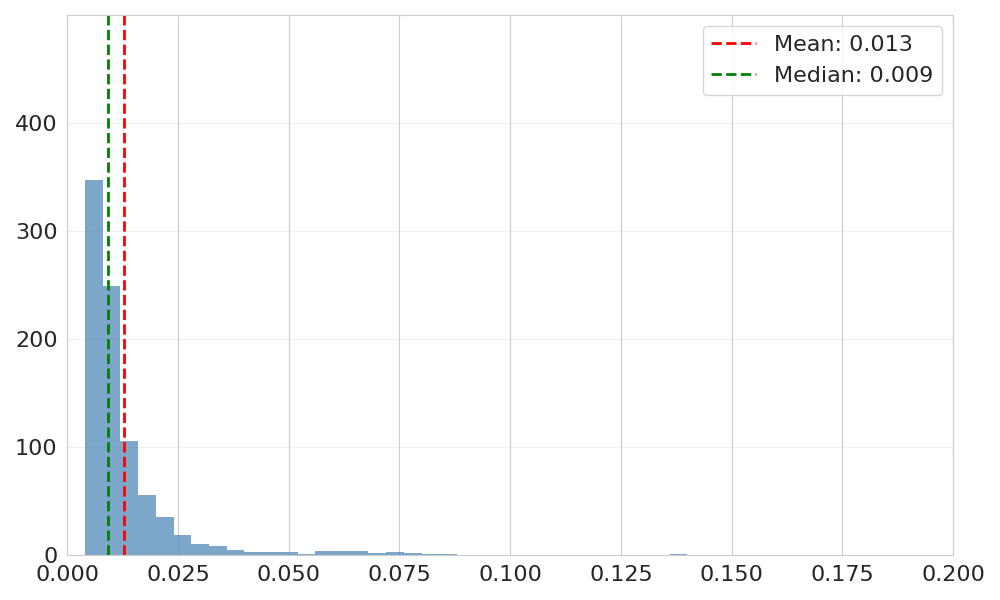}
& \includegraphics[width=\distrocolumnwidth]{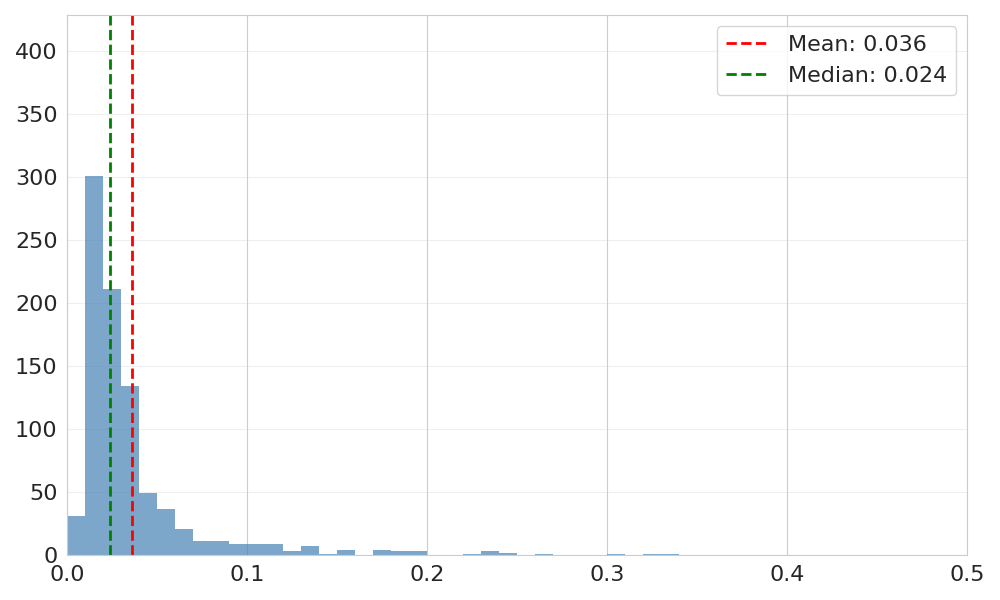}
& \includegraphics[width=\distrocolumnwidth]{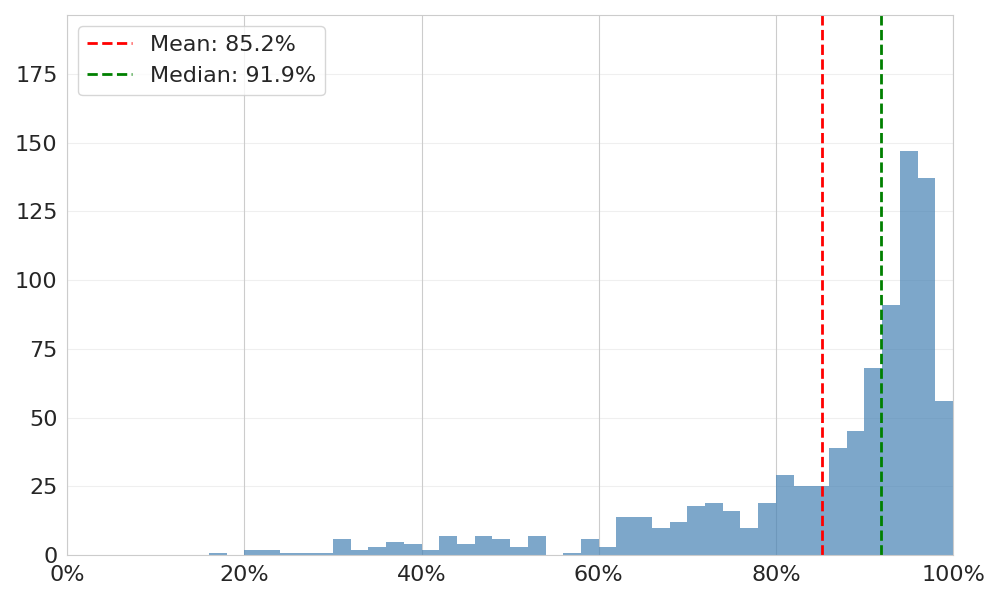}
& \includegraphics[width=\distrocolumnwidth]{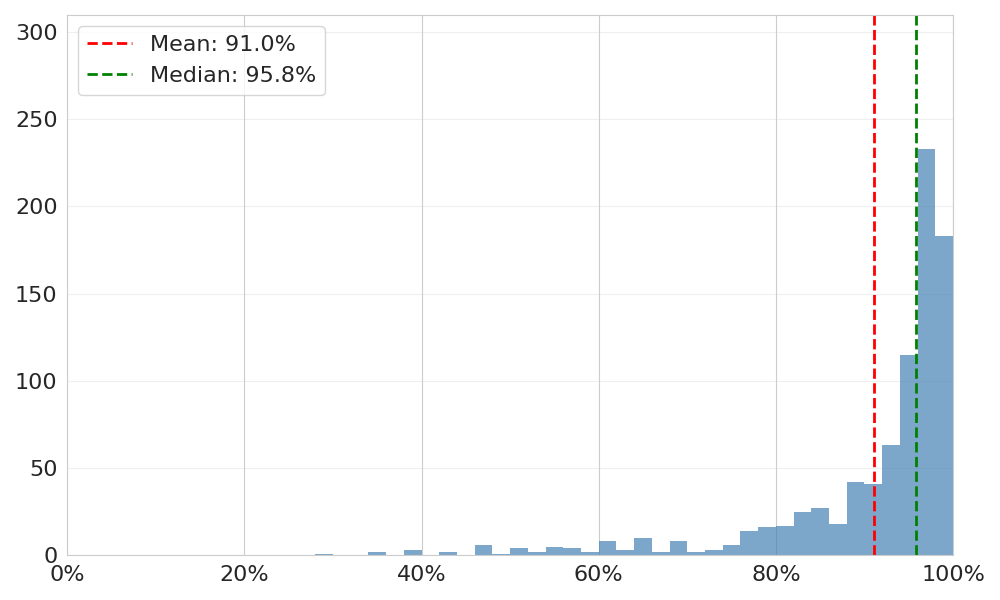} \\

\centering 3DMorph 
& \includegraphics[width=\distrocolumnwidth]{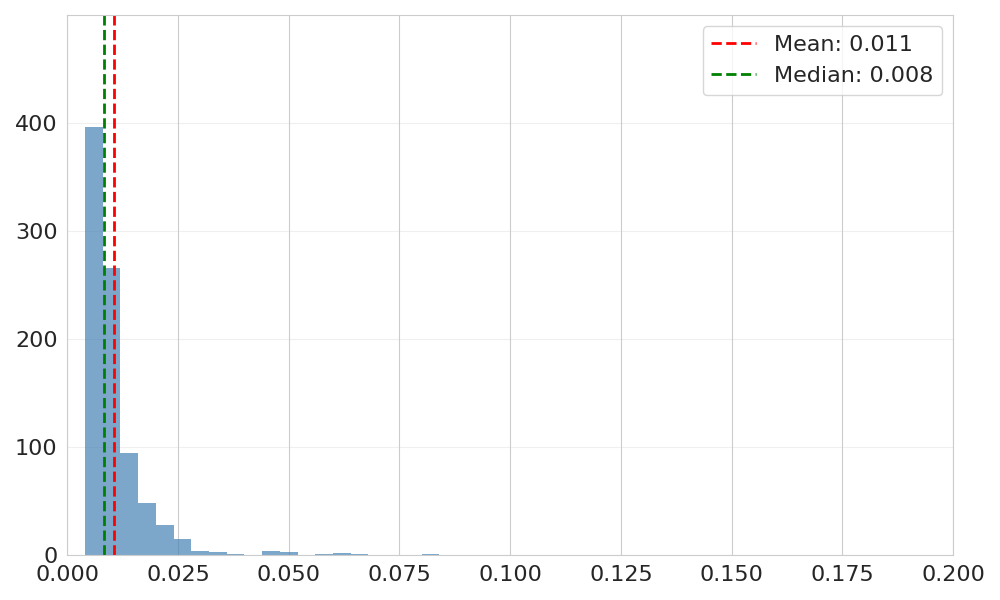}
& \includegraphics[width=\distrocolumnwidth]{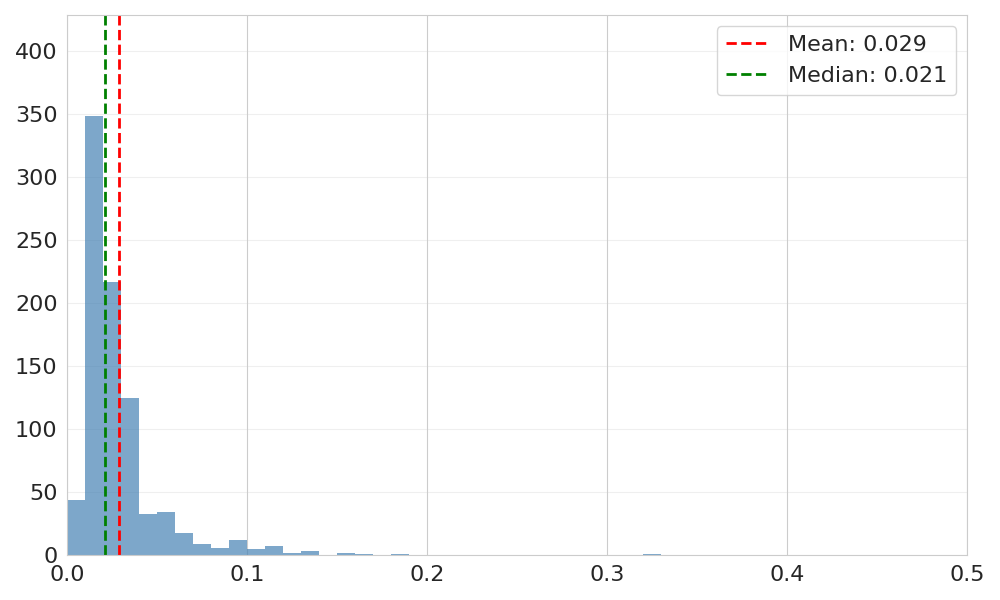}
& \includegraphics[width=\distrocolumnwidth]{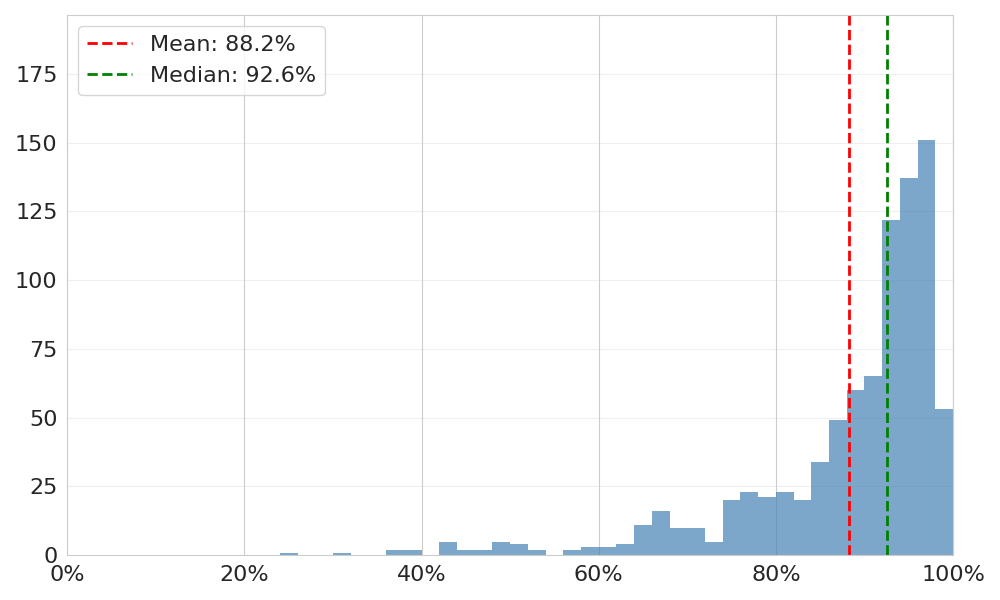}
& \includegraphics[width=\distrocolumnwidth]{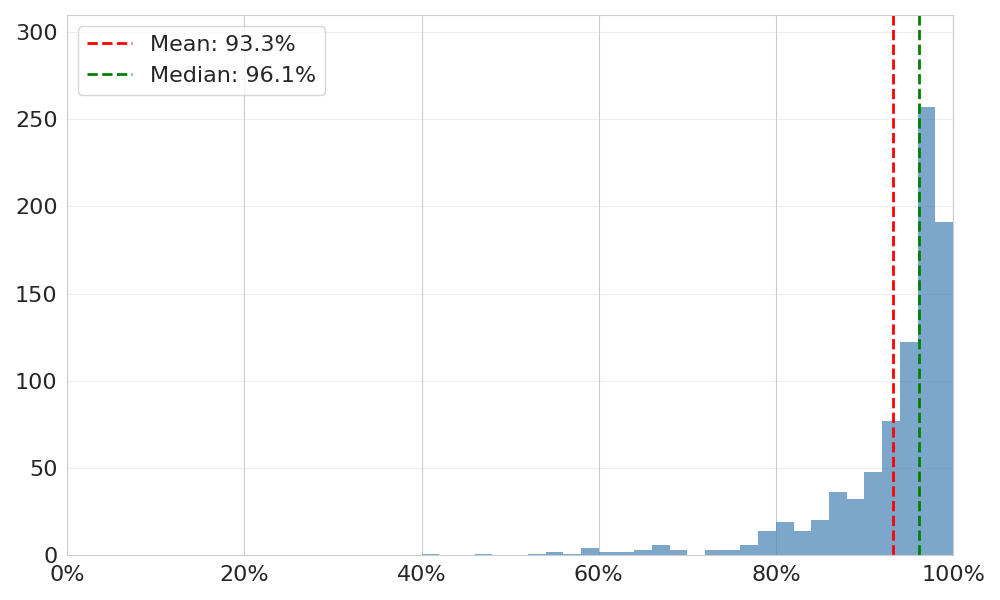} \\

\bottomrule

\end{tabular}
\captionof{figure}{Full per-sample performance distributions for the \textit{geometric metrics} reported in the main \textbf{quantitative comparison} (Tab.~\ref{table:quantitative}).}
\label{suppl-fig:distro-main-geo}
\end{center}

    \vspace{0.8cm} 
}]

\section{Metric Distribution Plots} 
Fig.~\ref{suppl-fig:distro-main-geo}, Fig.~\ref{suppl-fig:distro-main-visual}, Fig.~\ref{suppl-fig:distro-ablation-geo}, Fig.~\ref{suppl-fig:distro-ablation-visual}, and Fig.~\ref{suppl-fig:distro-edit-eval} present the full per-sample performance distributions for each metric and method corresponding to Tab.~\ref{table:quantitative}, Tab.~\ref{table:ablation}, and Tab.~\ref{table:inside-BB} in the main paper.
Each plot shows metric values on the x-axis and the number of samples per bin on the y-axis. We use 50 uniform bins, and the y-axis limits are fixed per metric to enable consistent comparison across methods. Both the mean and the median are indicated in each histogram. 
Note that Trellis-MV uses the available views jointly and therefore produces one reconstructed object per object pair, resulting in 74 objects rather than one output per rendered view.
For comparable y-axis scaling, we count each object as 12 samples, which affects only the axis scale and not the underlying distribution.




\newcommand{\distrocolumnwidth}{3.25 cm}

\begin{figure*}[t]
\centering
\begin{tabular}{m{2.2cm}M{\distrocolumnwidth}M{\distrocolumnwidth}M{\distrocolumnwidth}}
\hline
\centering Method & MS-SSIM\up & FSIM\up & GMSD\down \\
\hline

\centering TripoSG
& \includegraphics[width=\distrocolumnwidth]{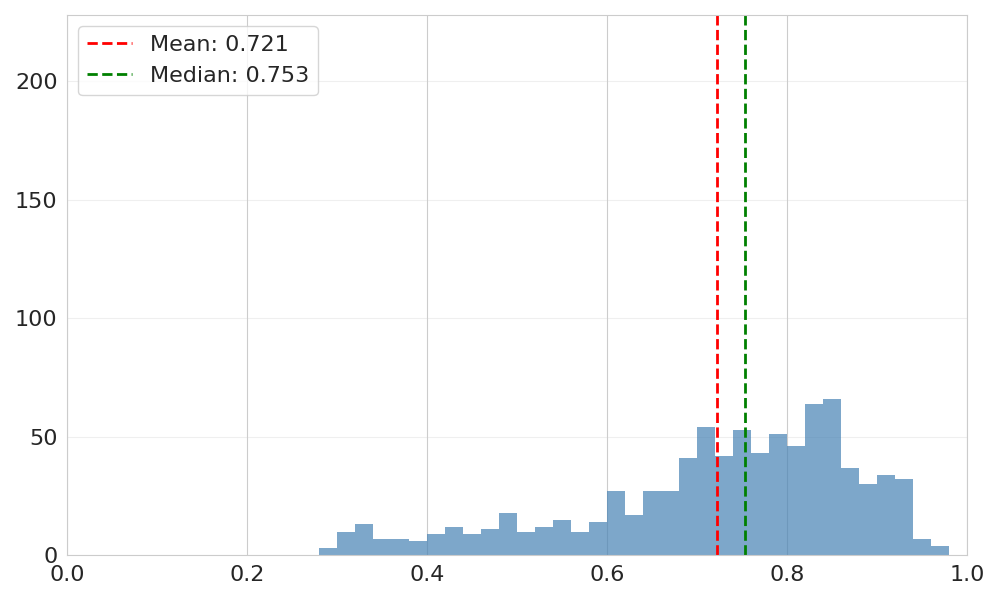}
& \includegraphics[width=\distrocolumnwidth]{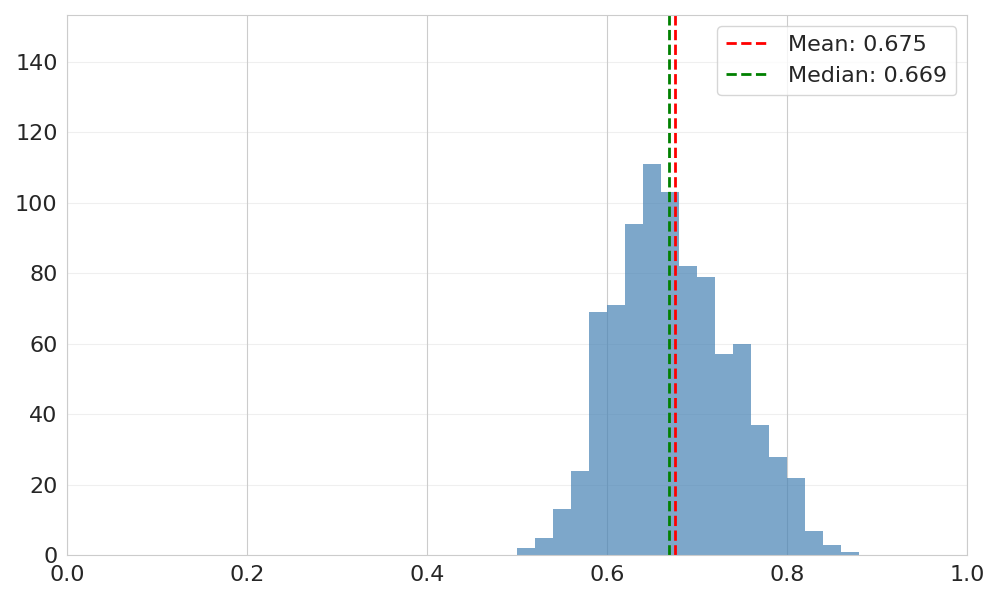}
& \includegraphics[width=\distrocolumnwidth]{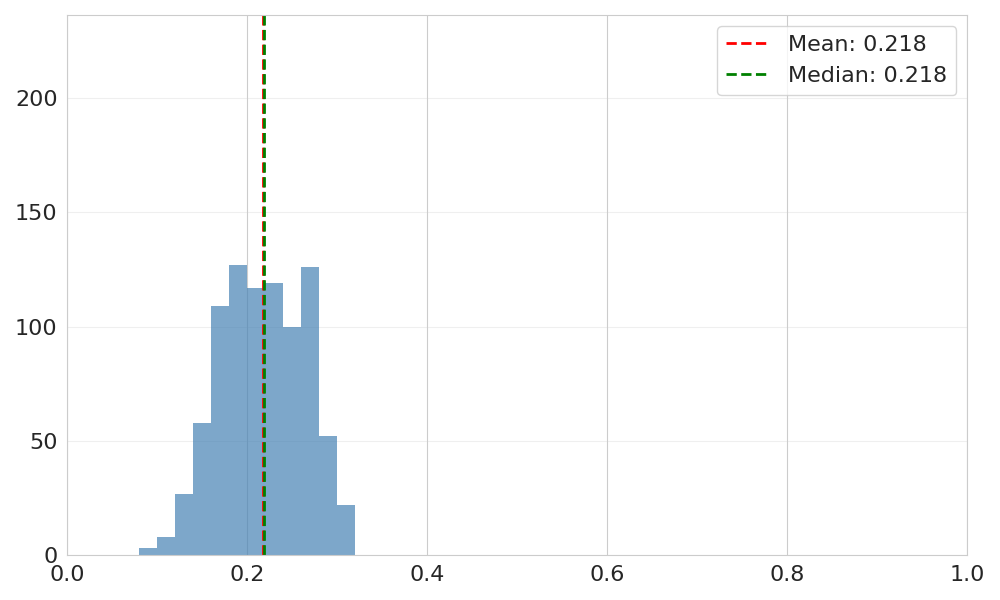} \\

\centering Trellis
& \includegraphics[width=\distrocolumnwidth]{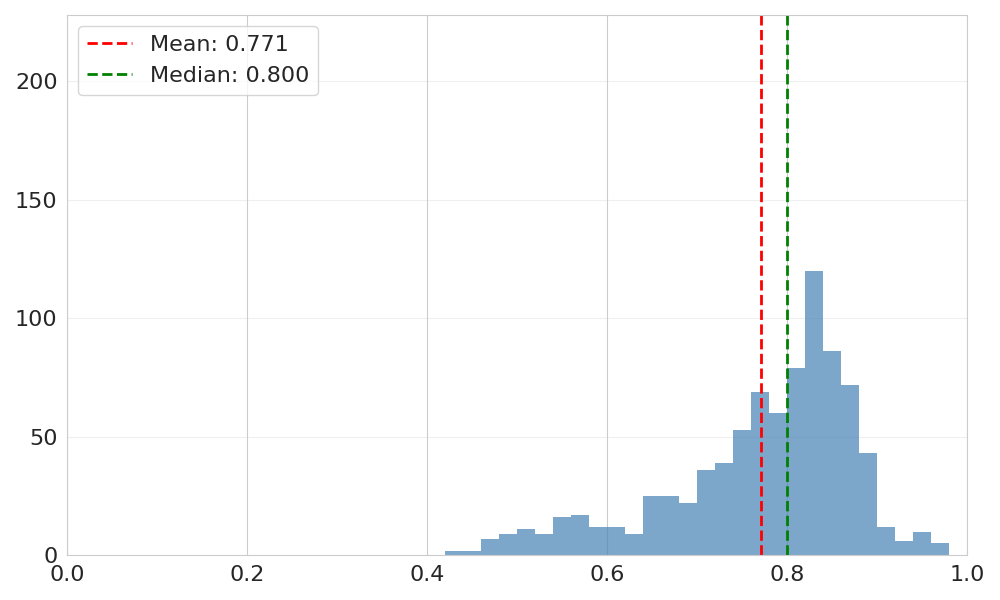}
& \includegraphics[width=\distrocolumnwidth]{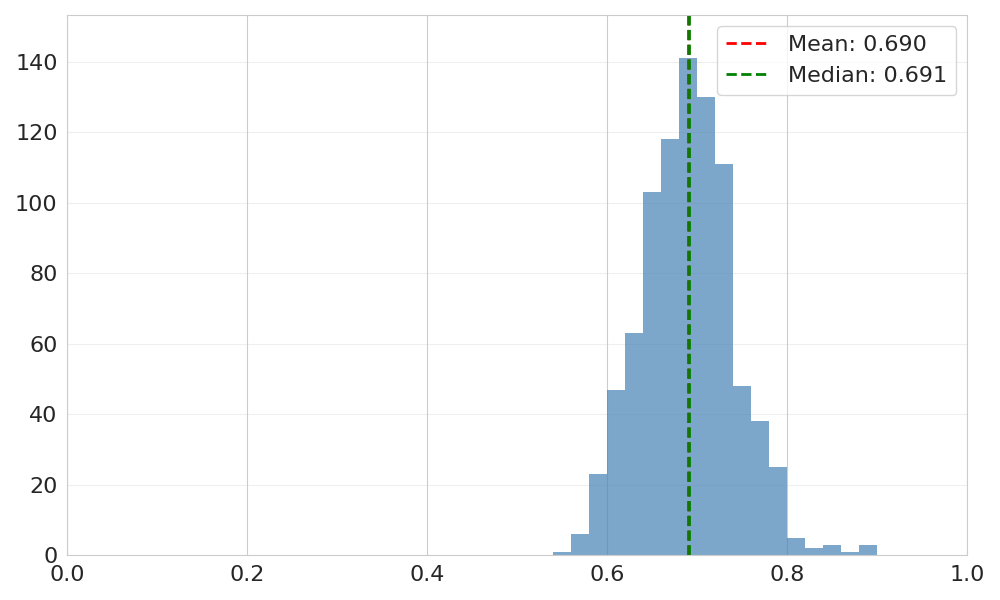}
& \includegraphics[width=\distrocolumnwidth]{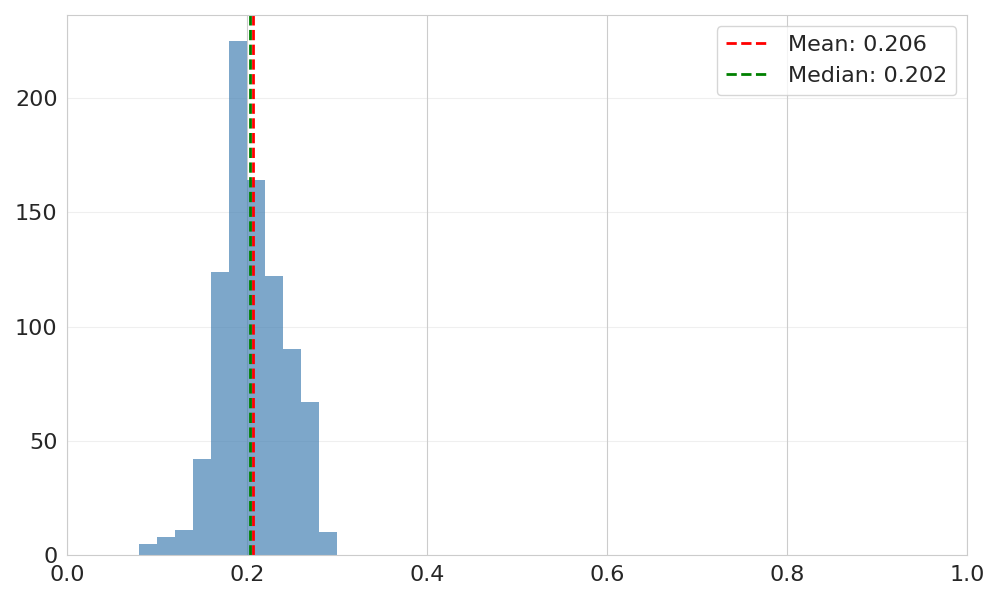} \\

\centering Trellis-MV
& \includegraphics[width=\distrocolumnwidth]{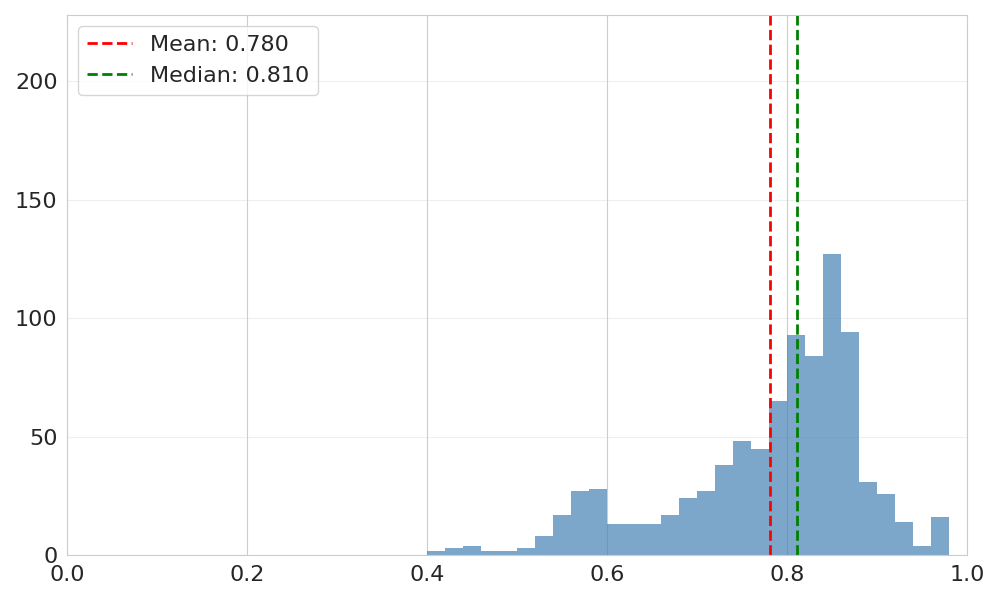}
& \includegraphics[width=\distrocolumnwidth]{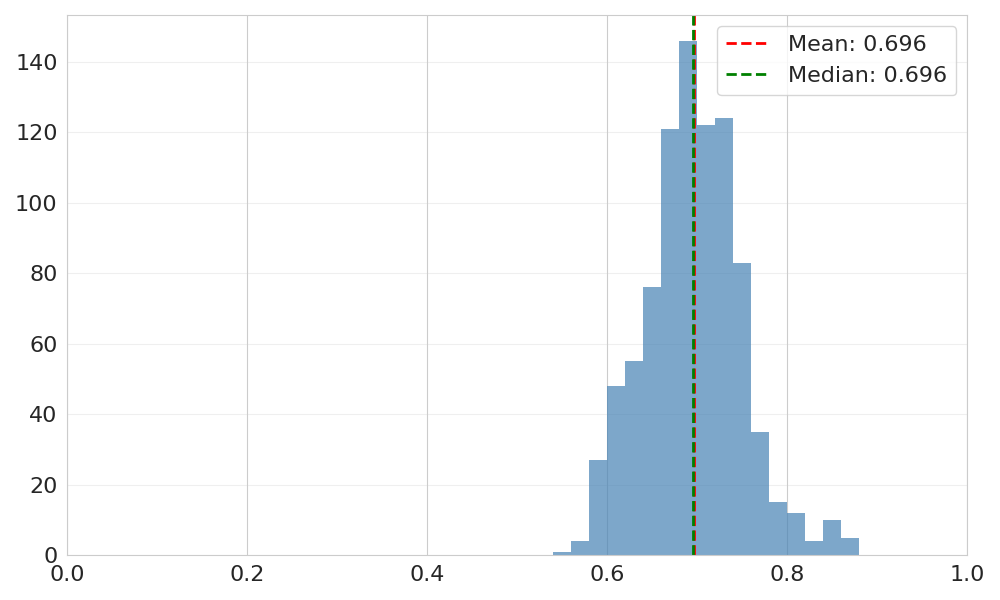}
& \includegraphics[width=\distrocolumnwidth]{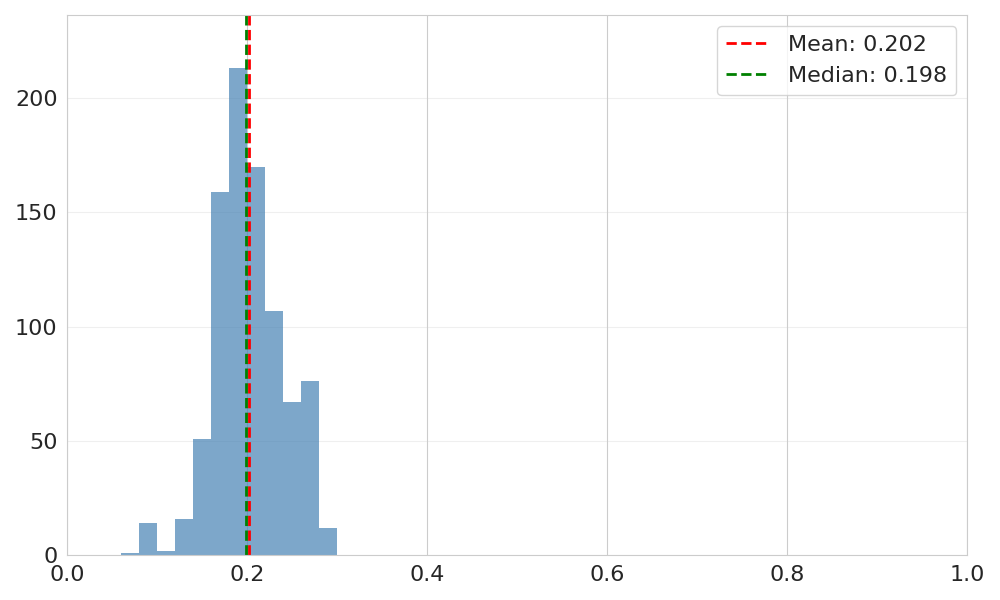} \\

\centering Hunyuan3D
& \includegraphics[width=\distrocolumnwidth]{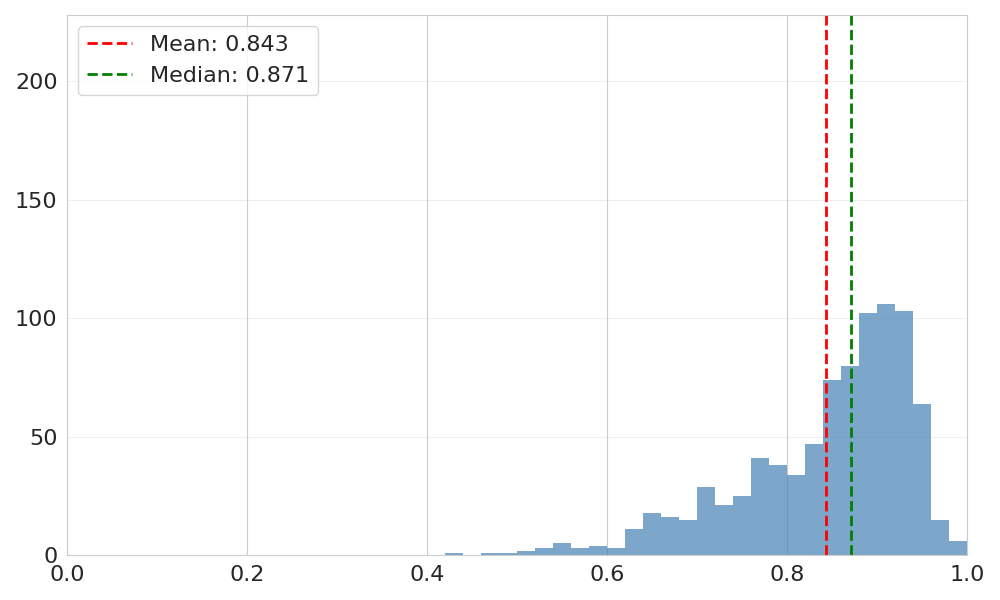}
& \includegraphics[width=\distrocolumnwidth]{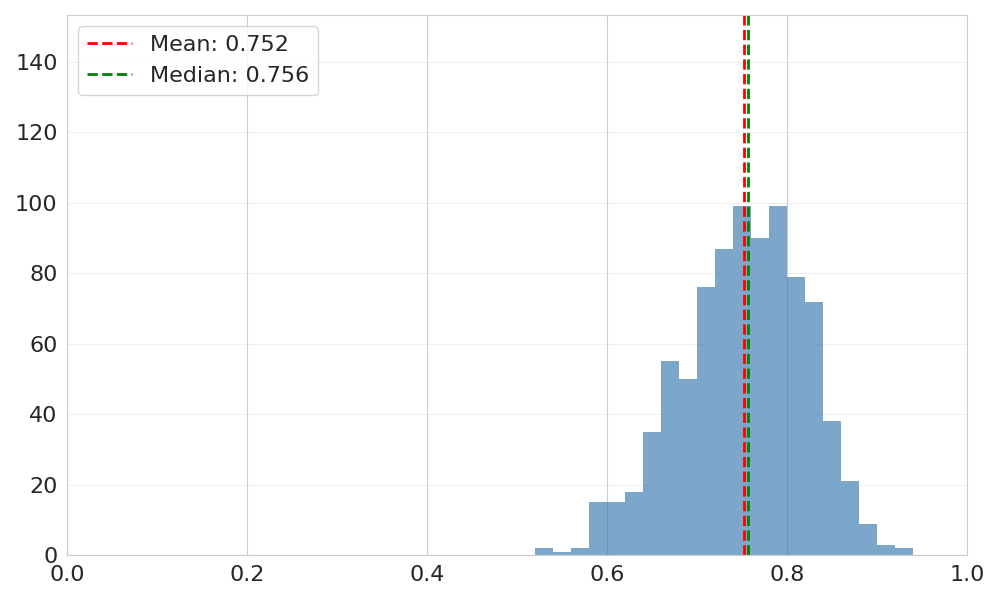}
& \includegraphics[width=\distrocolumnwidth]{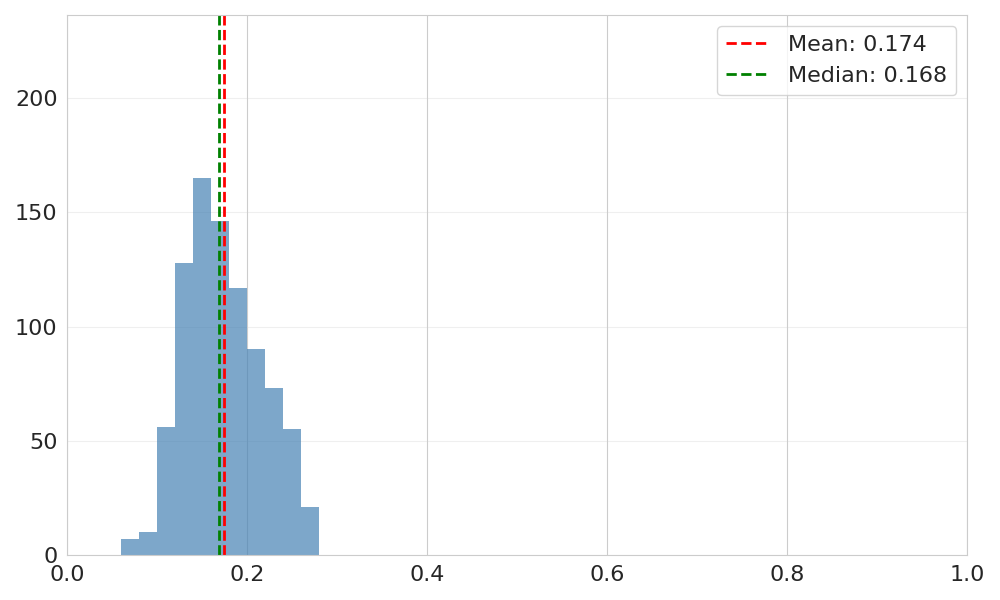} \\

\centering Trellis-RePaint
& \includegraphics[width=\distrocolumnwidth]{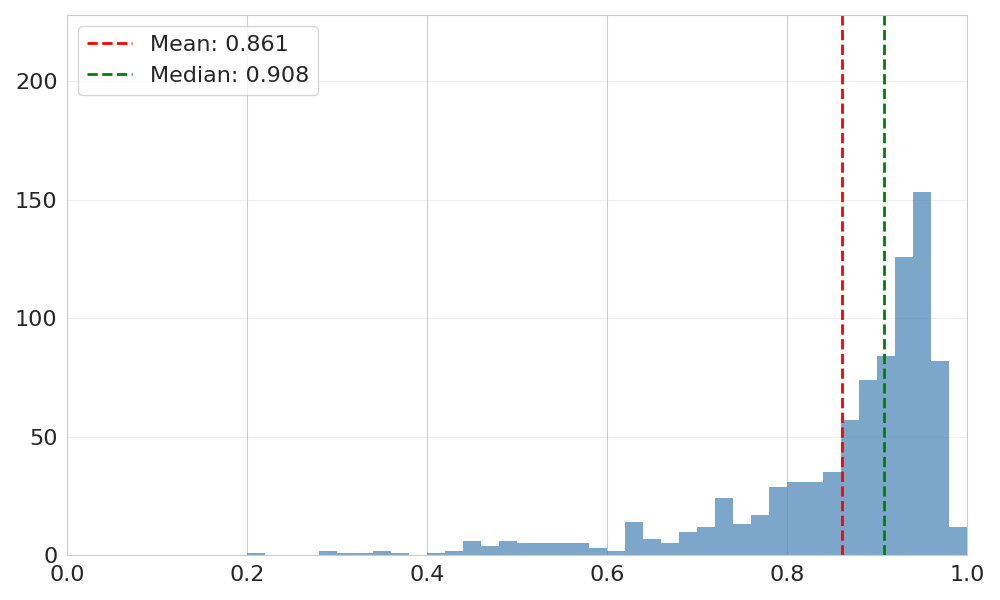}
& \includegraphics[width=\distrocolumnwidth]{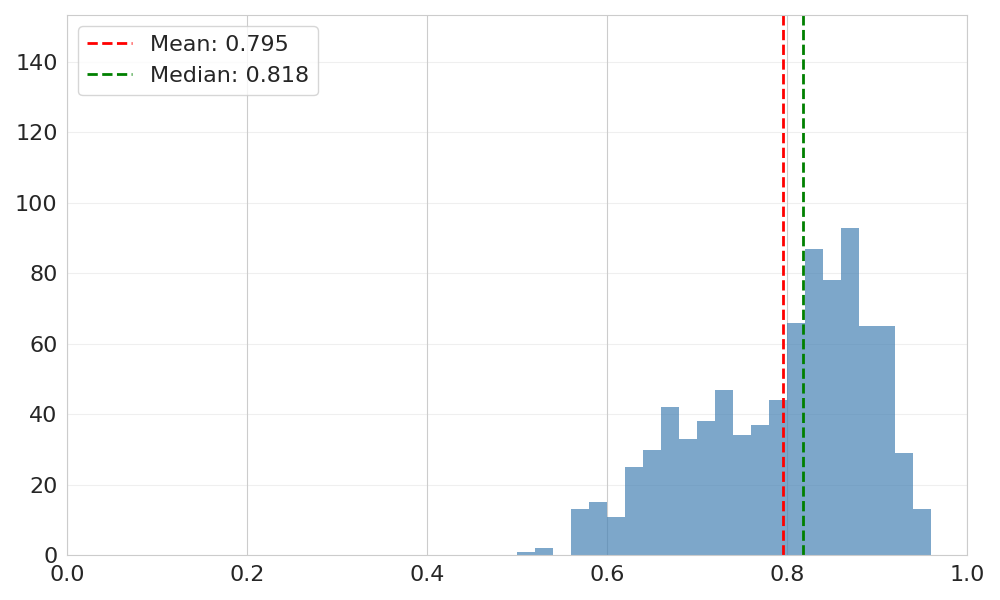}
& \includegraphics[width=\distrocolumnwidth]{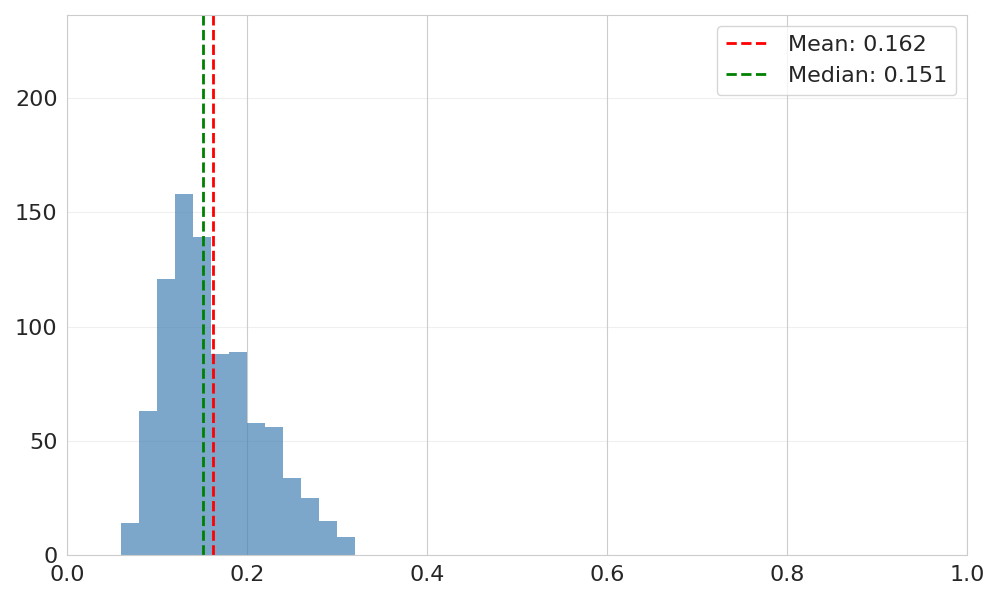} \\

\centering 3DMorph 
& \includegraphics[width=\distrocolumnwidth]{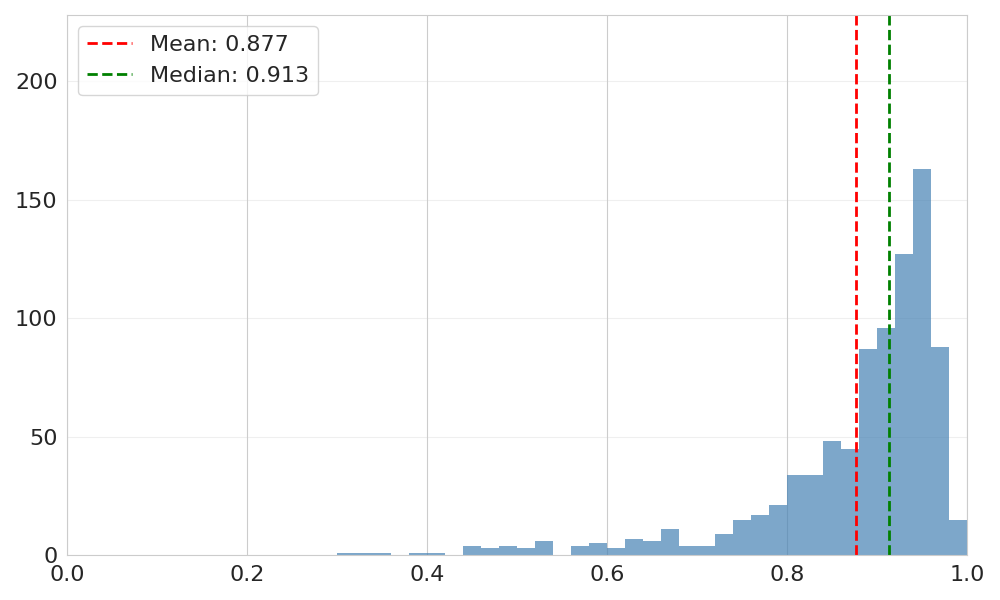}
& \includegraphics[width=\distrocolumnwidth]{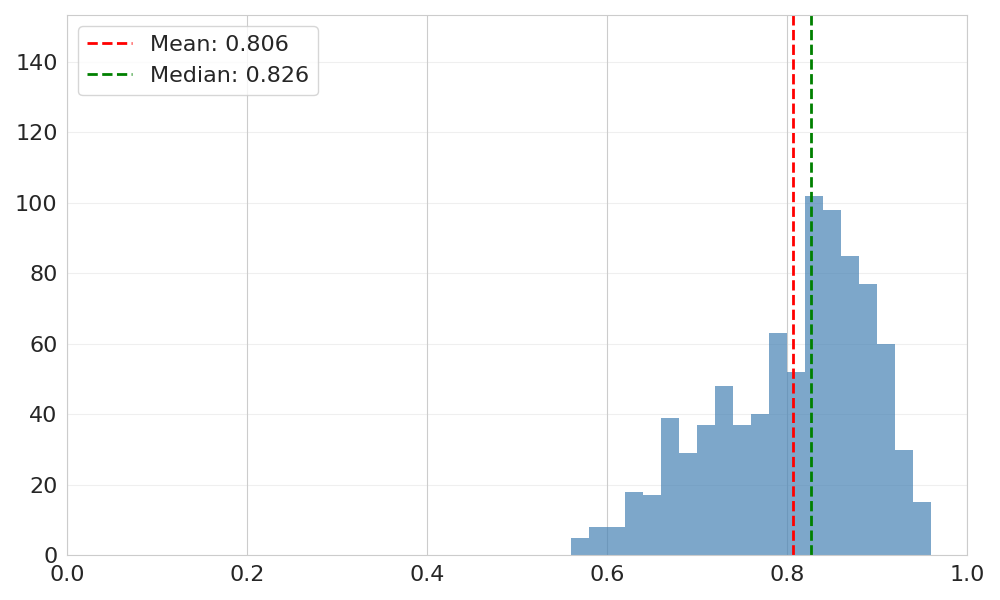}
& \includegraphics[width=\distrocolumnwidth]{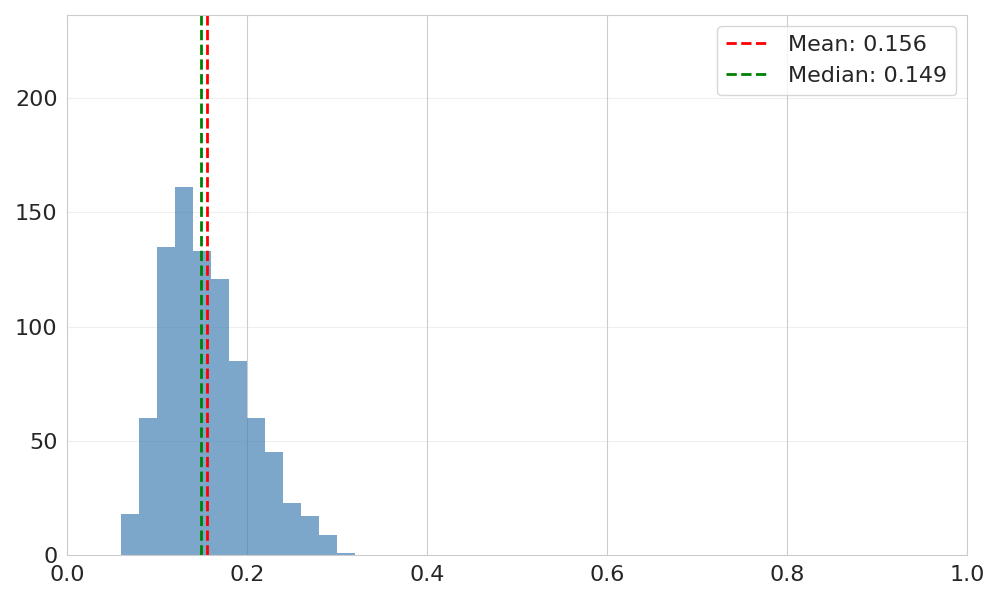} \\

\end{tabular}
\caption{Full per-sample performance distributions for the \textit{visual metrics} reported in the main \textbf{quantitative comparison} (Tab.~\ref{table:quantitative}).}
\label{suppl-fig:distro-main-visual}
\end{figure*}

\newcommand{\distrocolumnwidthablation}{3.1cm}

\begin{figure*}[!t]
\centering

\begin{tabular}{
    l                                    
    >{\centering\arraybackslash}m{1.2cm} 
    M{\distrocolumnwidthablation}                
    M{\distrocolumnwidthablation}                
    M{\distrocolumnwidthablation}                
    M{\distrocolumnwidthablation}                
}
\toprule
\centering Method & GT-BB &  CD\down & HD\down & IoU\up (\%)  & Dice\up (\%) \\
\toprule

\multirow{2}{*}[-5.0ex]{\shortstack{Trellis- \\ RePaint}}
& --
& \includegraphics[width=\distrocolumnwidthablation]{supplementary/distribution-v4/cd/Repaint.png}
& \includegraphics[width=\distrocolumnwidthablation]{supplementary/distribution-v4/hd/Repaint.png}
& \includegraphics[width=\distrocolumnwidthablation]{supplementary/distribution-v4/lou/Repaint.png}
& \includegraphics[width=\distrocolumnwidthablation]{supplementary/distribution-v4/dice/Repaint.png} \\

& \checkmark
& \includegraphics[width=\distrocolumnwidthablation]{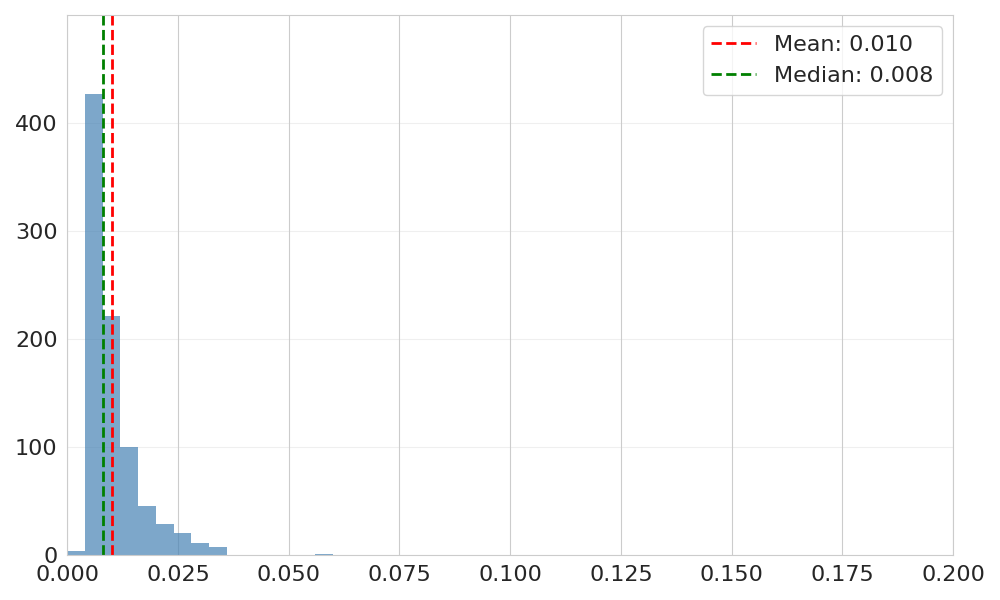}
& \includegraphics[width=\distrocolumnwidthablation]{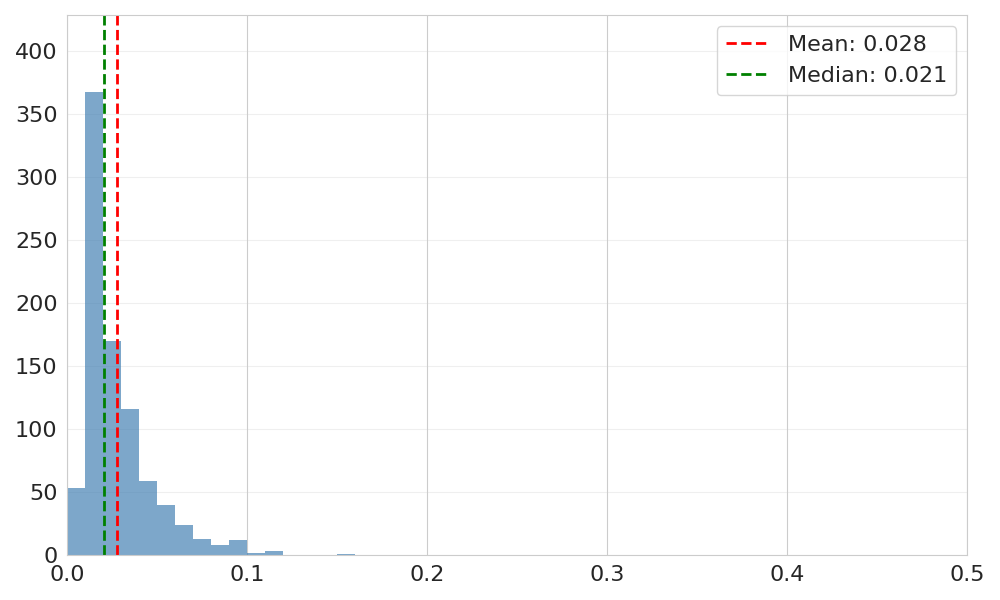}
& \includegraphics[width=\distrocolumnwidthablation]{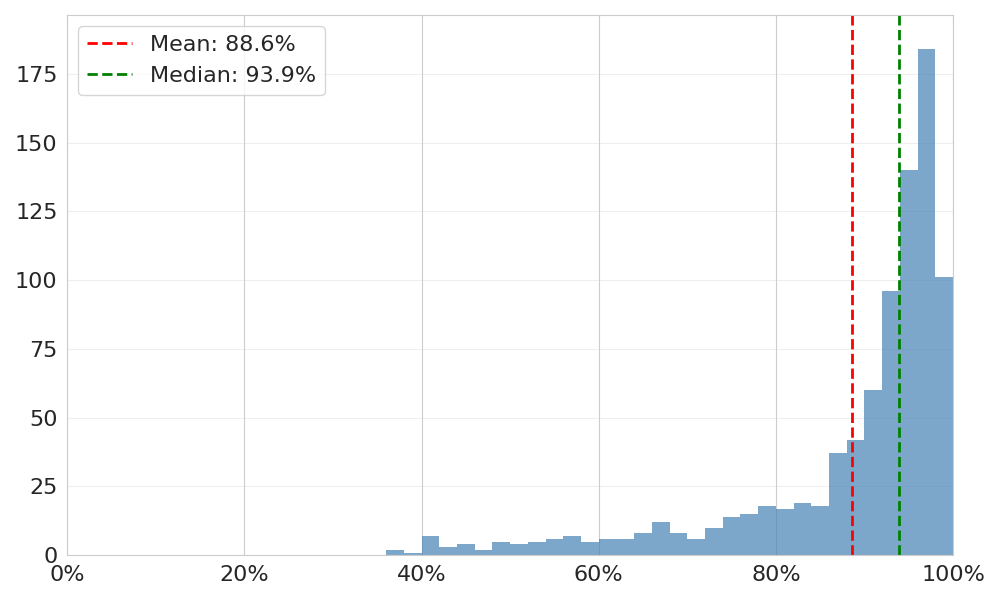}
& \includegraphics[width=\distrocolumnwidthablation]{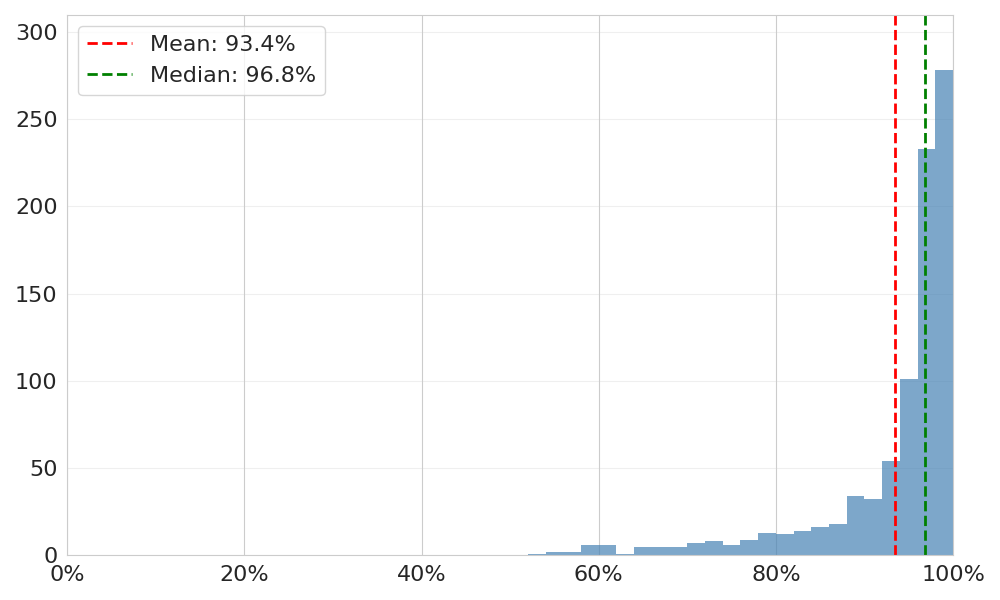} \\
\midrule
\multirow{2}{*}[-5.0ex]{3DMorph}
& --
& \includegraphics[width=\distrocolumnwidthablation]{supplementary/distribution-v4/cd/3DMorph.png}
& \includegraphics[width=\distrocolumnwidthablation]{supplementary/distribution-v4/hd/3DMorph.png}
& \includegraphics[width=\distrocolumnwidthablation]{supplementary/distribution-v4/lou/3DMorph.png}
& \includegraphics[width=\distrocolumnwidthablation]{supplementary/distribution-v4/dice/3DMorph.png} \\

& \checkmark
& \includegraphics[width=\distrocolumnwidthablation]{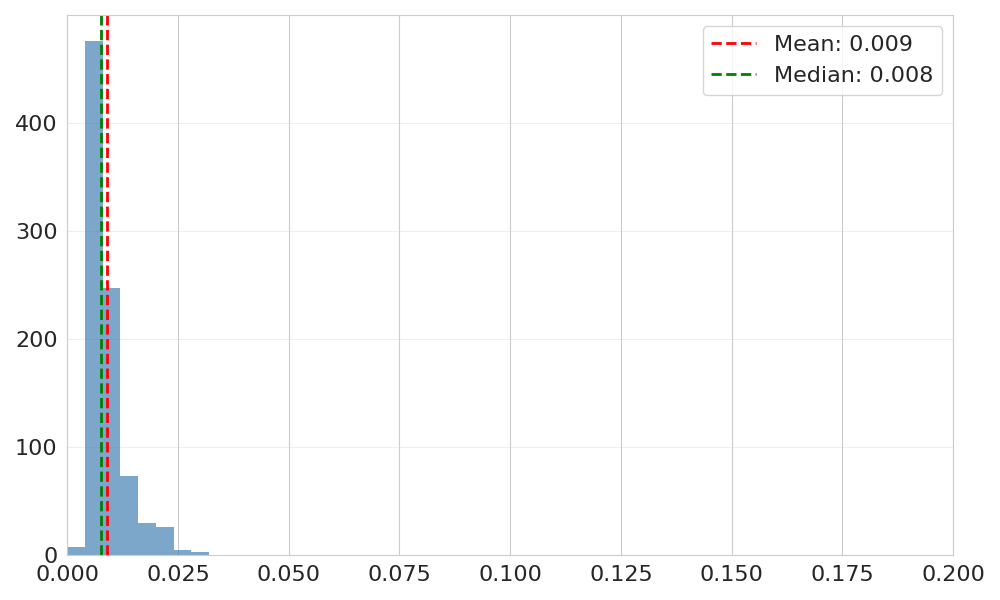}
& \includegraphics[width=\distrocolumnwidthablation]{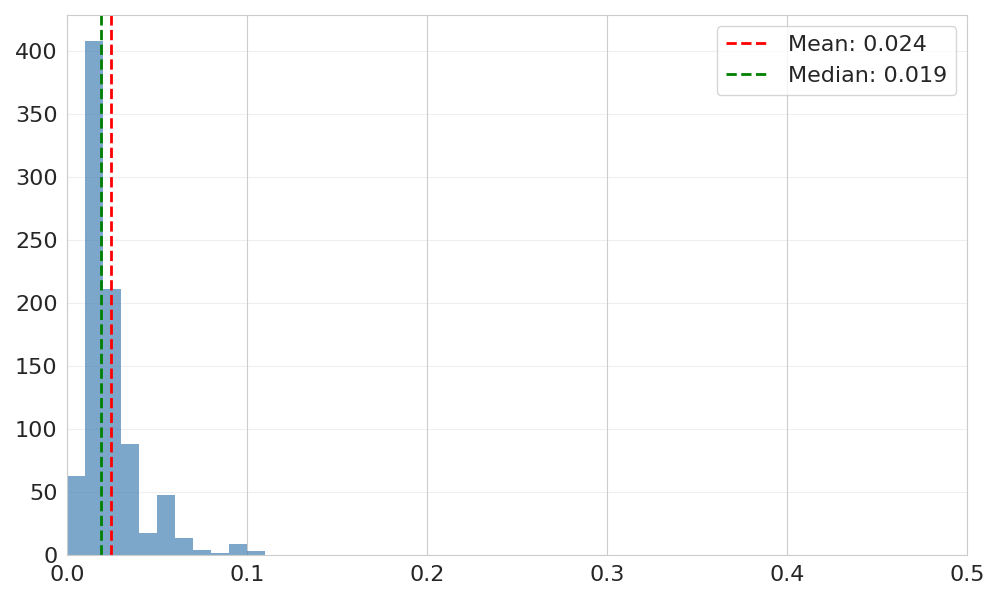}
& \includegraphics[width=\distrocolumnwidthablation]{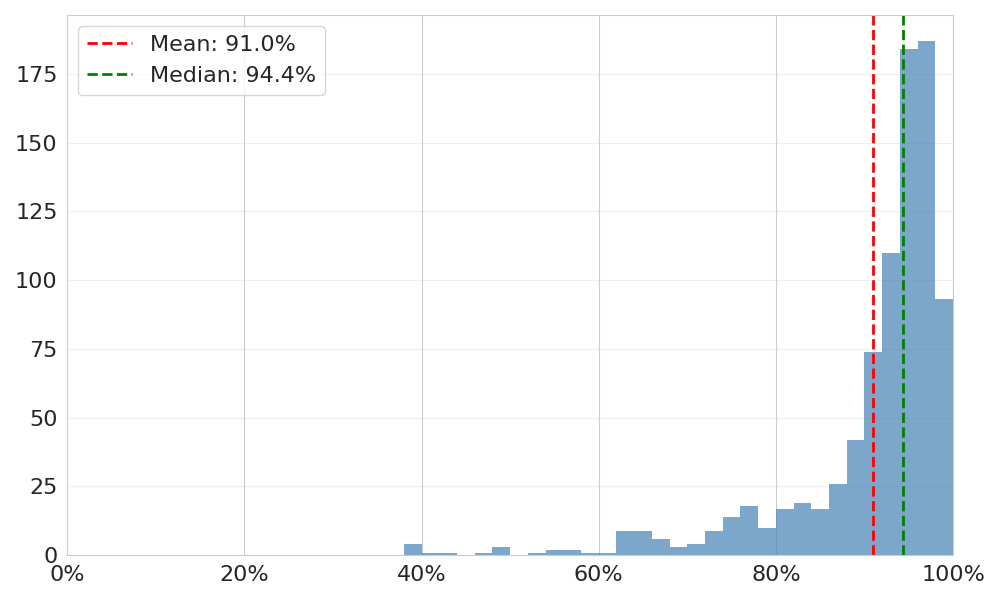}
& \includegraphics[width=\distrocolumnwidthablation]{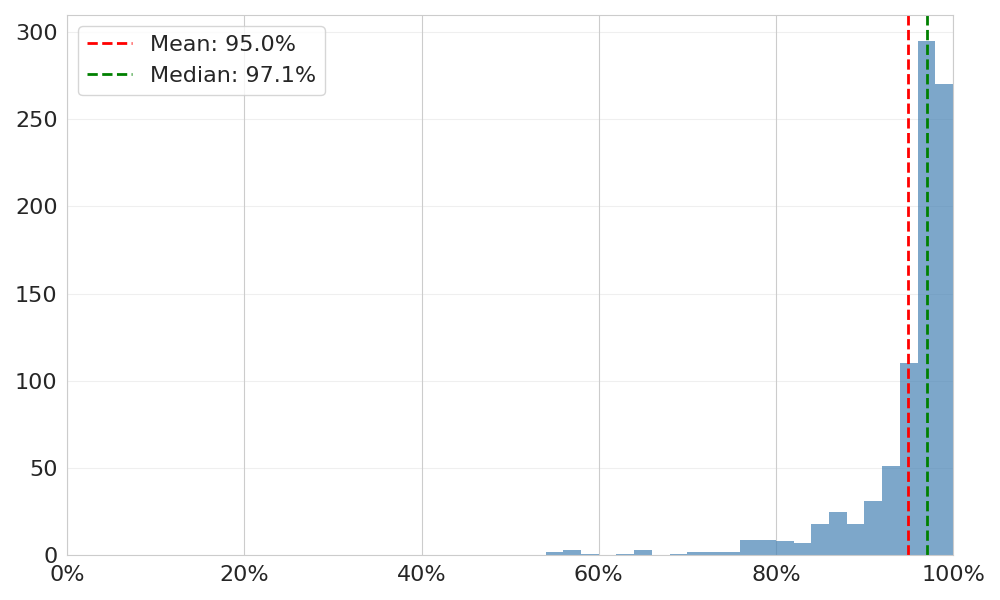} \\
\bottomrule

\end{tabular}

\caption{Full per-sample performance distributions for the \textit{geometric metrics} reported in the \textbf{ablation study} (Tab.~\ref{table:ablation}).}
\label{suppl-fig:distro-ablation-geo}
\end{figure*}

\newcommand{\distrocolumnwidthablationimg}{3.1cm}

\begin{figure*}[hbt]
\centering
\begin{tabular}{
    l                                    
    >{\centering\arraybackslash}m{1.2cm} 
    M{\distrocolumnwidthablationimg}                
    M{\distrocolumnwidthablationimg}                
    M{\distrocolumnwidthablationimg}                
}
\toprule
\centering Method & GT-BB & MS-SSIM\up & FSIM\up & GMSD\down  \\
\toprule
\multirow{2}{*}[-5.0ex]{\shortstack{Trellis- \\ RePaint}}
& --
& \includegraphics[width=\distrocolumnwidthablationimg]{supplementary/distribution-v4/ms-ssim/Repaint.png}
& \includegraphics[width=\distrocolumnwidthablationimg]{supplementary/distribution-v4/fsim/3DMorph.png}
& \includegraphics[width=\distrocolumnwidthablationimg]{supplementary/distribution-v4/gmsd/Repaint.png}\\

& \checkmark
& \includegraphics[width=\distrocolumnwidthablationimg]{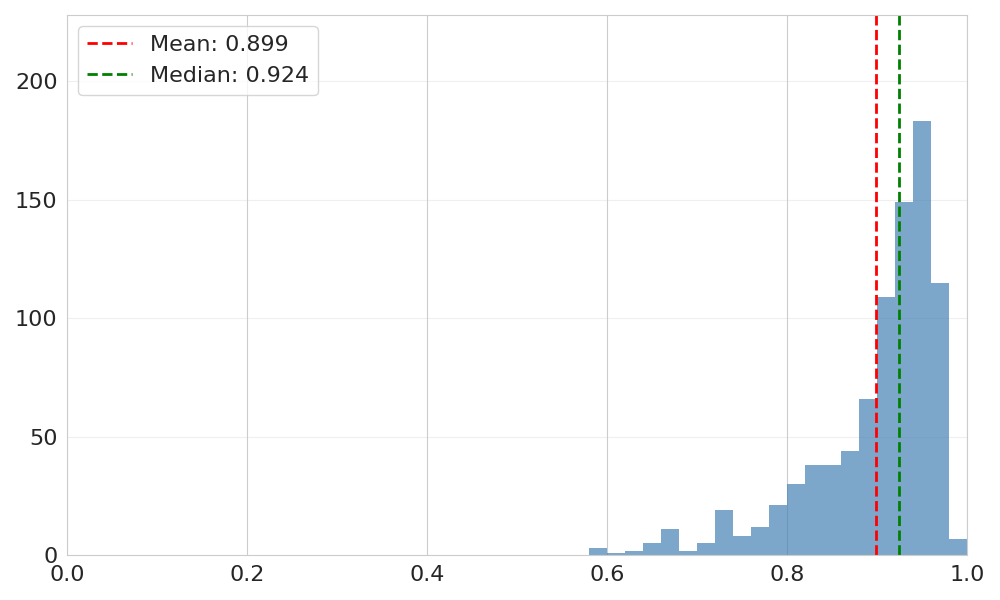}
& \includegraphics[width=\distrocolumnwidthablationimg]{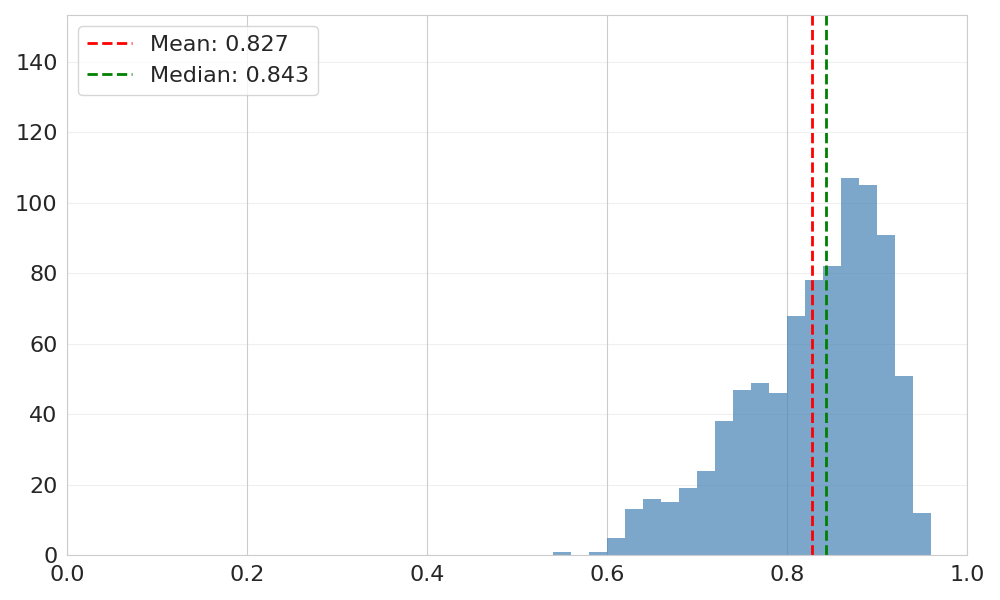}
& \includegraphics[width=\distrocolumnwidthablationimg]{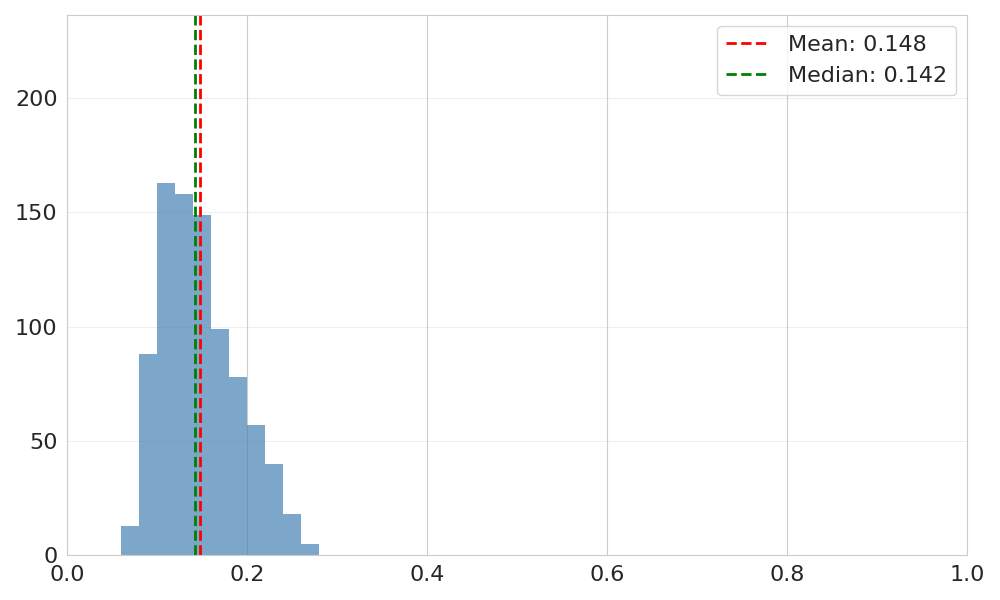} \\
\midrule

\multirow{2}{*}[-5.0ex]{3DMorph}
& --
& \includegraphics[width=\distrocolumnwidthablationimg]{supplementary/distribution-v4/ms-ssim/3DMorph.png}
& \includegraphics[width=\distrocolumnwidthablationimg]{supplementary/distribution-v4/fsim/3DMorph.png}
& \includegraphics[width=\distrocolumnwidthablationimg]{supplementary/distribution-v4/gmsd/3DMorph.png}\\

& \checkmark
& \includegraphics[width=\distrocolumnwidthablationimg]{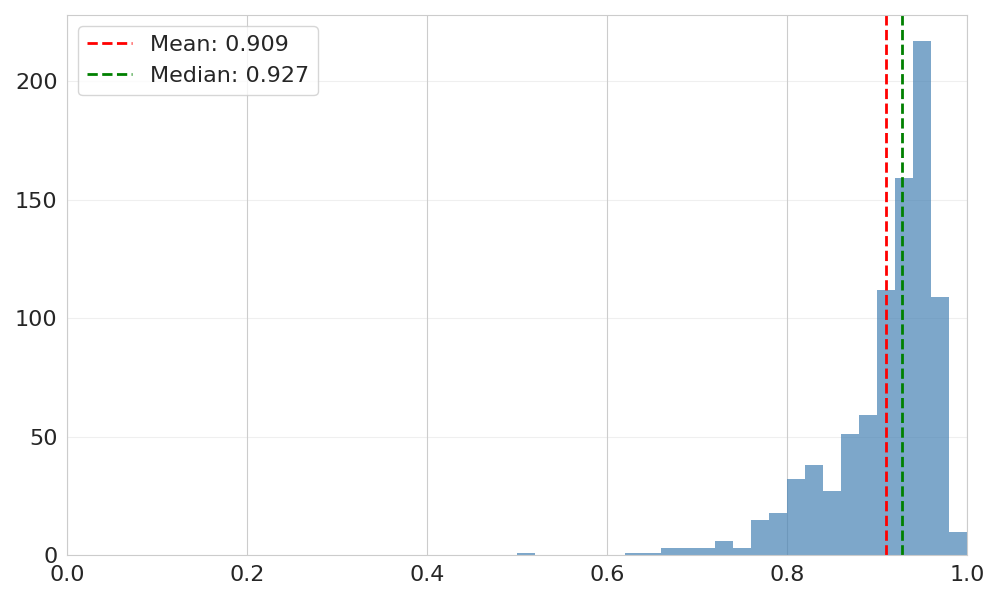}
& \includegraphics[width=\distrocolumnwidthablationimg]{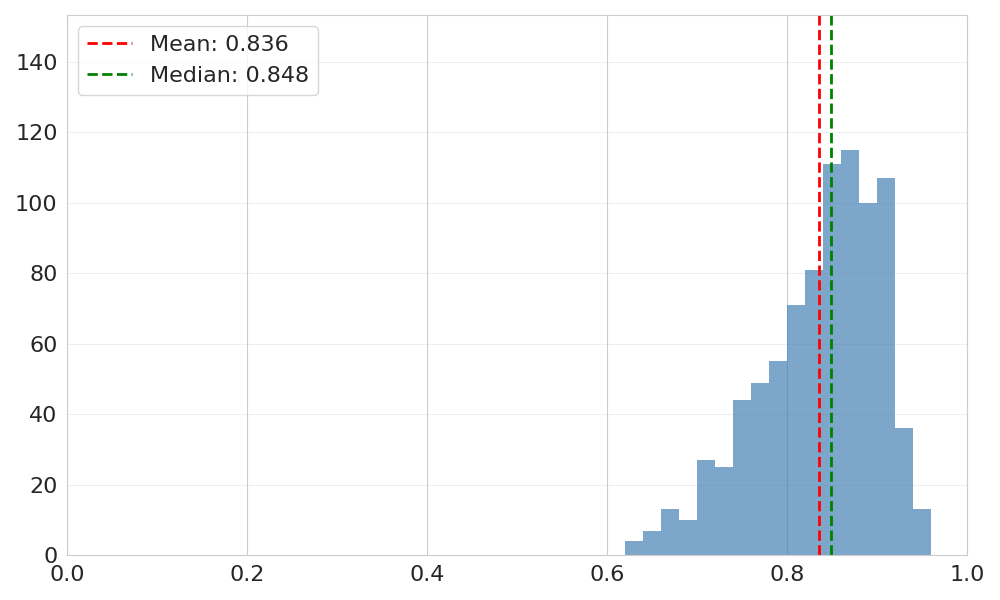}
& \includegraphics[width=\distrocolumnwidthablationimg]{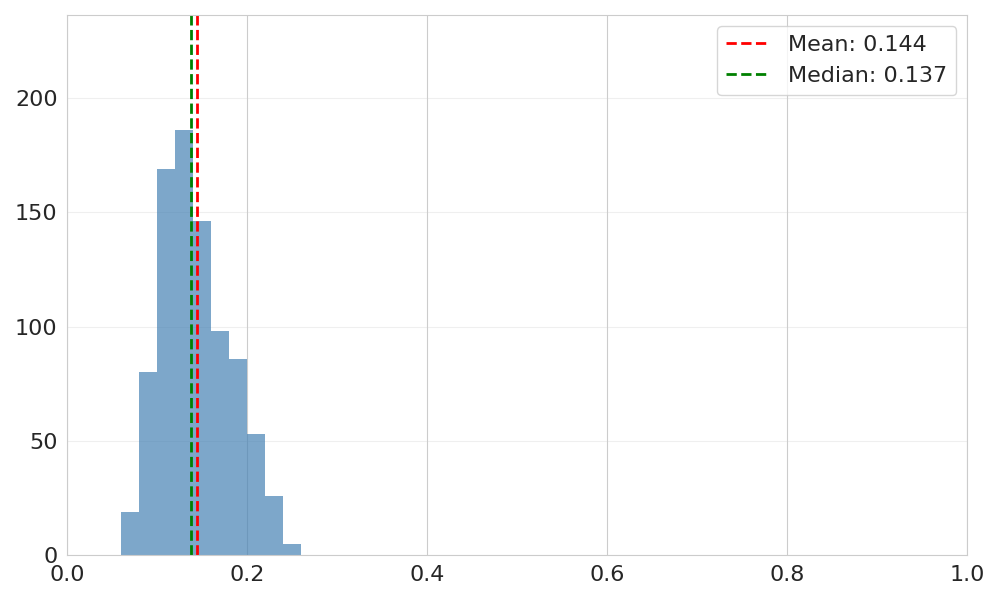}\\
\bottomrule

\end{tabular}
\caption{Full per-sample performance distributions for the \textit{visual metrics} reported in the \textbf{ablation study }(Tab.~\ref{table:ablation}).}
\label{suppl-fig:distro-ablation-visual}
\end{figure*}

\newcommand{\distroBB}{3.1cm}

\begin{figure*}[t]
\centering
\begin{tabular}{
    l                                    
    >{\centering\arraybackslash}m{1.2cm} 
    M{\distroBB}                
    M{\distroBB}                
    M{\distroBB}                
    M{\distroBB}                
}
\toprule
\centering Method & GT-BB &  CD\down & HD\down & IoU\up (\%)  & Dice\up (\%) \\
\toprule

\shortstack{Trellis- \\ RePaint}
& --
& \includegraphics[width=\distroBB]{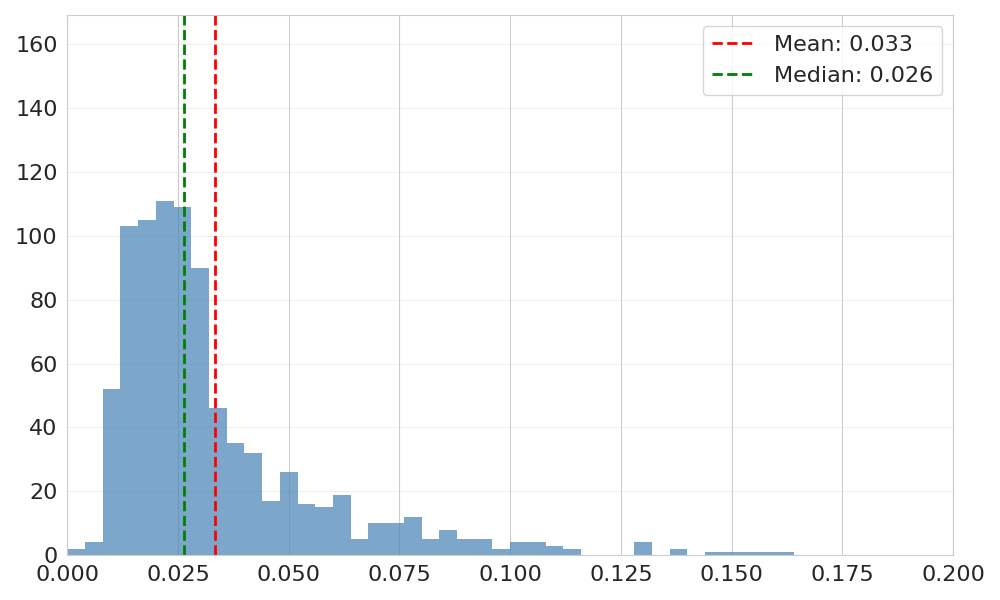}
& \includegraphics[width=\distroBB]{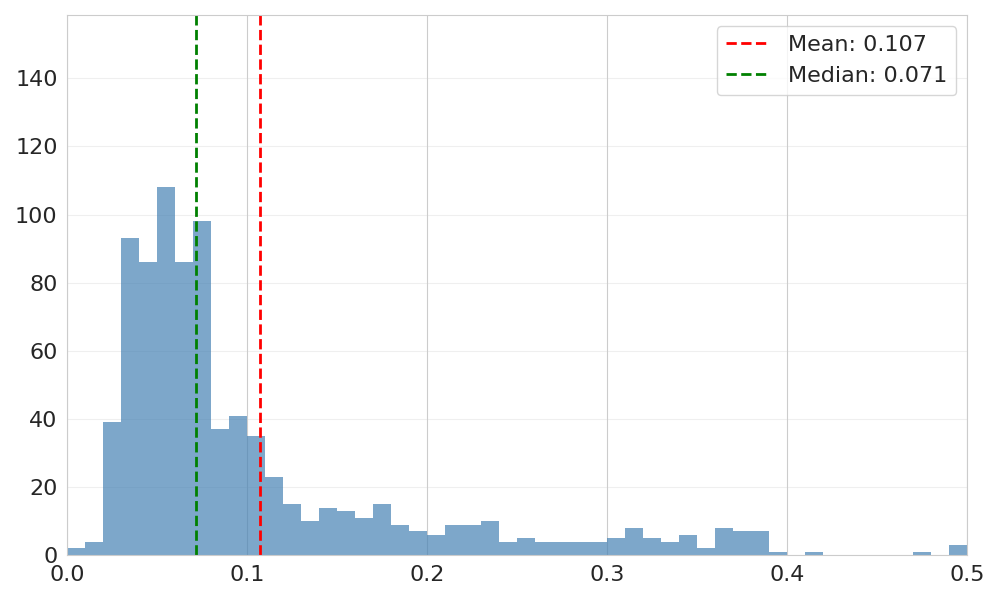}
& \includegraphics[width=\distroBB]{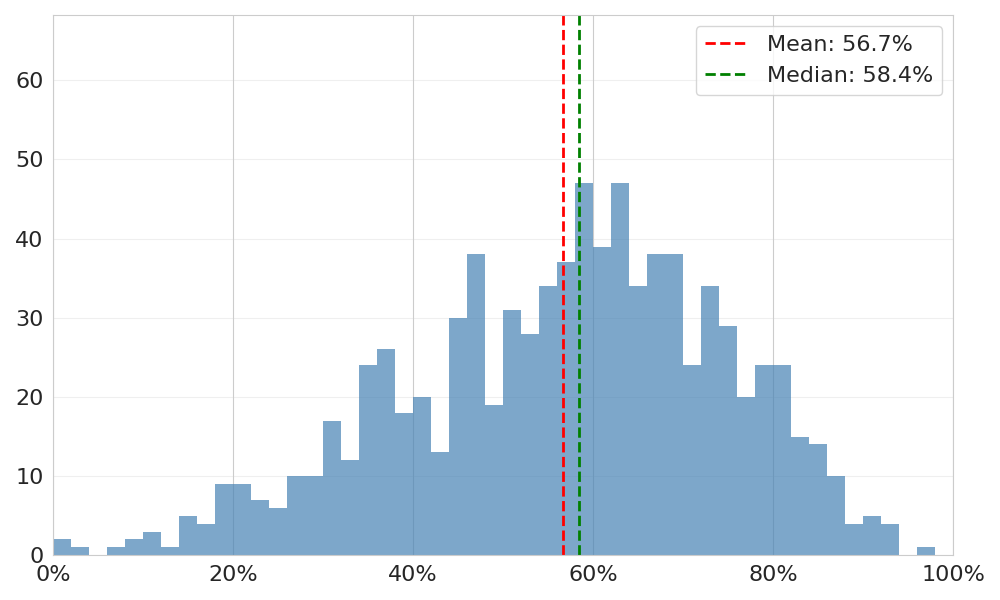}
& \includegraphics[width=\distroBB]{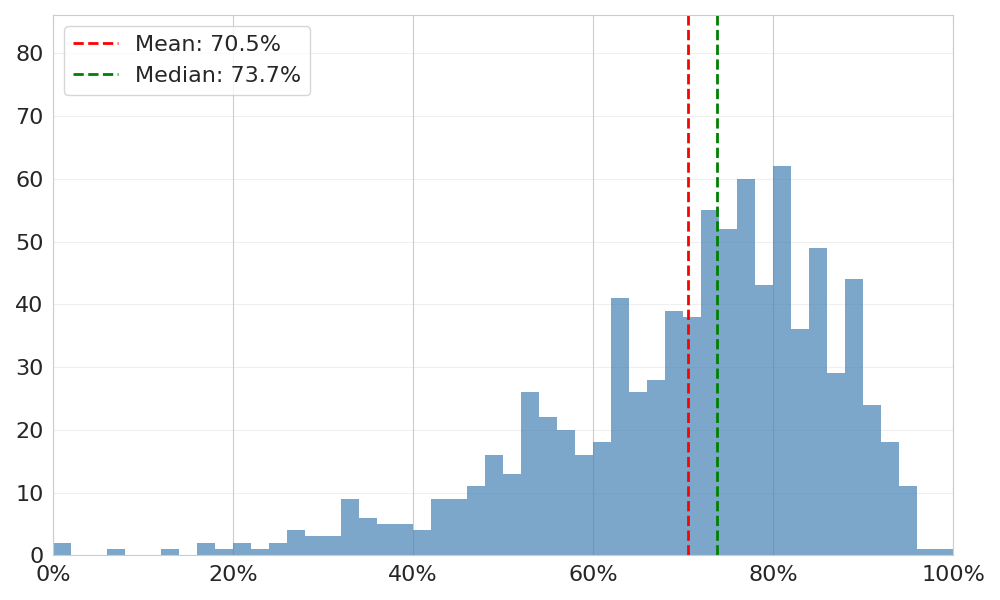} \\

3DMorph
& --
& \includegraphics[width=\distroBB]{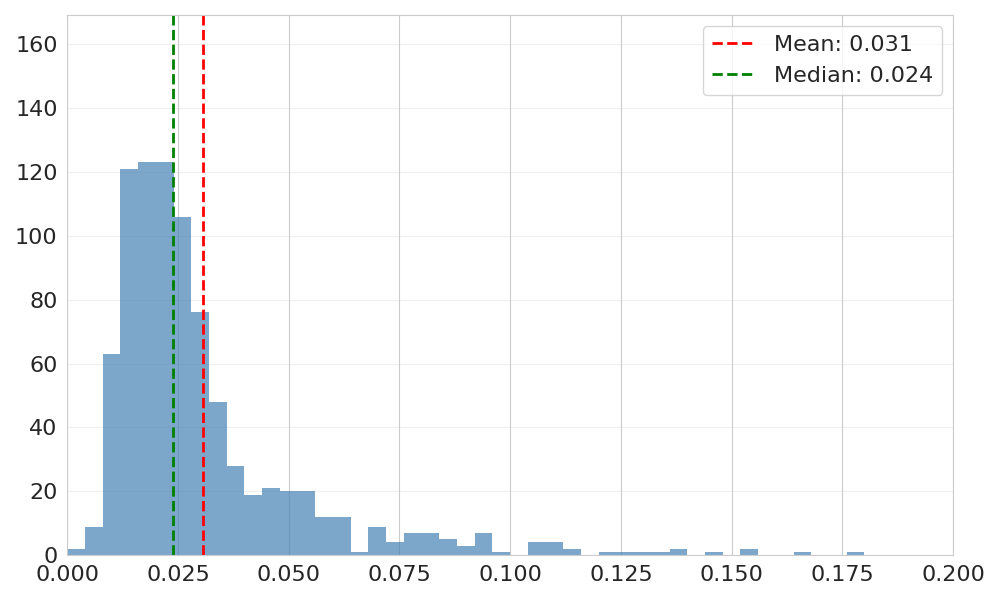}
& \includegraphics[width=\distroBB]{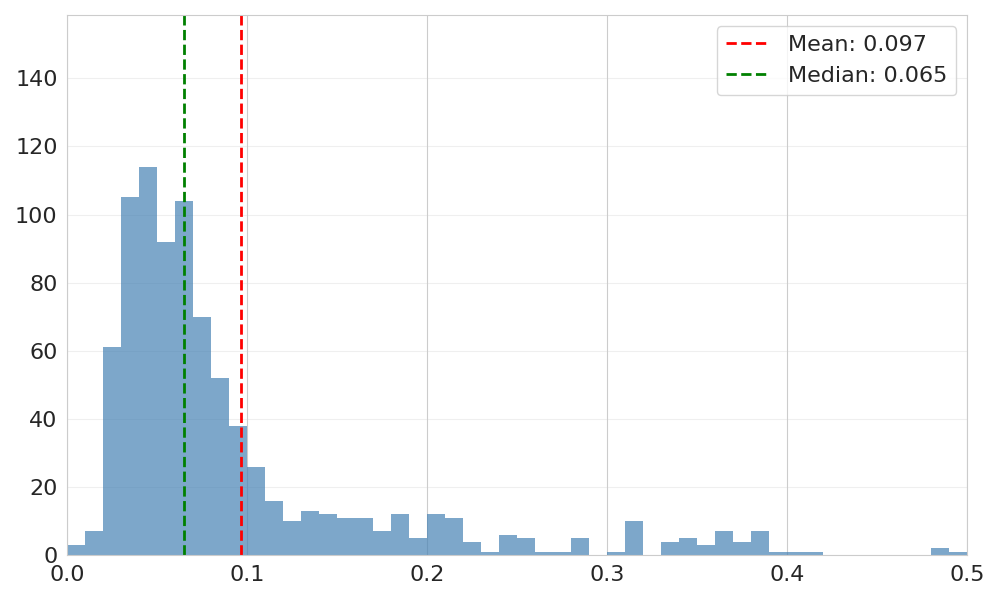}
& \includegraphics[width=\distroBB]{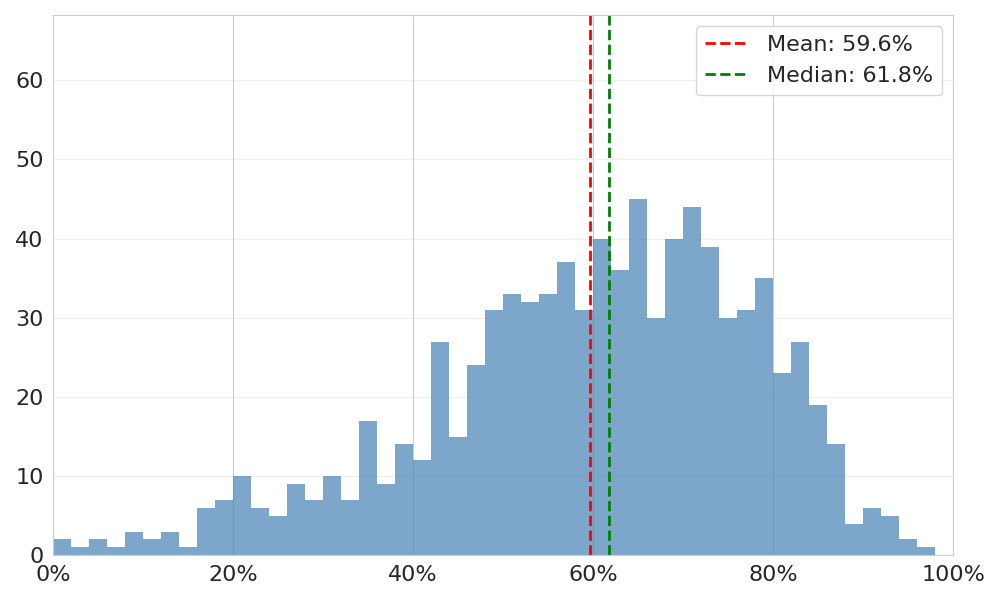}
& \includegraphics[width=\distroBB]{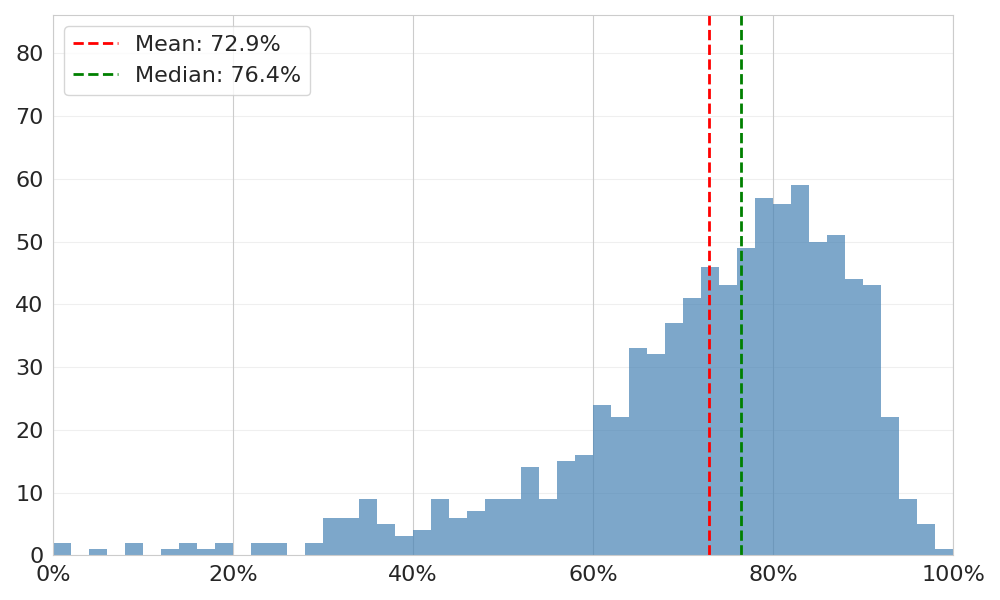} \\
\midrule

\shortstack{Trellis- \\ RePaint}
& \checkmark
& \includegraphics[width=\distroBB]{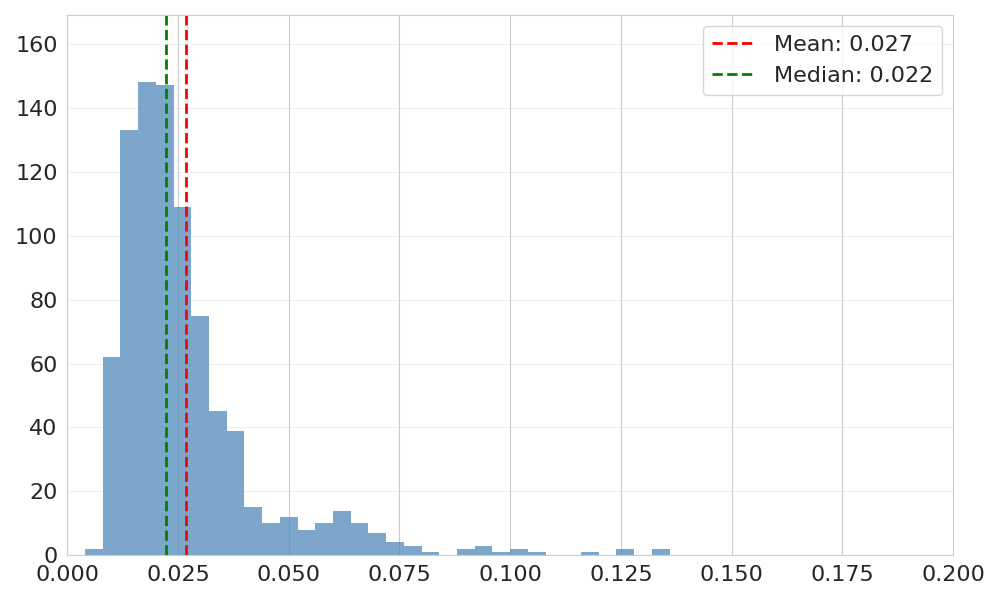}
& \includegraphics[width=\distroBB]{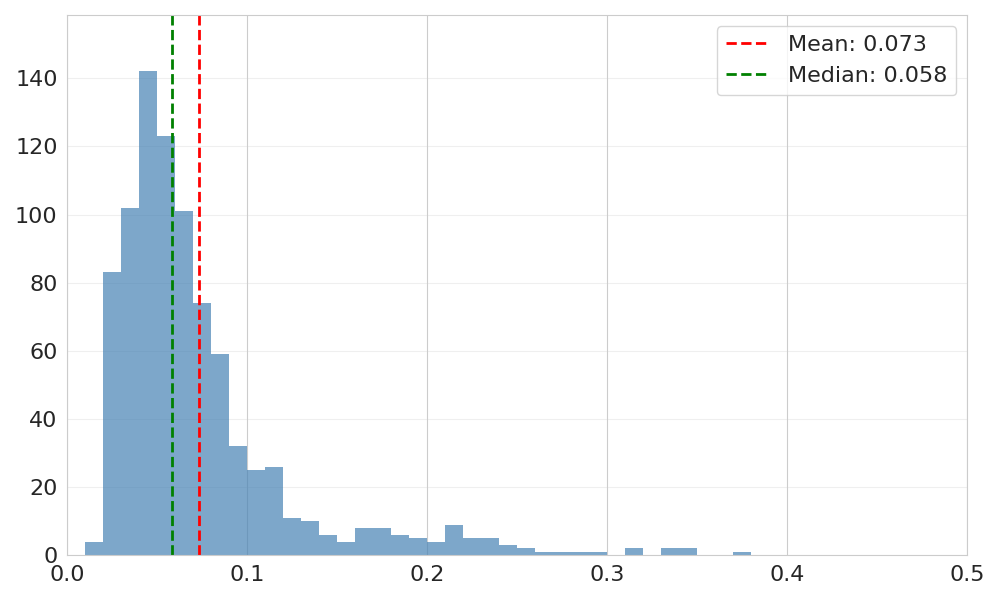}
& \includegraphics[width=\distroBB]{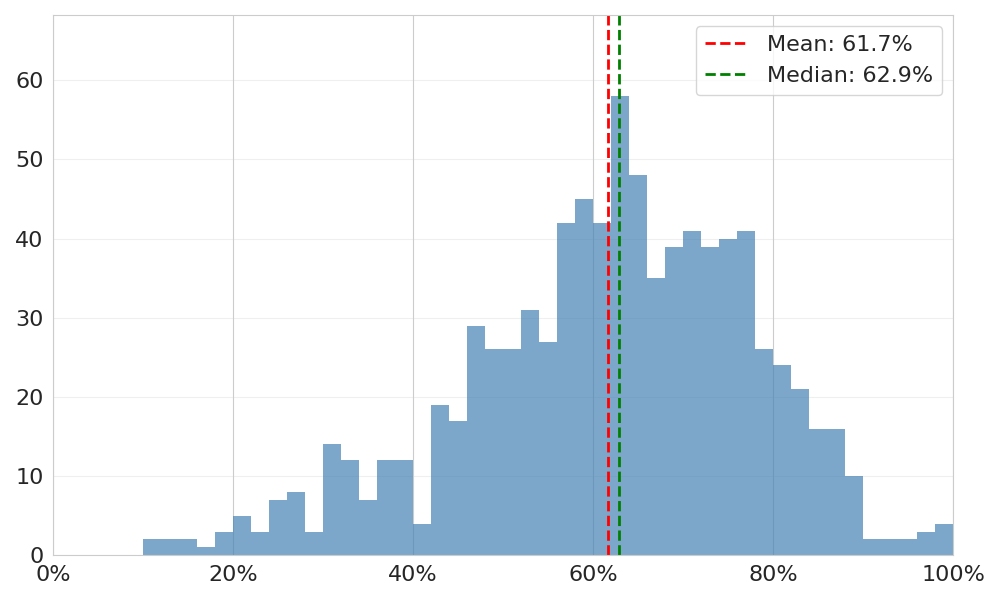}
& \includegraphics[width=\distroBB]{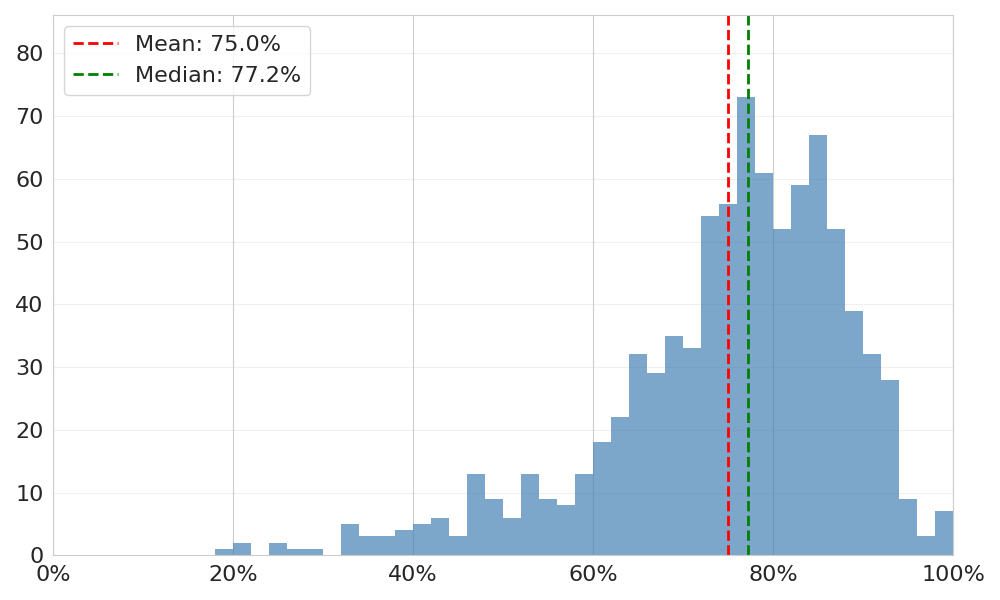} \\

3DMorph
& \checkmark
& \includegraphics[width=\distroBB]{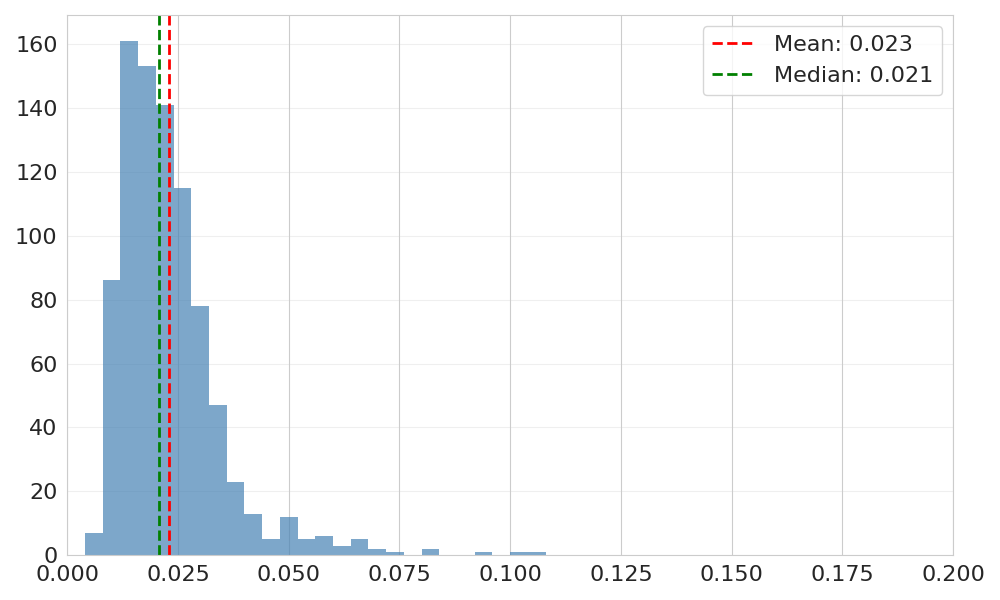}
& \includegraphics[width=\distroBB]{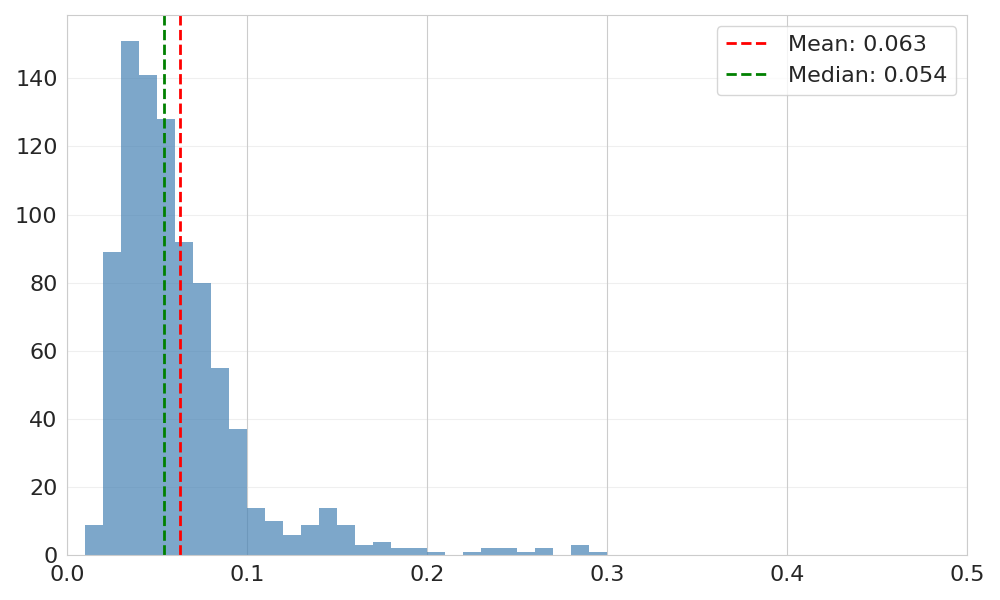}
& \includegraphics[width=\distroBB]{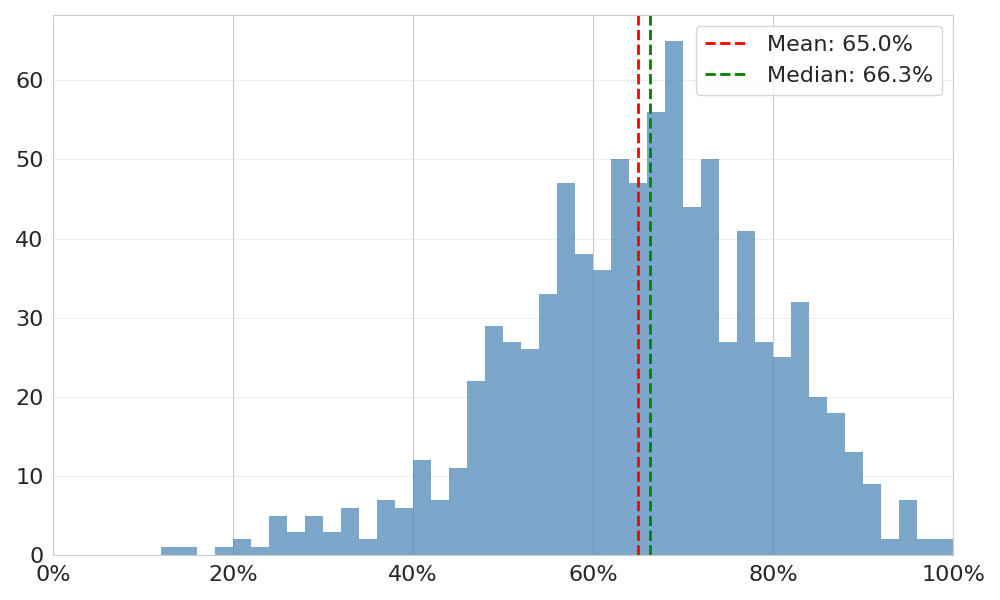}
& \includegraphics[width=\distroBB]{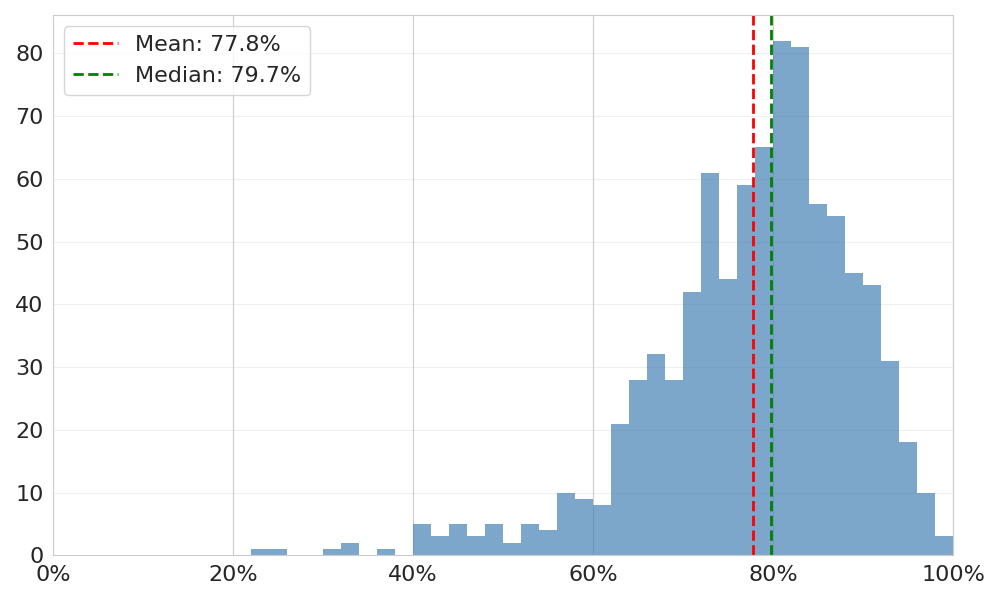} \\
\bottomrule

\end{tabular}
\caption{Full per-sample performance distributions for the \textbf{edit-region metrics} reported in Tab.~\ref{table:inside-BB}.}
\label{suppl-fig:distro-edit-eval}
\end{figure*}

\end{document}